\documentclass{article}

\usepackage{microtype}
\usepackage{graphicx}
\usepackage{subfigure}
\usepackage{booktabs}

\usepackage{url}
\usepackage[noend]{algpseudocode}
\usepackage{amsfonts}
\usepackage{amsthm}
\usepackage{amsbsy}
\usepackage{multirow}
\usepackage{makecell}
\usepackage{gensymb}
\usepackage{enumitem}
\usepackage{float}
\usepackage{varwidth}
\usepackage{tikz}
\usepackage{xcolor}
\usepackage{balance}
\usetikzlibrary{decorations.pathreplacing}
\usetikzlibrary{plotmarks}

\newcommand{\Reals}{\mathbb{R}}

\newcommand{\comment}[1]{}
\newcommand{\NA}[1]{#1}

\usepackage[nohyperref,accepted]{icml2020}

\icmltitlerunning{Linear Mode Connectivity and the Lottery Ticket Hypothesis}

\begin{document}

\twocolumn[
\icmltitle{Linear Mode Connectivity and the Lottery Ticket Hypothesis}




\begin{icmlauthorlist}
\icmlauthor{Jonathan Frankle}{mit}
\icmlauthor{Gintare Karolina Dziugaite}{element}
\icmlauthor{Daniel M. Roy}{toronto,vector}
\icmlauthor{Michael Carbin}{mit}
\end{icmlauthorlist}

\icmlaffiliation{mit}{MIT CSAIL}
\icmlaffiliation{element}{Element AI}
\icmlaffiliation{toronto}{University of Toronto}
\icmlaffiliation{vector}{Vector Institute}

\icmlcorrespondingauthor{Jonathan Frankle}{jfrankle@mit.edu}

\vskip 0.3in
]


\printAffiliationsAndNotice{}  

\begin{abstract}
We study whether a neural network optimizes to the same, linearly connected minimum under different samples of SGD noise (e.g., random data order and augmentation). We find that standard vision models become \emph{stable to SGD noise} in this way early in training. From then on, the outcome of optimization is determined to a linearly connected region.
We use this technique to study \emph{iterative magnitude pruning} (IMP), the procedure used by work on the lottery ticket hypothesis to identify subnetworks that could have trained in isolation to full accuracy. We find that these subnetworks only reach full accuracy when they are stable to SGD noise, which either occurs at initialization for small-scale settings (MNIST) or early in training for large-scale settings (ResNet-50 and Inception-v3 on ImageNet).
\end{abstract}

\section{Introduction}

When training a neural network with mini-batch stochastic gradient descent (SGD), training examples are presented to the network in a random order within each epoch.
In many cases, each example also undergoes random data augmentation.
This randomness can be seen as \emph{noise} that varies from training run to training run and alters the network's trajectory through the optimization landscape, even when the initialization and hyperparameters are fixed.
In this paper, we investigate how this SGD noise affects the outcome of optimizing neural networks and the role this effect plays in sparse, \emph{lottery ticket} networks \citep{lth}.

\def\normaltwo{\x,{1/exp(((5*\x)^2)/2)-3}}
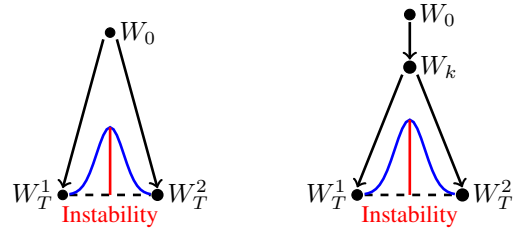
\begin{figure}
\centering
    \begin{tikzpicture}[scale=0.9]
    \filldraw[black] (0, -0.6) circle (2pt) node[anchor=west] {$W_0$};
    \draw[->, line width=1] (-0.1, -0.7) -> (-0.7, -2.9);
    \filldraw[black] (-0.7, -3) circle (2pt) node[anchor=east] {$W_T^1$};
    \draw[->, line width=1] (0.1, -0.7) -> (0.7, -2.9);
    \filldraw[black, line width=1] (0.7, -3) circle (2pt) node[anchor=west] {$W_T^2$};
    \draw[dashed, line width=1] (-0.6, -3) -- (0.6, -3);
    \draw[color=blue,domain=-0.6:0.6,line width=1] plot (\normaltwo) node[right] {};
    \draw[color=red,line width=1] (0, -2) -- (0, -3) node[label={[yshift=-0.7cm]:{\small Instability}},red] {};
    \end{tikzpicture}%
    \hspace{3em}%
    \begin{tikzpicture}[scale=1]
    \filldraw[black] (0, -0.6) circle (2pt) node[anchor=west] {$W_0$};
    \draw[->, line width=1] (0, -0.7) -> (0, -1.2);
    \filldraw[black, line width=1] (0, -1.3) circle (2pt) node[anchor=west] {$W_k$};
    \draw[->, line width=1] (-0.1, -1.4) -> (-0.7, -2.9);
    \filldraw[black] (-0.7, -3) circle (2pt) node[anchor=east] {$W_T^1$};
    \draw[->, line width=1] (0.1, -1.4) -> (0.7, -2.9);
    \filldraw[black, line width=1] (0.7, -3) circle (2pt) node[anchor=west] {$W_T^2$};
    \draw[dashed, line width=1] (-0.6, -3) -- (0.6, -3);
    \draw[color=blue,domain=-0.6:0.6,line width=1] plot (\normaltwo) node[right] {};
    \draw[color=red,line width=1] (0, -2) -- (0, -3) node[label={[yshift=-0.7cm]:{\small Instability}},red] {};
    \end{tikzpicture}
    \vspace{-1.3em}
    \caption{A diagram of instability analysis from step 0 (left) and step $k$ (right) when comparing networks using linear interpolation.
    }
    \label{fig:stability-visualization}   
    \vspace{-6mm} 
\end{figure}

\textbf{Instability analysis.}
To study these questions, we propose \emph{instability analysis}.
The goal of instability analysis is to determine whether the outcome of optimizing a particular neural network is \emph{stable to SGD noise}.
Figure \ref{fig:stability-visualization} (left) visualizes instability analysis.
First, we create a network $\mathcal{N}$ with random initialization $W_0$.
We then train two copies of $\mathcal{N}$ with different samples of SGD noise (i.e., different random data orders and augmentations).
Finally, we compare the resulting networks to measure the effect of these different samples of SGD noise on the outcome of optimization.
If the networks are sufficiently similar according to a criterion, we determine $\mathcal{N}$ to be stable to SGD noise.
We also study this behavior starting from the state of $\mathcal{N}$ at step $k$ of training (Figure \ref{fig:stability-visualization} right).
Doing so allows us to determine \emph{when} the outcome of optimization becomes stable to SGD noise.

There are many possible ways in which to compare the networks that result from instability analysis (Appendix \ref{app:alternate-distance-metrics}).
We use the behavior of the optimization landscape along the line between these networks (blue in Figure \ref{fig:stability-visualization}).
Does error remain flat or even decrease (meaning the networks are in the same, linearly connected minimum), or is there a barrier of increased error?
We define the \emph{linear interpolation instability} of $\mathcal{N}$ to SGD noise as the maximum increase in error along this path (red).
We consider $\mathcal{N}$ stable to SGD noise if error does not increase along this path, i.e., instability $\approx$ 0.
This means $\mathcal{N}$ will find the same, linearly connected minimum regardless of the sample of SGD noise.

By linearly interpolating at the end of training in this fashion, we assess a linear form of \emph{mode connectivity}, a phenomenon where the minima found by two networks are connected by a path of nonincreasing error. 
\citet{freeman2016topology}, \citet{draxler2018essentially}, and \citet{garipov2018loss} show that the modes of standard vision networks trained from different initializations are connected by nonlinear paths of constant error or loss.
Based on this work, we expect that all networks we examine are connected by such paths.
However, the modes found by \citeauthor{draxler2018essentially} and \citeauthor{garipov2018loss} are not connected by \emph{linear} paths.
The only extant example of linear mode connectivity is by \citet{kolter}, who train MLPs from the same initialization on disjoint subsets of MNIST and find that the resulting networks are connected by linear paths of constant test error.
\NA{In contrast, we explore linear mode connectivity from points throughout training, we do so at larger scales, and we focus on different samples of SGD noise rather than disjoint samples of data}.

\begin{table*}
\scriptsize
\centering
\begin{tabular}{l@{\ }l@{\ }|@{\ }c@{\ }c@{\ } c@{\ }c@{\ \ }c@{\ \ }c@{\ \ }c@{\ }c@{\ } c@{\ } c@{\ }|@{\ }c@{\ }c@{\ }}
\toprule
Network & Variant & Dataset & Params& Train Steps & Batch & Accuracy & Optimizer & Rate & Schedule & Warmup & BatchNorm & Pruned Density & Style \\ \midrule
LeNet & & MNIST & 266K & 24K Iters & 60 & 98.3 $\pm$ 0.1\% & adam &  12e-4 & constant & 0 & No &  3.5\% & Iterative \\ \midrule
ResNet-20 & Standard &  &  &  &  & 91.7 $\pm$ 0.1\% &  &  0.1 & \multirow{3}{*}{\makecell{10x drop at\\32K, 48K}}&  0 & \multirow{3}{*}{\makecell{Yes}} & 16.8\% & \multirow{3}{*}{\makecell{Iterative}} \\
ResNet-20 & Low & CIFAR-10 & 274K & 63K Iters & 128 & 88.8 $\pm$ 0.1\% & momentum & 0.01 & & 0 &  & 8.6\% \\
ResNet-20 & Warmup &  & & & & 89.7 $\pm$ 0.3\% &  & 0.03 &  &  30K & & 8.6\% \\ \midrule
VGG-16 & Standard & & & & & 93.7 $\pm$ 0.1\% &  & 0.1 & \multirow{3}{*}{\makecell{10x drop at\\32K, 48K}} & 0 & \multirow{3}{*}{\makecell{Yes}} & 1.5\% & \multirow{3}{*}{\makecell{Iterative}} \\
VGG-16 & Low & CIFAR-10 & 14.7M & 63K Iters & 128& 91.7 $\pm$ 0.1\% & momentum & 0.01 & &  0 &  & 5.5\% \\
VGG-16 & Warmup & & & & & 93.4 $\pm$ 0.1\% &  & 0.1 & & 30K & & 1.5\% \\ \midrule
ResNet-50 & & ImageNet & 25.5M & 90 Eps & 1024 & 76.1 $\pm$ 0.1\% & momentum & 0.4 & 10x drop at 30,60,80 & 5 Eps & Yes &  30\% & One-Shot \\
Inception-v3 & & ImageNet & 27.1M & 171 Eps & 1024 & 78.1 $\pm$ 0.1\% & momentum & 0.03 & linear decay to 0.005 & 0 & Yes & 30\% & One-Shot \\
\bottomrule
\end{tabular}
\caption{Our networks and hyperparameters.
Accuracies are the means and standard deviations across three initializations.
Hyperparameters for ResNet-20 standard are from \citet{resnet}. Hyperparameters for VGG-16 standard are from \citet{rethinking-pruning}.
Hyperparameters for \emph{low}, \emph{warmup}, and LeNet are adapted from \citet{lth}.
Hyperparameters for ImageNet networks are from Google's reference TPU code \citep{tpu-implementation}.
Note: \citeauthor{lth} mistakenly refer to ResNet-20 as ``ResNet-18,'' which is a separate network.}
\label{fig:small-networks}
\vspace{-1mm}
\end{table*}

We perform instability analysis on standard networks for MNIST, CIFAR-10, and ImageNet.
All but the smallest MNIST network are unstable to SGD noise at initialization according to linear interpolation.
However, by a point early in training (3\% for ResNet-20 on CIFAR-10 and 20\% for ResNet-50 on ImageNet), all networks become stable to SGD noise.
From this point on, the outcome of optimization is determined to a linearly connected minimum.

\textbf{The lottery ticket hypothesis.}
Finally, we show that instability analysis and linear interpolation are valuable scientific tools for understanding other phenomena in deep learning.
Specifically, we study the sparse networks discussed by the recent \emph{lottery ticket hypothesis} \citep[LTH;][]{lth}.
The LTH conjectures that, at initialization, neural networks contain sparse subnetworks that can train in isolation to full accuracy.

Empirical evidence for the LTH consists of experiments using a procedure called \emph{iterative magnitude pruning} (IMP).
On small networks for MNIST and CIFAR-10, IMP retroactively finds subnetworks at initialization that can train to the same accuracy as the full network (we call such subnetworks \emph{matching}).
Importantly, IMP finds matching subnetworks at \emph{nontrivial} sparsity levels, i.e., those beyond which subnetworks found by trivial random pruning are matching.
In more challenging settings, however, there is no empirical evidence for the LTH:
IMP subnetworks of VGGs and ResNets on CIFAR-10 and ImageNet are not matching at nontrivial sparsities \citep{rethinking-pruning, gale}.

We show that instability analysis distinguishes known cases where IMP succeeds and fails to find matching subnetworks at nontrivial sparsities, providing the first basis for understanding the mixed results in the literature.
Namely, IMP subnetworks are only matching when they are stable to SGD noise according to linear interpolation.
Using this insight, we identify new scenarios where we can find sparse, matching subnetworks at nontrivial sparsities, including in more challenging settings (e.g., ResNet-50 on ImageNet).
In these settings, sparse IMP subnetworks become stable to SGD noise \emph{early} in training rather than at initialization, just as we find with the unpruned networks.
Moreover, these stable IMP subnetworks are also matching.
In other words, early in training (if not at initialization), sparse subnetworks emerge that can complete training in isolation and reach full accuracy.
These findings shed new light on neural network training dynamics, hint at possible mechanisms underlying lottery ticket phenomena, and extend the lottery ticket observations to larger scales.

\textbf{Contributions.} We make the following contributions:
\vspace{-1em}
\begin{itemize}[itemsep=-0.25em,leftmargin=1em]
\item We introduce \emph{instability analysis} to determine whether the outcome of optimizing a neural network is stable to SGD noise, and we suggest linear mode connectivity for making this determination.
\item On a range of image classification benchmarks including standard networks on ImageNet, we observe that networks become stable to SGD noise early in training.
\item We use instability analysis to distinguish successes and failures of IMP (the method behind extant lottery ticket results).
Namely, sparse IMP subnetworks are matching only when they are stable to SGD noise.
\item We generalize IMP to find subnetworks early in training rather than at initialization.
We show that IMP subnetworks become stable and matching when set to their weights from early in training, making it possible to extend the lottery ticket observations to larger scales.
\end{itemize}
\vspace{-0.6em}

\begin{figure*}
\centering
\vspace{-3mm}
\begin{tikzpicture}[x=\textwidth,y=\textwidth, every node/.style = {anchor=north west}]
\node[anchor=center] at (0.5, -0.2) {\includegraphics[width=0.25\textwidth]{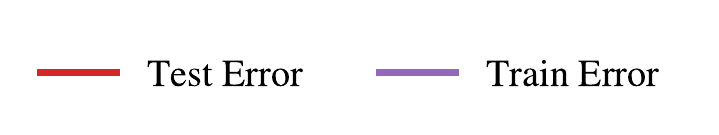}};
\node at (0.0, 0) {\includegraphics[width=0.19\textwidth]{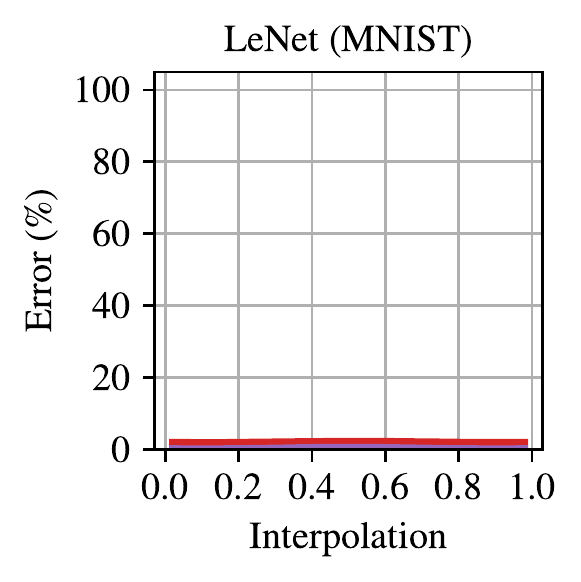}};
\node at (0.2, 0) {\includegraphics[width=0.19\textwidth]{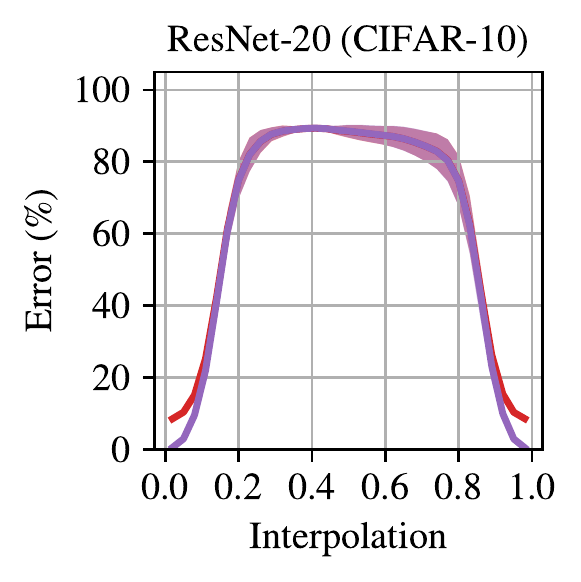}};
\node at (0.4, 0) {\includegraphics[width=0.19\textwidth]{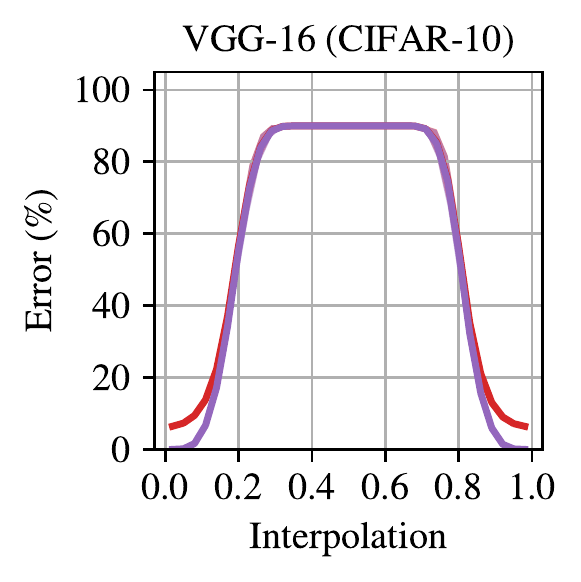}};
\node at (0.6, 0) {\includegraphics[width=0.19\textwidth]{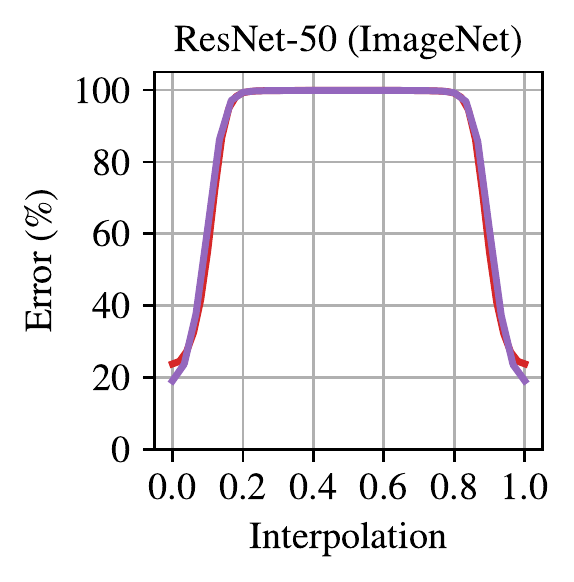}};
\node at (0.8, 0) {\includegraphics[width=0.19\textwidth]{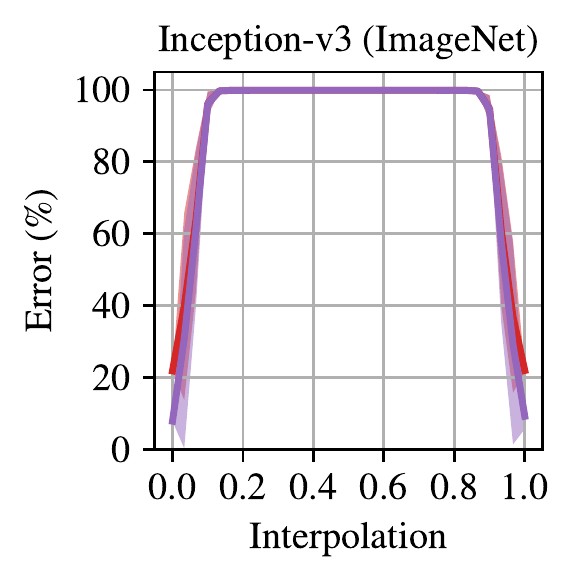}};
\end{tikzpicture}
\vspace{-11.5mm}
\caption{Error when linearly interpolating between networks trained from the same initialization with different SGD noise. Lines are means and standard deviations over three initializations and three data orders (nine samples total). Trained networks are at 0.0 and 1.0.}
\vspace{-5mm}
\label{fig:full-instability-at-init}
\end{figure*}

\vspace{-0.2em}
\section{Preliminaries and Methodology}
\label{sec:methodology}

\textbf{Instability analysis.}
To perform instability analysis on a network $\mathcal{N}$ with weights $W$, we make two copies of $\mathcal{N}$ and train them with different random samples of SGD noise (i.e., different data orders and augmentations), producing trained weights $W_T^1$ and $W_T^2$.
We compare these weights with a function, producing a value we call the \emph{instability} of $\mathcal{N}$ to SGD noise.
We then determine whether this value satisfies a criterion indicating that $\mathcal{N}$ is stable to SGD noise.
The weights of $\mathcal{N}$ could be randomly initialized ($W = W_0$ in Figure \ref{fig:stability-visualization}) or the result of $k$ training steps ($W = W_k$).

Formally, we model SGD by function $\mathcal{A}^{s \rightarrow t}: \Reals^D \times U \to \Reals^D$ that maps weights $W_{s} \in \Reals^D$ at step $s$ and SGD randomness $u \sim U$ to weights $W_{t} \in \Reals^D$ at step $t$ by training for $t-s$ steps ($s, t \in \{0,..,T\}$).
Algorithm \ref{alg:stability} outlines instability analysis with a function $f : \Reals^D \times \Reals^D \rightarrow \Reals$.
\vspace{-1em}
\begin{algorithm}[H]
\small
\caption{Compute instability of $W_k$ with function $f$.}
\begin{algorithmic}[1]
\State Train $W_k$ to $W_T^1$ with noise $u_1 \sim U$: $W_T^1 = \mathcal{A}^{k \rightarrow T}(W_k, u_1)$
\State Train $W_k$ to $W_T^2$ with noise $u_2 \sim U$: $W_T^2 = \mathcal{A}^{k \rightarrow T}(W_k, u_2)$
\State Return $f(W_T^1, W_T^2)$, i.e., the \emph{instability} of $W_k$ to SGD noise.
\end{algorithmic}
\label{alg:stability}
\end{algorithm}
\vspace{-1.5em}

\textbf{Linear interpolation.}
Consider a path $p$ on the optimization landscape between networks $W_1$ and $W_2$.
We define the \emph{error barrier height} of $p$ as the maximum increase in error from that of $W_1$ and $W_2$ along path $p$.
For instability analysis, we use as our function $f$ the error barrier height along the \emph{linear} path between two networks $W_1$ and $W_2$.\footnote{See Appendix \ref{app:alternate-distance-metrics} for alternate ways of comparing the networks.}

Formally, let $\mathcal{E}(W)$ be the (train or test) error of a network with weights $W$.
Let $\mathcal{E}_\alpha(W_1, W_2) = \mathcal{E}(\alpha W_1 + (1-\alpha)W_2)$ for $\alpha \in [0, 1]$ be the error of the network created by linearly interpolating between $W_1$ and $W_2$.
Let $\mathcal{E}_{\mathsf{sup}}(W_1, W_2) = \sup_{\alpha} \mathcal{E}_\alpha(W_1, W_2)$ be the highest error when interpolating in this way.
Finally, let $\bar{\mathcal{E}}(W_1, W_2) = \mathsf{mean}(\mathcal{E}(W_1), \mathcal{E}(W_2))$.
The error barrier height on the linear path between $W_1$ and $W_2$ (which is our function $f$ for instability analysis) is $\mathcal{E}_{\mathsf{sup}}(W_1, W_2) - \bar{\mathcal{E}}(W_1, W_2)$ (red line in Figure \ref{fig:stability-visualization}).
When we perform instability analysis on a network $\mathcal{N}$ with this function, we call this quantity the \emph{linear interpolation instability} (shorthand: \emph{instability}) of $\mathcal{N}$ to SGD noise.

\textbf{Linear mode connectivity.}
Two networks $W_1$ and $W_2$ are \emph{mode connected} if there exists a path between them along which the error barrier height $\approx 0$ \citep{draxler2018essentially, garipov2018loss}.
They are \emph{linearly mode connected} if this is true along the linear path.
For instability analysis, we consider a network $\mathcal{N}$ stable to SGD noise (shorthand: \emph{stable}) when the networks that result from instability analysis are linearly mode connected; that is, when the linear interpolation instability of $\mathcal{N} \approx 0$.
Otherwise, it is unstable to SGD noise (shorthand: \emph{unstable}).
Empirically, we consider instability $<$ 2\% to be stable; this margin accounts for noise and matches increases in error along paths found by \citet[Table B.1]{draxler2018essentially} and \citet[Table 2]{garipov2018loss}.
We use 30 evenly-spaced values of $\alpha$, and we average instability from three initializations and three runs per initialization (nine combinations total).

\textbf{Networks and datasets.}
We study image classification networks on MNIST, CIFAR-10, and ImageNet as listed in Table \ref{fig:small-networks}.
All hyperparameters are standard values from reference code or prior work as cited in Table \ref{fig:small-networks}.
The \emph{warmup} and \emph{low} variants of ResNet-20 and VGG-16 are adapted from hyperparameters in \citet{lth}.

\section{Instability Analysis of Unpruned Networks}
\label{sec:full-networks}

In this section, we perform instability analysis on the standard networks in Table \ref{fig:small-networks} from many points during training.
We find that, although only LeNet is stable to SGD noise at initialization, every network becomes stable early in training, meaning the outcome of optimization from that point forward is determined to a linearly connected minimum.

\begin{figure*}
\centering
\begin{tikzpicture}[x=\textwidth,y=\textwidth, every node/.style = {anchor=north west}]
\node[anchor=center] at (0.5, -0.195) {\includegraphics[width=0.25\textwidth]{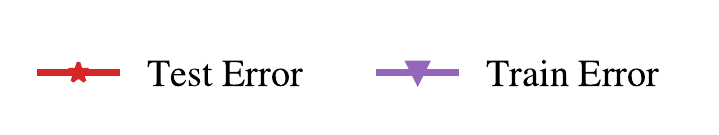}};
\node at (0.0, 0) {\includegraphics[width=0.19\textwidth]{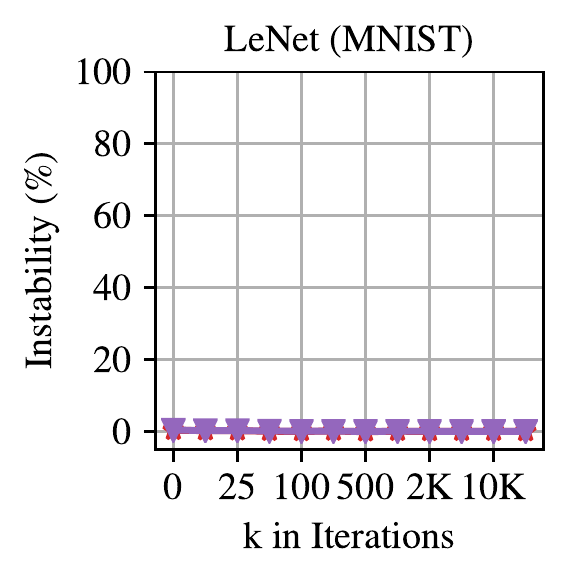}};
\node at (0.2, 0) {\includegraphics[width=0.19\textwidth]{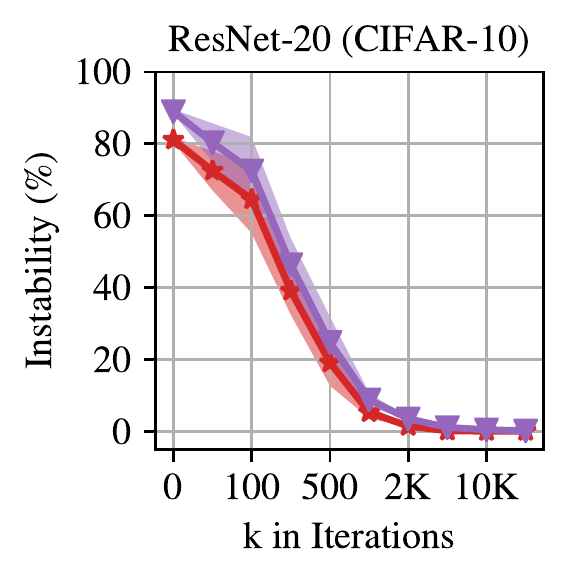}};
\node at (0.4, 0) {\includegraphics[width=0.19\textwidth]{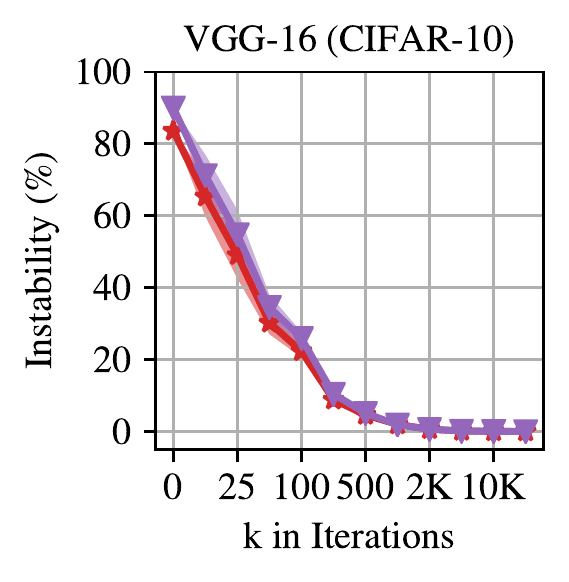}};
\node at (0.6, 0) {\includegraphics[width=0.19\textwidth]{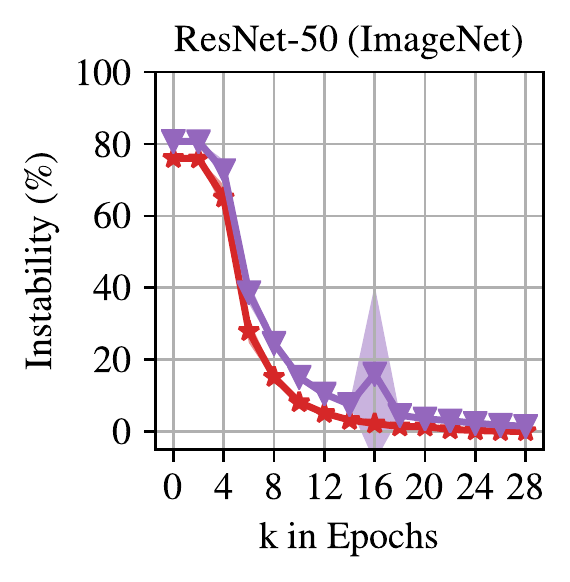}};
\node at (0.8, 0) {\includegraphics[width=0.19\textwidth]{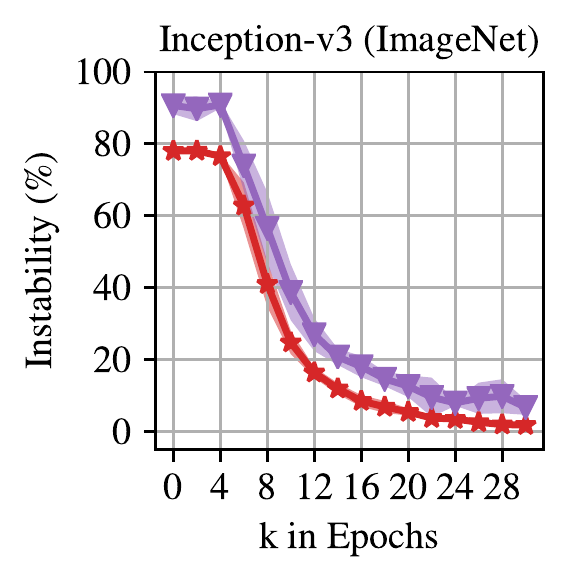}};
\end{tikzpicture}
\vspace{-11.5mm}
\caption{Linear interpolation instability when starting from step $k$.
 Each line is the mean and standard deviation across three initializations and three data orders (nine samples in total).}
\label{fig:full-instability-later}
\vspace{-4mm}
\end{figure*}

\begin{figure}
\centering
\vspace{-2mm}
\begin{tikzpicture}[x=\textwidth,y=\textwidth, every node/.style = {anchor=north west}]
\node[anchor=center] at (0.2, -0.2) {\includegraphics[width=0.8\columnwidth]{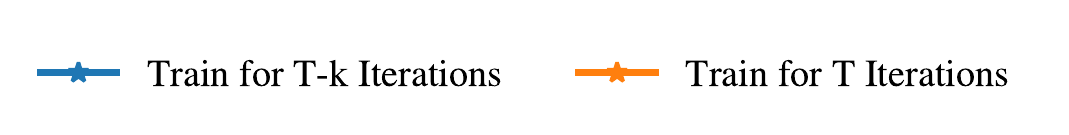}};
\node at (0, 0) {\includegraphics[width=0.4\columnwidth]{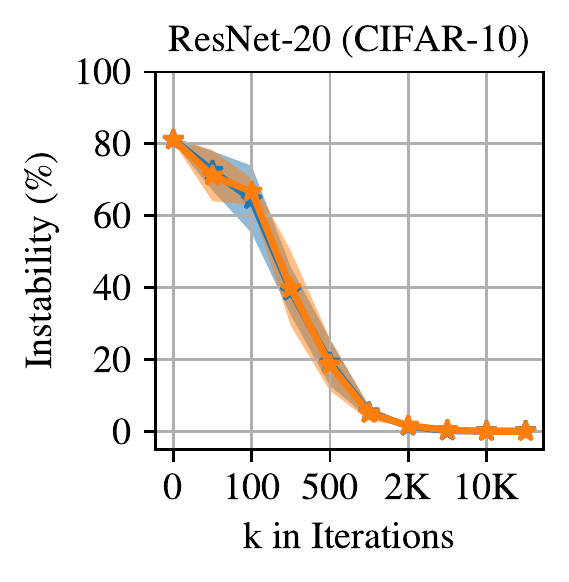}};
\node at (0.2, 0) {\includegraphics[width=0.4\columnwidth]{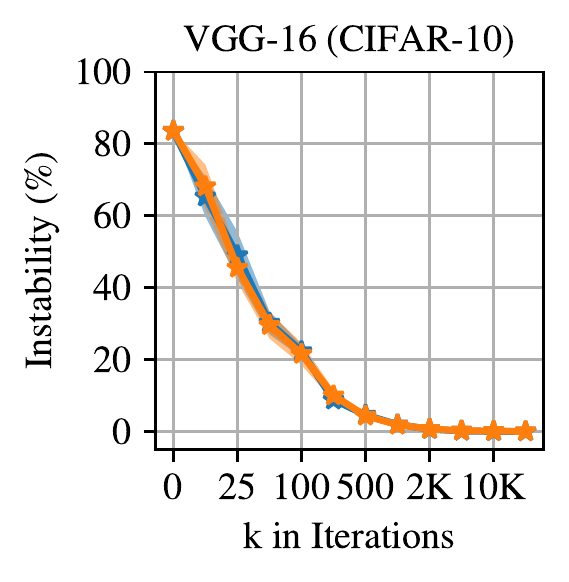}};
\end{tikzpicture}
\vspace{-6mm}
\caption{Linear interpolation instability on the test set when making two copies of the state of the network at step $k$ and either (1) training for the remaining $T-k$ steps (blue) or (2) training for $T$ steps with the learning rate schedule reset to step 0 (orange).}
\label{fig:disentangling}
\vspace{-4mm}
\end{figure}

\textbf{Instability analysis at initialization.}
We first perform instability analysis from initialization.
We use Algorithm \ref{alg:stability} with $W_0$ (visualized in Figure \ref{fig:stability-visualization} left): train two copies of the same, randomly initialized network with different samples of SGD noise.
Figure \ref{fig:full-instability-at-init} shows the train (purple) and test (red) error when linearly interpolating between the minima found by these copies.
Except for LeNet (MNIST), none of the networks are stable at initialization.
In fact, train and test error rise to the point of random guessing when linearly interpolating.
LeNet's error rises slightly, but by less than a percentage point.
We conclude that, in general, larger-scale image classification networks are unstable at initialization according to linear interpolation.

\textbf{Instability analysis during training.}
Although larger networks are unstable at initialization, they may become stable at some point afterwards;
for example, in the limit, they will be stable trivially after the last step of training.
To investigate when networks become stable, we perform instability analysis using the state of the network at various training steps.
That is, we train a network for $k$ steps, make two copies, train them to completion with different samples of SGD noise, and linearly interpolate (Figure \ref{fig:stability-visualization} right).
We do so for many values of $k$, assessing whether there is a point after which the outcome of optimization is determined to a linearly connected minimum regardless of SGD noise.

For each $k$, Figure \ref{fig:full-instability-later} shows the linear interpolation instability of the network at step $k$, i.e., the maximum error during interpolation (the peaks in Figure \ref{fig:full-instability-at-init}) minus the mean of the errors of the two networks (the endpoints in Figure \ref{fig:full-instability-at-init}).
In all cases, test set instability decreases as $k$ increases, culminating in stable networks.
The steps at which networks become stable are early in training.
For example, they do so at iterations 2000 for ResNet-20 and 1000 VGG-16; in other words, after 3\% and 1.5\% of training, SGD noise does not affect the final linearly connected minimum.
ResNet-50 and Inception-v3 become stable later: at epoch 18 (20\% into training) and 28 (16\%), respectively, using the test set.

For LeNet, ResNet-20, and VGG-16, instability is essentially identical when measured in terms of train or test error, and the networks become stable to SGD noise at the same time for both quantities.
For ResNet-50 and Inception-v3, train instability follows the same trend as test instability but is slightly higher at all points, meaning train set stability occurs later for ResNet-50 and does not occur in our range of analysis for Inception-v3.
Going forward, we present all results with respect to test error for simplicity and include corresponding train error data in the appendices.
3
\begin{algorithm}
    \small
    \caption{IMP rewinding to step $k$ and $N$ iterations.}
    \begin{algorithmic}[1]
    \State Create a network with randomly initialization $W_0 \in \mathbb{R}^d$.
    \State Initialize pruning mask to $m = 1^{d}$.
    \State Train $W_0$ to $W_k$ with noise $u \sim U$: $W_k = \mathcal{A}^{0 \rightarrow k}(W_0, u)$.
    \For{$n \in \{1, \ldots, N\}$}
    \State \begin{varwidth}[t]{\linewidth}Train $m \odot W_k$ to $m \odot W_T$ with noise $u' \sim U$:\\$W_T = \mathcal{A}^{k \rightarrow T}(m \odot W_k, u')$.\end{varwidth}
    \State \begin{varwidth}[t]{\linewidth}Prune the lowest magnitude entries of $W_T$ that remain.\\Let $m[i] = 0$ if $W_T[i]$ is pruned.\end{varwidth}
    \EndFor
    \State Return $W_k, m$ 
    \end{algorithmic}
    \label{alg:imp}
\end{algorithm}

\textbf{Disentangling instability from training time.}
Varying the step $k$ from which we run instability analysis has two effects.
First, it changes the state of the network from which we train two copies to completion on different SGD noise.
Second, it changes the number of steps for which those copies are trained: when we run instability analysis from step $k$, we train the copies under different SGD noise for $T-k$ steps.
As $k$ increases, the copies have fewer steps during which to potentially find linearly unconnected minima.
It is possible that the gradual decrease in instability as $k$ increases and the eventual emergence of linear mode connectivity is just an artifact of these shorter training times.

To disentangle the role of training time in our experiments, we modify instability analysis to train the copies for $T$ iterations no matter the value of $k$.
When doing so, we reset the learning rate schedule to iteration 0 after making the copies.
In Figure \ref{fig:disentangling}, we compare instability with and without this modification for ResNet-20 and VGG-16 on CIFAR-10.
Instability is indistinguishable in both cases, indicating that the different numbers of training steps did not play a role in the earlier results.
Going forward, we present all results by training copies for $T-k$ steps as in Algorithm \ref{alg:stability}.

\begin{table}
\scriptsize
\centering
\begin{tabular}{@{\ }l@{\ }|@{\ }c@{\ \ }c@{\ \ }c@{\ }c@{\ }|@{\ }c@{\ }c@{\ }}
\toprule
Network & Full & IMP & Rand Prune & Rand Reinit & $\Delta$ IMP & Matching? \\
\midrule
LeNet & 98.3 & 98.2  & 96.7 & 97.5 &  0.1 & Y\\ \midrule
ResNet-20 & 91.7 & 88.5 & 88.6 & 88.8 & 3.2 & N\\
ResNet-20 Low & 88.8 & 89.0 & 85.7 & 84.7 & -0.2 & Y\\
ResNet-20 Warmup & 89.7 & 89.6 & 85.7 & 85.6 & 0.1 & Y \\ \midrule
VGG-16 & 93.7  & 90.9 & 89.4 & 91.0 & 2.8 & N\\
VGG-16 Low & 91.7  &  91.6 & 90.1 & 90.2 & 0.1 & Y\\
VGG-16 Warmup & 93.4 & 93.2 & 90.1 &  90.7 & 0.2 & Y\\ \midrule
ResNet-50 & 76.1 & 73.7 & 73.1 & 73.4 & 2.4 & N \\
Inception-v3 & 78.1 & 75.7 & 75.2 & 75.5 & 2.4 & N \\
\bottomrule
\end{tabular}
\vspace{-2mm}
\caption{Accuracy of IMP and random subnetworks when rewinding to $k=0$ at the sparsities in Table \ref{fig:small-networks}. Accuracies are means across three initializations. All standard deviations are $< 0.2$.}
\vspace{-4mm}
\label{tab:lth-at-zero}
\end{table}

\newpage
\section{Instability Analysis of Lottery Tickets}
\label{sec:pruned-networks}

In this section, we leverage instability analysis and our observations about linear mode connectivity to gain new insights into the behavior of sparse \emph{lottery ticket} networks.

\vspace{-0.75mm}
\subsection{Overview}
\vspace{-0.75mm}

We have long known that it is possible to \emph{prune} neural networks after training, often removing 90\% of weights without reducing accuracy after some additional training \citep[e.g.,][]{reed1993pruning,han-pruning,gale}.
However, sparse networks are more difficult to train from scratch.
Beyond \emph{trivial} sparsities where many weights remain and random subnetworks can train to full accuracy, sparse networks trained in isolation are generally less accurate than the corresponding dense networks \citep{han-pruning,pruning-filters,rethinking-pruning, lth}.

However, there is a known class of sparse networks that remain accurate at nontrivial sparsities.
On small vision tasks, an algorithm called \emph{iterative magnitude pruning} (IMP) retroactively finds sparse subnetworks that were capable of training in isolation from initialization to full accuracy at the sparsities attained by pruning \citep{lth}.
The existence of such subnetworks raises the prospect of replacing conventional, dense networks with sparse ones, creating new opportunities to reduce the cost of training.
However, in more challenging settings, IMP subnetworks perform no better than subnetworks chosen randomly, meaning they only train to full accuracy at trivial sparsities \citep{rethinking-pruning, gale}.

We find that instability analysis offers new insights into the behavior of IMP subnetworks and a potential explanation for their successes and failures.
Namely, the sparsest IMP subnetworks only train to full accuracy when they are stable to SGD noise.
In other words, when different samples of SGD noise cause an IMP subnetwork to find minima that are not linearly connected, then test accuracy is lower.

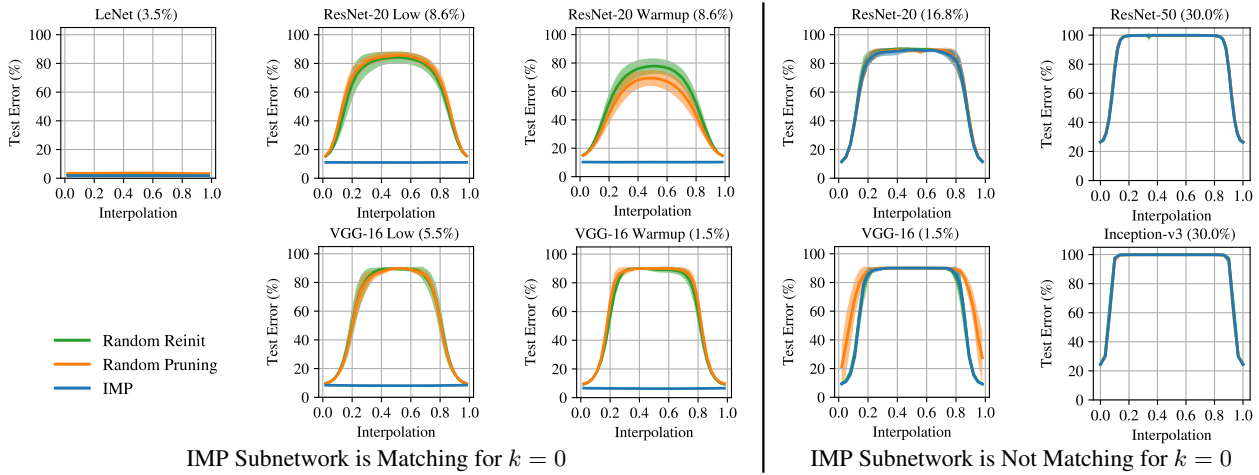
\begin{figure*}
\begin{tikzpicture}[x=\textwidth,y=\textwidth, every node/.style = {anchor=north west}]
\node at (0.0, 0) {\includegraphics[width=0.18\textwidth]{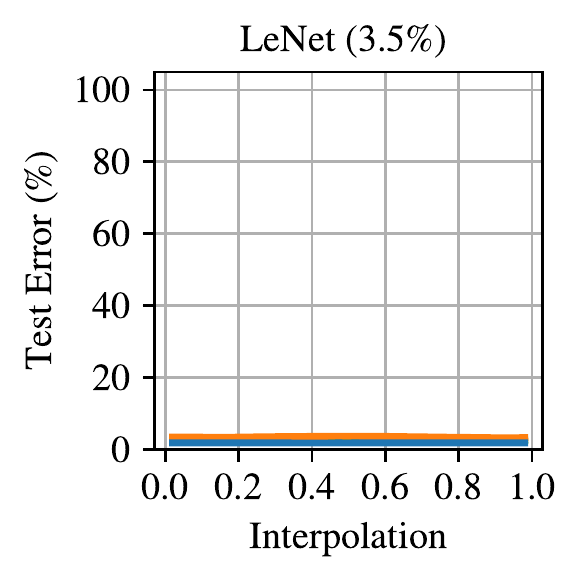}};
\node at (0.2, 0) {\includegraphics[width=0.18\textwidth]{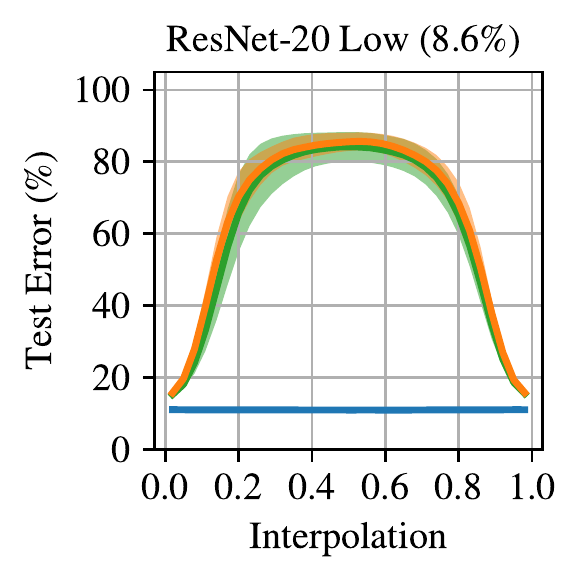}};
\node at (0.4, 0) {\includegraphics[width=0.18\textwidth]{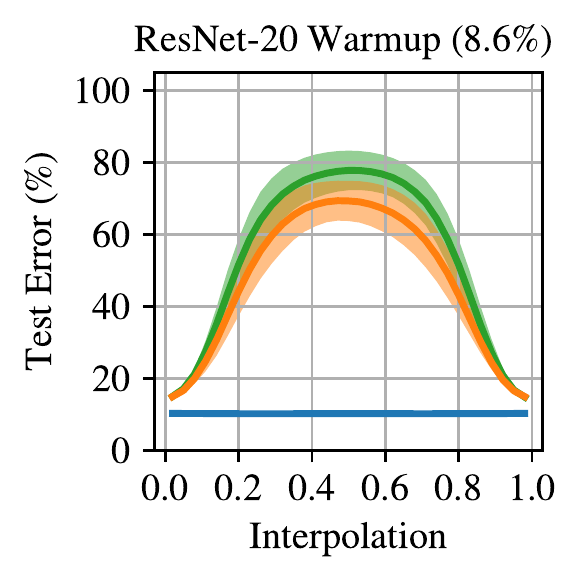}};
\node at (0.6, 0) {\includegraphics[width=0.18\textwidth]{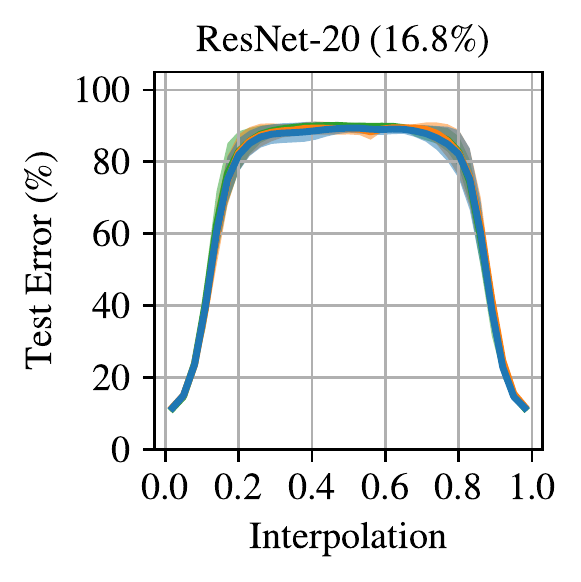}};
\node at (0.8, 0) {\includegraphics[width=0.18\textwidth]{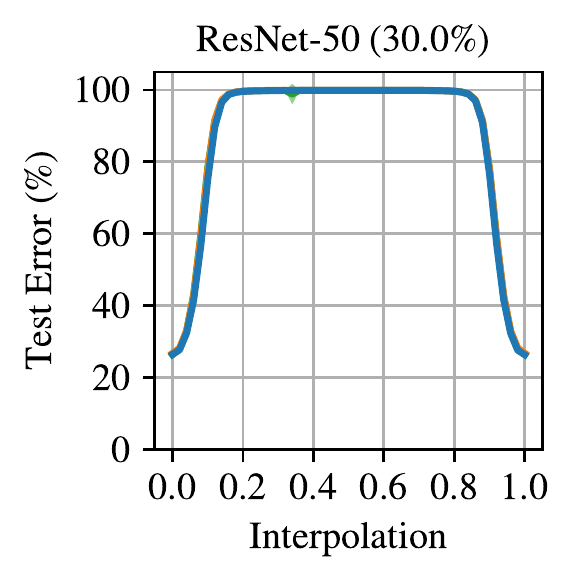}};

\node at (0.03, -0.24) {\includegraphics[width=0.16\textwidth]{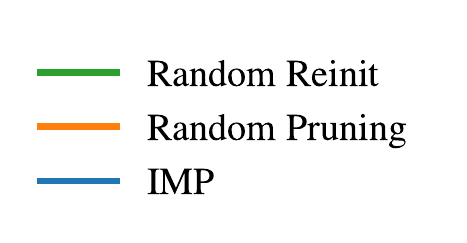}};
\node at (0.2,  -0.17) {\includegraphics[width=0.18\textwidth]{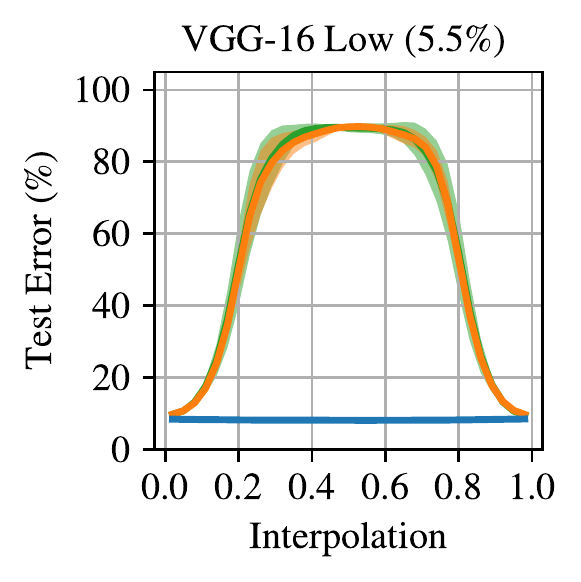}};
\node at (0.4,  -0.17) {\includegraphics[width=0.18\textwidth]{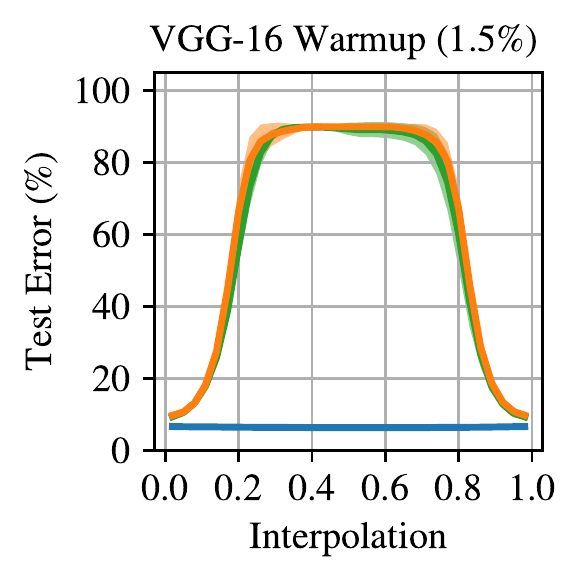}};
\node at (0.6,  -0.17) {\includegraphics[width=0.18\textwidth]{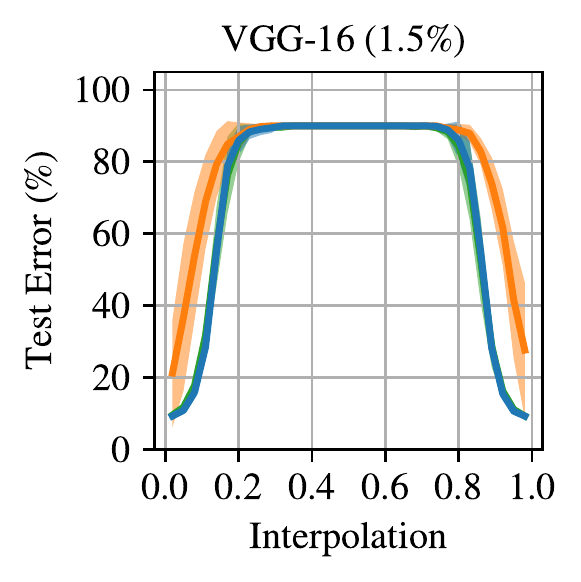}};
\node at (0.8,  -0.17) {\includegraphics[width=0.18\textwidth]{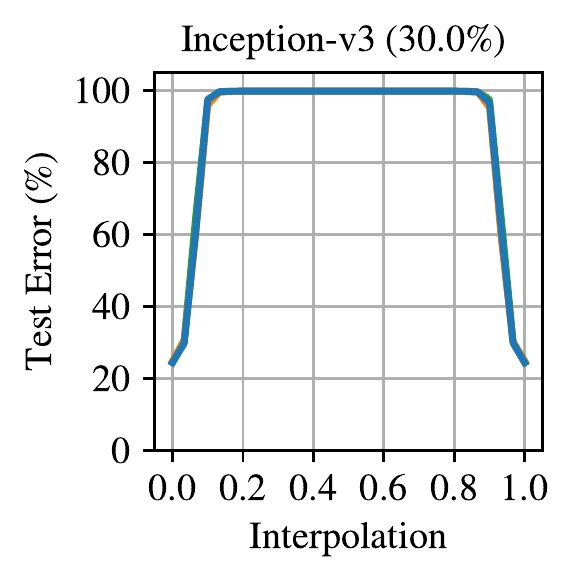}};

\node[anchor=center] at (0.3, -0.365) {\small IMP Subnetwork is Matching for $k=0$};
\node[anchor=center] at (0.8, -0.365) {\small IMP Subnetwork is Not Matching for $k=0$};
\draw[line width=0.3mm] (0.6, -0.01) -- (0.6, -0.375);
\end{tikzpicture}
\vspace{-1em}
\caption{Test error when linearly interpolating between subnetworks trained from the same initialization with different SGD noise. 
Lines are means and standard deviations over three initializations and three data orders (nine in total). Percents are weights remaining.}
\label{fig:imp-instability-at-zero}
\vspace{-3mm}
\end{figure*}

\begin{figure*}
\vspace{-2mm}
\begin{tikzpicture}[x=\textwidth,y=\textwidth, every node/.style = {anchor=north west}]
\node at (0.0, 0) {\includegraphics[width=0.18\textwidth]{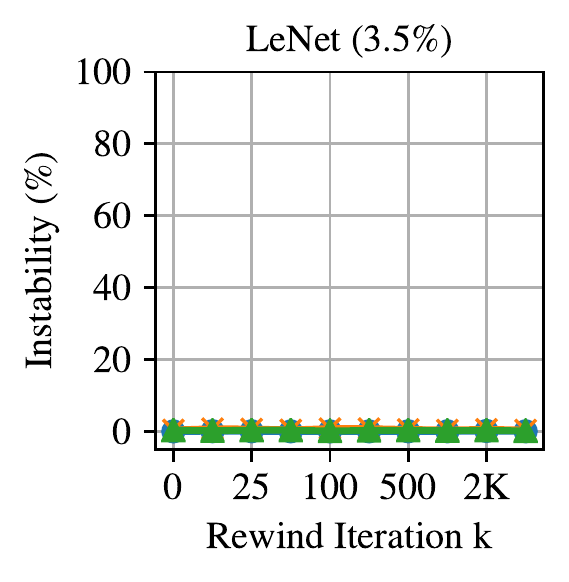}};
\node at (0.2, 0) {\includegraphics[width=0.18\textwidth]{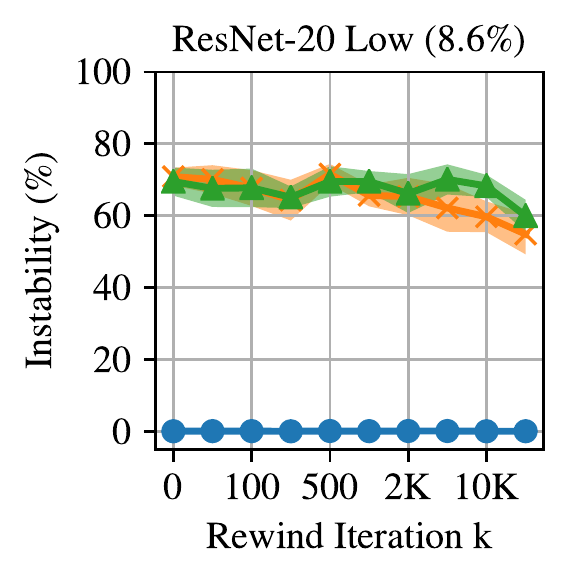}};
\node at (0.4, 0) {\includegraphics[width=0.18\textwidth]{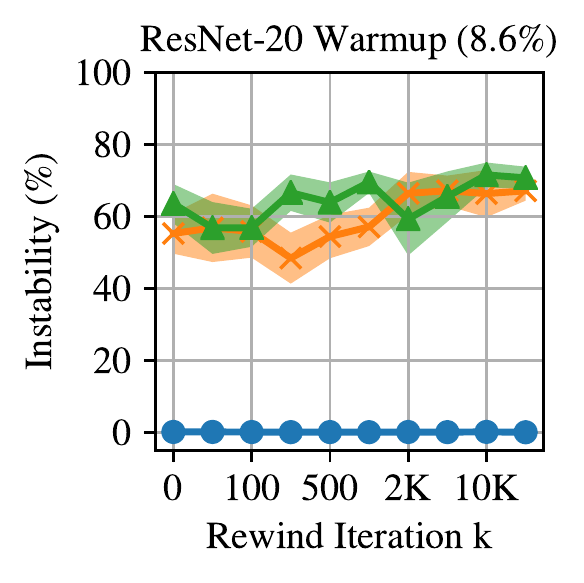}};
\node at (0.6, 0) {\includegraphics[width=0.18\textwidth]{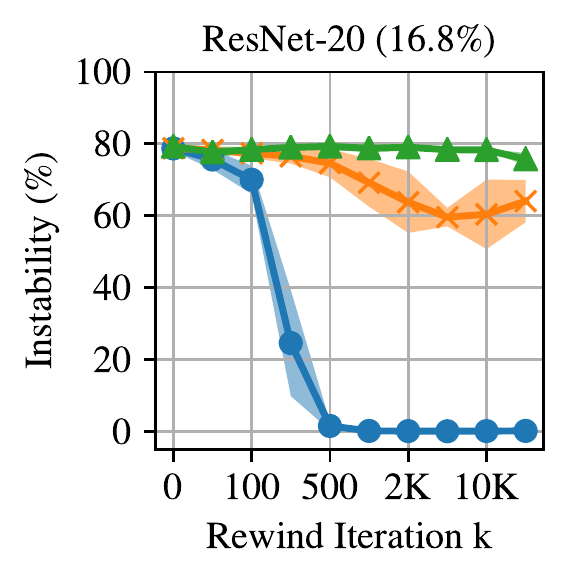}};
\node at (0.8, 0) {\includegraphics[width=0.18\textwidth]{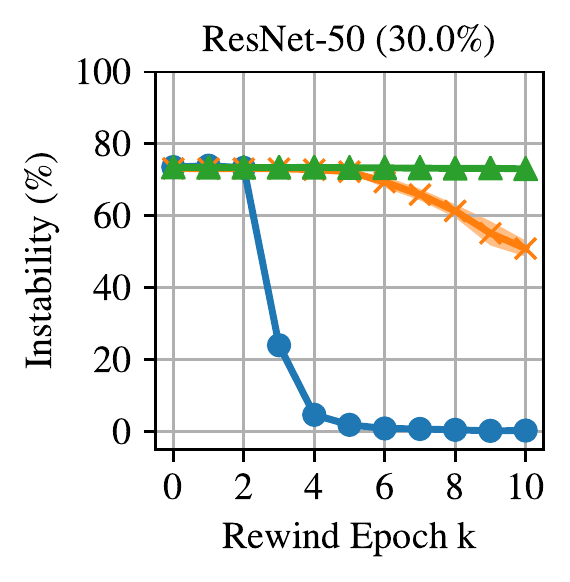}};

\node at (0.03, -0.24) {\includegraphics[width=0.16\textwidth]{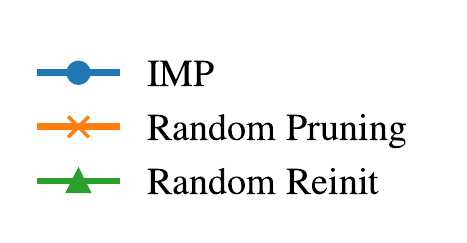}};
\node at (0.2,  -0.17) {\includegraphics[width=0.18\textwidth]{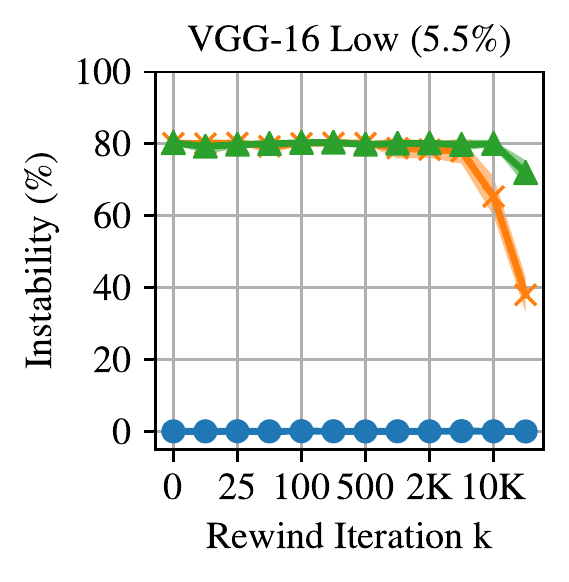}};
\node at (0.4,  -0.17) {\includegraphics[width=0.18\textwidth]{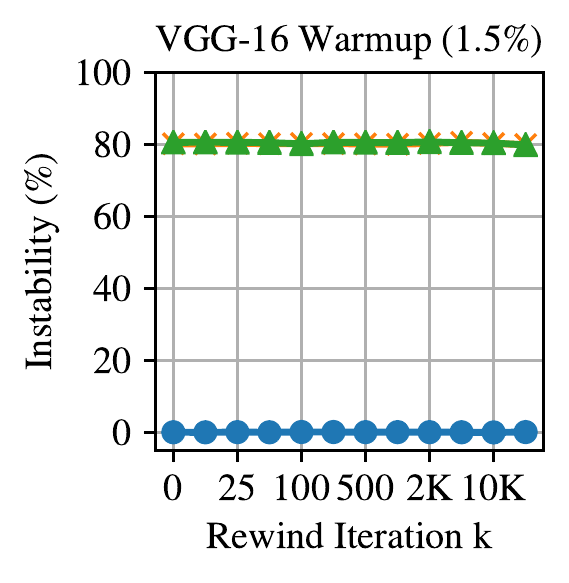}};
\node at (0.6,  -0.17) {\includegraphics[width=0.18\textwidth]{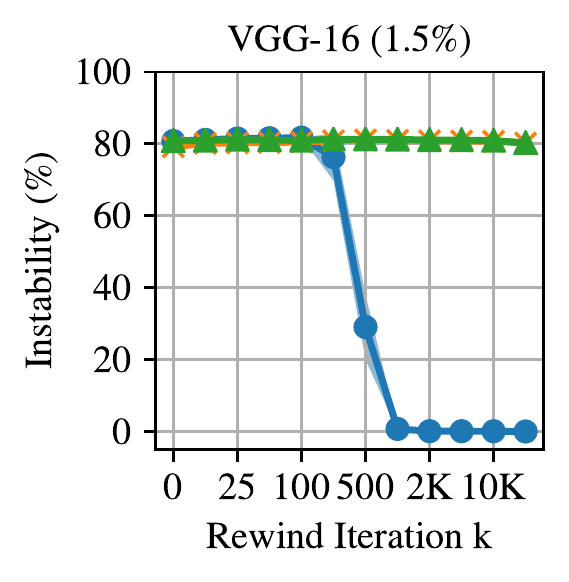}};
\node at (0.8,  -0.17) {\includegraphics[width=0.18\textwidth]{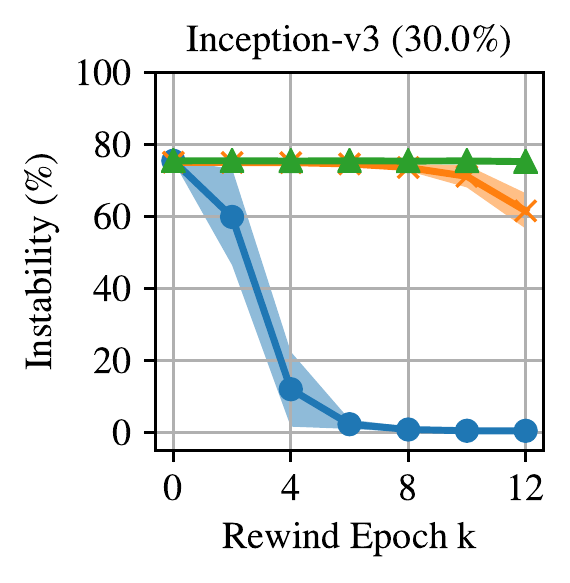}};

\node[anchor=center] at (0.3, -0.365) {\small IMP Subnetwork is Matching for $k=0$};
\node[anchor=center] at (0.8, -0.365) {\small IMP Subnetwork is Not Matching for $k=0$};
\draw[line width=0.3mm] (0.6, -0.01) -- (0.6, -0.375);
\end{tikzpicture}
\vspace{-1em}
\caption{Linear interpolation instability of subnetworks created using the state of the full network at step $k$ and applying a pruning mask.
Lines are means and standard deviations over three initializations and three data orders (nine in total). Percents are weights remaining.}
\label{fig:sparse-instability-later}
\vspace{-3mm}
\end{figure*}
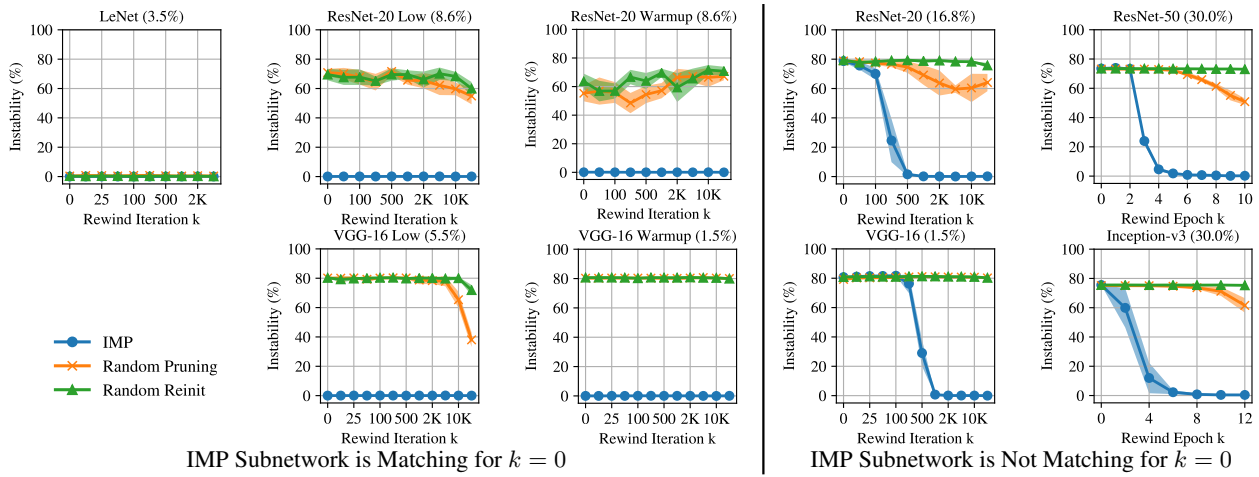

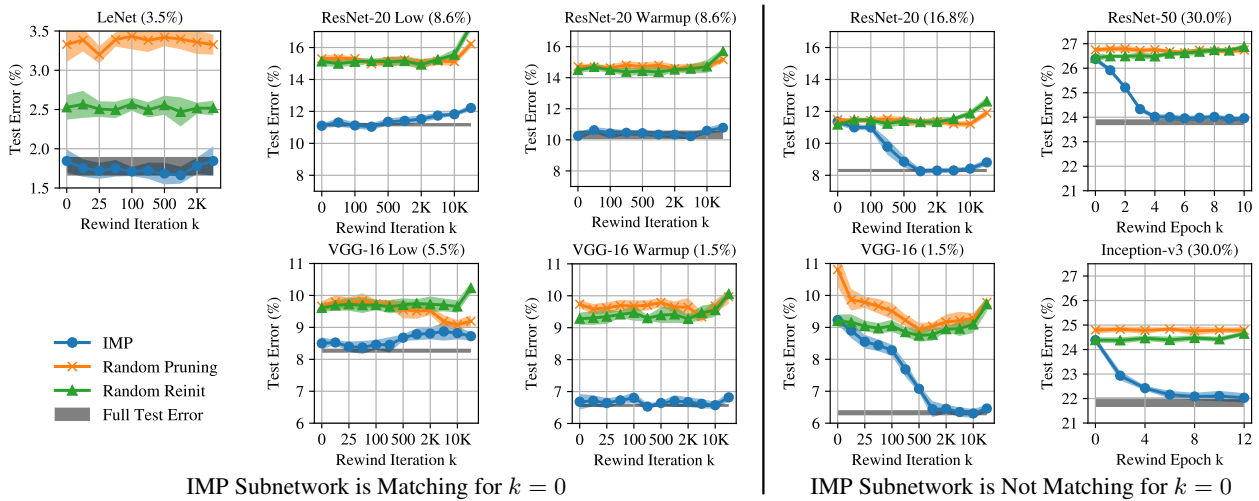
\begin{figure*}
\vspace{-2mm}
\begin{tikzpicture}[x=\textwidth,y=\textwidth, every node/.style = {anchor=north west}]
\node at (0.0, 0) {\includegraphics[width=0.18\textwidth]{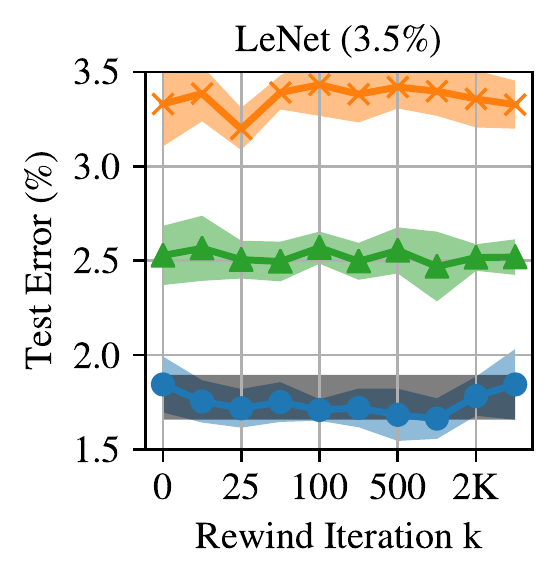}};
\node at (0.2, 0) {\includegraphics[width=0.18\textwidth]{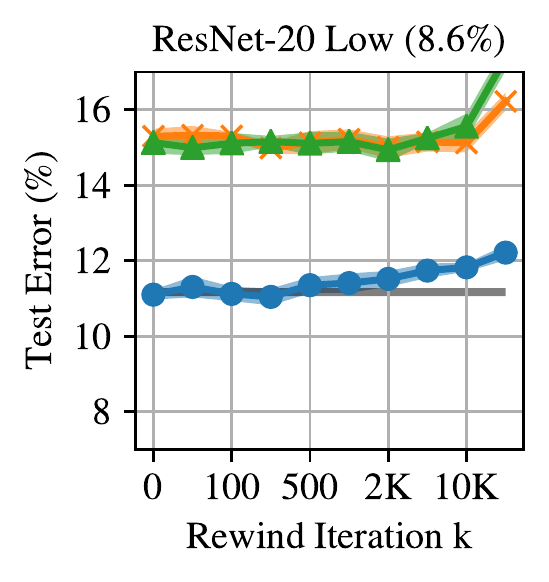}};
\node at (0.4, 0) {\includegraphics[width=0.18\textwidth]{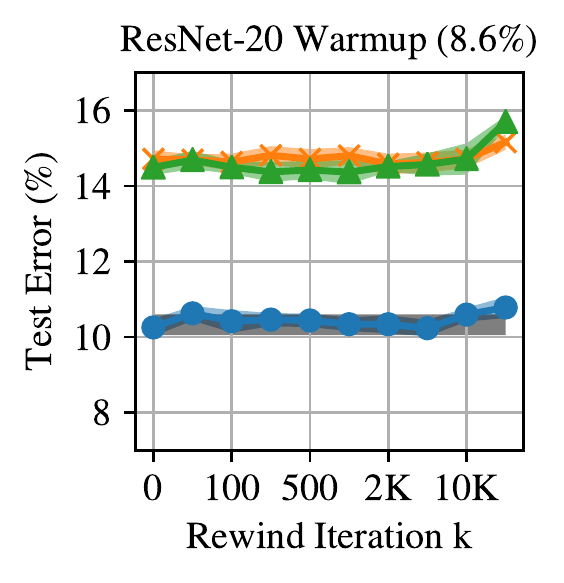}};
\node at (0.6, 0) {\includegraphics[width=0.18\textwidth]{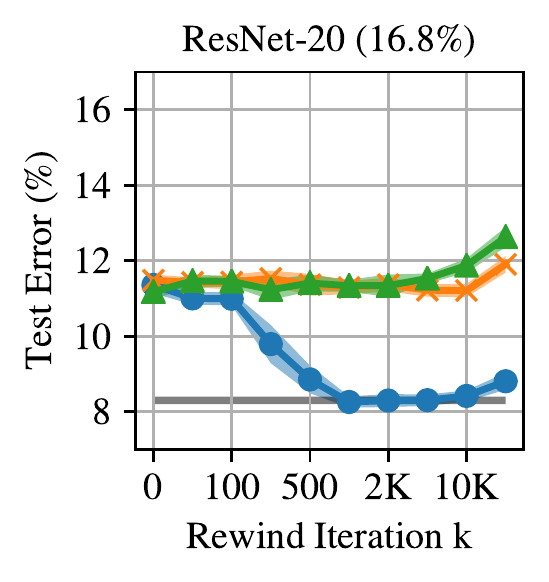}};
\node at (0.8, 0) {\includegraphics[width=0.18\textwidth]{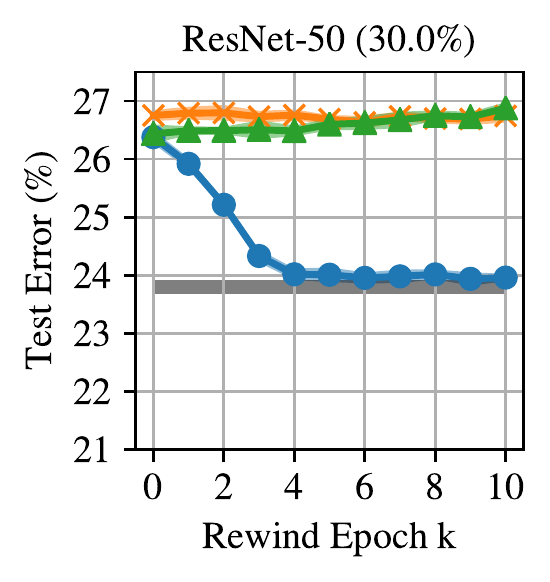}};

\node at (0.03, -0.24) {\includegraphics[width=0.16\textwidth]{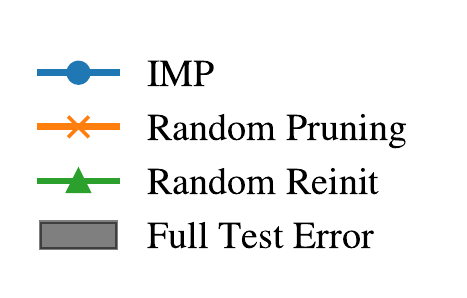}};
\node at (0.2,  -0.18) {\includegraphics[width=0.18\textwidth]{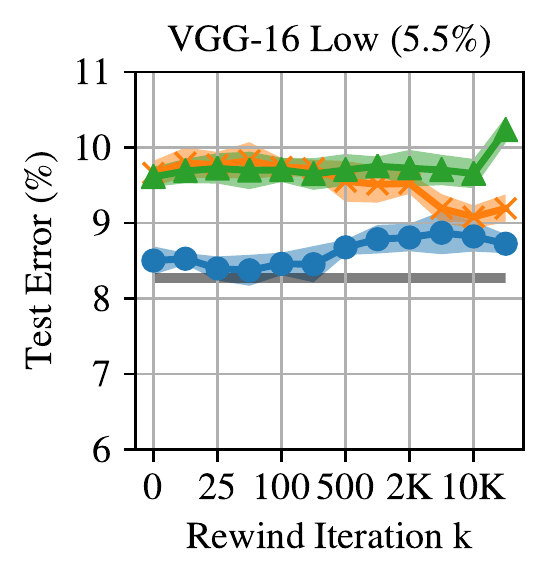}};
\node at (0.4,  -0.18) {\includegraphics[width=0.18\textwidth]{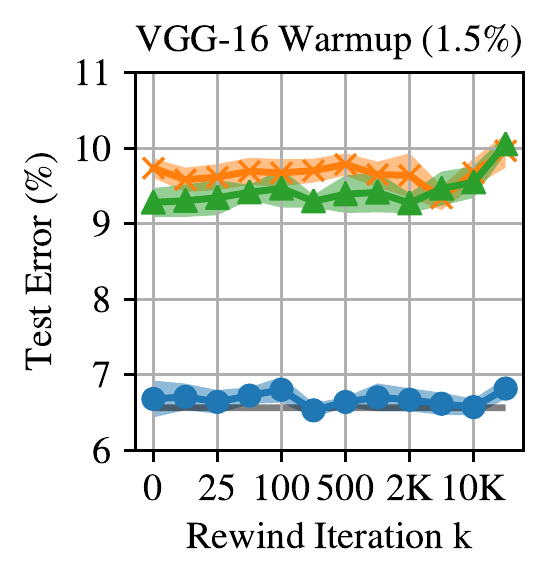}};
\node at (0.6,  -0.18) {\includegraphics[width=0.18\textwidth]{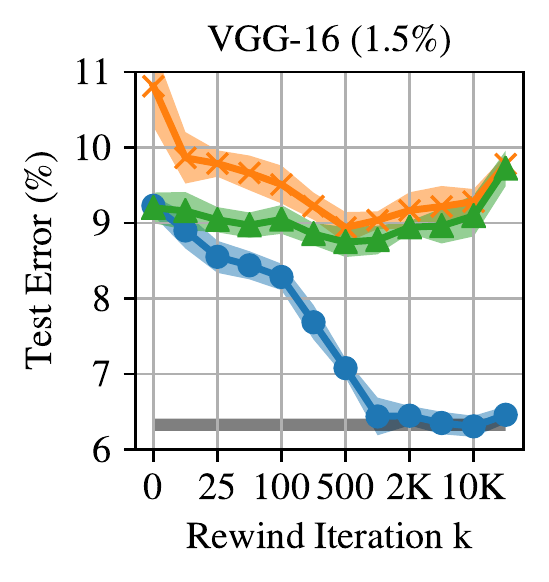}};
\node at (0.8,  -0.18) {\includegraphics[width=0.18\textwidth]{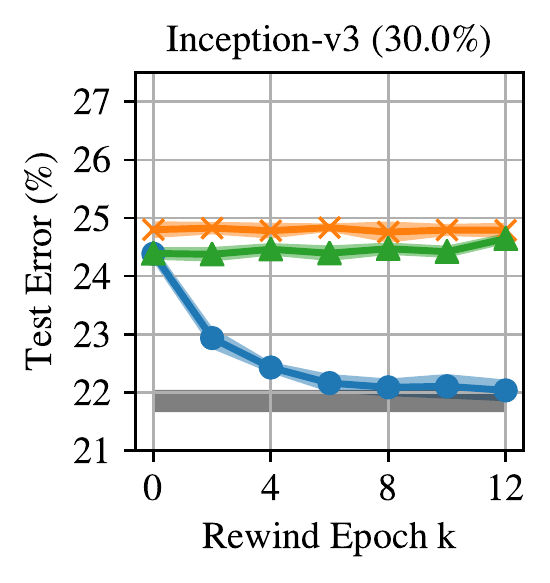}};

\node[anchor=center] at (0.3, -0.385) {\small IMP Subnetwork is Matching for $k=0$};
\node[anchor=center] at (0.8, -0.385) {\small IMP Subnetwork is Not Matching for $k=0$};
\draw[line width=0.3mm] (0.6, -0.01) -- (0.6, -0.39);
\end{tikzpicture}
\vspace{-1em}
\caption{Test error of subnetworks created using the state of the full network at step $k$ and applying a pruning mask.
Lines are means and standard deviations over three initializations and three data orders (nine in total). Percents are weights remaining.}
\label{fig:sparse-error-later}
\end{figure*}

\vspace{-0.75mm}
\subsection{Methodology}
\vspace{-0.75mm}

\textbf{Iterative magnitude pruning.}
Iterative magnitude pruning (IMP) is a procedure to retroactively find a subnetwork of the state of the full network at step $k$ of training.
To do so, IMP trains a network to completion, prunes weights with the lowest magnitudes globally, and \emph{rewinds} the remaining weights back to their values at iteration $k$ (Algorithm \ref{alg:imp}).
The result is a subnetwork $(W_k, m)$ where $W_k \in \mathbb{R}^d$ is the state of the full network at step $k$ and $m \in \{0, 1\}^d$ is a mask such that $m \odot W_k$ (where $\odot$ is the element-wise product) is a pruned network.
We can run IMP iteratively (pruning 20\% of weights \citep{lth}, rewinding, and repeating until a target sparsity) or in one shot (pruning to a target sparsity at once).
We one-shot prune ImageNet networks for efficiency and iteratively prune otherwise (Table \ref{fig:small-networks}).

\citet{lth} focus on finding sparse subnetworks at initialization; as such, they only use IMP to ``reset'' unpruned weights to their values at initialization.
One of our contributions is to generalize IMP to \emph{rewind} to any step $k$.
\citeauthor{lth} refer to subnetworks that match the accuracy of the full network as \emph{winning tickets} because they have ``won the initialization lottery'' with weights that make attaining this accuracy possible.
When we rewind to iteration $k > 0$, subnetworks are no longer randomly initialized, so the term \emph{winning ticket} is no longer appropriate.
Instead, we refer to such subnetworks simply as \emph{matching}.

\textbf{Sparsity levels.}
In this section, we focus on the most extreme sparsity levels for which IMP returns a matching subnetwork at any rewinding step $k$.
These levels are in Table \ref{fig:small-networks}, and Appendix \ref{app:sparsity} explains these choices.
These sparsities provide the best contrast between sparse networks that are matching and (1) the full, overparameterized networks and (2) other classes of sparse networks.
Appendix \ref{app:stability-over-sparsity} includes the analyses from this section for all sparsities for ResNet-20 and VGG-16, which we summarize in Section \ref{sec:stability-across-sparsities}; due to the computational costs of these experiments, we only collected data across all sparsities for these networks.

\subsection{Experiments and Results}
\label{sec:sparsity-exps}

\begin{figure*}
\begin{tikzpicture}[x=\textwidth,y=\textwidth, every node/.style = {anchor=north west}]
\node at (0.0, 0) {\includegraphics[width=0.18\textwidth]{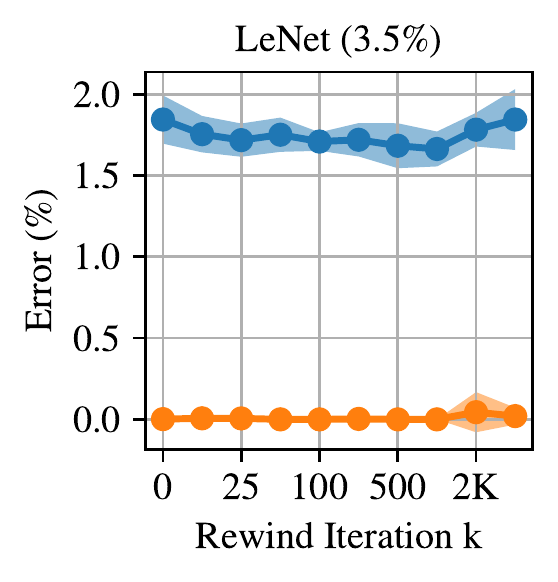}};
\node at (0.2, 0) {\includegraphics[width=0.18\textwidth]{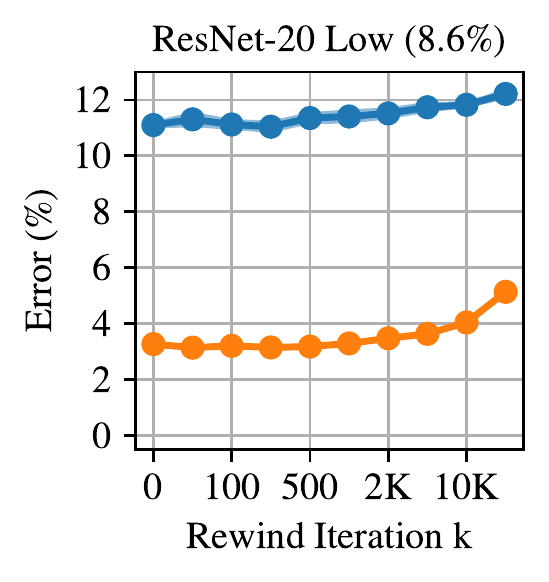}};
\node at (0.4, 0) {\includegraphics[width=0.18\textwidth]{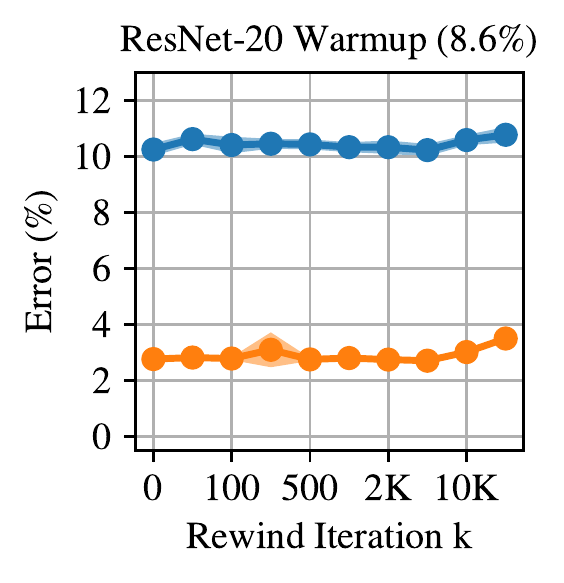}};
\node at (0.6, 0) {\includegraphics[width=0.18\textwidth]{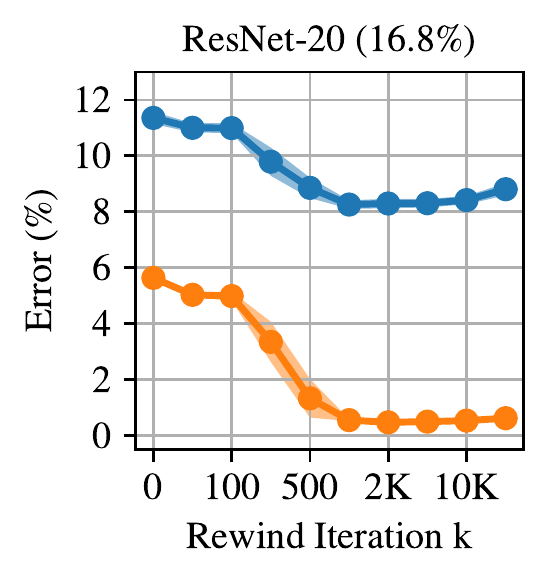}};
\node at (0.8, 0) {\includegraphics[width=0.18\textwidth]{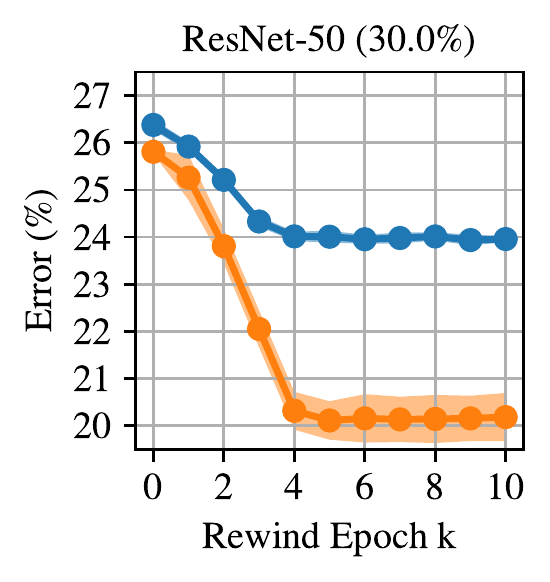}};

\node at (0.0,  -0.24) {\includegraphics[width=0.14\textwidth]{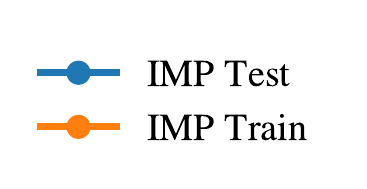}};
\node at (0.2,  -0.18) {\includegraphics[width=0.18\textwidth]{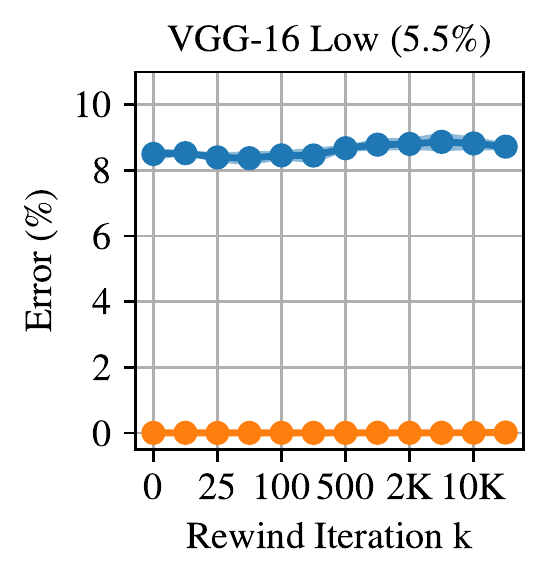}};
\node at (0.4,  -0.18) {\includegraphics[width=0.18\textwidth]{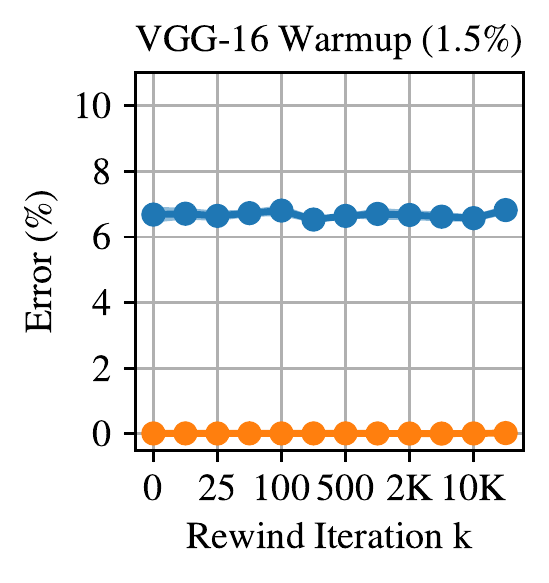}};
\node at (0.6,  -0.18) {\includegraphics[width=0.18\textwidth]{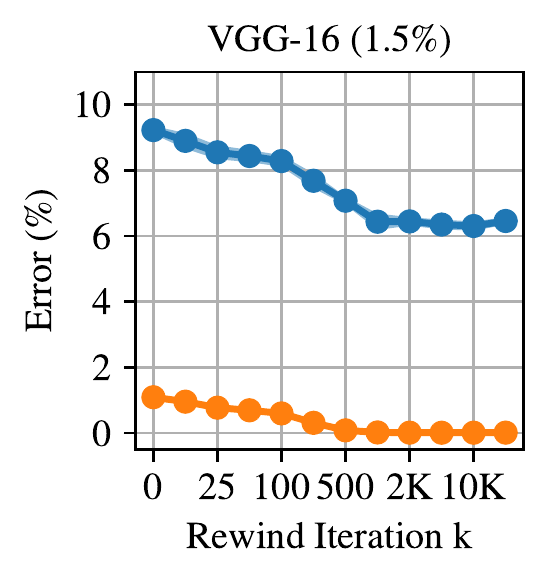}};
\node at (0.8,  -0.18) {\includegraphics[width=0.18\textwidth]{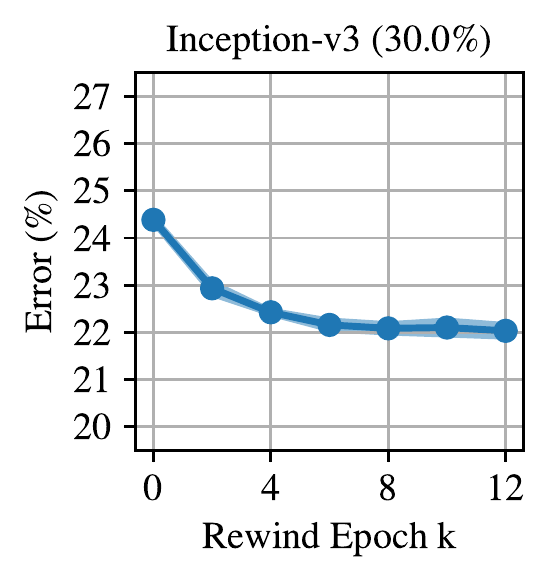}};

\node[anchor=center] at (0.3, -0.385) {\small IMP Subnetwork is Matching for $k=0$};
\node[anchor=center] at (0.8, -0.385) {\small IMP Subnetwork is Not Matching for $k=0$};
\draw[line width=0.3mm] (0.6, -0.01) -- (0.6, -0.395);
\end{tikzpicture}
\vspace{-1em}
\caption{Train error of subnetworks created using the state of the full network at step $k$ and apply a pruning mask.
Lines are means and standard deviations over three initializations and three data orders (nine in total). Percents are weights remaining.
We did not compute the train set quantities for Inception-v3 due to computational limitations.}
\label{fig:train-sparse-instability-later-error}
\vspace{-3mm}
\end{figure*}

\textbf{Recapping the lottery ticket hypothesis.}
We begin by studying sparse subnetworks rewound to initialization ($k = 0$).
This is the lottery ticket experiment from \citet{lth}.
As Table \ref{tab:lth-at-zero} shows, when rewinding to step 0, IMP subnetworks of LeNet are matching, as are variants of ResNet-20 and VGG-16 with lower learning rates or learning rate warmup (changes proposed by \citeauthor{lth} to make it possible for IMP to find matching subnetworks).
However, IMP subnetworks of standard ResNet-20, standard VGG-16, ResNet-50, and Inception-v3 are not matching.
In fact, they are no more accurate than subnetworks generated by randomly pruning or reinitializing the IMP subnetworks, suggesting that neither the structure nor the initialization uncovered by IMP provides a performance advantage.
For full details on the accuracy of these subnetworks at all levels of sparsity, see Appendix \ref{app:sparsity}.

\textbf{Instability analysis of subnetworks at initialization.}
When we perform instability analysis on these subnetworks, we find that they are only matching when they are stable to SGD noise (Figure \ref{fig:imp-instability-at-zero}).
The IMP subnetworks of LeNet, ResNet-20 (low, warmup), and VGG-16 (low, warmup) are stable and matching (Figure \ref{fig:imp-instability-at-zero}, left).
In all other cases, IMP subnetworks are neither stable nor matching (Figure \ref{fig:imp-instability-at-zero}, left).
The low and warmup results are notable because \citeauthor{lth} selected these hyperparameters specifically for IMP to find matching subnetworks; that this change also makes the subnetworks stable adds further evidence of a connection between instability and accuracy in IMP subnetworks.

No randomly pruned or reinitialized subnetworks are stable or matching at these sparsities except those of LeNet: LeNet subnetworks are not matching but error only rises slightly when interpolating.
For all other networks, error approaches that of random guessing when interpolating.

\textbf{Instability analysis of subnetworks during training.}
We just saw that IMP subnetworks are matching from initialization only when they are stable.
In Section \ref{sec:full-networks}, we found that unpruned networks become stable only after a certain amount of training.
Here, we combine these observations: we study whether IMP subnetworks become stable later in training and, if so, whether improved accuracy follows.

Concretely, we perform IMP where we rewind to iteration $k > 0$ after pruning.
Doing so produces a subnetwork ($W_k$, $m$) of the state of the full network at iteration $k$.
We then run instability analysis using this subnetwork.
Another way of looking at this experiment is that it simulates training the full network to iteration $k$, generating a pruning mask, and evaluating the instability of the resulting subnetwork; the underlying mask-generation procedure involves training the network many times in the course of performing IMP.

The blue dots in Figure \ref{fig:sparse-instability-later} show the instability of the IMP subnetworks at many rewinding iterations.
Networks whose IMP subnetworks were stable when rewinding to iteration 0 remain stable at all other rewinding points (Figure \ref{fig:sparse-instability-later}, left).
Notably, networks whose IMP subnetworks were \emph{unstable} when rewinding to iteration 0 become stable when rewinding later.
IMP subnetworks of ResNet-20 and VGG-16 become stable at iterations 500 (0.8\% into training) and 1000 (1.6\%).
Likewise, IMP subnetworks of ResNet-50 and Inception-v3 become stable at epochs 5 (5.5\% into training) and 6 (3.5\%).
In all cases, the IMP subnetworks become stable sooner than the unpruned networks, substantially so for ResNet-50 (epoch 5 vs. 18) and Inception-v3 (epoch 6 vs. 28).

\begin{figure*}
\centering
\begin{tikzpicture}[x=\textwidth,y=\textwidth, every node/.style = {anchor=north west}]
\node[anchor=center] at (0.5, -0.245) {\includegraphics[width=1.2\columnwidth]{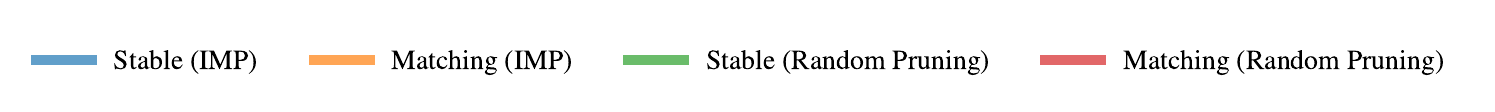}};
\node at (0, 0) {\includegraphics[width=\columnwidth]{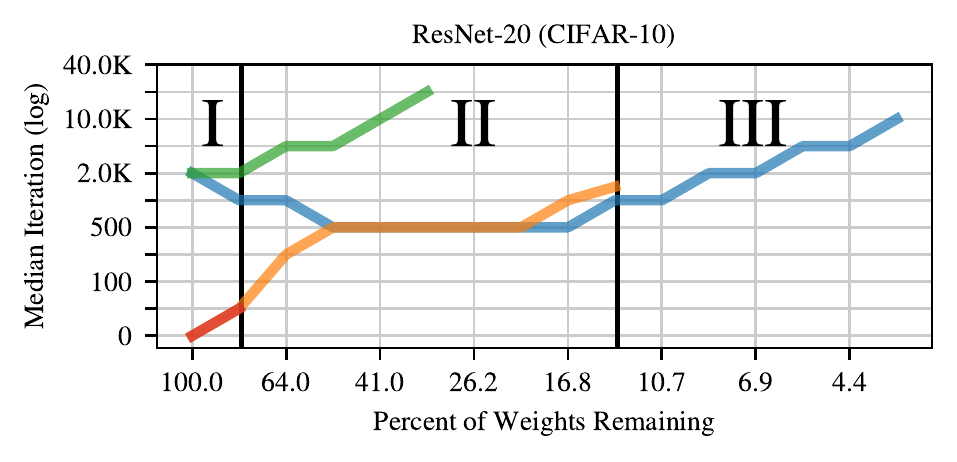}};
\node at (0.5, 0) {\includegraphics[width=\columnwidth]{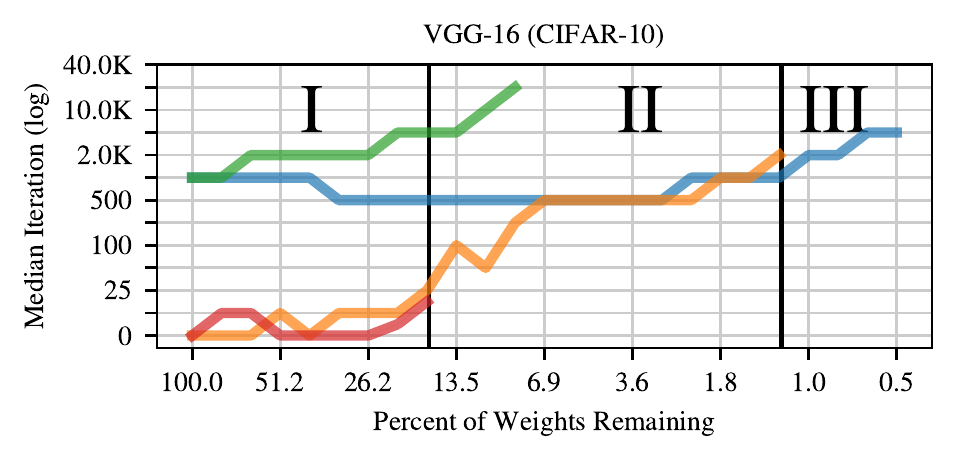}};
\end{tikzpicture}
\vspace{-5mm}
\caption{The median rewinding iteration at which IMP subnetworks and randomly pruned subnetworks of ResNet-20 and VGG-16 become stable and matching.
A subnetwork is stable if instability $< 2\%$. A subnetwork is matching if the accuracy drop $< 0.2\%$; we only include points where a majority of subnetworks are matching at any rewinding iteration. Each line is the median across three initializations and three data orders (nine samples in total).}
\label{fig:sparsity-stability-summary}
\end{figure*}

The test error of the IMP subnetworks behaves similarly.
The blue line in Figure \ref{fig:sparse-error-later} plots the error of the IMP subnetworks and the gray line plots the error of the full networks to one standard deviation; subnetworks are matching when the lines cross.
Networks whose IMP subnetworks were matching when rewinding to step 0 (Figure \ref{fig:sparse-error-later}, left) generally remain matching at later iterations (except for ResNet-20 low and VGG-16 low at the latest rewinding points).
Notably, networks whose IMP subnetworks were \emph{not} matching when rewinding to iteration 0 (Figure \ref{fig:sparse-error-later}, right) become matching when rewinding later.
Moreover, these rewinding points closely coincide with those where the subnetworks become stable.
In summary, at these extreme sparsities, IMP subnetworks are matching when they are stable.

\textbf{Other observations.}
Interestingly, the same pattern holds for the train error: for those networks whose IMP subnetworks were not matching at step 0, train error decreases when rewinding later.
For ResNet-20 and VGG-16, rewinding makes it possible for the IMP subnetworks to converge to 0\% train error.
These results suggest stable IMP subnetworks also optimize better.

Randomly pruned and reinitialized subnetworks are unstable and non-matching at all rewinding points (with LeNet again an exception).
Although it is beyond the scope of our study, this behavior suggests a potential broader link between subnetwork stability and accuracy: IMP subnetworks are matching and become stable at least as early as the full networks, while other subnetworks are less accurate and unstable for the sparsities and rewinding points we consider.

\subsection{Results at Other Sparsity Levels}
\label{sec:stability-across-sparsities}

Thus far, we have performed instability analysis at only two sparsities: unpruned networks (Section \ref{sec:full-networks}) and an extreme sparsity (Section \ref{sec:sparsity-exps}).
Here, we examine sparsities between these levels and beyond for ResNet-20 and VGG-16.
Figure \ref{fig:sparsity-stability-summary} presents the median iteration at which IMP and randomly pruned subnetworks become stable (instability $< 2\%$) and matching (accuracy drop $< 0.2\%$, allowing a small margin for noise) across sparsity levels.%
\footnote{In Appendix \ref{app:stability-over-sparsity}, we present the full instability and error data that we used to produce this summary.}

\textbf{Stability behavior.}
As sparsity increases, the iteration at which the IMP subnetworks become stable decreases, plateaus, and eventually increases.
In contrast, the iteration at which randomly pruned subnetworks become stable only increases until the subnetworks are no longer stable at any rewinding iteration.

\textbf{Matching behavior.}
We separate the sparsities into three \emph{ranges} where different sparse networks are matching.

In sparsity range I, the networks are so overparameterized that even randomly pruned subnetworks are matching (red).
These are sparsities we refer to as \emph{trivial}.
This range occurs when more than 80.0\% and 16.8\% of weights remain for ResNet-20 and VGG-16.

In sparsity range II, the networks are sufficiently sparse that only IMP subnetworks are matching (orange).
This range occurs when 80.0\%-13.4\% and 16.8\%-1.2\% of weights remain in ResNet-20 and VGG-16.
For part of this range, IMP subnetworks become matching and stable at approximately the same rewinding iteration; namely, when 51.2\%-13.4\% and 6.9\%-1.5\% of weights remain for ResNet-20 and VGG-16.
In Section \ref{sec:sparsity-exps}, we observed this behavior for a single, extreme sparsity level for each network.
Based on Figure \ref{fig:sparsity-stability-summary}, we conclude that there are many sparsities where these rewinding iterations coincide for ResNet-20 and VGG-16.

In sparsity range III, the networks are so sparse that even IMP subnetworks are not matching at any rewinding iteration we consider.
This range occurs when fewer than 13.4\% and 1.2\% of weights remain for ResNet-20 and VGG-16.
According to Appendix \ref{app:stability-over-sparsity}, the error of IMP subnetworks still decreases when they become stable (although not to the point that they are matching), potentially suggesting a broader relationship between instability and accuracy.

\section{Discussion}

\textbf{Instability analysis.}
We introduce instability analysis as a novel way to study the sensitivity of a neural network's optimization trajectory to SGD noise.
In doing so, we uncover a class of situations in which \emph{linear} mode connectivity emerges, whereas previous examples of mode connectivity (e.g., between networks trained from different initializations) at similar scales required nonlinear paths \citep{draxler2018essentially, garipov2018loss}.

Our full network results divide training into two phases: an unstable phase where the network finds linearly unconnected minima due to SGD noise and a stable phase where the linearly connected minimum is determined.
Our finding that stability emerges early in training adds to work suggesting that training comprises a noisy first phase and a less stochastic second phase.
For example, the Hessian eigenspectrum settles into a few large values and a bulk \citep{gur2018gradient}, and
large-batch training at high learning rates benefits from learning rate warmup \citep{goyal2017accurate}.

One way to exploit our findings is to explore changing aspects of optimization (e.g., learning rate schedule or optimizer) similar to \citet{goyal2017accurate} once the network becomes stable to improve performance; instability analysis can evaluate the consequences of doing so.
We also believe instability analysis provides a scientific tool for topics related to the scale and distribution of SGD noise, e.g.,
the relationship between batch size, learning rate, and generalization \citep{lecun2012efficient,keskar2016large,goyal2017accurate,smith2018a,smith2018dont} and the efficacy of alternative learning rate schedules \citep{smith2017cyclical,smith2018superconvergence, li2019exponential}.

\textbf{The lottery ticket hypothesis.}
The lottery ticket hypothesis \citep{lth} conjectures that any ``randomly initialized, dense neural network contains a subnetwork that---when trained in isolation---matches the accuracy of the original network.''
This work is among several recent papers to propose that merely sparsifying at initialization can produce high performance neural networks \citep{mallya2018piggyback, zhou2019deconstructing,ramanujan2020s,evci2020rigging}.
\citeauthor{lth} support the lottery ticket hypothesis by using IMP to find matching subnetworks at initialization in small vision networks.
However, follow-up studies show \citep{rethinking-pruning, gale} and we confirm that IMP does not find matching subnetworks at nontrivial sparsities in more challenging settings.
We use instability analysis to distinguish the successes and failures of IMP as identified in previous work.
In doing so, we make a new connection between the lottery ticket hypothesis and the optimization dynamics of neural networks.

\textbf{Practical impact of rewinding.}
By extending IMP with rewinding, we show how to find matching subnetworks in much larger settings than in previous work, albeit from \emph{early} in training rather than initialization.
Our technique has already been adopted for practical purposes.
\citet{morcos2019one} show that subnetworks found by IMP with rewinding transfer between vision tasks, meaning the effort of finding a subnetworks can be amortized by reusing it many times.
\citet{renda2020comparing} show that IMP with rewinding prunes to state-of-the-art sparsities, matching or exceeding the performance of standard techniques that fine-tune at a low learning rate after pruning \citep[e.g.,][]{han-pruning, he2018amc}.
Other efforts use rewinding to further study lottery tickets \citep{yu2020playing, frankle2020early, caron2020finding, savarese2020winning, yin2020the}.

\textbf{Pruning.}
In larger-scale settings, IMP subnetworks only become stable and matching after the full network has been trained for some number of steps.
Recent proposals attempt to prune networks at initialization \citep{snip, wang2020picking}, but our results suggest that the best time to do so may be after some training.
Likewise, most pruning methods only begin to sparsify networks late in training or after training \citep{han-pruning, gale, he2018amc}.
The existence of matching subnetworks early in training suggests that there is an unexploited opportunity to prune networks much earlier than current methods.


\newpage
\section{Conclusions}

We propose instability analysis as a way to shed light on how SGD noise affects the outcome of optimizing neural networks.
We find that standard networks for MNIST, CIFAR-10, and ImageNet become \emph{stable} to SGD noise early in training, after which the outcome of optimization is determined to a linearly connected minimum.

We then apply instability analysis to better understand a key question at the center of the lottery ticket hypothesis: why does iterative magnitude pruning find sparse networks that can train from initialization to full accuracy in smaller-scale settings (e.g., MNIST) but not on more challenging tasks (e.g., ImageNet)?
We find that extremely sparse IMP subnetworks only train to full accuracy when they are stable to SGD noise, which occurs at initialization in some settings but only after some amount of training in others.

Instability analysis and our linear mode connectivity criterion contribute to a growing range of empirical tools for studying and understanding the behavior of neural networks in practice.
In this paper, we show that it has already yielded new insights into neural network training dynamics and lottery ticket phenomena.

\section*{Acknowledgements}

We gratefully acknowledge the support of IBM, which provided us with the GPU resources necessary to conduct experiments on CIFAR-10 through the MIT-IBM Watson AI Lab. In particular, we express our gratitude to David Cox and John Cohn.

We gratefully acknowledge the support of Google, which provided us with the TPU resources necessary to conduct experiments on ImageNet through the TensorFlow Research Cloud. In particular, we express our gratitude to Zak Stone.

This work was supported in part by cloud credits from the MIT Quest for Intelligence.

This work was supported in part by the Office of Naval Research (ONR N00014-17-1-2699).

This work was supported in part by DARPA Award \#HR001118C0059.

Daniel M. Roy was supported, in part, by an NSERC Discovery Grant and Ontario Early Researcher Award.

This research was carried out in part while Gintare Karolina Dziugaite and Daniel M. Roy participated in the Foundations of Deep Learning program at the Simons Institute for the Theory of Computing.

\newpage

\bibliography{local}

\begin{thebibliography}{35}
\providecommand{\natexlab}[1]{#1}
\providecommand{\url}[1]{\texttt{#1}}
\expandafter\ifx\csname urlstyle\endcsname\relax
  \providecommand{\doi}[1]{doi: #1}\else
  \providecommand{\doi}{doi: \begingroup \urlstyle{rm}\Url}\fi

\bibitem[Caron et~al.(2020)Caron, Morcos, Bojanowski, Mairal, and
  Joulin]{caron2020finding}
Caron, M., Morcos, A., Bojanowski, P., Mairal, J., and Joulin, A.
\newblock Finding winning tickets with limited (or no) supervision, 2020.

\bibitem[Draxler et~al.(2018)Draxler, Veschgini, Salmhofer, and
  Hamprecht]{draxler2018essentially}
Draxler, F., Veschgini, K., Salmhofer, M., and Hamprecht, F.~A.
\newblock Essentially no barriers in neural network energy landscape.
\newblock In \emph{International Conference on Machine Learning}, 2018.

\bibitem[Evci et~al.(2020)Evci, Gale, Menick, Castro, and
  Elsen]{evci2020rigging}
Evci, U., Gale, T., Menick, J., Castro, P.~S., and Elsen, E.
\newblock Rigging the lottery: Making all tickets winners, 2020.

\bibitem[Frankle \& Carbin(2019)Frankle and Carbin]{lth}
Frankle, J. and Carbin, M.
\newblock The lottery ticket hypothesis: Finding sparse, trainable neural
  networks.
\newblock In \emph{International Conference on Learning Representations}, 2019.

\bibitem[Frankle et~al.(2020)Frankle, Schwab, and Morcos]{frankle2020early}
Frankle, J., Schwab, D.~J., and Morcos, A.~S.
\newblock The early phase of neural network training.
\newblock In \emph{International Conference on Learning Representations}, 2020.

\bibitem[Freeman \& Bruna(2017)Freeman and Bruna]{freeman2016topology}
Freeman, C.~D. and Bruna, J.
\newblock Topology and geometry of half-rectified network optimization.
\newblock In \emph{International Conference on Learning Representations}, 2017.

\bibitem[Gale et~al.(2019)Gale, Elsen, and Hooker]{gale}
Gale, T., Elsen, E., and Hooker, S.
\newblock The state of sparsity in deep neural networks, 2019.
\newblock arXiv:1902.09574.

\bibitem[Garipov et~al.(2018)Garipov, Izmailov, Podoprikhin, Vetrov, and
  Wilson]{garipov2018loss}
Garipov, T., Izmailov, P., Podoprikhin, D., Vetrov, D.~P., and Wilson, A.~G.
\newblock Loss surfaces, mode connectivity, and fast ensembling of dnns.
\newblock In \emph{Advances in Neural Information Processing Systems}, pp.\
  8789--8798, 2018.

\bibitem[Google(2018)]{tpu-implementation}
Google.
\newblock Networks for {I}magenet on {TPU}s, 2018.
\newblock URL \url{https://github.com/tensorflow/tpu/tree/master/models/}.

\bibitem[Goyal et~al.(2017)Goyal, Doll{\'a}r, Girshick, Noordhuis, Wesolowski,
  Kyrola, Tulloch, Jia, and He]{goyal2017accurate}
Goyal, P., Doll{\'a}r, P., Girshick, R., Noordhuis, P., Wesolowski, L., Kyrola,
  A., Tulloch, A., Jia, Y., and He, K.
\newblock Accurate, large minibatch {SGD}: training {I}magenet in 1 hour, 2017.

\bibitem[Gur-Ari et~al.(2018)Gur-Ari, Roberts, and Dyer]{gur2018gradient}
Gur-Ari, G., Roberts, D.~A., and Dyer, E.
\newblock Gradient descent happens in a tiny subspace.
\newblock \emph{arXiv preprint arXiv:1812.04754}, 2018.

\bibitem[Han et~al.(2015)Han, Pool, Tran, and Dally]{han-pruning}
Han, S., Pool, J., Tran, J., and Dally, W.
\newblock Learning both weights and connections for efficient neural network.
\newblock In \emph{Advances in Neural Information Processing Systems}, pp.\
  1135--1143, 2015.

\bibitem[He et~al.(2016)He, Zhang, Ren, and Sun]{resnet}
He, K., Zhang, X., Ren, S., and Sun, J.
\newblock Deep residual learning for image recognition.
\newblock In \emph{Proceedings of the IEEE Conference on Computer Vision and
  Pattern Recognition}, pp.\  770--778, 2016.

\bibitem[He et~al.(2018)He, Lin, Liu, Wang, Li, and Han]{he2018amc}
He, Y., Lin, J., Liu, Z., Wang, H., Li, L.-J., and Han, S.
\newblock Amc: Automl for model compression and acceleration on mobile devices.
\newblock In \emph{Proceedings of the European Conference on Computer Vision
  (ECCV)}, pp.\  784--800, 2018.

\bibitem[Keskar et~al.(2017)Keskar, Mudigere, Nocedal, Smelyanskiy, and
  Tang]{keskar2016large}
Keskar, N.~S., Mudigere, D., Nocedal, J., Smelyanskiy, M., and Tang, P. T.~P.
\newblock On large-batch training for deep learning: Generalization gap and
  sharp minima.
\newblock In \emph{International Conference on Learning Representations}, 2017.

\bibitem[LeCun et~al.(2012)LeCun, Bottou, Orr, and
  M{\"u}ller]{lecun2012efficient}
LeCun, Y.~A., Bottou, L., Orr, G.~B., and M{\"u}ller, K.-R.
\newblock Efficient backprop.
\newblock In \emph{Neural networks: Tricks of the trade}, pp.\  9--48.
  Springer, 2012.

\bibitem[Lee et~al.(2019)Lee, Ajanthan, and Torr]{snip}
Lee, N., Ajanthan, T., and Torr, P. H.~S.
\newblock {SNIP}: Single-shot network pruning based on connection sensitivity.
\newblock In \emph{International Conference on Learning Representations}, 2019.

\bibitem[Li et~al.(2017)Li, Kadav, Durdanovic, Samet, and
  Graf]{pruning-filters}
Li, H., Kadav, A., Durdanovic, I., Samet, H., and Graf, H.~P.
\newblock Pruning filters for efficient convnets.
\newblock In \emph{International Conference on Learning Representations}, 2017.

\bibitem[Li \& Arora(2020)Li and Arora]{li2019exponential}
Li, Z. and Arora, S.
\newblock An exponential learning rate schedule for deep learning.
\newblock In \emph{International Conference on Learning Representations}, 2020.

\bibitem[Liu et~al.(2019)Liu, Sun, Zhou, Huang, and
  Darrell]{rethinking-pruning}
Liu, Z., Sun, M., Zhou, T., Huang, G., and Darrell, T.
\newblock Rethinking the value of network pruning.
\newblock In \emph{International Conference on Learning Representations}, 2019.

\bibitem[Mallya et~al.(2018)Mallya, Davis, and Lazebnik]{mallya2018piggyback}
Mallya, A., Davis, D., and Lazebnik, S.
\newblock Piggyback: Adapting a single network to multiple tasks by learning to
  mask weights.
\newblock In \emph{Proceedings of the European Conference on Computer Vision
  (ECCV)}, pp.\  67--82, 2018.

\bibitem[Morcos et~al.(2019)Morcos, Yu, Paganini, and Tian]{morcos2019one}
Morcos, A., Yu, H., Paganini, M., and Tian, Y.
\newblock One ticket to win them all: generalizing lottery ticket
  initializations across datasets and optimizers.
\newblock In \emph{Advances in Neural Information Processing Systems}, pp.\
  4932--4942, 2019.

\bibitem[Nagarajan \& Kolter(2019)Nagarajan and Kolter]{kolter}
Nagarajan, V. and Kolter, J.~Z.
\newblock Uniform convergence may be unable to explain generalization in deep
  learning.
\newblock In \emph{Advances in Neural Information Processing Systems}, pp.\
  11615--11626, 2019.

\bibitem[Ramanujan et~al.(2020)Ramanujan, Wortsman, Kembhavi, Farhadi, and
  Rastegari]{ramanujan2020s}
Ramanujan, V., Wortsman, M., Kembhavi, A., Farhadi, A., and Rastegari, M.
\newblock What's hidden in a randomly weighted neural network?
\newblock In \emph{Proceedings of the IEEE/CVF Conference on Computer Vision
  and Pattern Recognition}, pp.\  11893--11902, 2020.

\bibitem[Reed(1993)]{reed1993pruning}
Reed, R.
\newblock Pruning algorithms: A survey.
\newblock \emph{IEEE transactions on Neural Networks}, 4\penalty0 (5):\penalty0
  740--747, 1993.

\bibitem[Renda et~al.(2020)Renda, Frankle, and Carbin]{renda2020comparing}
Renda, A., Frankle, J., and Carbin, M.
\newblock Comparing fine-tuning and rewinding in neural network pruning.
\newblock In \emph{International Conference on Learning Representations}, 2020.

\bibitem[Savarese et~al.(2020)Savarese, Silva, and Maire]{savarese2020winning}
Savarese, P., Silva, H., and Maire, M.
\newblock Winning the lottery with continuous sparsification.
\newblock \emph{arXiv preprint arXiv:1912.04427}, 2020.

\bibitem[Smith(2017)]{smith2017cyclical}
Smith, L.~N.
\newblock Cyclical learning rates for training neural networks.
\newblock In \emph{2017 IEEE Winter Conference on Applications of Computer
  Vision (WACV)}, pp.\  464--472. IEEE, 2017.

\bibitem[Smith \& Topin(2018)Smith and Topin]{smith2018superconvergence}
Smith, L.~N. and Topin, N.
\newblock Super-convergence: Very fast training of residual networks using
  large learning rates.
\newblock In \emph{International Conference on Learning Representations}, 2018.

\bibitem[Smith \& Le(2018)Smith and Le]{smith2018a}
Smith, S.~L. and Le, Q.~V.
\newblock A bayesian perspective on generalization and stochastic gradient
  descent.
\newblock In \emph{International Conference on Learning Representations}, 2018.

\bibitem[Smith et~al.(2018)Smith, Kindermans, and Le]{smith2018dont}
Smith, S.~L., Kindermans, P.-J., and Le, Q.~V.
\newblock Don't decay the learning rate, increase the batch size.
\newblock In \emph{International Conference on Learning Representations}, 2018.

\bibitem[Wang et~al.(2020)Wang, Zhang, and Grosse]{wang2020picking}
Wang, C., Zhang, G., and Grosse, R.
\newblock Picking winning tickets before training by preserving gradient flow.
\newblock In \emph{International Conference on Learning Representations}, 2020.

\bibitem[Yin et~al.(2020)Yin, Kim, Oh, Wang, Serrano, Seo, and
  Choi]{yin2020the}
Yin, S., Kim, K.-H., Oh, J., Wang, N., Serrano, M., Seo, J.-S., and Choi, J.
\newblock The sooner the better: Investigating structure of early winning
  lottery tickets, 2020.

\bibitem[Yu et~al.(2020)Yu, Edunov, Tian, and Morcos]{yu2020playing}
Yu, H., Edunov, S., Tian, Y., and Morcos, A.~S.
\newblock Playing the lottery with rewards and multiple languages: Lottery
  tickets in rl and nlp.
\newblock In \emph{International Conference on Learning Representations}, 2020.

\bibitem[Zhou et~al.(2019)Zhou, Lan, Liu, and Yosinski]{zhou2019deconstructing}
Zhou, H., Lan, J., Liu, R., and Yosinski, J.
\newblock Deconstructing lottery tickets: Zeros, signs, and the supermask.
\newblock In \emph{Advances in Neural Information Processing Systems}, pp.\
  3592--3602, 2019.

\end{thebibliography}
\bibliographystyle{icml2020}

\balance

\begin{appendix}

\newpage

\section*{Overview and Contents}

In this supplementary material, we include data that either (1) we processed to produce the plots in the paper or (2) that we were not able to fit in the main body of the paper. The contents of these appendices are as follows:

\textbf{Appendix \ref{app:sparsity}.} The process by which we chose the ``extreme'' sparsity levels used in Section \ref{sec:pruned-networks}.

\textbf{Appendix \ref{app:state}.} Details about the states of the unpruned networks and IMP subnetworks at the rewinding iterations, including full network accuracy, $L_2$ distance from initialization, $L_2$ distance to the trained weights, and the $L_2$ distance between trained weights under different data orders.

\textbf{Appendix \ref{app:throughout}.} Linear interpolation instability data \emph{throughout} training for ResNet-20 and VGG-16; that is, interpolating between the states at each epoch of networks trained on different dataorders.

\textbf{Appendix \ref{app:stability-over-sparsity}.} Linear interpolation instability and test error across rewinding iterations for ResNet-20 and VGG-16 at all levels of sparsity (not just the extreme sparsity we analyzed in Section \ref{sec:sparsity-exps}). This data was used to create Figure \ref{fig:sparsity-stability-summary}.

\textbf{Appendix \ref{app:hills}.} The error when linearly interpolating for all networks in all configurations (unpruned and sparse) at all rewinding iterations. This data was used to create the instability plots in Figures \ref{fig:full-instability-later} and \ref{fig:sparse-instability-later}.

\textbf{Appendix \ref{app:sparse-train-data}.} The training set instability for the sparse networks corresponding to the test set instability data that we present in Section \ref{sec:pruned-networks} Figure \ref{fig:sparse-instability-later}.

\textbf{Appendix \ref{app:alternate-distance-metrics}.} Functions other than linear mode connectivity for comparing the networks that result from our instability analysis experiments: $L_2$ distance, cosine distance, number of identical classifications, and $L_2$ distance of losses.

\section{Selecting Extreme Sparsity Levels for IMP}
\label{app:sparsity}

In this appendix, we describe how we select the extreme sparsity level that we examine in Section \ref{sec:sparsity-exps} for each IMP subnetwork.
For each network and hyperparameter configuration, our goal is to study the most extreme sparsity level at which matching subnetworks are known to exist early in training.
To do so, we use IMP to generate subnetworks at many different sparsities for many different rewinding iterations.
We then select the most extreme sparsity level at which any IMP under any rewinding iteration produces a matching subnetwork.

\begin{figure*}
\centering
\includegraphics[width=0.33\textwidth]{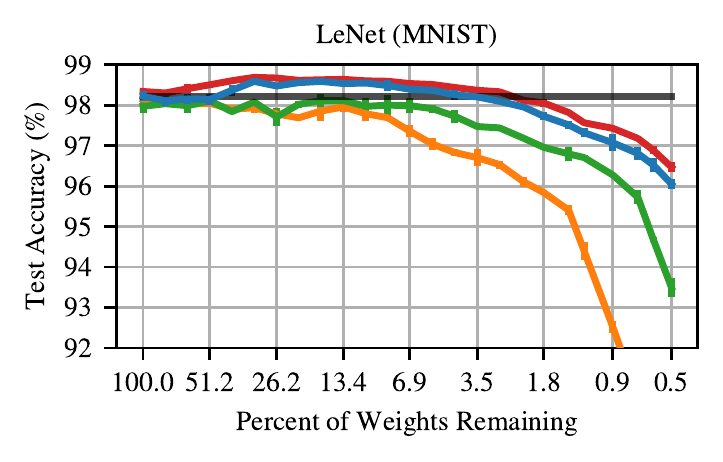}%
\includegraphics[width=0.33\textwidth]{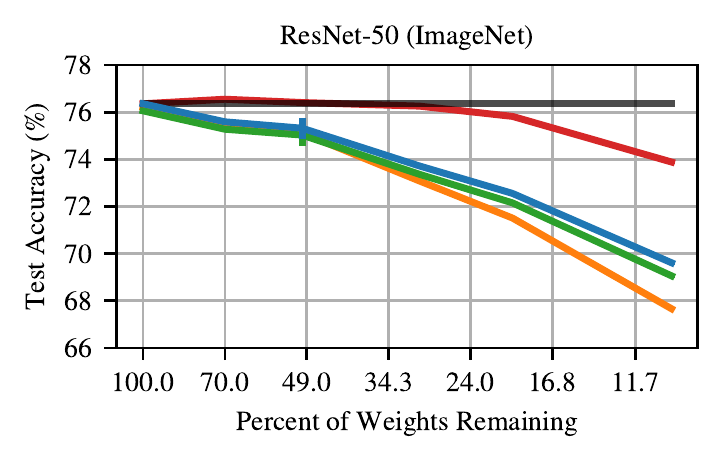}%
\includegraphics[width=0.33\textwidth]{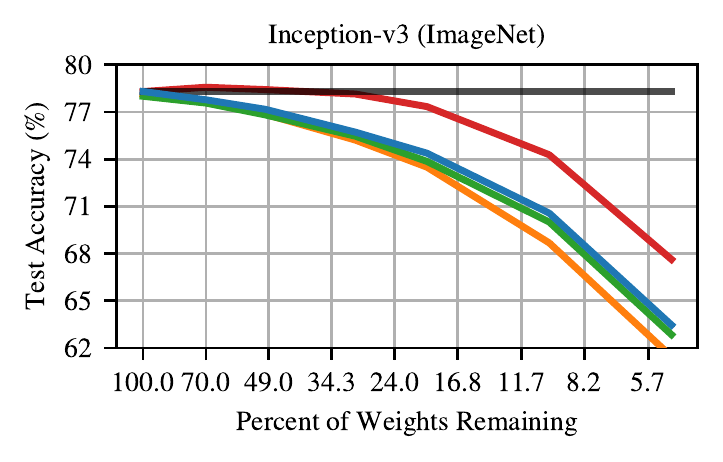}

\includegraphics[width=0.33\textwidth]{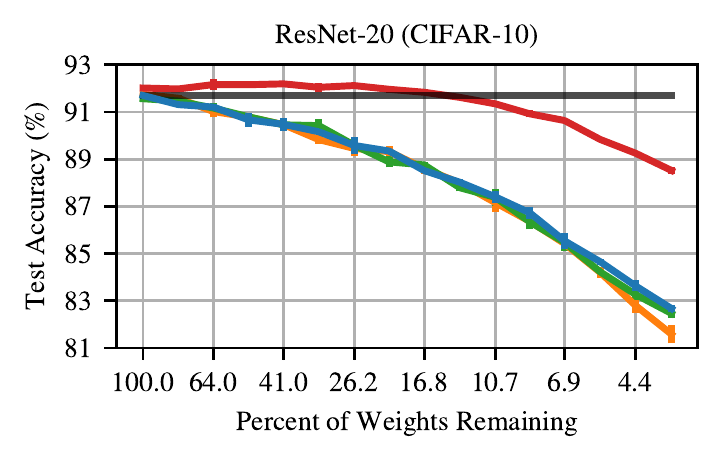}%
\includegraphics[width=0.33\textwidth]{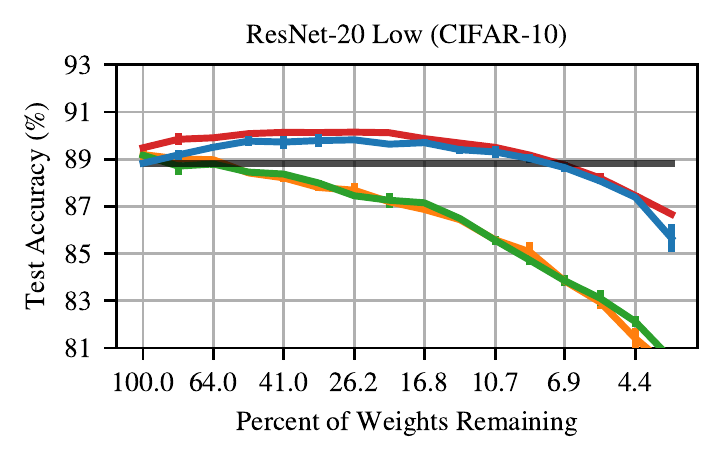}%
\includegraphics[width=0.33\textwidth]{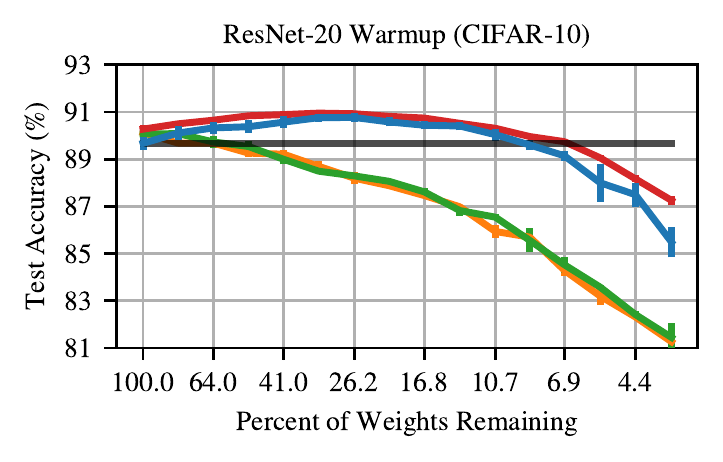}

\includegraphics[width=0.33\textwidth]{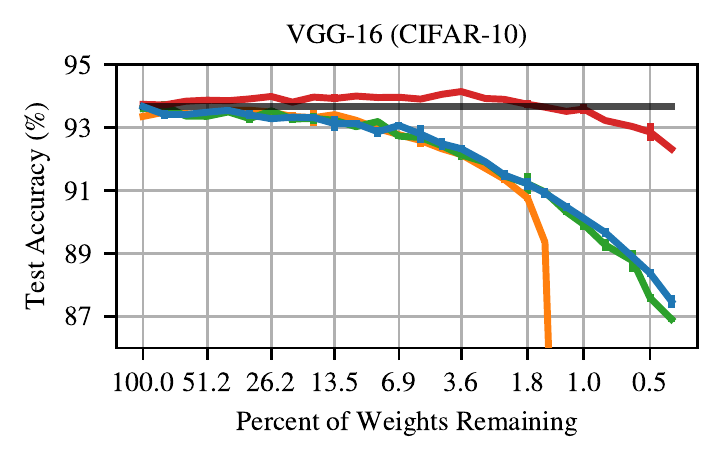}%
\includegraphics[width=0.33\textwidth]{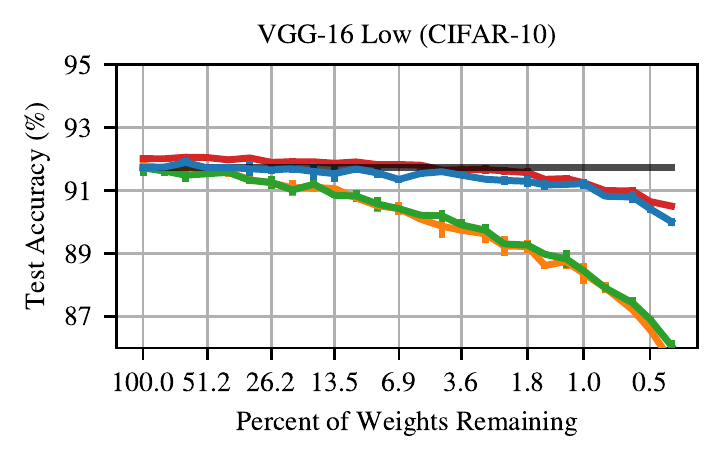}%
\includegraphics[width=0.33\textwidth]{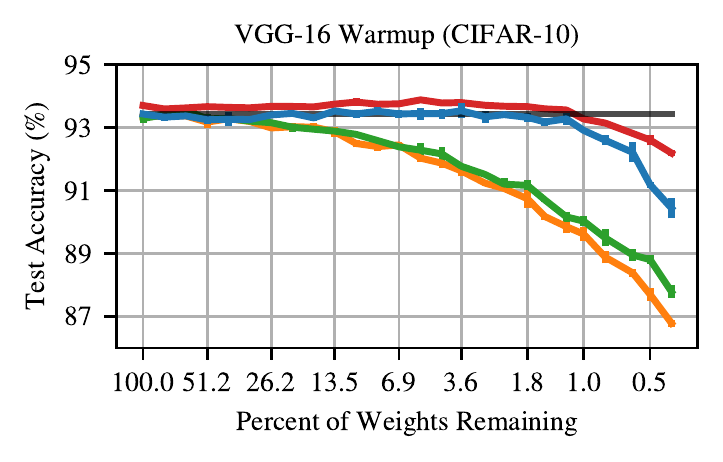}

\includegraphics[width=0.5\textwidth]{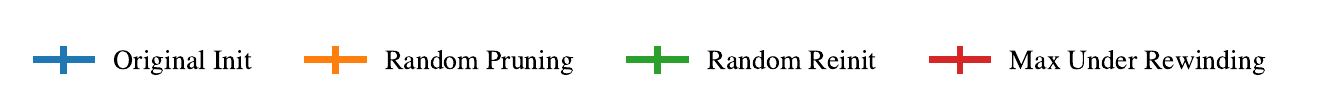}

\caption{An illustration of the methodology by which we select the extreme sparsity levels that we study in Section \ref{sec:pruned-networks}. The red line is the maximum accuracy achieved by any IMP subnetwork under any rewinding iteration. The black line is the accuracy of the full network. We use the most extreme sparsity level for which the red and black lines overlap. Each line is the mean and standard deviation across three runs with different initializations.}
\label{fig:selecting-sparsity}
\end{figure*}

In Figure \ref{fig:selecting-sparsity}, each plot contains the maximum accuracy found by any rewinding iteration in red.
The black line is the accuracy of the unpruned network to one standard deviation. 
For each network, we select the most extreme sparsity for which the red and black lines intersect.
As a basis for comparison, these plots also include the result of performing IMP with $k= 0$ (blue line), random pruning (orange line), and random reinitialization of the IMP subnetworks with $k= 0$ (green line).

Note that, for computational reasons, ResNet-50 and Inception-v3 are pruned using one-shot pruning, meaning the networks are pruned to the target sparsity all at once.
All other networks are pruned using iterative pruning, meaning the networks are pruned by 20\% after each iteration of IMP until they reach the target sparsity.
Pruning 20\% per iteration was the practice adopted by \citet{lth}.
This information is specified in the rightmost Table \ref{fig:small-networks}.

\section{The State of the Network at Rewinding}
\label{app:state}

\subsection{Methodology}

In the main body of the paper, we perform instability analysis by training to step $k$, making two copies of the network, optionally apply a pruning mask (as in Section \ref{sec:pruned-networks}), and training these two copies to completion under different samples of SGD noise.
We find that, for a sufficiently large value of $k$, the trained networks will find the same, linearly connected minimum.
In this appendix, we address the following question: what is the state of the network at the step $k$ from which this linear connectivity results?
Are the networks so far along in training that they are virtually fully optimized?
Have they traveled the vast majority of the distance from initialization to the eventual minimum?
In this sense, is the iteration at which the network becomes stable ``trivial?''
We address these questions in two ways.

\textbf{Error at rewinding.}
In Figure \ref{fig:state-error}, we present the error of the unpruned network at each rewinding iteration we consider in the main body of the paper.
With this data, we investigate how close the network has come to its full accuracy when it becomes stable.

\begin{figure*}
\begin{tikzpicture}[x=\textwidth,y=\textwidth, every node/.style = {anchor=north west}]
\node at (0.0, 0) {\includegraphics[width=0.19\textwidth]{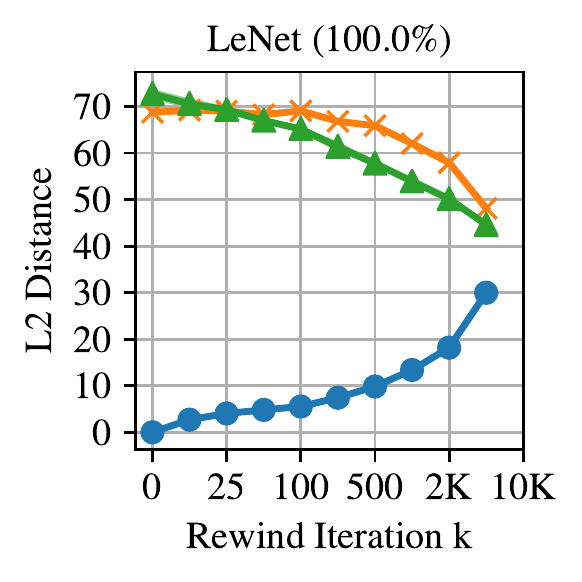}};
\node at (0.2, 0) {\includegraphics[width=0.19\textwidth]{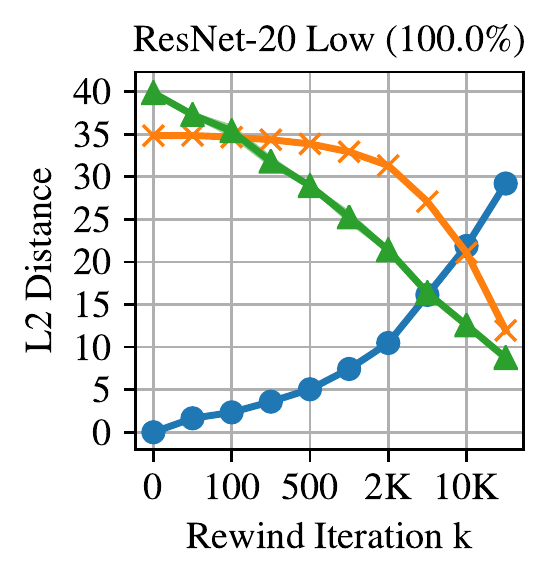}};
\node at (0.4, 0) {\includegraphics[width=0.19\textwidth]{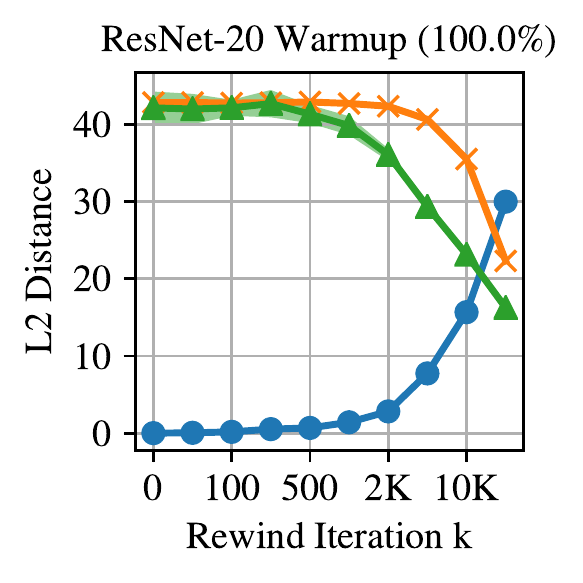}};
\node at (0.6, 0) {\includegraphics[width=0.19\textwidth]{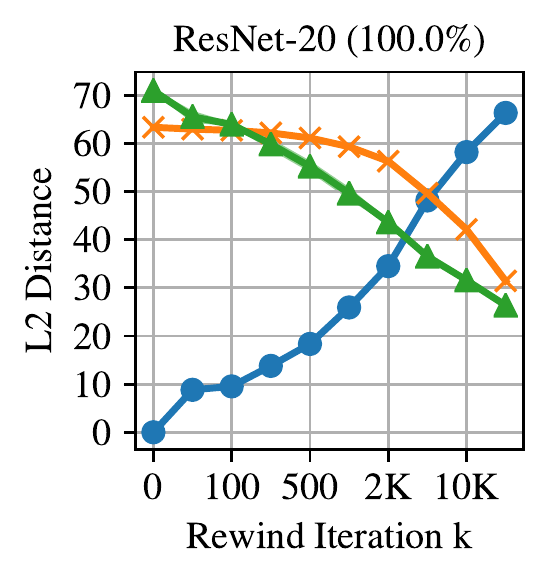}};
\node at (0.8, 0) {\includegraphics[width=0.19\textwidth]{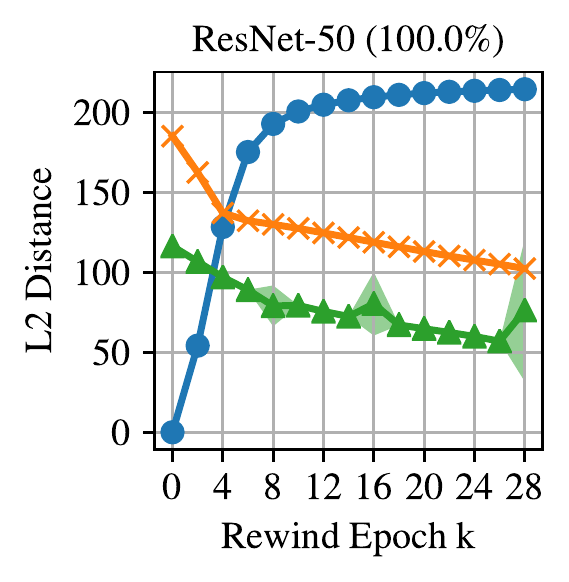}};

\node at (0.2,  -0.19) {\includegraphics[width=0.19\textwidth]{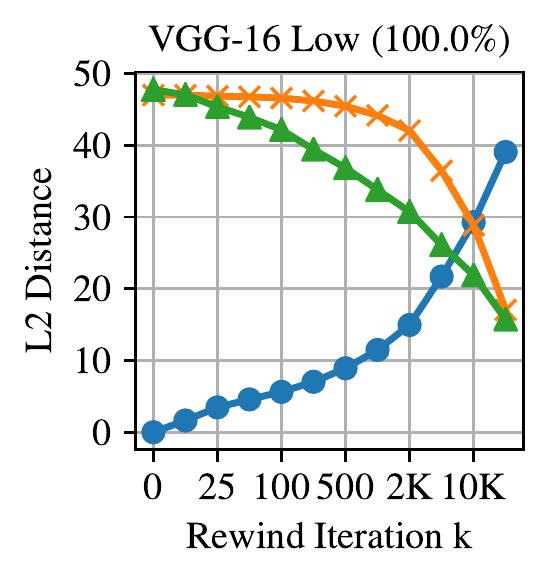}};
\node at (0.4,  -0.19) {\includegraphics[width=0.19\textwidth]{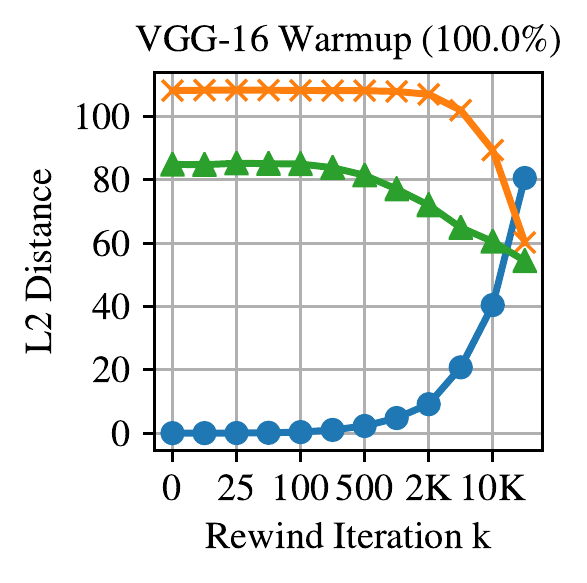}};
\node at (0.6,  -0.19) {\includegraphics[width=0.19\textwidth]{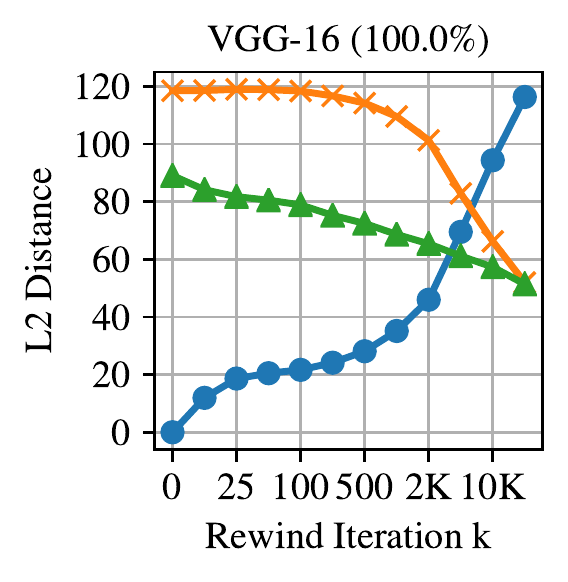}};
\node at (0.8,  -0.19) {\includegraphics[width=0.19\textwidth]{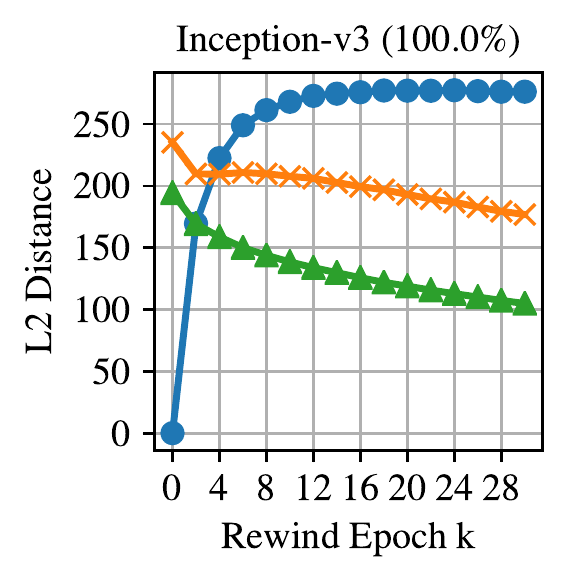}};

\end{tikzpicture}
\centering
\includegraphics[width=\textwidth]{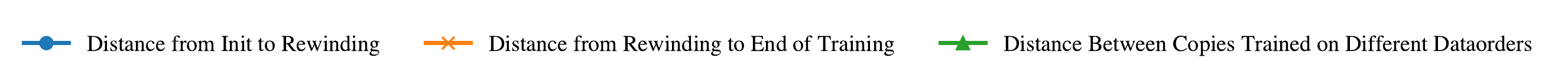}
\vspace{-2.3em}
\caption{Various $L_2$ distances for the full networks at the rewinding iteration specified on the x-axis.
         Each line is the mean and standard deviation across three initializations.}
\label{fig:state-l2-full}
\end{figure*}

\begin{figure*}
\begin{tikzpicture}[x=\textwidth,y=\textwidth, every node/.style = {anchor=north west}]
\node at (0.0, 0) {\includegraphics[width=0.19\textwidth]{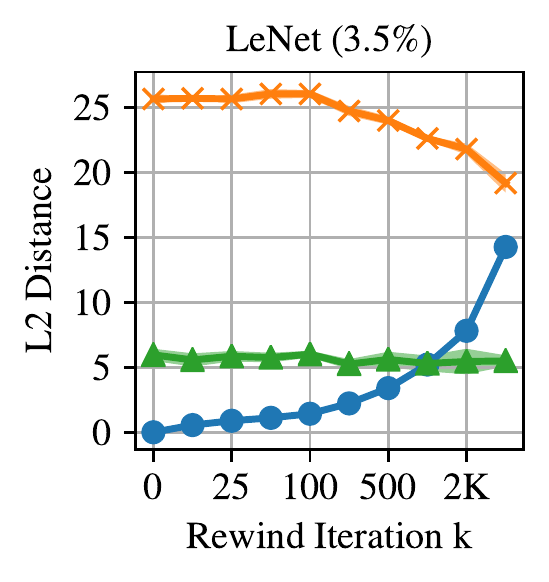}};
\node at (0.2, 0) {\includegraphics[width=0.19\textwidth]{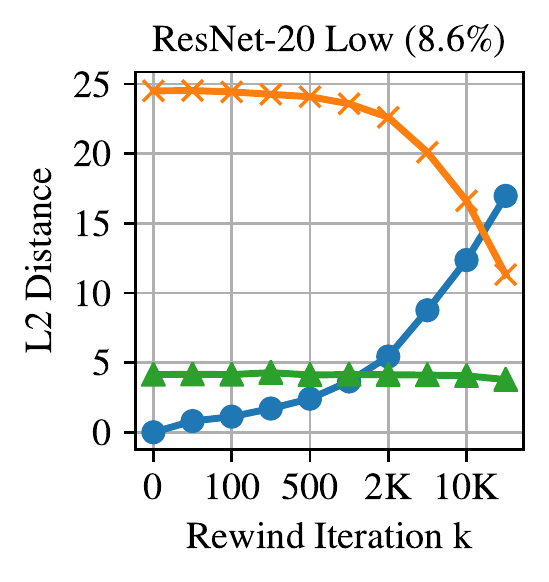}};
\node at (0.4, 0) {\includegraphics[width=0.19\textwidth]{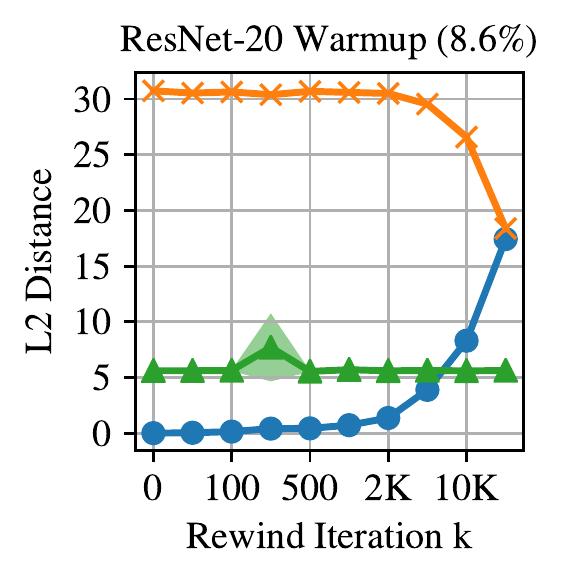}};
\node at (0.6, 0) {\includegraphics[width=0.19\textwidth]{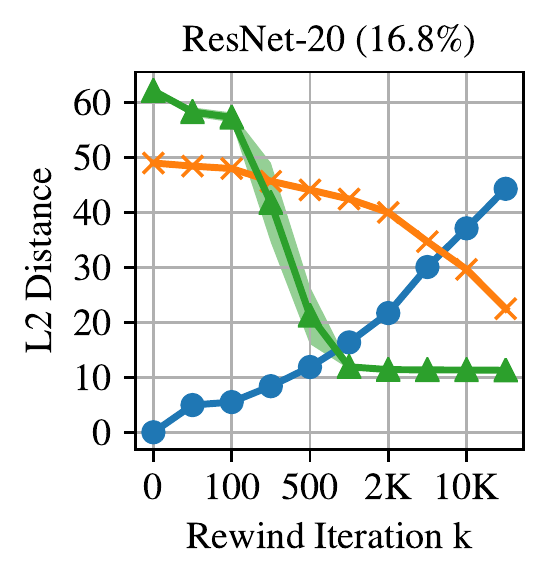}};
\node at (0.8, 0) {\includegraphics[width=0.19\textwidth]{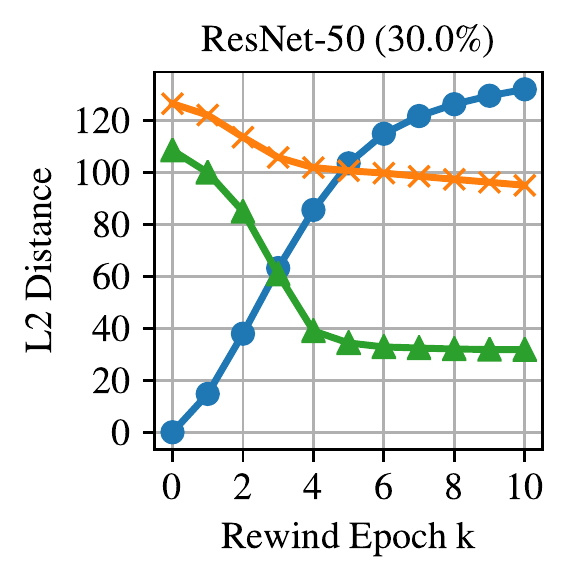}};

\node at (0.2,  -0.19) {\includegraphics[width=0.19\textwidth]{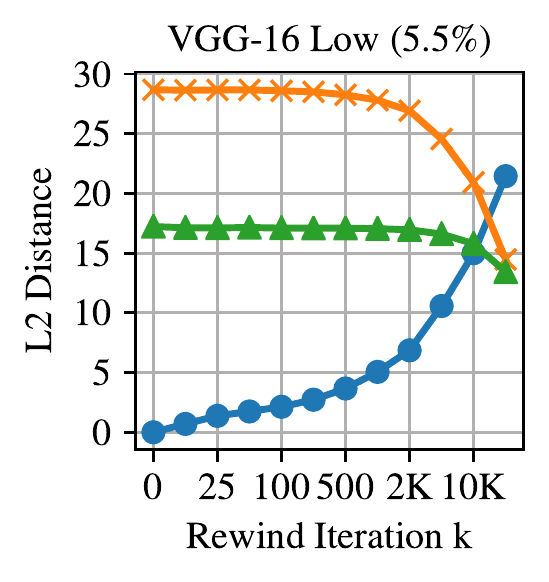}};
\node at (0.4,  -0.19) {\includegraphics[width=0.19\textwidth]{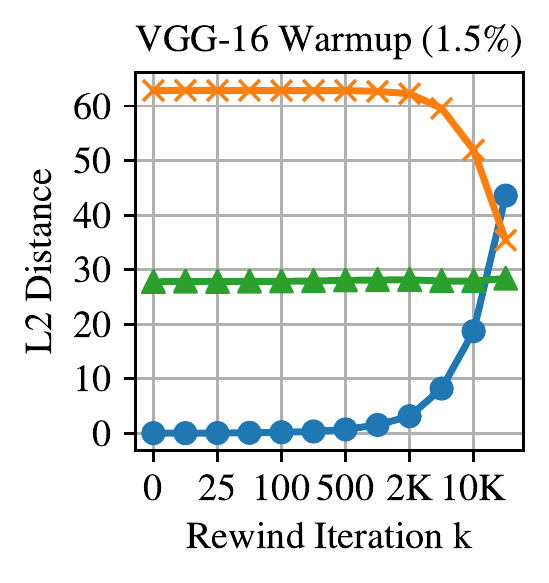}};
\node at (0.6,  -0.19) {\includegraphics[width=0.19\textwidth]{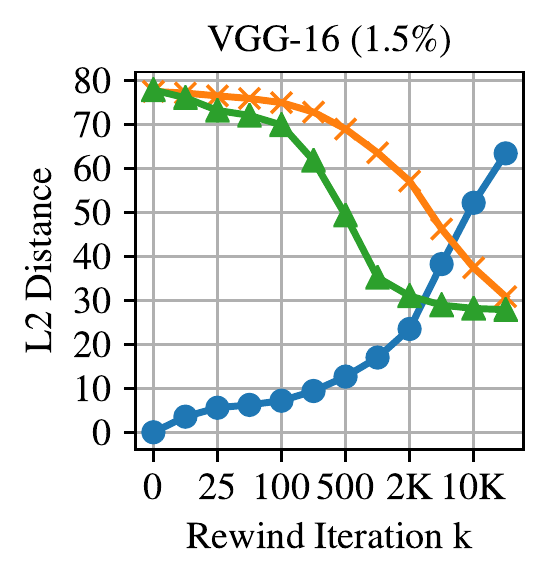}};
\node at (0.8,  -0.19) {\includegraphics[width=0.19\textwidth]{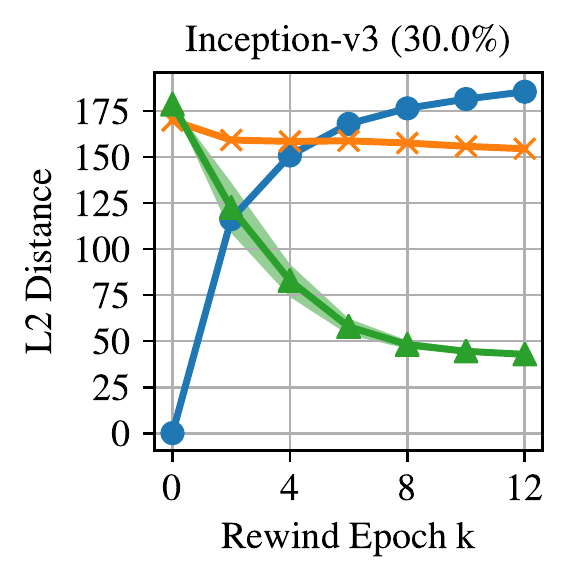}};

\end{tikzpicture}
\centering
\includegraphics[width=\textwidth]{figures/full-l2-at-lr/lenet-level0-dataorder-legend}
\vspace{-2.3em}
\caption{Various $L_2$ distances for the IMP subnetworks at the rewinding iteration specified on the x-axis.
         Each line is the mean and standard deviation across three initializations. Each $L_2$ distance is computed after applying the pruning mask to the states of the networks in question.}
\label{fig:state-l2-sparse}
\end{figure*}

\textbf{$L_2$ distances.}
In Figures \ref{fig:state-l2-full} and \ref{fig:state-l2-sparse}, we measure various $L_2$ distances that capture how close the network is to initialization and to the end of training.
In particular, we measure three distances as shown in the diagram below (which is an annotated version of Figure \ref{fig:stability-visualization}).

\begin{center}
\begin{tikzpicture}[scale=1]

    \filldraw[black] (0, -0.2) circle (2pt) node[anchor=west] {$W_0$};
    \draw[->, line width=1] (0, -0.3) -> (0, -1.2);
    \filldraw[black, line width=1] (0, -1.3) circle (2pt) node[anchor=west] {$W_k$};
    \draw[->, line width=1] (-0.1, -1.4) -> (-0.7, -2.9);
    \filldraw[black] (-0.7, -3) circle (2pt) node[anchor=east] {$W_T^1$};
    \draw[->, line width=1] (0.1, -1.4) -> (0.7, -2.9);
    \filldraw[black, line width=1] (0.7, -3) circle (2pt) node[anchor=west] {$W_T^2$};
    \definecolor{mygreen}{HTML}{2CA02C}
    \draw[color=mygreen, line width=1, |-|] (-0.5, -3) -- (0.5, -3) node[label={[yshift=-1.1cm,xshift=-0.5cm]:{\parbox{5cm}{\raggedright \small Distance Between Copies Trained on Different Dataorders}}},mygreen] {};
    \draw[fill=mygreen,mygreen] (0.0, -3) node[mark size=1.5mm] {\pgfuseplotmark{triangle*}};
    
    \definecolor{myorange}{HTML}{FF7F0E}
    \draw[color=myorange, line width=1,|-|] (-0.35, -1.25) -> (-1, -2.75) node[label={[xshift=-0.5cm, yshift=0.2cm]:\parbox{2.8cm}{\raggedright \small Distance from Rewinding to\\End of Training}},myorange] {};
    \draw[fill=myorange,myorange] (-0.84, -2) node[mark size=1.75mm] {\pgfuseplotmark{x}};
    
    \definecolor{myblue}{HTML}{1f77b4}
    \draw[fill=myblue,myblue] (-0.2, -0.75) circle (1mm);
    \draw[color=myblue,|-|, line width=1] (-0.2, -0.3) -> (-0.2, -1.2) node[label={[xshift=-0.2cm, yshift=-0.1cm]:\parbox{3cm}{\raggedright \small Distance\\from Init to\\ Rewinding}},myblue] {};
\end{tikzpicture}
\end{center}

First, we measure the $L_2$ distance in parameter space from initialization to the state of the network at step $k$ (blue circle in the diagram above and in Figures \ref{fig:state-l2-full} and \ref{fig:state-l2-sparse}); for the sparse IMP subnetworks, we measure the $L_2$ distance after applying the pruning mask to both initialization and the state of the network at iteration $k$.
This quantity captures the distance that the network has traversed from initialization by step $k$.

Second, we measure the distance from the state of the network at step $k$ to its state at the end of training under one data order (orange x in the diagram above and in Figures \ref{fig:state-l2-full} and \ref{fig:state-l2-sparse}).
This quantity captures the distance that the network traverses after step $k$.
If the network is very close to the minimum by the time it becomes stable, then we expect this quantity to be small compared to the $L_2$ distance between initialization and iteration $k$; that would indicate that the network has already traversed a large distance and has a relatively smaller distance to go.

Finally, we measure the distance between the final states of networks trained from step $k$ under different data orders (green triangle in the diagram above and in Figures \ref{fig:state-l2-full} and \ref{fig:state-l2-sparse}).
This quantity captures the size of the linearly connected minimum found by the networks.
We are interested in how this distance compares to the distance traveled by the networks and how this quantity changes as the rewinding iteration varies.

\subsection{Results}

\begin{figure*}
\begin{tikzpicture}[x=\textwidth,y=\textwidth, every node/.style = {anchor=north west}]
\node at (0.0, 0) {\includegraphics[width=0.19\textwidth]{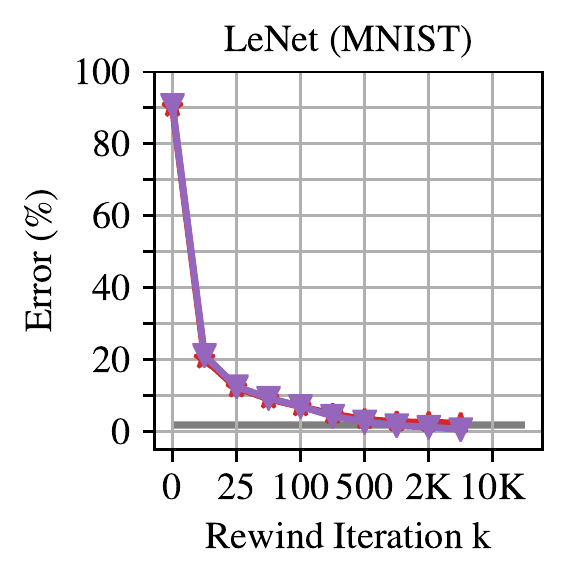}};
\node at (0.2, 0) {\includegraphics[width=0.19\textwidth]{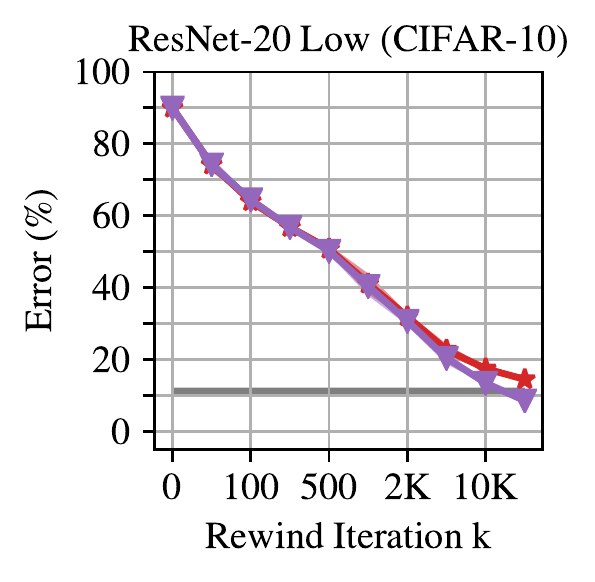}};
\node at (0.4, 0) {\includegraphics[width=0.19\textwidth]{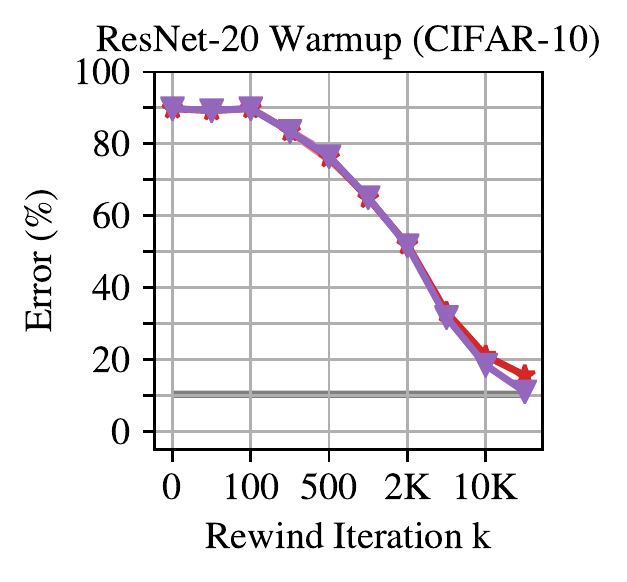}};
\node at (0.6, 0) {\includegraphics[width=0.19\textwidth]{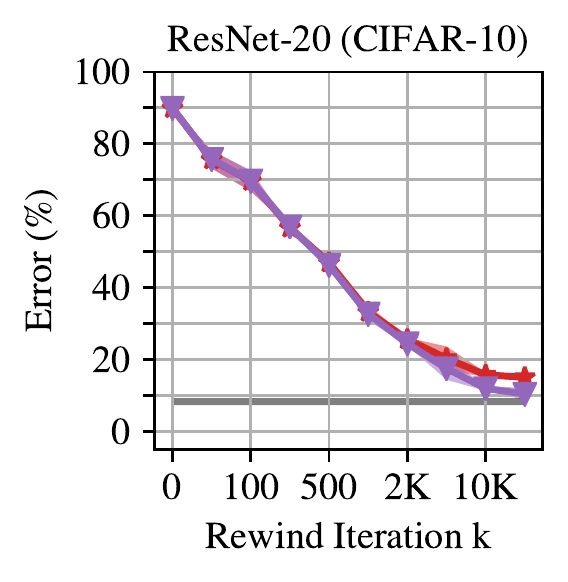}};
\node at (0.8, 0) {\includegraphics[width=0.19\textwidth]{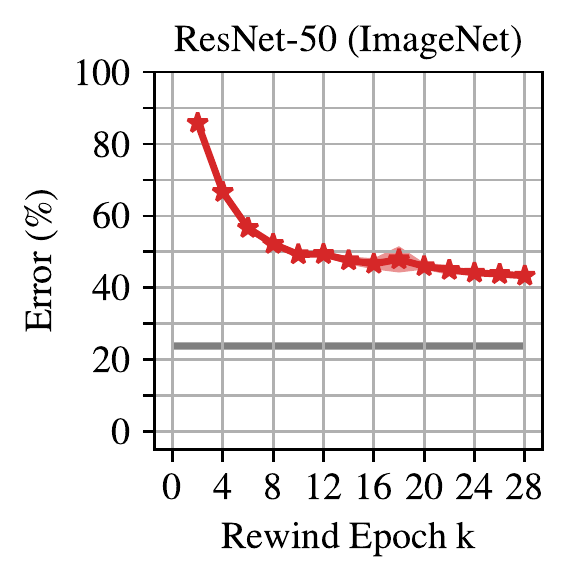}};

\node at (0.03, -0.24) {\includegraphics[width=0.16\textwidth]{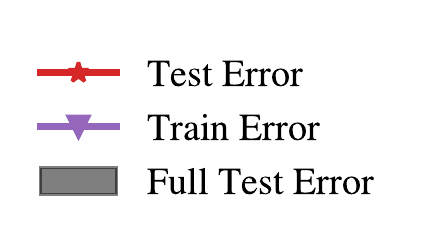}};
\node at (0.2,  -0.19) {\includegraphics[width=0.19\textwidth]{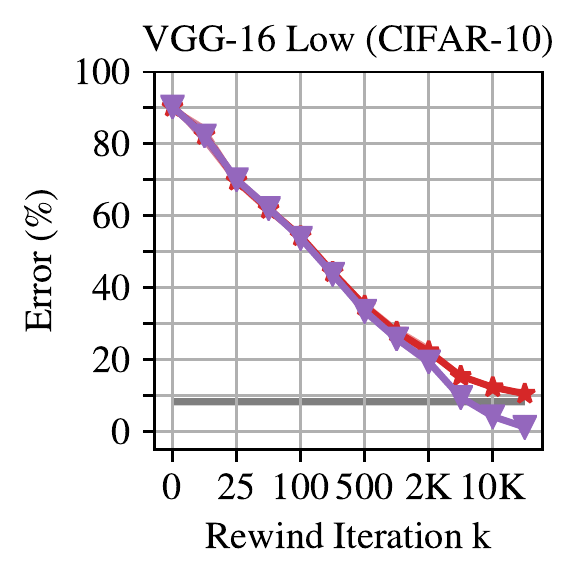}};
\node at (0.4,  -0.19) {\includegraphics[width=0.19\textwidth]{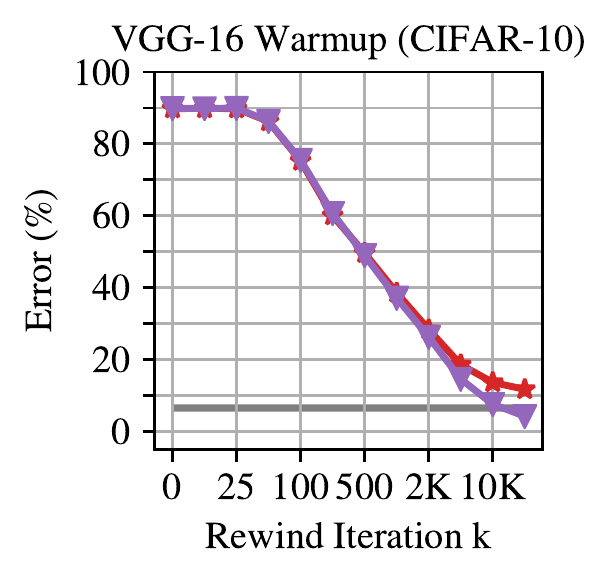}};
\node at (0.6,  -0.19) {\includegraphics[width=0.19\textwidth]{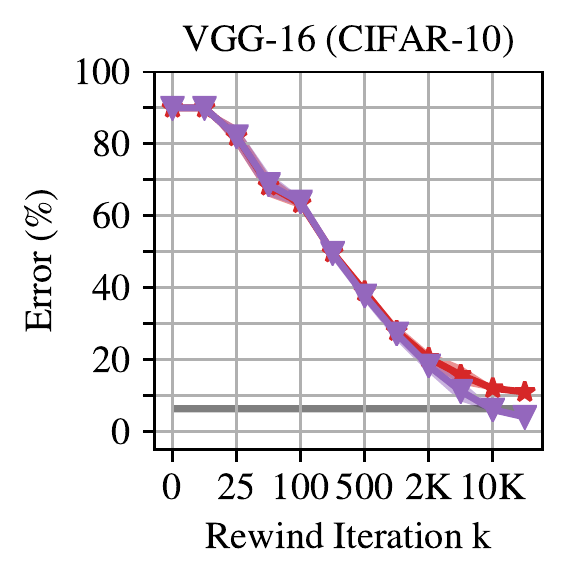}};
\node at (0.8,  -0.19) {\includegraphics[width=0.19\textwidth]{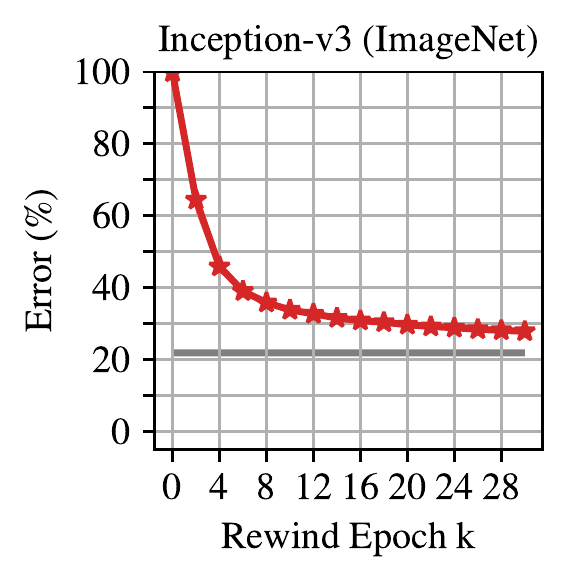}};

\end{tikzpicture}
\vspace{-2.3em}
\caption{The error of the full networks at the rewinding iteration specified on the x-axis.
         For clarity, this is the error of the network at that specific iteration of training, before any copies are made or further training occurs.
         Each line is the mean and standard deviation across three initializations.}
\label{fig:state-error}
\end{figure*}

\textbf{Error at rewinding.}
These results appear in Figure \ref{fig:state-error}.
Recall that the unpruned networks become stable at a different (typically later) iteration than the IMP subnetworks, so we consider two rewinding points for each network.

\textit{Unpruned networks.}
ResNet-20 and VGG-16 become stable to SGD noise at iterations 2000 and 1000, at which point test error is about 25\% (compared to final error 8.3\%) for ResNet-20 and 20\% (compared to final error 6.3\%) for VGG-16.
Train error is at a similar value to test error at these points; in both cases, train error eventually converges to 0\%.
We conclude that, at the iteration at which they become stable to SGD noise, these networks have not fully converged but are much closer to their final errors than to random guessing.

We see similar behavior for the unpruned ResNet-50 and Inception-v3 networks, which become stable to SGD noise at epochs 18 and 28.
At these points, test error is 55\% (compared to final error 24\%) for ResNet-50 and 33\% (compared to final error 22\%) for Inception-v3.
Both networks are most of the way to their final performance.

\textit{IMP pruned subnetworks.}
The IMP pruned subnetworks become stable to SGD noise earlier than the unpruned networks. 
ResNet-20 and VGG-16 become stable to SGD noise at iterations 500 and 1000, at which point error is 30\% (compared to final error 8.3\%) for ResNet-20 and 35\% (compared to final error 6.3\%) for VGG-16.
These networks have not fully converged but are closer to their final errors than to random guessing.
IMP subnetworks of ResNet-50 and Inception-v3 become stable to SGD noise much earlier than the unpruned networks---at epoch 5 and epoch 6, respectively.
At these points, error is much higher---55\% for ResNet-50 and 40\% for Inception-v3---leaving these networks substantial room to further train.
We did not evaluate the train accuracy at these checkpoints for the ImageNet networks due to storage and computational limitations.

\textbf{$L_2$ distances.}
These results appear in Figures \ref{fig:state-l2-full} and \ref{fig:state-l2-sparse}.

\textit{Unpruned networks.}
ResNet-20 and VGG-16 become stable to SGD noise at iterations 2000 and 1000, at which point they are closer to their initial weights than to their final weights.
This indicates that they still have a substantial distance to travel on the optimization landscape and are still far from their final weights.
This result is particularly remarkable considering our observation in Appendix \ref{app:overtime} that stable networks follow the same, linearly connected trajectory throughout training (according to test error); the $L_2$ distance data suggests that they do so for a substantial distance.

The unpruned ResNet-50 and Inception-v3 networks are closer to their final weights than their initial weights when they become stable to SGD noise.
In fact, it appears that distance from initialization begins to plateau and distance to the final weights only decreases slowly.
This may indicate that the networks will make much slower progress for the remaining 80\% of training iterations.

The green triangles in these plots show the distance between the weights of copies of the network trained from a rewinding iteration to completion on different data orders.
In all cases, the distance between these copies is substantial, even after the networks become stable.
As a point of comparison, we use the distance that the networks travel between initialization and the final weights, which is captured by the orange $x$ for rewinding iteration 0.
For ResNet-20, the distance between copies trained on different data orders from iteration 2000 (when it becomes stable) is more than half the distance that the network travels during the entirety of training.
The same is true for VGG-16 from iteration 1000 (when it becomes stable).
For ResNet-50 and Inception-v3, this distance is about a quarter and half (respectively) of the distance the networks travel over the course of training.
These are remarkably large distances considering that any network on this line segment reaches full test accuracy.

\textit{IMP pruned subnetworks.}
We show the same data for the IMP subnetworks in Figure \ref{fig:state-l2-sparse}.
Each $L_2$ distance in this figure is measured after applying the pruning mask to all weights.
When ResNet-20 and VGG-16 become stable to SGD noise (iterations 2000 and 1000, respectively), they are about 2x (ResNet-20) and 3x (VGG-16) closer to their initial weights than their final weights.
ResNet-50 and Inception-v3 are about equal distances from both points for the epochs at which they become stable.

Unique to the IMP subnetworks, we observe here and in Appendix \ref{app:alternate-distance-metrics} that the $L_2$ distance between copies trained on different data orders drops alongside instability, plateauing at a lower value when training from the rewinding iteration at which the subnetworks becomes stable.
Even this lower distance is still a substantial fraction of the overall distance the network travels: 25\%, 45\%, 27\%, and 28\% for ResNet-20, VGG-16, ResNet-50, and Inception-v3.

\section{Instability Throughout Training}
\label{app:throughout}
\label{app:overtime}

In Section \ref{sec:full-networks}, we find stable networks that arrive at minima that are linearly connected.
In this appendix, we study whether the trajectories they follow are also linearly connected.
In other words, when training two copies of the same network with different noise, are the states of the network at each step $t$ connected by a linear path over which test error does not increase?
In the main body of the paper, we study this quantity only at the end of training (i.e., $t = T$).
Here, we study it for all iterations $t$ \emph{throughout} training.
To study this behavior, we linearly interpolate between the networks at each epoch of training and compute instability.

\begin{figure*}
\centering
\includegraphics[width=\columnwidth]{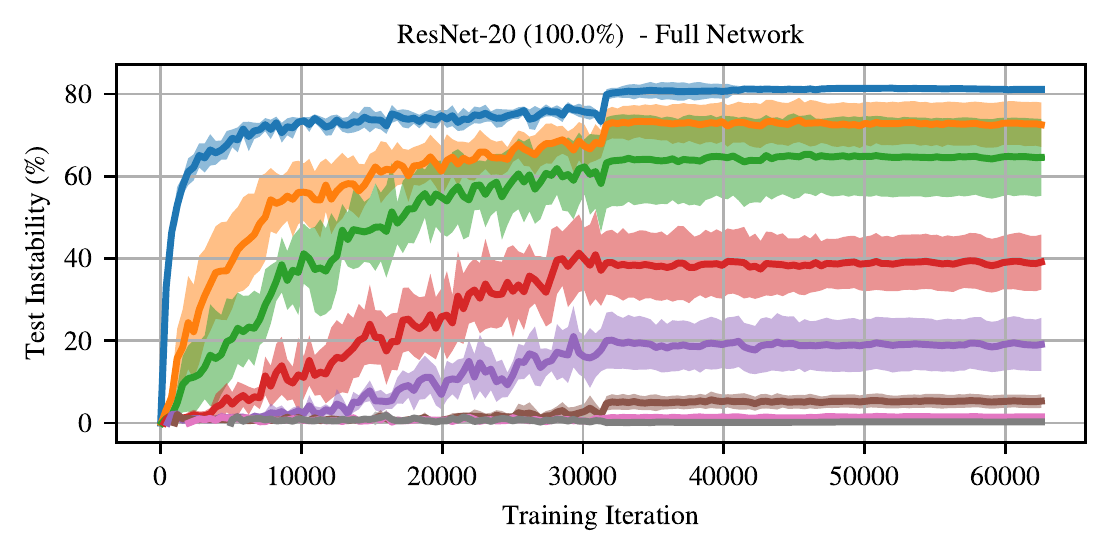}
\includegraphics[width=\columnwidth]{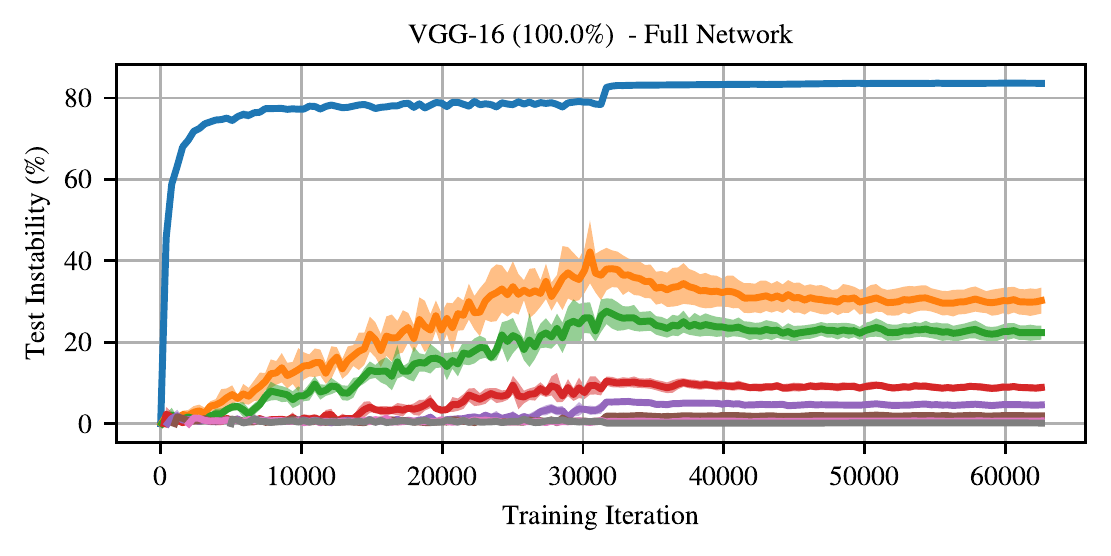}

\includegraphics[width=\columnwidth]{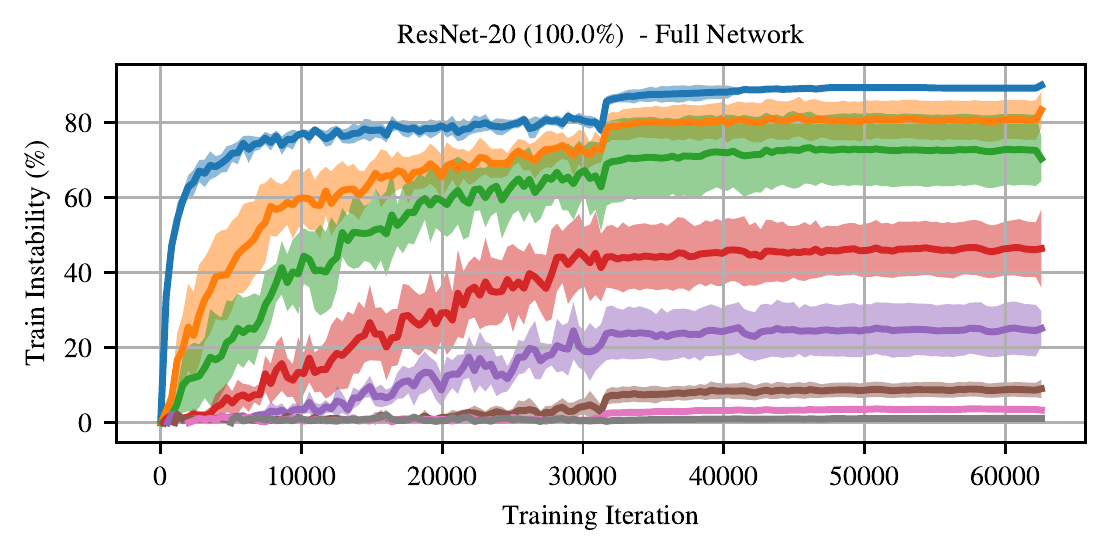}
\includegraphics[width=\columnwidth]{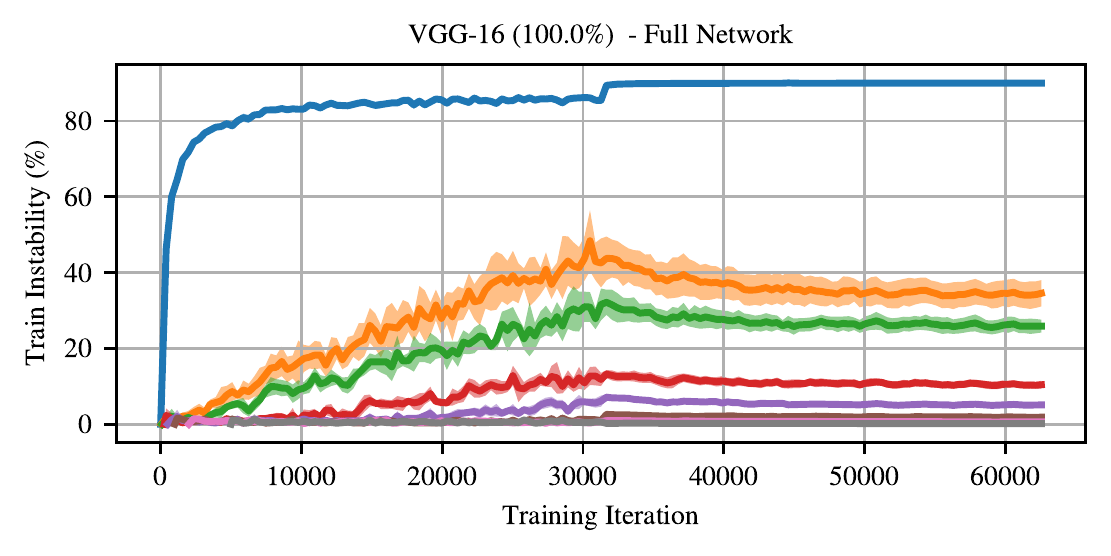}

\includegraphics[width=\columnwidth]{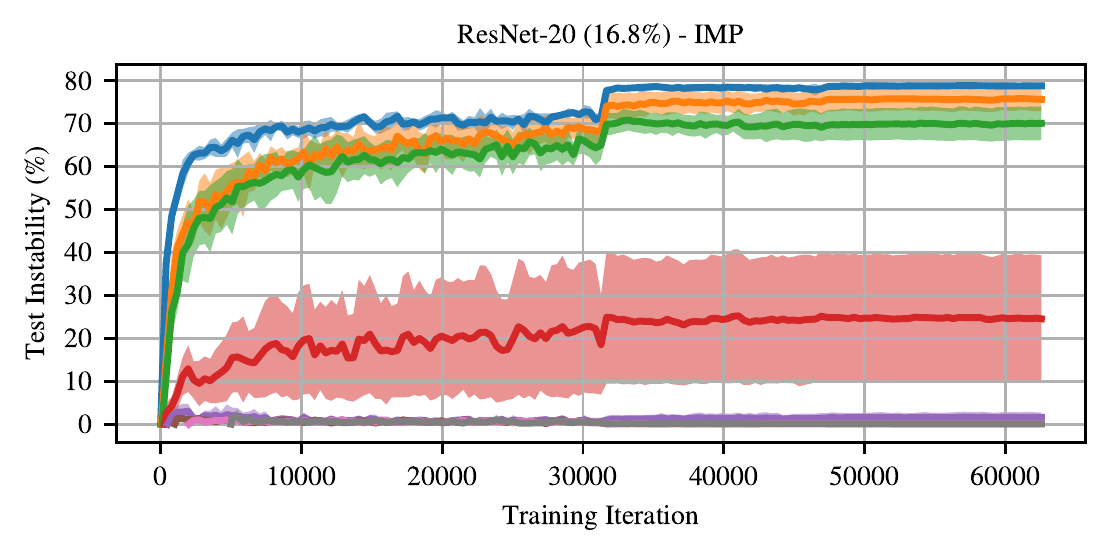}
\includegraphics[width=\columnwidth]{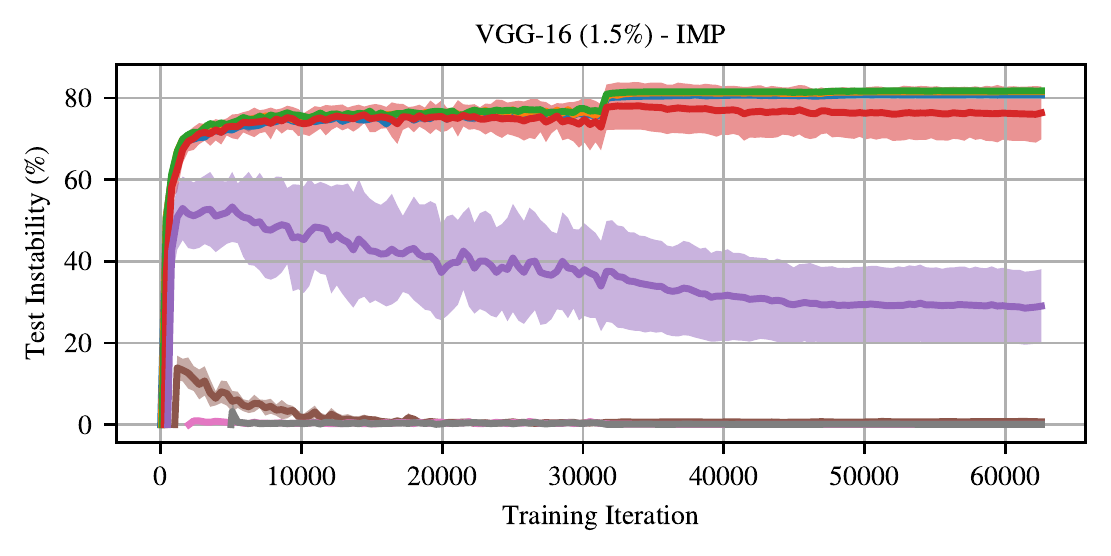}

\includegraphics[width=\columnwidth]{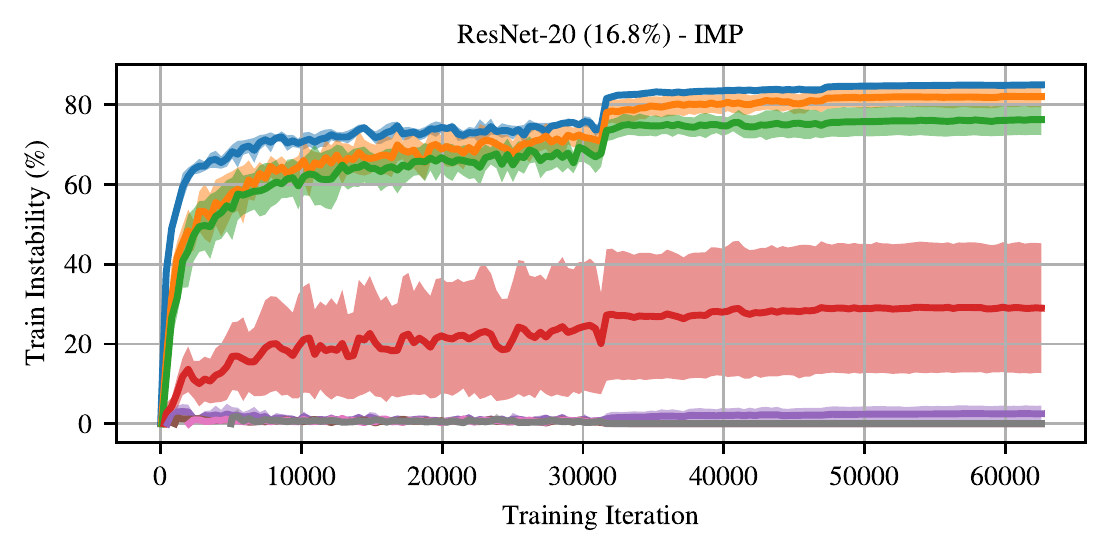}
\includegraphics[width=\columnwidth]{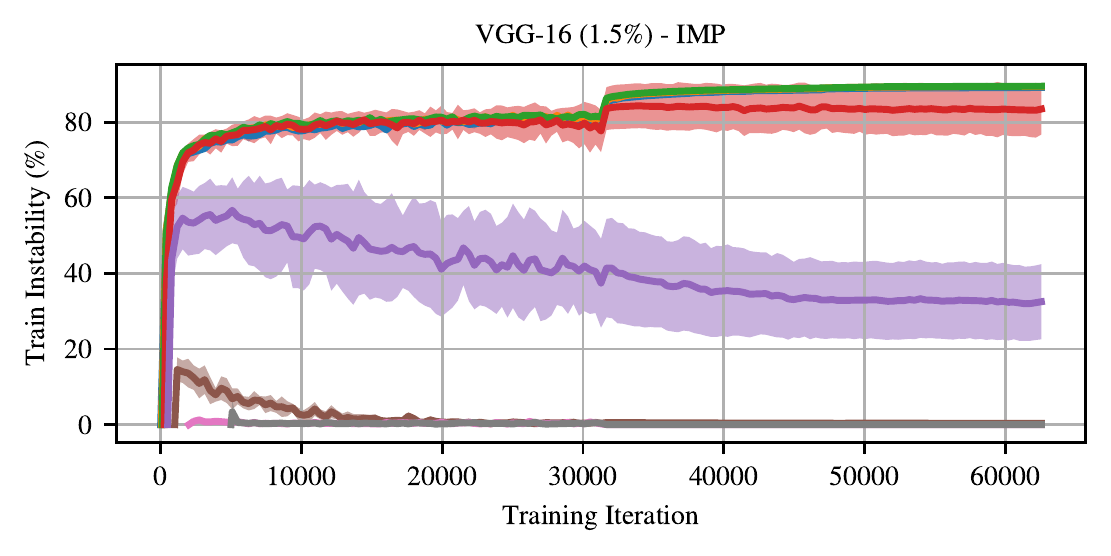}

\includegraphics[width=.6\textwidth]{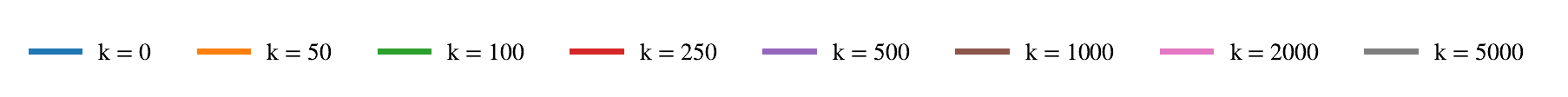}\vspace{-1em}
\caption{Instability throughout training for ResNet-20 and VGG-16 using both the unpruned networks and the IMP-pruned networks as computed on both the test set and train set. Each line involves training to iteration $k$ and then training two copies on different data orders after.
Each point is the instability when interpolating between the states of the networks at the training iteration on the x-axis.}
\label{fig:instability-overtime-appendix}
\end{figure*}

Figure \ref{fig:instability-overtime-appendix} plots instability throughout training for ResNet-20 and VGG-16 from different rewinding iterations $k$ for both train and test error for the unpruned networks and the IMP subnetworks.
We begin with the unpruned networks.
For $k = 0$ (blue line), instability increases rapidly.
In fact, it follows the same pattern as error: as the train or test error of each network decreases, the maximum possible instability increases (since instability never exceeds random guessing).
With larger values of $k$, instability increases more slowly throughout training.
When $k$ is sufficiently large that the networks are stable at the end of training, they are generally stable at every epoch of training ($k = 2000$, pink line).
In other words, after iteration 2000, the networks follow identical optimization trajectories \NA{modulo linear interpolation}.

The IMP subnetworks of ResNet-20 exhibit the same behavior as the unpruned network: when the network is stable at the end of training, it is stable throughout training, meaning two copies of the same network follow the same optimization trajectory up to linear interpolation.
The IMP subnetworks of VGG-16 exhibit sightly different behavior at rewinding iterations 500 and 1000: instability initially spikes (meaning the networks rapidly become separated by a loss barrier) but decreases gradually thereafter.
For rewinding iteration 1000, it decreases to 0, meaning the networks are stable by the end of training.
For all other rewinding iterations, being stable at the end of training corresponds to being stable throughout training, so it is possible that rewinding iteration 1000 represents a transition point between the unstable rewinding iterations earlier and the stable rewinding iterations later.

\section{Instability Data at All Sparsities}
\label{app:stability-over-sparsity}

\begin{figure*}
\begin{tikzpicture}[x=\textwidth,y=\textwidth, every node/.style = {anchor=north west}]
\node at (0.0, 0) {\includegraphics[width=0.19\textwidth]{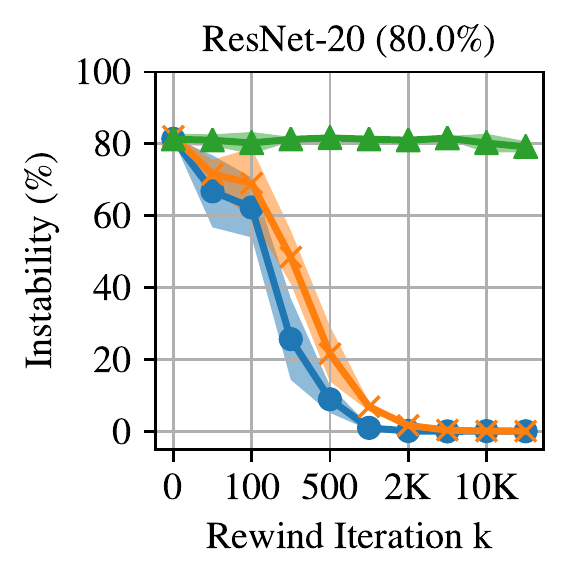}};
\node at (0.2, 0) {\includegraphics[width=0.19\textwidth]{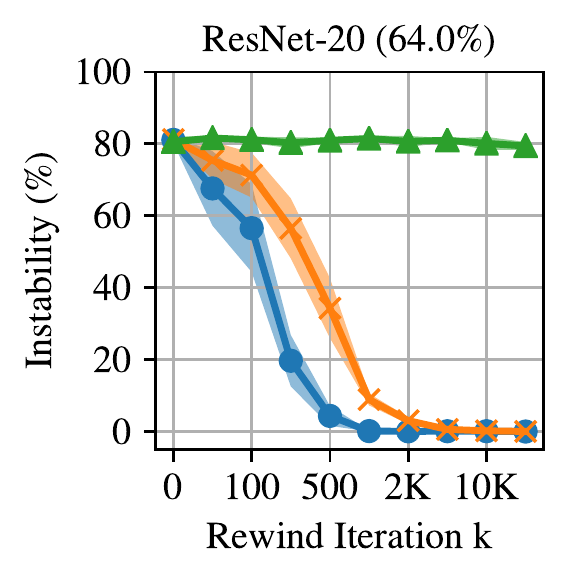}};
\node at (0.4, 0) {\includegraphics[width=0.19\textwidth]{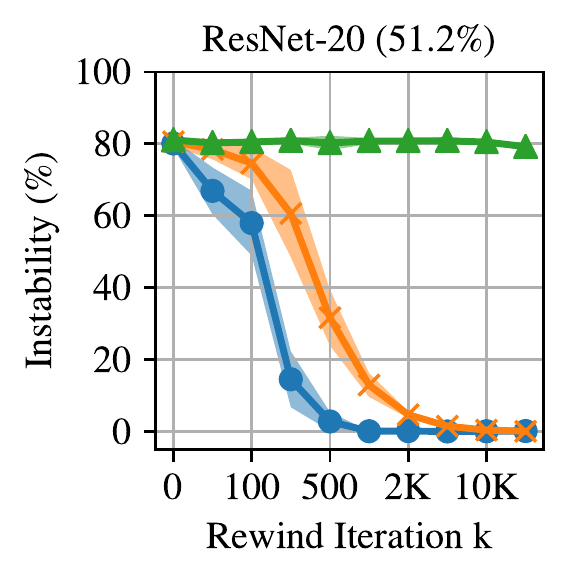}};
\node at (0.6, 0) {\includegraphics[width=0.19\textwidth]{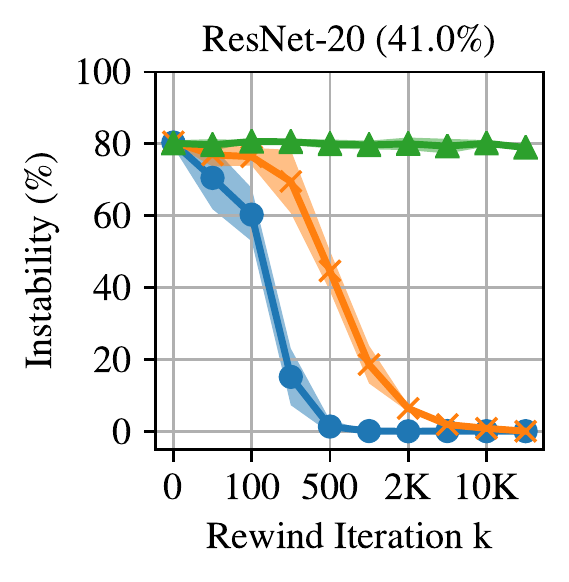}};
\node at (0.8, 0) {\includegraphics[width=0.19\textwidth]{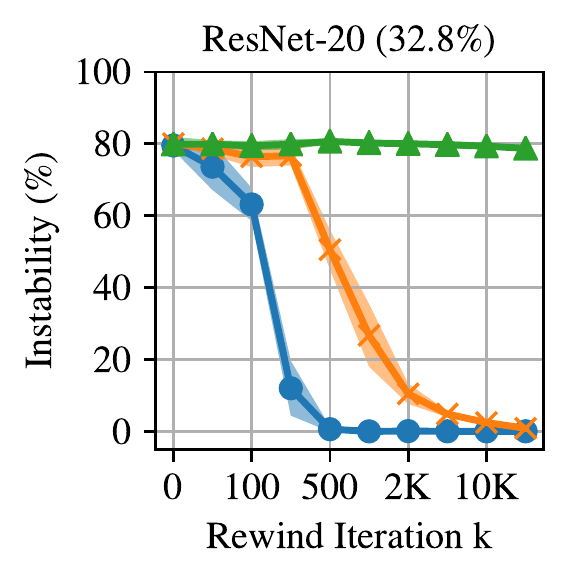}};

\node at (0.0, -0.19) {\includegraphics[width=0.19\textwidth]{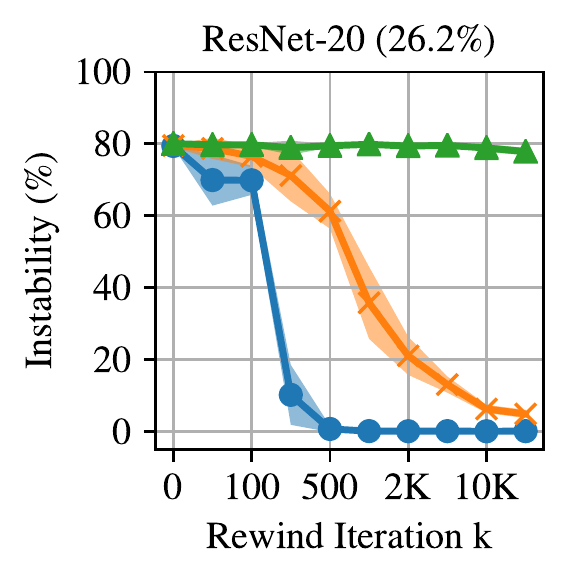}};
\node at (0.2, -0.19) {\includegraphics[width=0.19\textwidth]{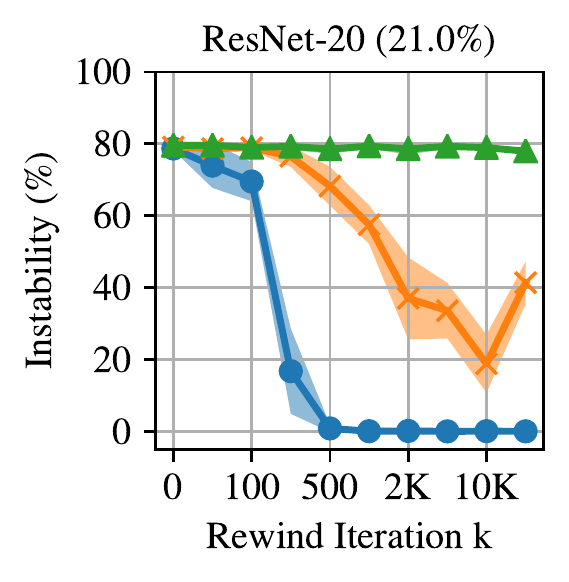}};
\node at (0.4, -0.19) {\includegraphics[width=0.19\textwidth]{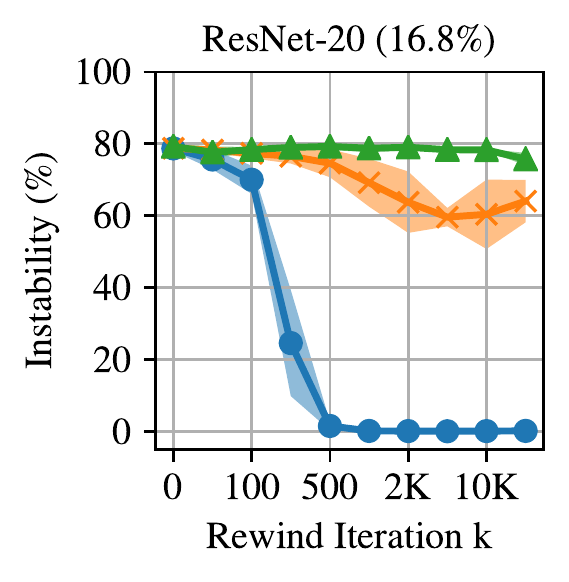}};
\node at (0.6, -0.19) {\includegraphics[width=0.19\textwidth]{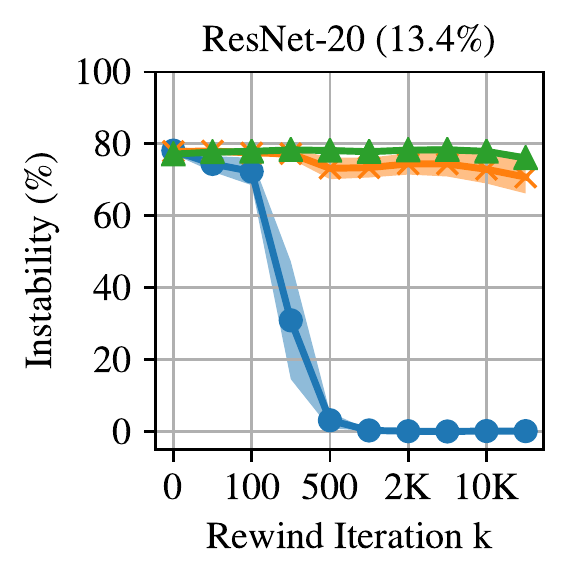}};
\node at (0.8, -0.19) {\includegraphics[width=0.19\textwidth]{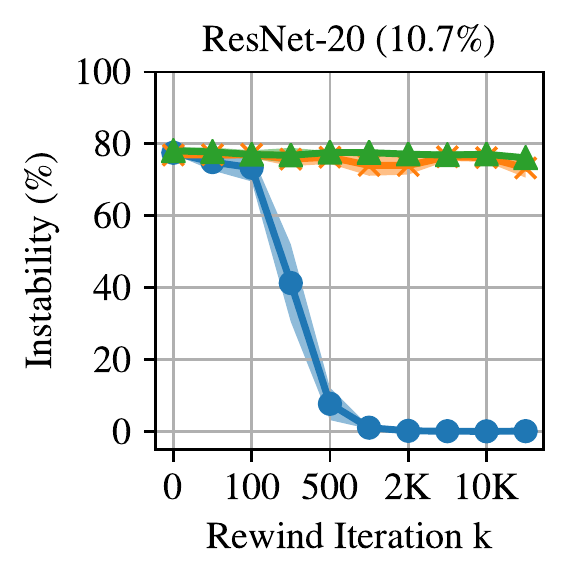}};

\node at (0.0, -0.38) {\includegraphics[width=0.19\textwidth]{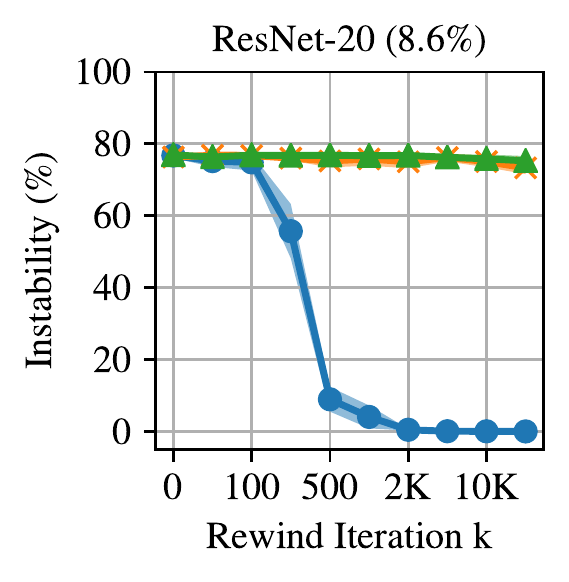}};
\node at (0.2, -0.38) {\includegraphics[width=0.19\textwidth]{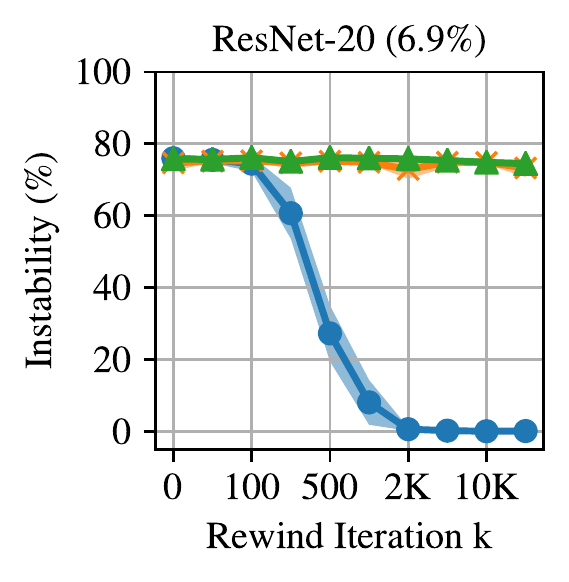}};
\node at (0.4, -0.38) {\includegraphics[width=0.19\textwidth]{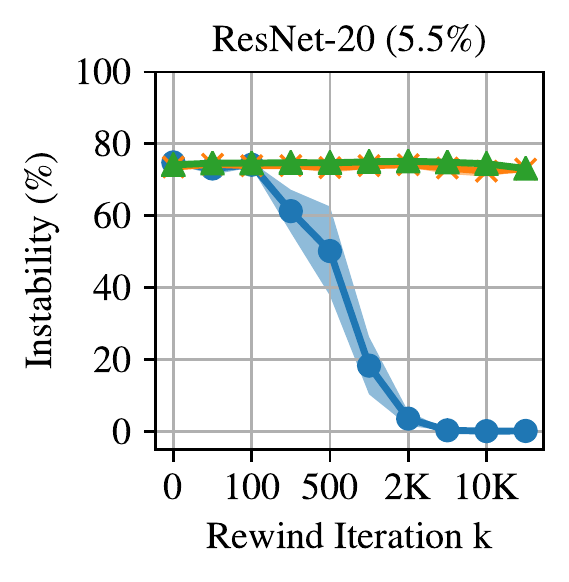}};
\node at (0.6, -0.38) {\includegraphics[width=0.19\textwidth]{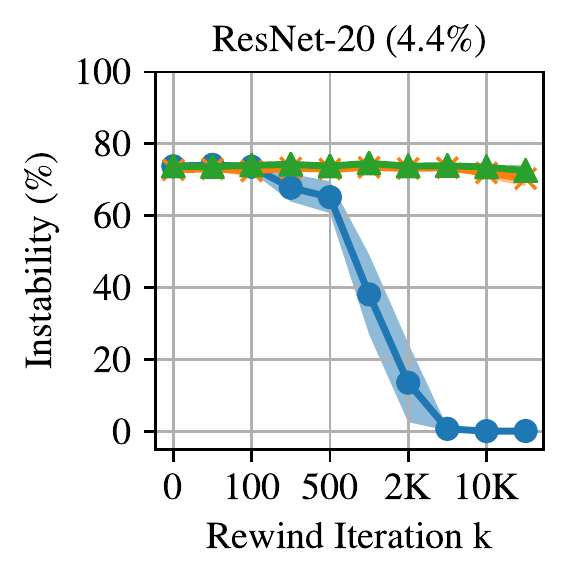}};
\node at (0.8, -0.38) {\includegraphics[width=0.19\textwidth]{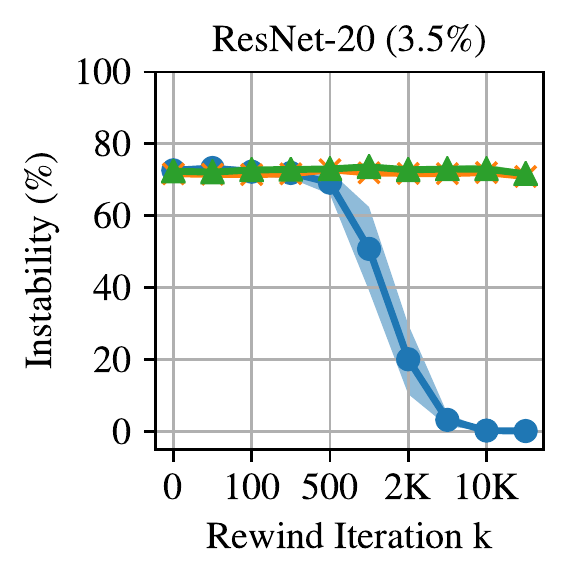}};

\end{tikzpicture}
\centering
\vspace{-1em}%
\includegraphics[width=0.5\textwidth]{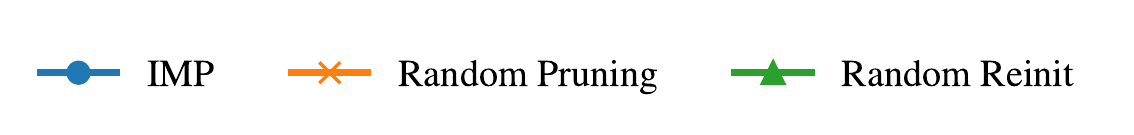}
\vspace{-0.5em}
\caption{The instability of subnetworks of ResNet-20 created using the state of the full network at iteration $k$ and trained on different data orders from there. Each line is the mean and standard deviation across three initializations and three data orders (nine samples total). Percents are percents of weights remaining.}
\label{fig:resnet20-across-sparsities-stability}
\end{figure*}

\begin{figure*}
\begin{tikzpicture}[x=\textwidth,y=\textwidth, every node/.style = {anchor=north west}]
\node at (0.0, 0) {\includegraphics[width=0.19\textwidth]{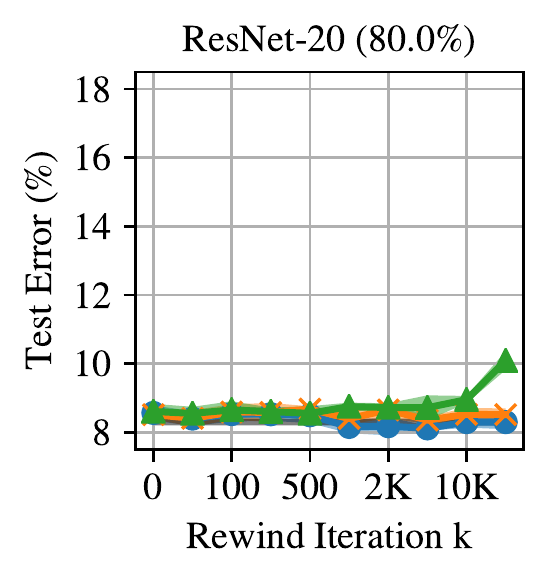}};
\node at (0.2, 0) {\includegraphics[width=0.19\textwidth]{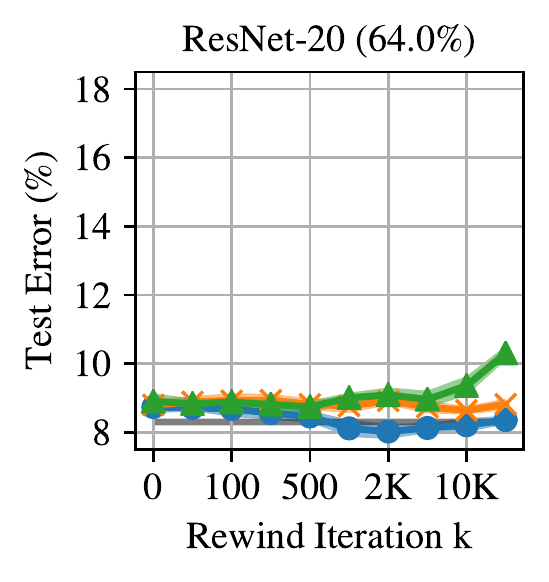}};
\node at (0.4, 0) {\includegraphics[width=0.19\textwidth]{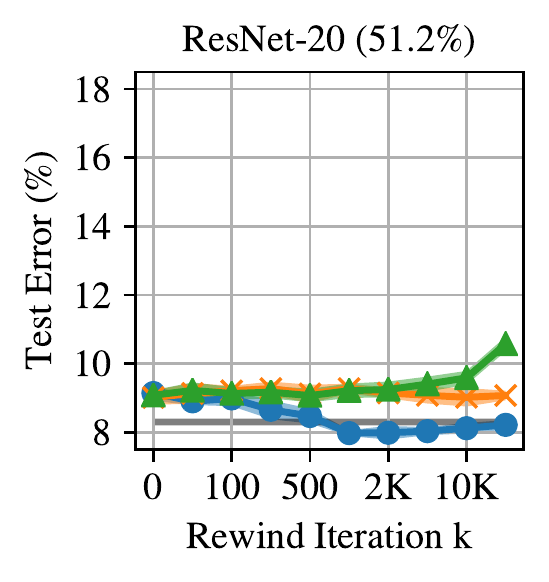}};
\node at (0.6, 0) {\includegraphics[width=0.19\textwidth]{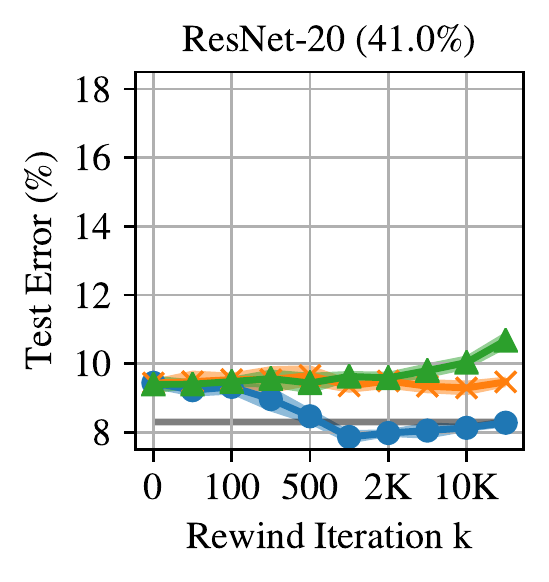}};
\node at (0.8, 0) {\includegraphics[width=0.19\textwidth]{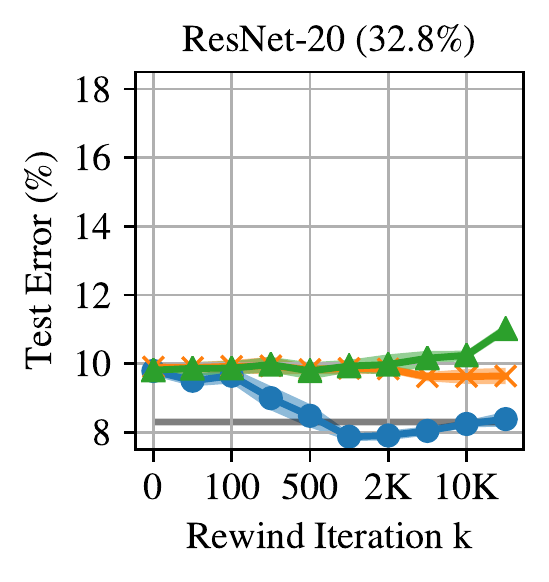}};

\node at (0.0, -0.19) {\includegraphics[width=0.19\textwidth]{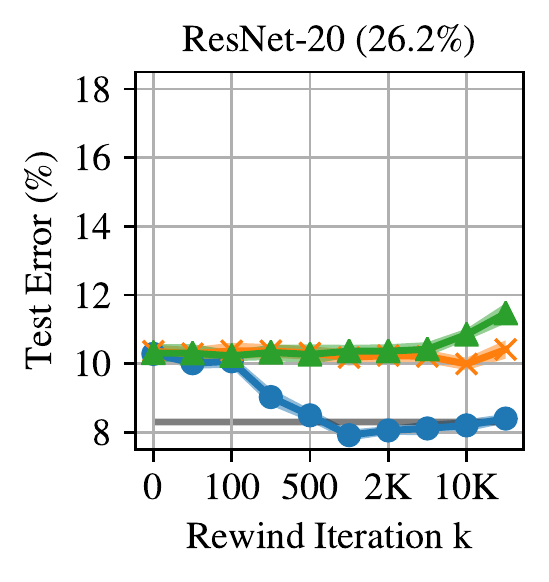}};
\node at (0.2, -0.19) {\includegraphics[width=0.19\textwidth]{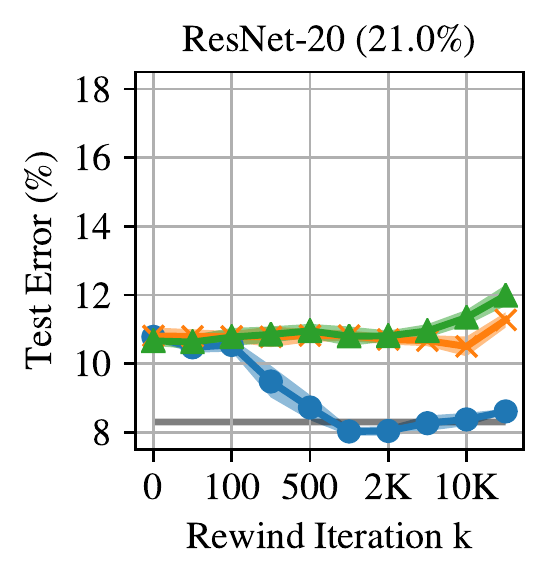}};
\node at (0.4, -0.19) {\includegraphics[width=0.19\textwidth]{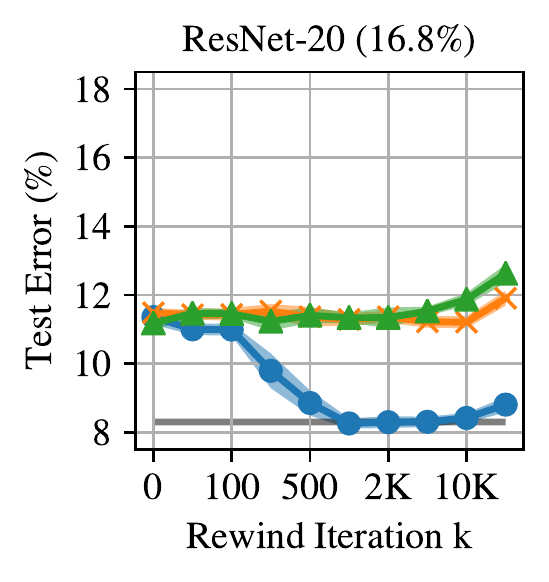}};
\node at (0.6, -0.19) {\includegraphics[width=0.19\textwidth]{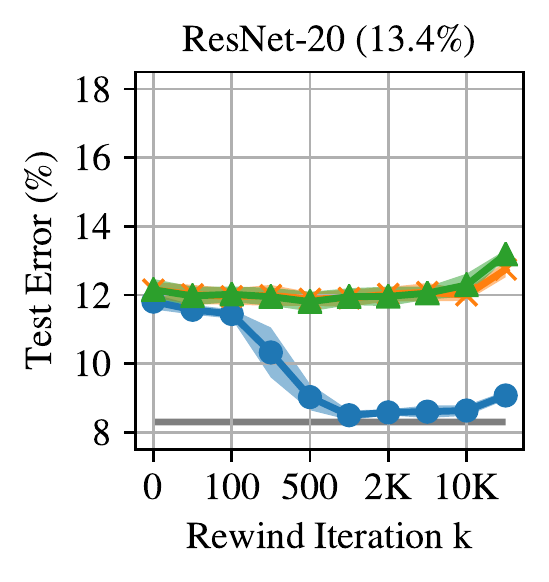}};
\node at (0.8, -0.19) {\includegraphics[width=0.19\textwidth]{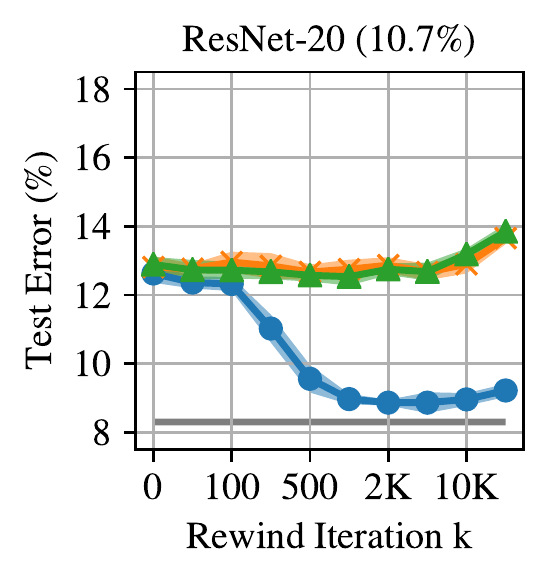}};

\node at (0.0, -0.38) {\includegraphics[width=0.19\textwidth]{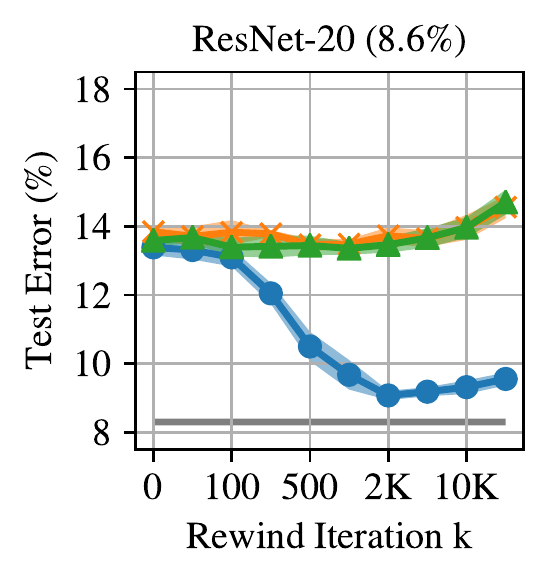}};
\node at (0.2, -0.38) {\includegraphics[width=0.19\textwidth]{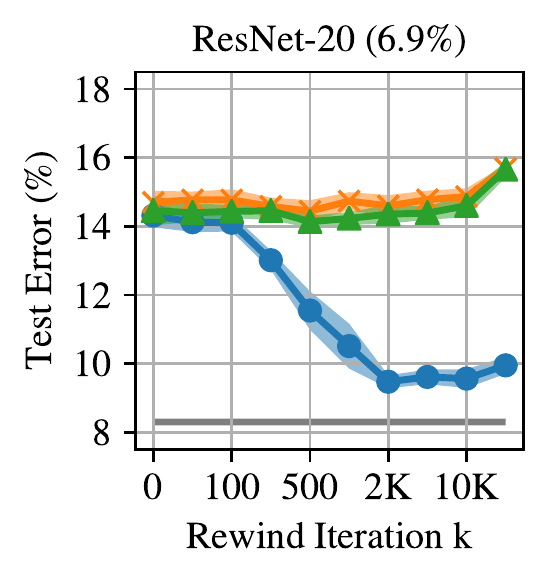}};
\node at (0.4, -0.38) {\includegraphics[width=0.19\textwidth]{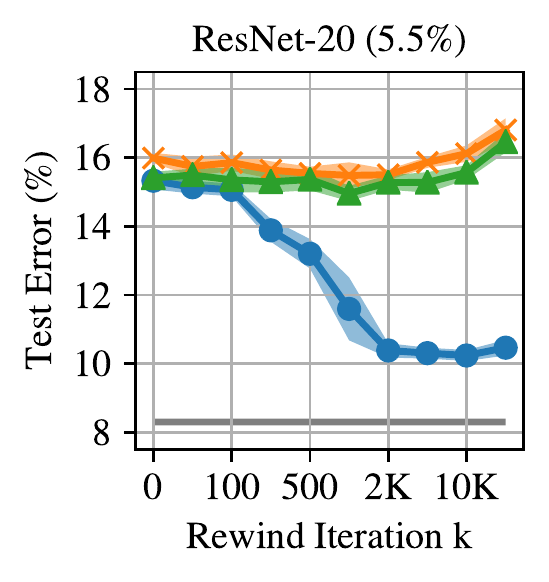}};
\node at (0.6, -0.38) {\includegraphics[width=0.19\textwidth]{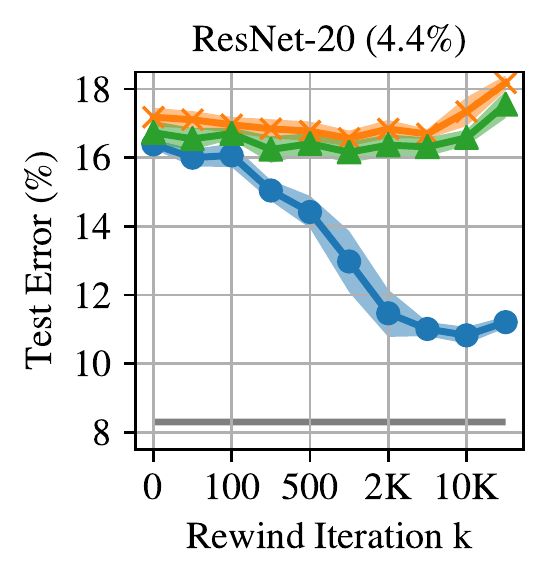}};
\node at (0.8, -0.38) {\includegraphics[width=0.19\textwidth]{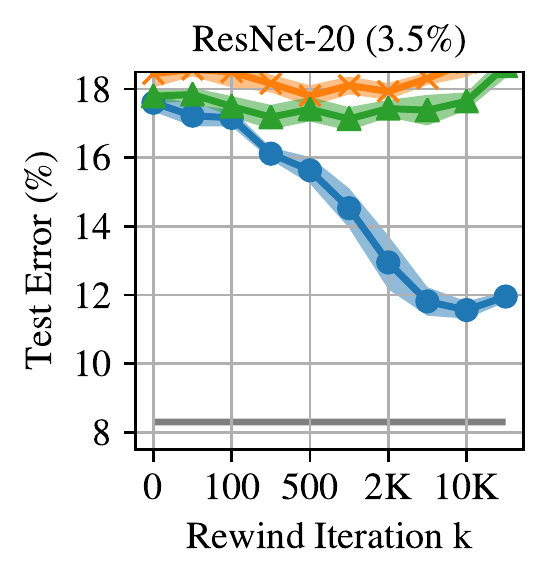}};

\end{tikzpicture}
\centering
\vspace{-1em}%
\includegraphics[width=0.6\textwidth]{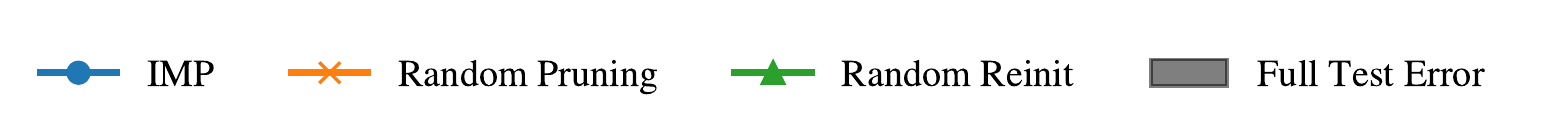}
\vspace{-0.5em}
\caption{The test error of subnetworks of ResNet-20 created using the state of the full network at iteration $k$ and trained on different data orders from there. Each line is the mean and standard deviation across three initializations. Gray lines are the accuracies of the full networks to one standard deviation. Percents are percents of weights remaining.}
\label{fig:resnet20-across-sparsities-error}
\end{figure*}

\begin{figure*}
\begin{tikzpicture}[x=\textwidth,y=\textwidth, every node/.style = {anchor=north west}]
\node at (0.0, 0) {\includegraphics[width=0.19\textwidth]{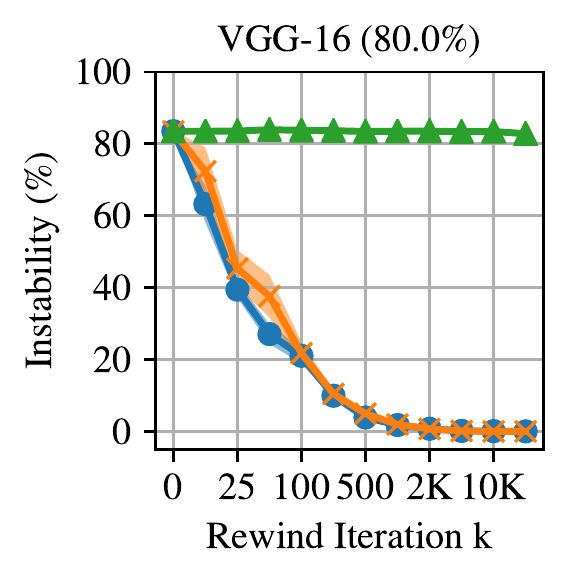}};
\node at (0.2, 0) {\includegraphics[width=0.19\textwidth]{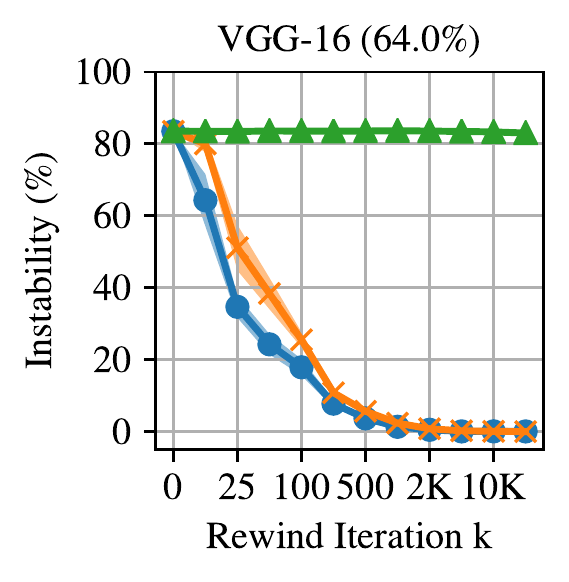}};
\node at (0.4, 0) {\includegraphics[width=0.19\textwidth]{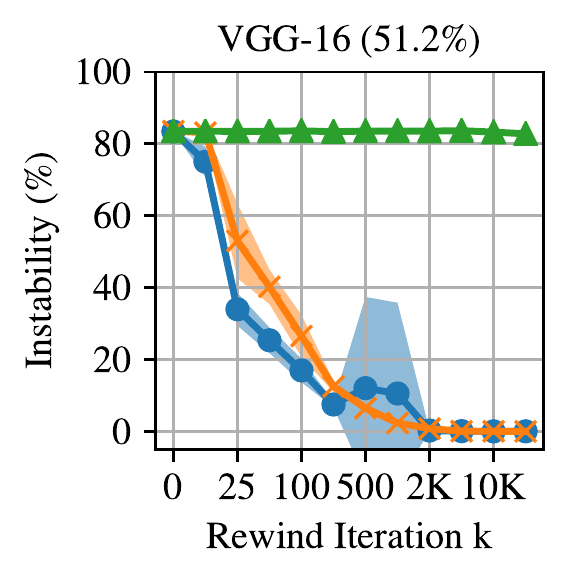}};
\node at (0.6, 0) {\includegraphics[width=0.19\textwidth]{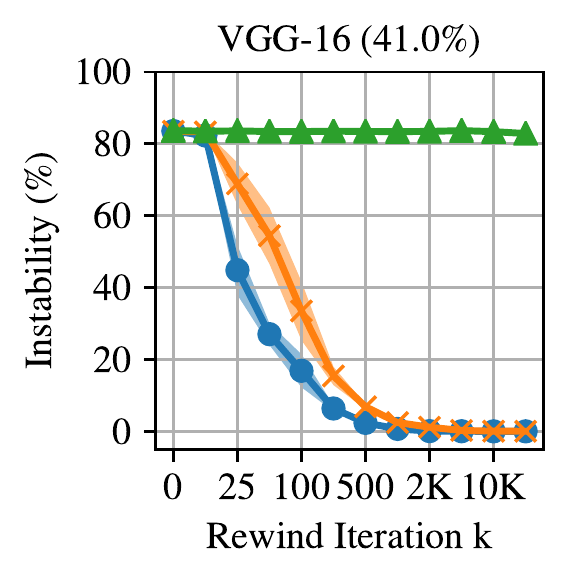}};
\node at (0.8, 0) {\includegraphics[width=0.19\textwidth]{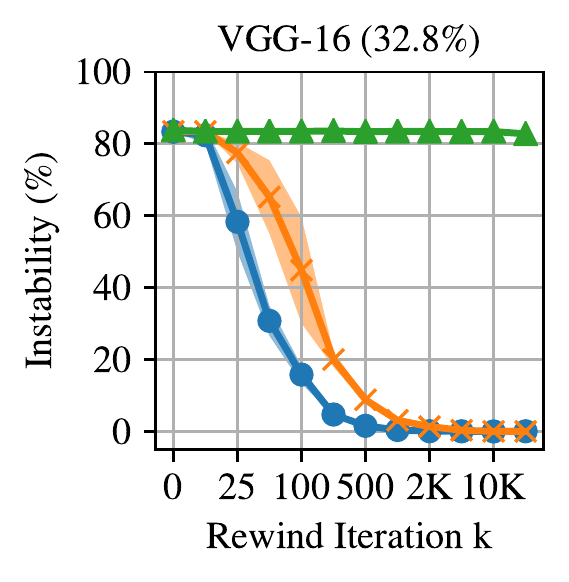}};

\node at (0.0, -0.19) {\includegraphics[width=0.19\textwidth]{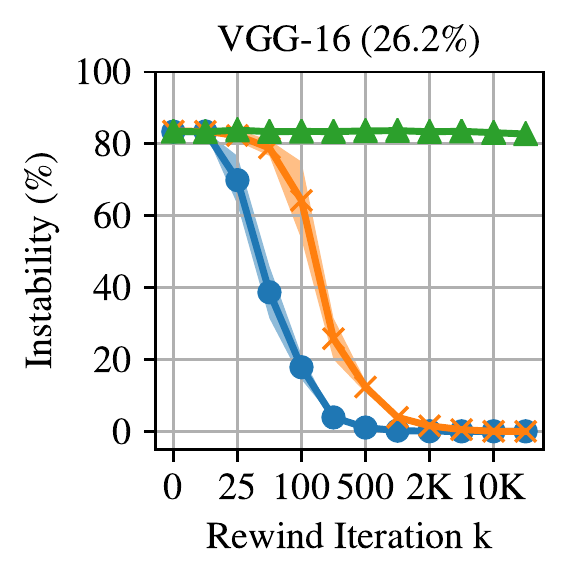}};
\node at (0.2, -0.19) {\includegraphics[width=0.19\textwidth]{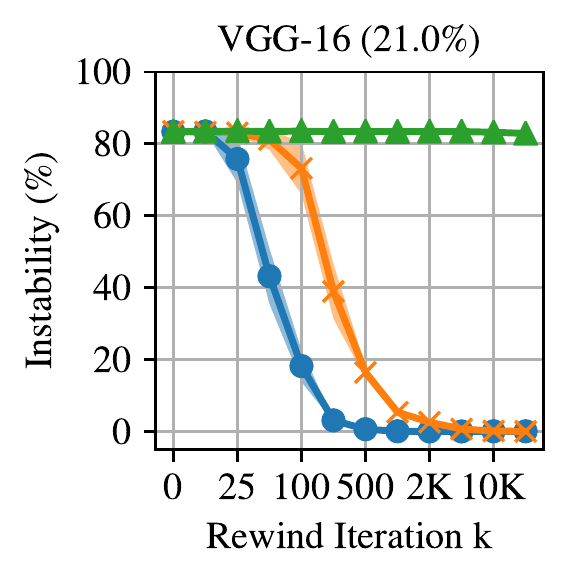}};
\node at (0.4, -0.19) {\includegraphics[width=0.19\textwidth]{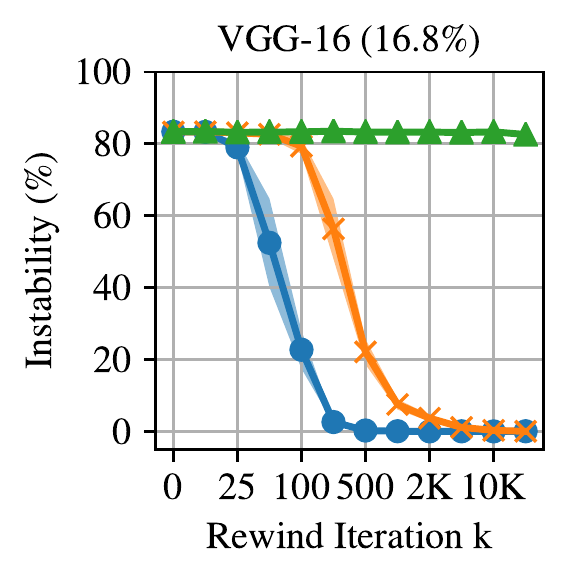}};
\node at (0.6, -0.19) {\includegraphics[width=0.19\textwidth]{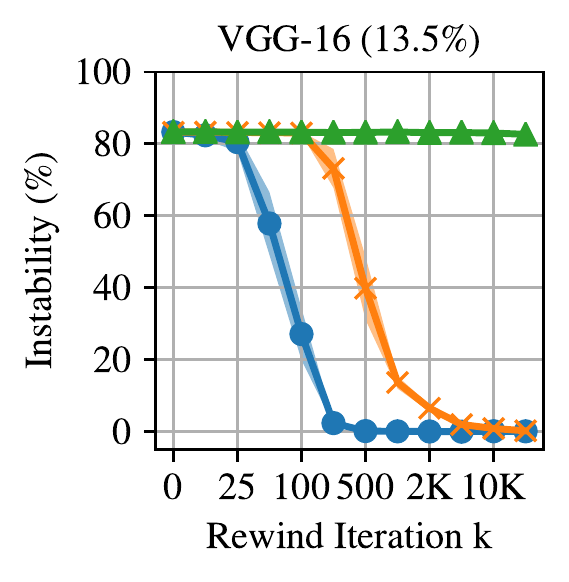}};
\node at (0.8, -0.19) {\includegraphics[width=0.19\textwidth]{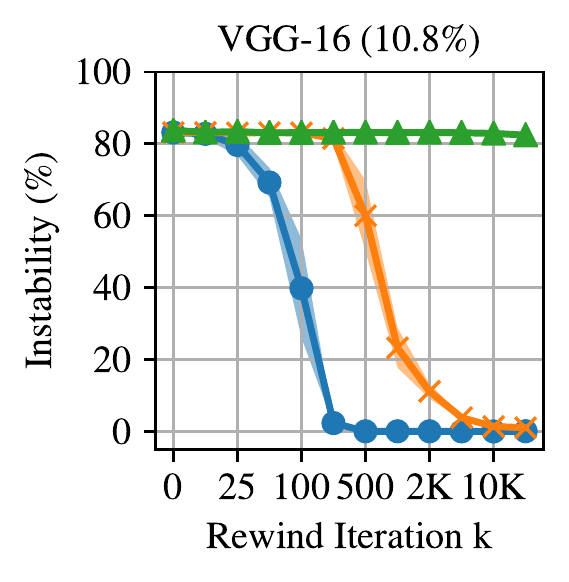}};

\node at (0.0, -0.38) {\includegraphics[width=0.19\textwidth]{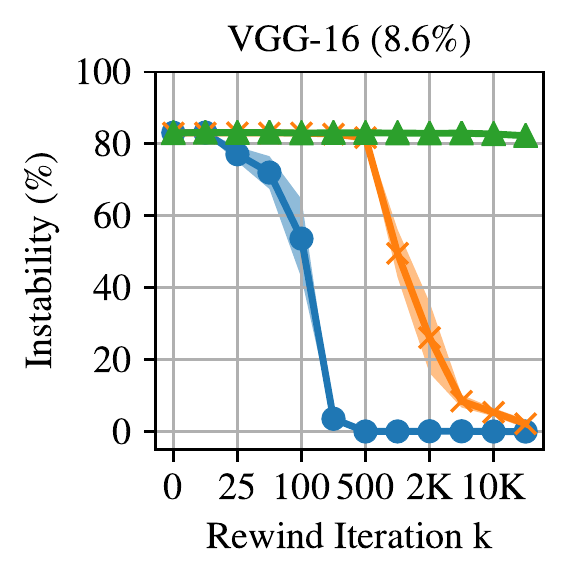}};
\node at (0.2, -0.38) {\includegraphics[width=0.19\textwidth]{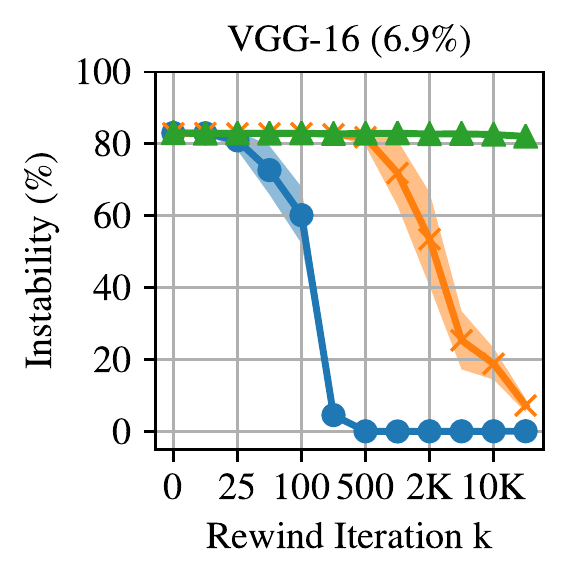}};
\node at (0.4, -0.38) {\includegraphics[width=0.19\textwidth]{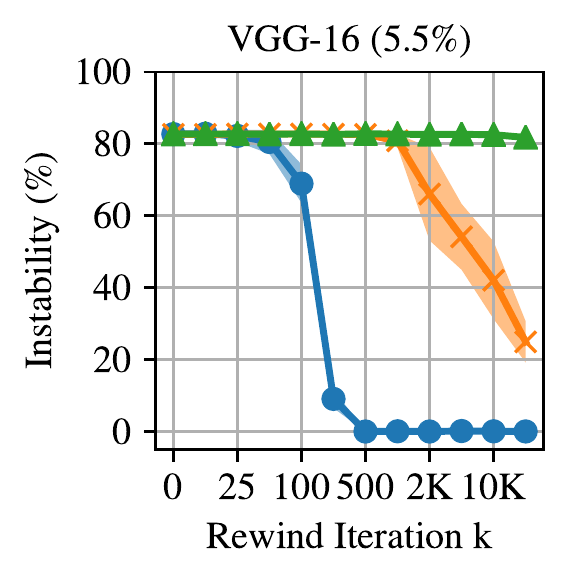}};
\node at (0.6, -0.38) {\includegraphics[width=0.19\textwidth]{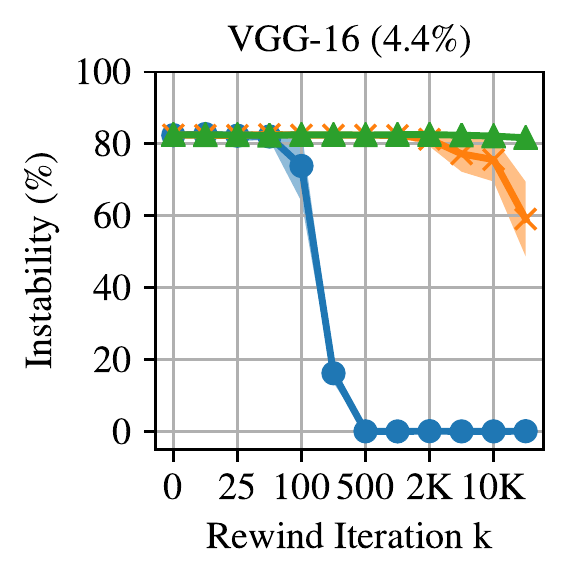}};
\node at (0.8, -0.38) {\includegraphics[width=0.19\textwidth]{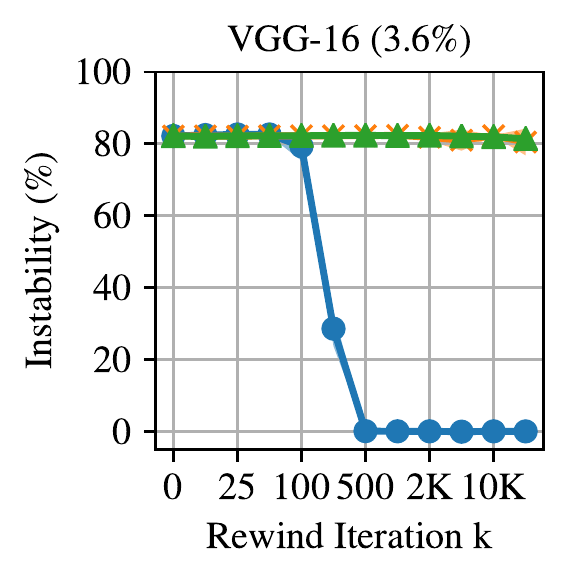}};

\node at (0.0, -0.57) {\includegraphics[width=0.19\textwidth]{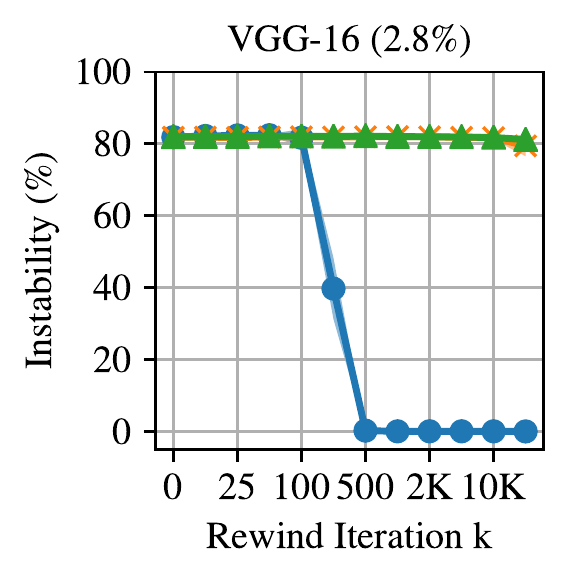}};
\node at (0.2, -0.57) {\includegraphics[width=0.19\textwidth]{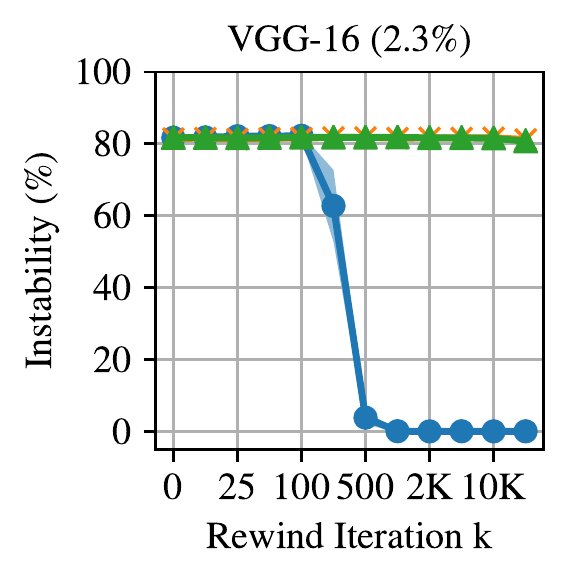}};
\node at (0.4, -0.57) {\includegraphics[width=0.19\textwidth]{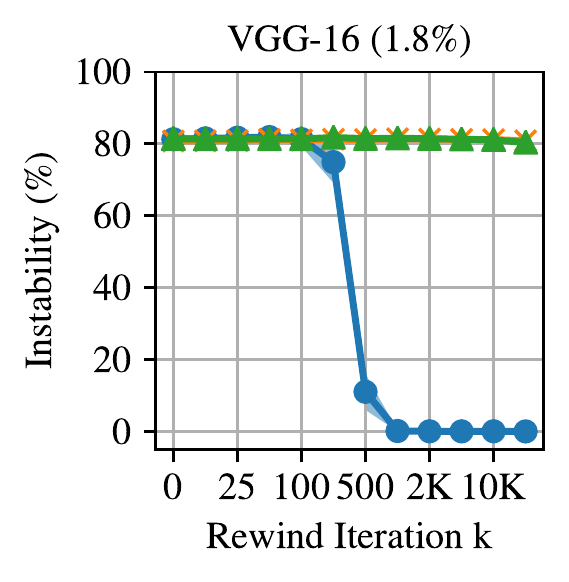}};
\node at (0.6, -0.57) {\includegraphics[width=0.19\textwidth]{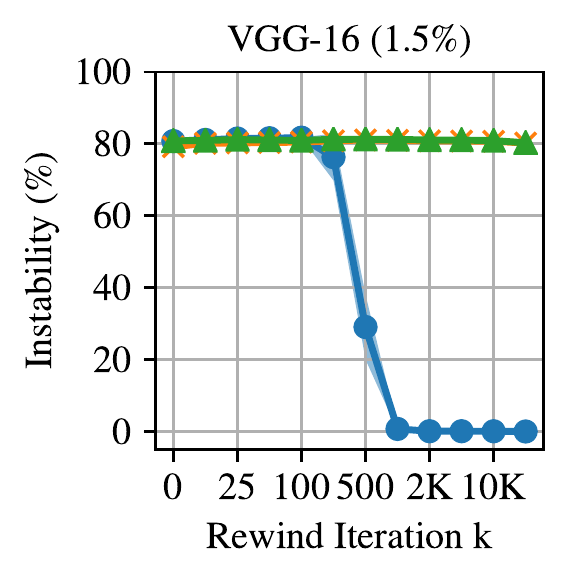}};
\node at (0.8, -0.57) {\includegraphics[width=0.19\textwidth]{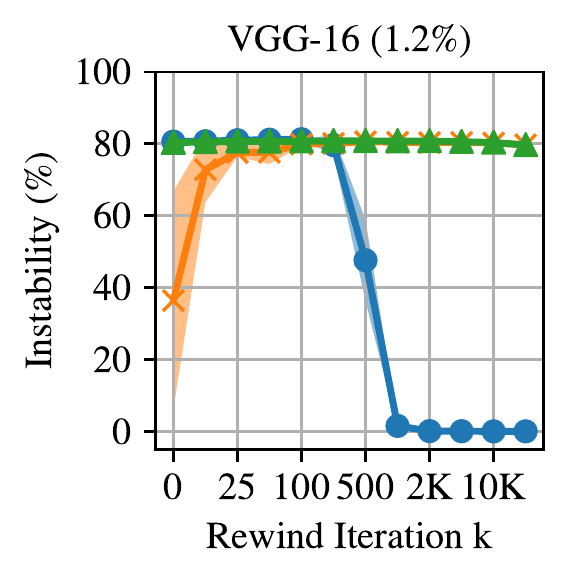}};

\node at (0.0, -0.76) {\includegraphics[width=0.19\textwidth]{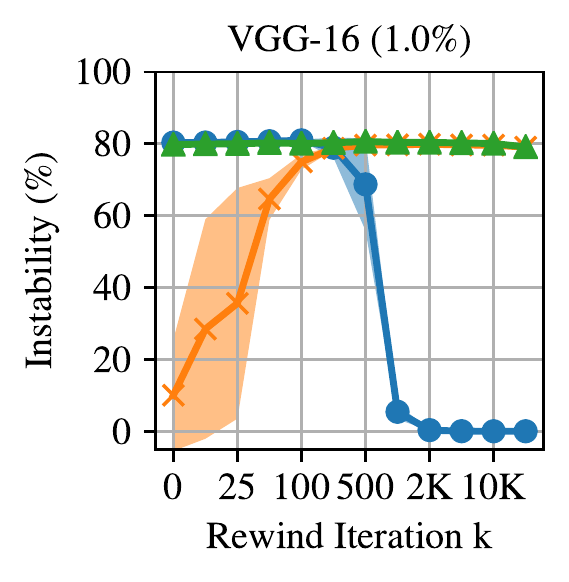}};
\node at (0.2, -0.76) {\includegraphics[width=0.19\textwidth]{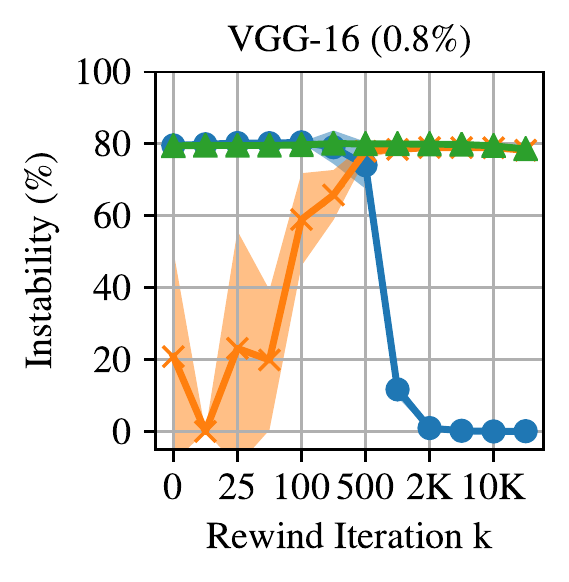}};
\node at (0.4, -0.76) {\includegraphics[width=0.19\textwidth]{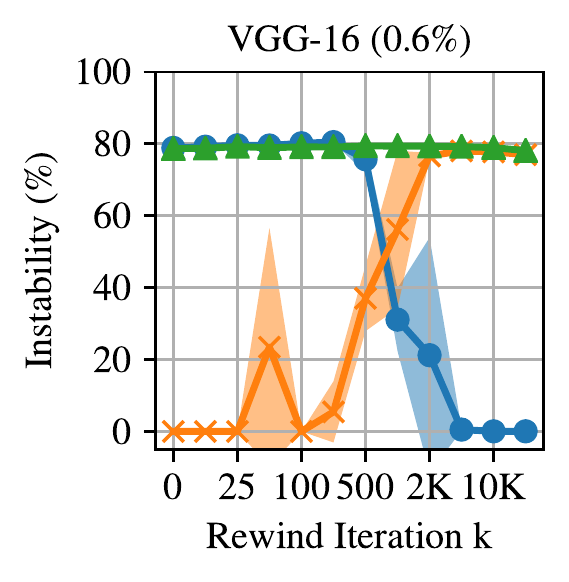}};
\end{tikzpicture}
\centering
\vspace{-1em}%
\includegraphics[width=0.6\textwidth]{figures/sparse-stability-alllevels/resnet20-level1-dataorder-legend}
\vspace{-0.5em}
\caption{The instability of subnetworks of VGG-16 created using the state of the full network at iteration $k$ and trained on different data orders from there. Each line is the mean and standard deviation across three initializations and three data orders (nine samples total) Percents are percents of weights remaining.}
\label{fig:vgg16-across-sparsities-stability}
\end{figure*}

\begin{figure*}
\begin{tikzpicture}[x=\textwidth,y=\textwidth, every node/.style = {anchor=north west}]
\node at (0.0, 0) {\includegraphics[width=0.19\textwidth]{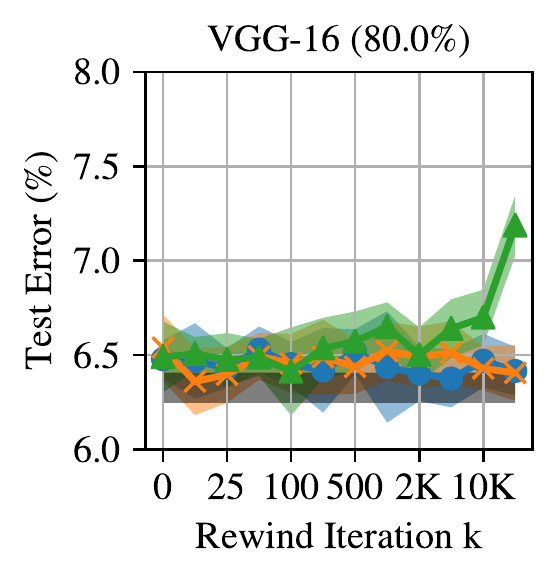}};
\node at (0.2, 0) {\includegraphics[width=0.19\textwidth]{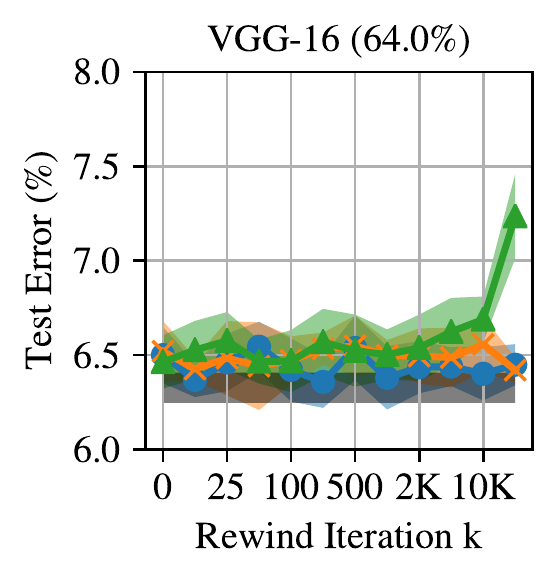}};
\node at (0.4, 0) {\includegraphics[width=0.19\textwidth]{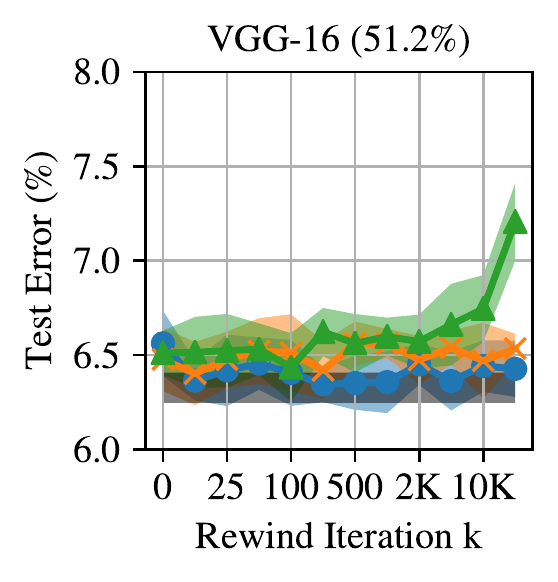}};
\node at (0.6, 0) {\includegraphics[width=0.19\textwidth]{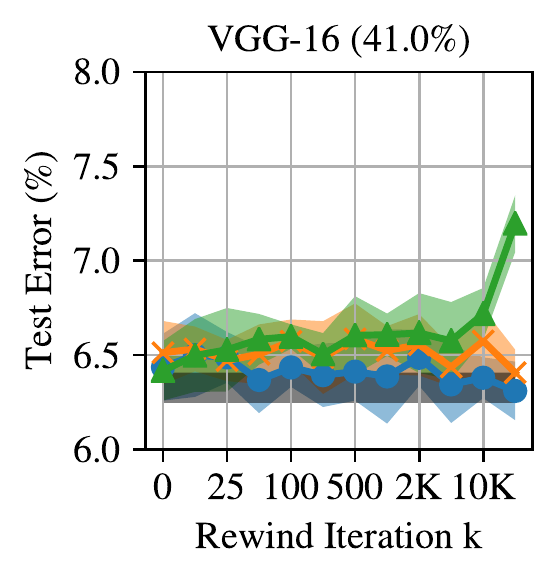}};
\node at (0.8, 0) {\includegraphics[width=0.19\textwidth]{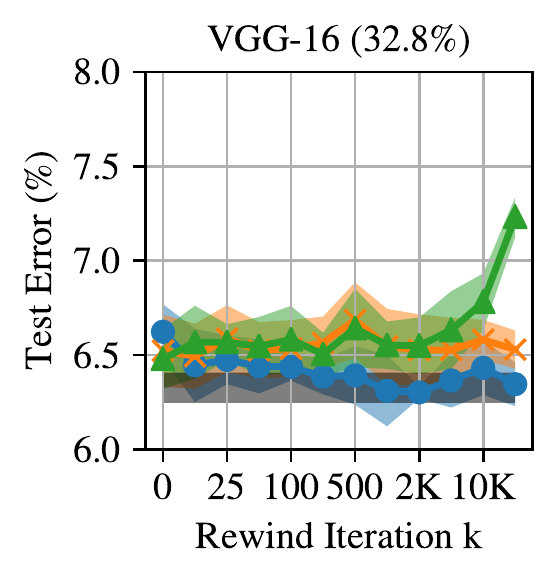}};

\node at (0.0, -0.19) {\includegraphics[width=0.19\textwidth]{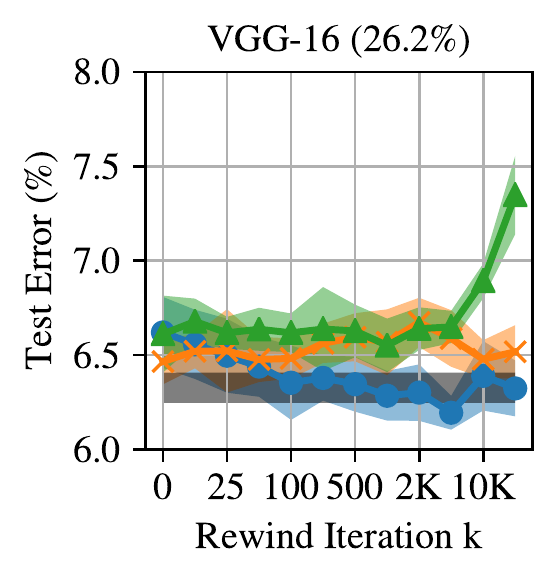}};
\node at (0.2, -0.19) {\includegraphics[width=0.19\textwidth]{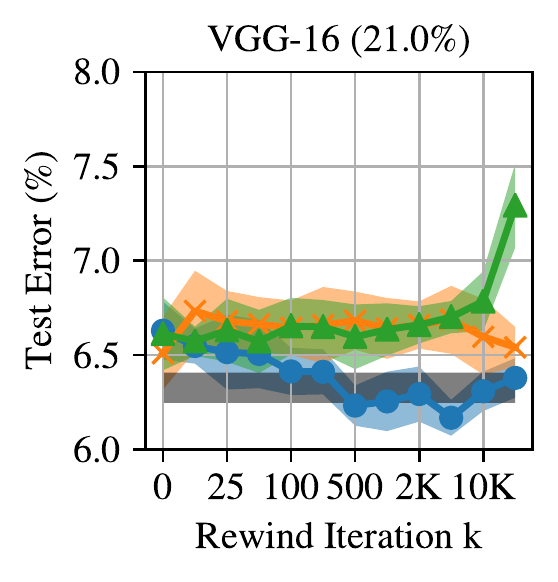}};
\node at (0.4, -0.19) {\includegraphics[width=0.19\textwidth]{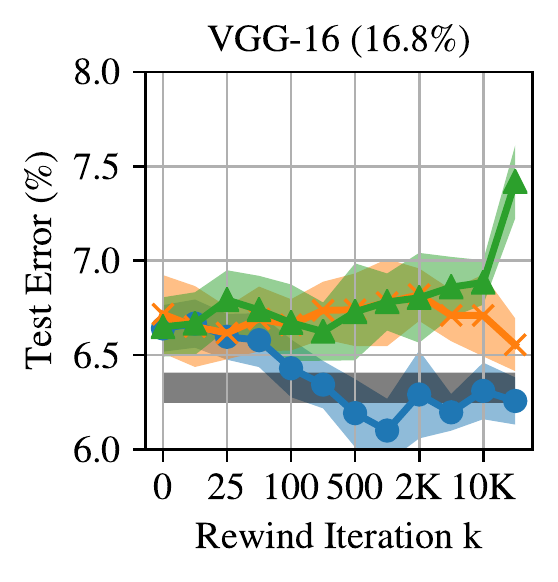}};
\node at (0.6, -0.19) {\includegraphics[width=0.19\textwidth]{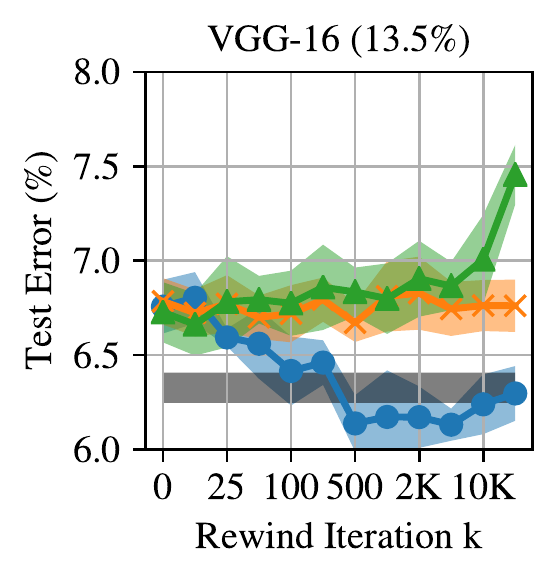}};
\node at (0.8, -0.19) {\includegraphics[width=0.19\textwidth]{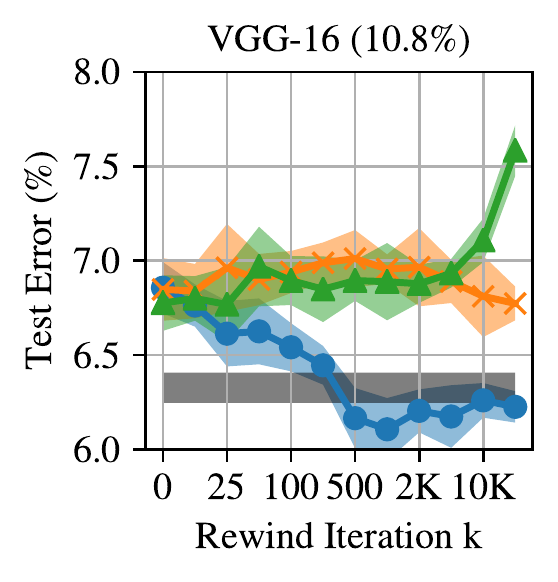}};

\node at (0.0, -0.38) {\includegraphics[width=0.19\textwidth]{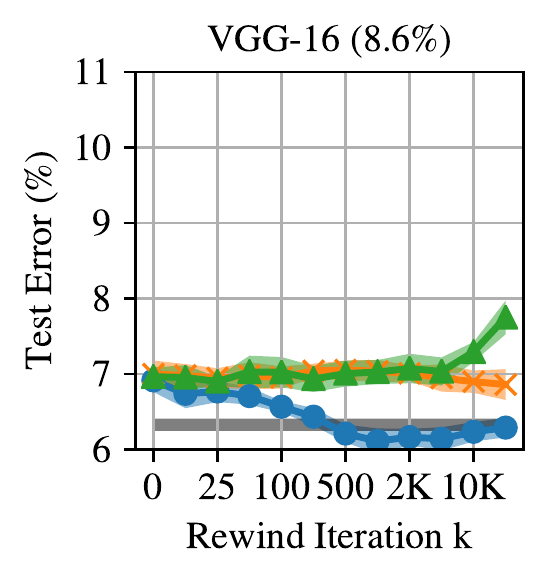}};
\node at (0.2, -0.38) {\includegraphics[width=0.19\textwidth]{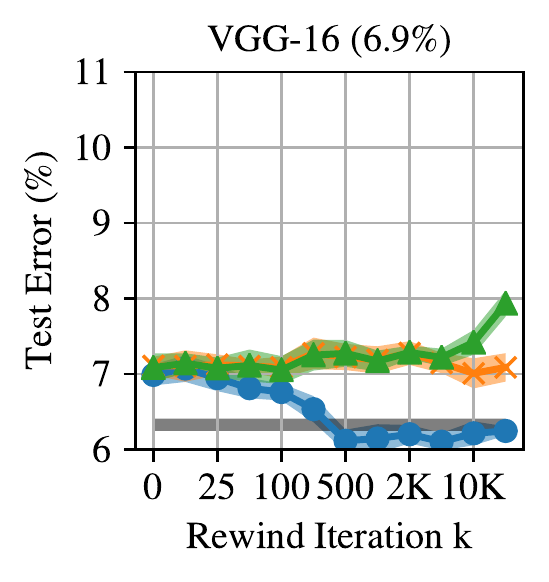}};
\node at (0.4, -0.38) {\includegraphics[width=0.19\textwidth]{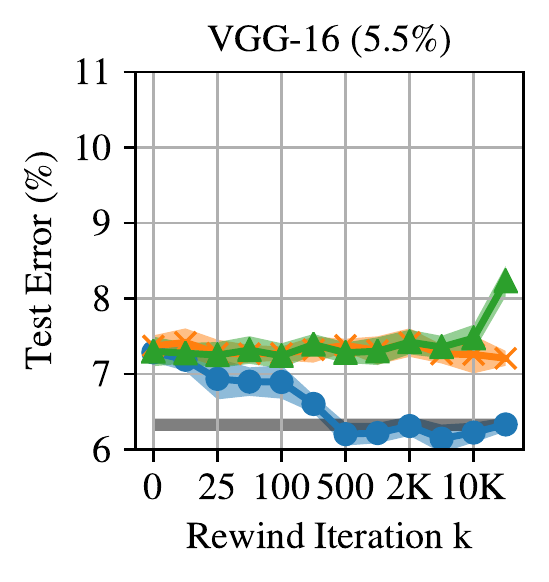}};
\node at (0.6, -0.38) {\includegraphics[width=0.19\textwidth]{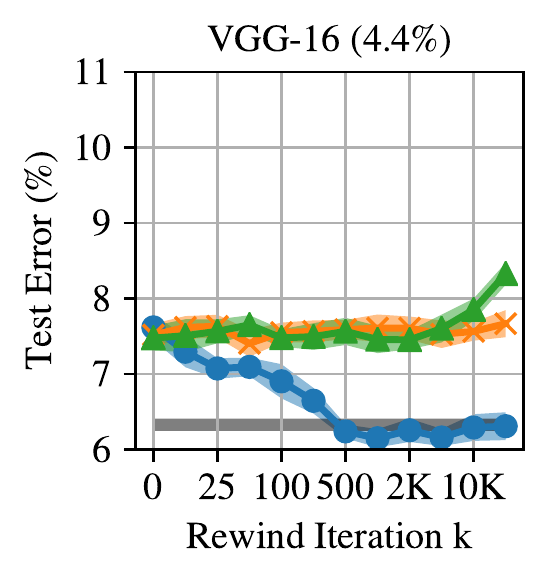}};
\node at (0.8, -0.38) {\includegraphics[width=0.19\textwidth]{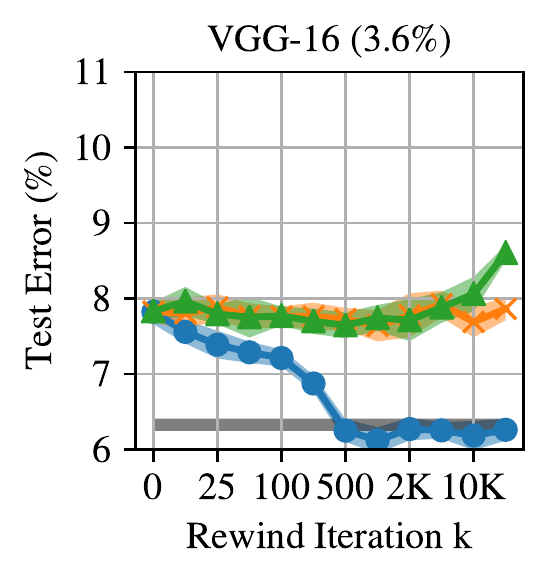}};

\node at (0.0, -0.57) {\includegraphics[width=0.19\textwidth]{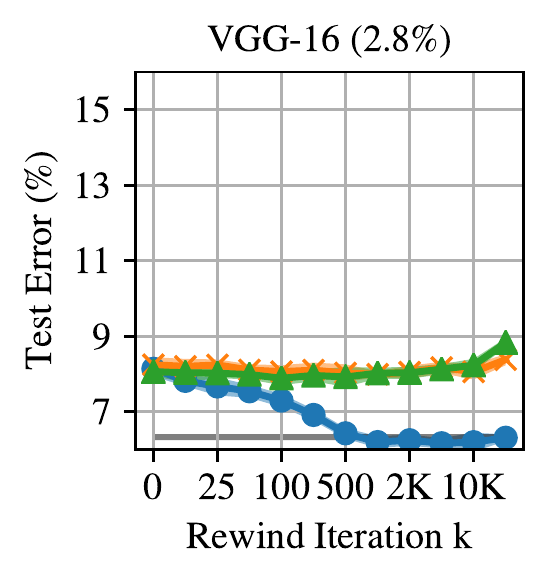}};
\node at (0.2, -0.57) {\includegraphics[width=0.19\textwidth]{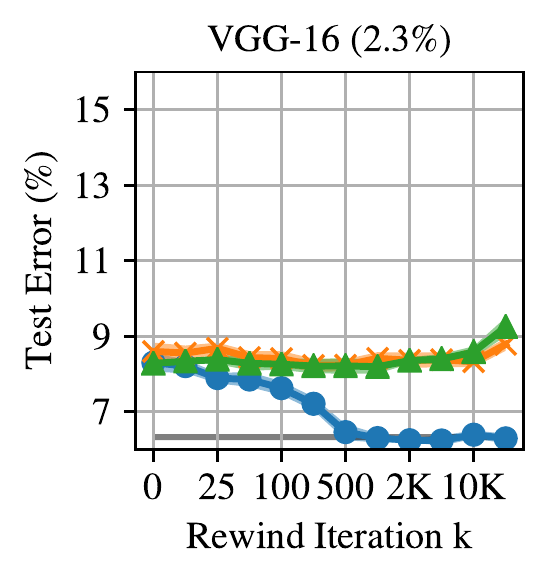}};
\node at (0.4, -0.57) {\includegraphics[width=0.19\textwidth]{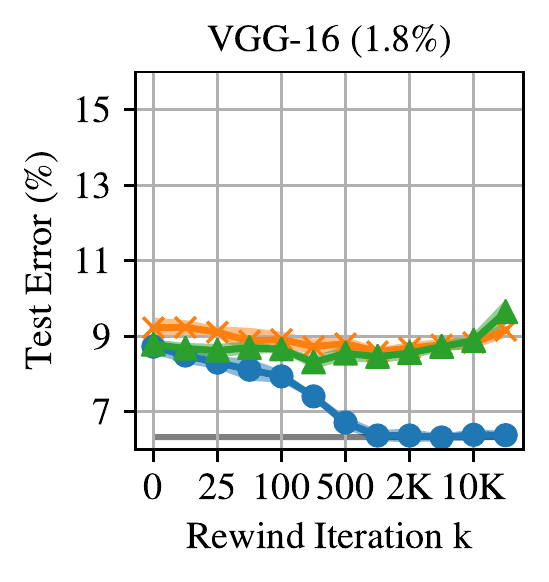}};
\node at (0.6, -0.57) {\includegraphics[width=0.19\textwidth]{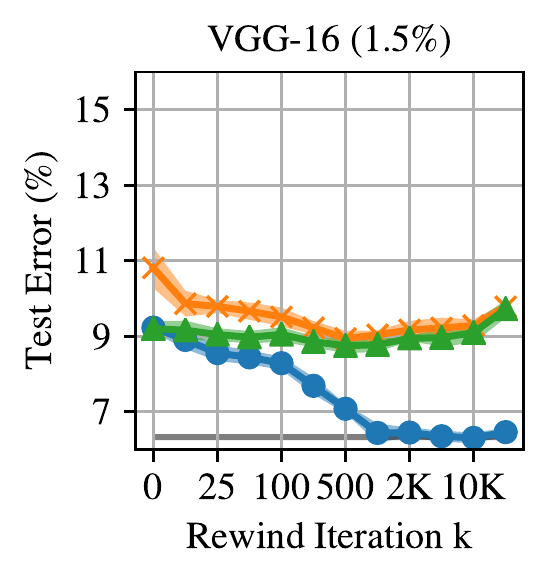}};
\node at (0.8, -0.57) {\includegraphics[width=0.19\textwidth]{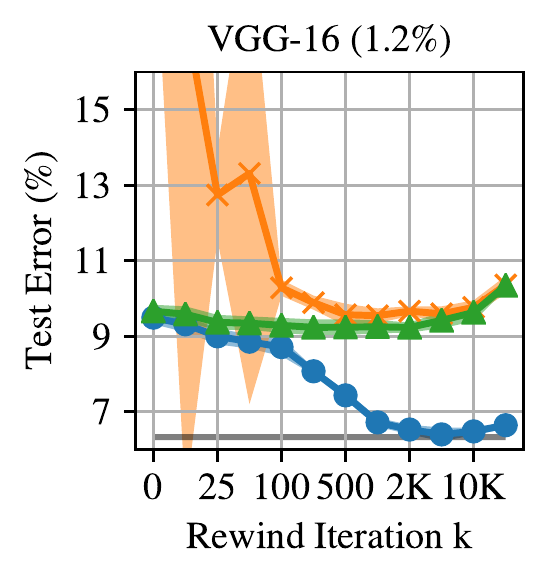}};

\node at (0.0, -0.76) {\includegraphics[width=0.19\textwidth]{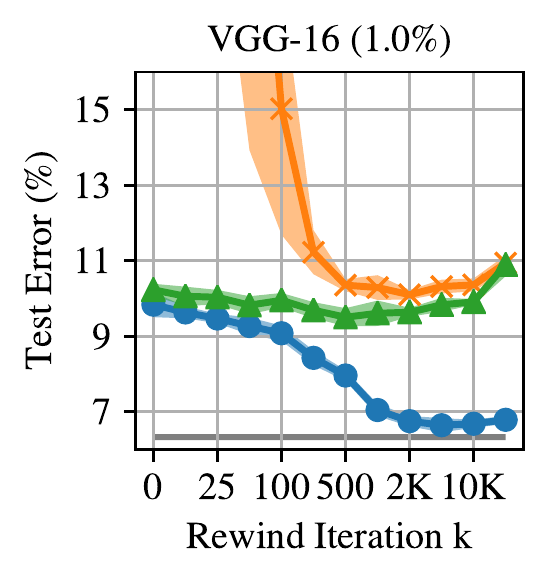}};
\node at (0.2, -0.76) {\includegraphics[width=0.19\textwidth]{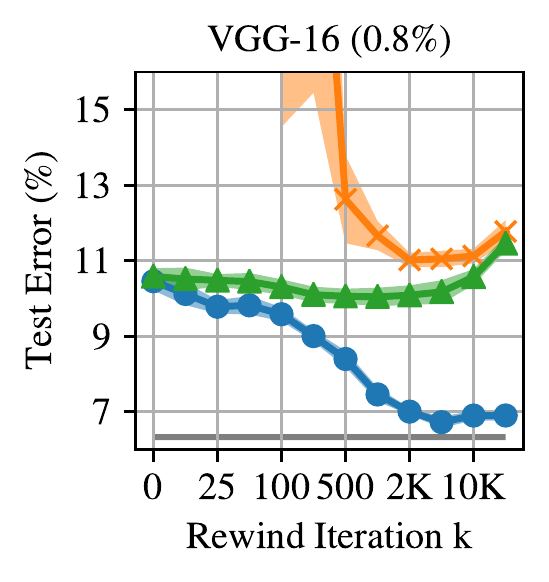}};
\node at (0.4, -0.76) {\includegraphics[width=0.19\textwidth]{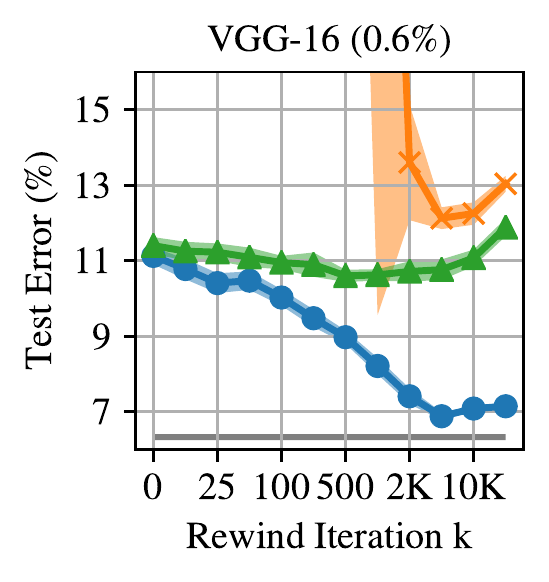}};

\end{tikzpicture}
\centering
\vspace{-1em}%
\includegraphics[width=0.6\textwidth]{figures/sparse-error-alllevels/resnet20-level1-dataorder-legend}
\vspace{-0.5em}
\caption{The test error of subnetworks of VGG-16 created using the state of the full network at iteration $k$ and trained on different data orders from there. Each line is the mean and standard deviation across three initializations. Gray lines are the accuracies of the full networks to one standard deviation. Percents are percents of weights remaining.}
\label{fig:vgg16-across-sparsities-error}
\end{figure*}

In Figure \ref{fig:sparse-instability-later} in Section \ref{sec:sparsity-exps}, we show the effect of rewinding iteration on instability and test error for sparse subnetworks.
We specifically focus on the most extreme level of sparsity for which IMP at any rewinding iteration is matching (as selected in Appendix \ref{app:sparsity}).
In this appendix, we present the relationship between rewinding iteration and instability/test error for all levels of sparsity for standard ResNet-20 (Figures \ref{fig:resnet20-across-sparsities-stability} and \ref{fig:resnet20-across-sparsities-error}) and VGG-16 (Figures \ref{fig:vgg16-across-sparsities-stability} and \ref{fig:vgg16-across-sparsities-error}) on CIFAR-10.
Section \ref{sec:stability-across-sparsities} and Figure \ref{fig:sparsity-stability-summary} summarize this data, so we defer analysis of this data to that section.

This data begins with 80\% of weights remaining and includes sparsities attained by repeatedly pruning 20\% of weights (e.g., 64\% of weights remaining, 51\% of weights remaining, etc.).
We include these levels in particular because we use IMP to prune 20\% of weights per iteration, meaning we have sparse IMP subnetworks for each of these levels.
We include data for every sparsity level displayed in Appendix \ref{app:sparsity}, including those beyond the extreme sparsities we study in Section \ref{sec:sparsity-exps}.

We only collected this data for standard ResNet-20 and VGG-16 on CIFAR-10.
We determined that it was more valuable to spend our limited computational resources on these networks (whose instability and accuracy are sensitive to rewinding at the extreme sparsity level) than for the \emph{low} and \emph{warmup} variants (which are consistently stable and matching at the extreme sparsity level).
We did not have the computational resources to compute this data on the ImageNet networks for all sparsities.

\section{Full Linear Interpolation Data}
\label{app:hills}

\begin{figure*}
\centering
\includegraphics[width=0.19\textwidth]{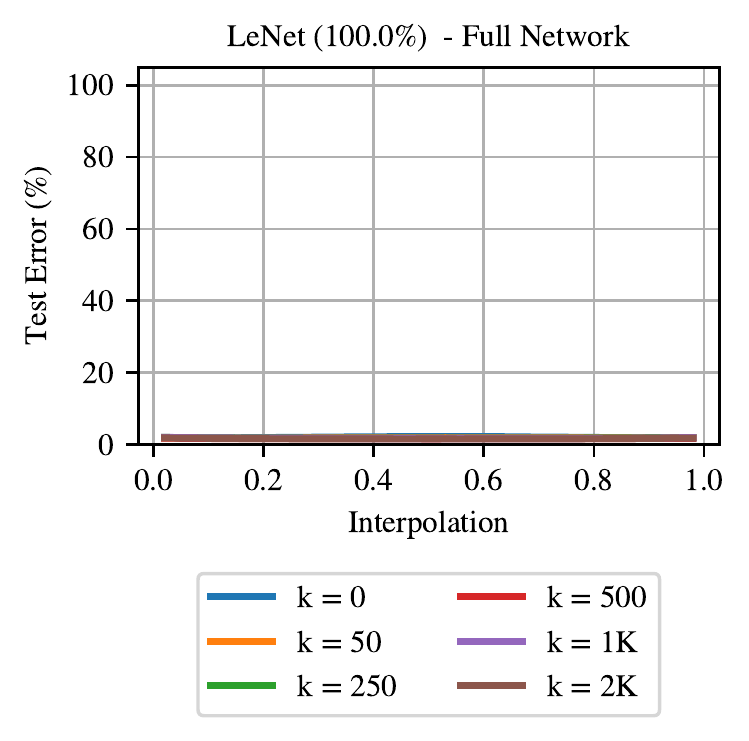}
\includegraphics[width=0.19\textwidth]{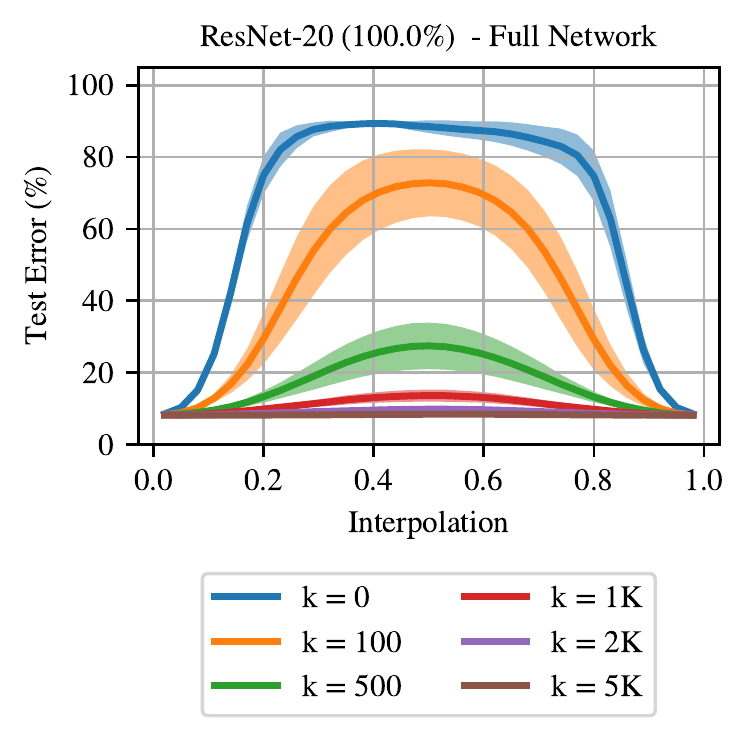}
\includegraphics[width=0.19\textwidth]{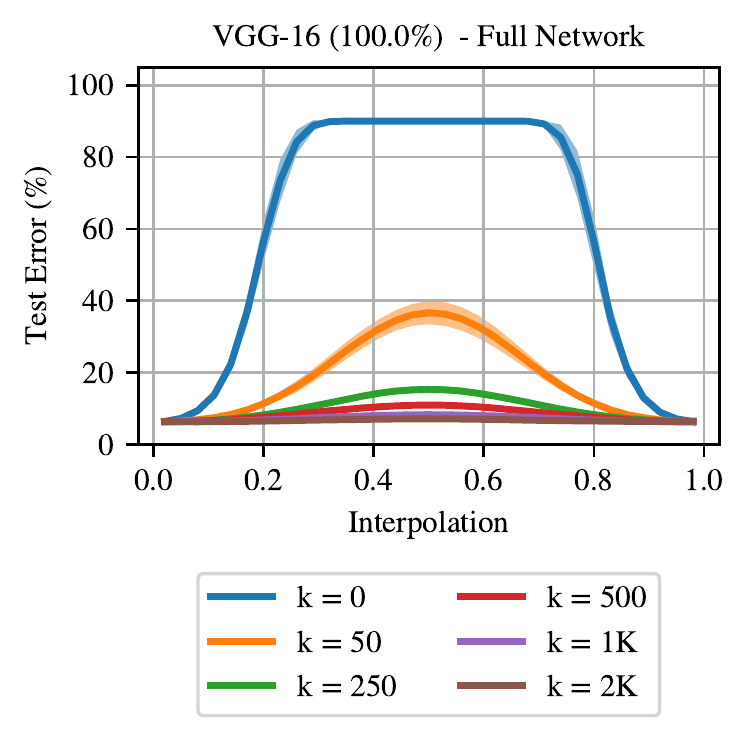}
\includegraphics[width=0.19\textwidth]{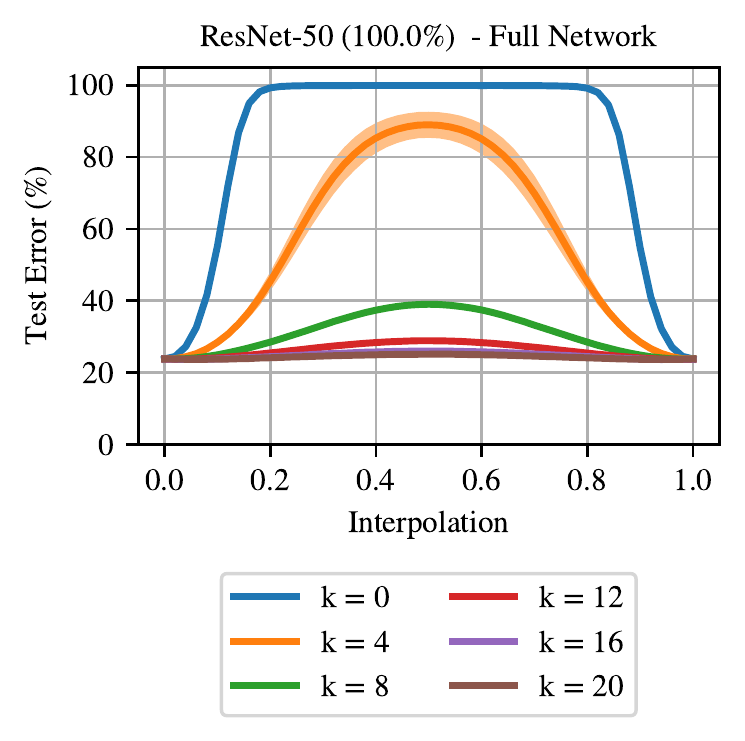}
\includegraphics[width=0.19\textwidth]{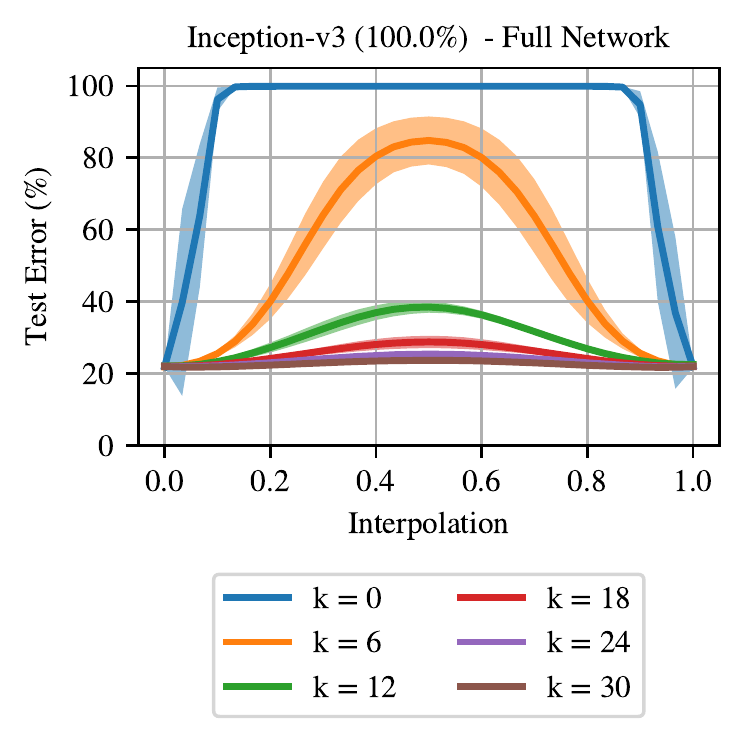}

\includegraphics[width=0.19\textwidth]{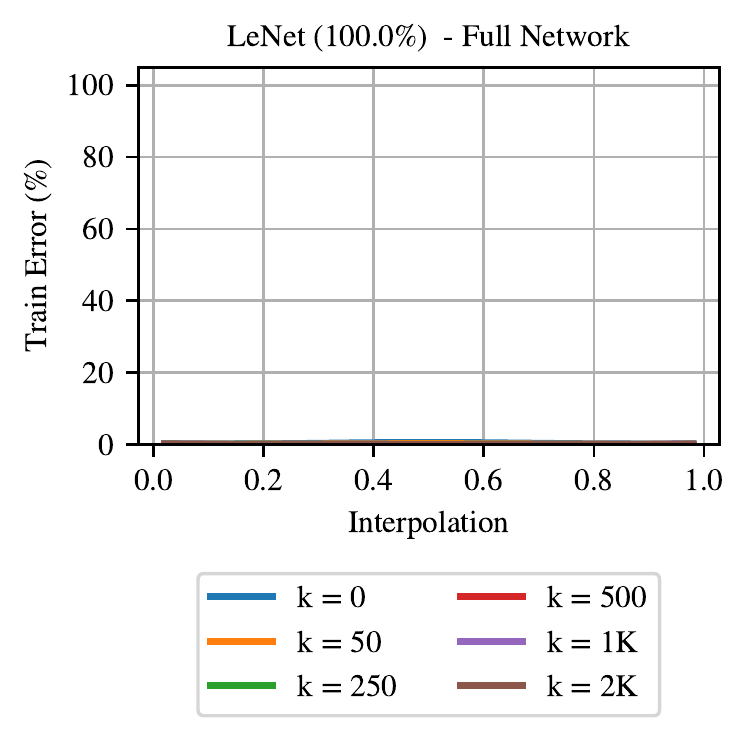}
\includegraphics[width=0.19\textwidth]{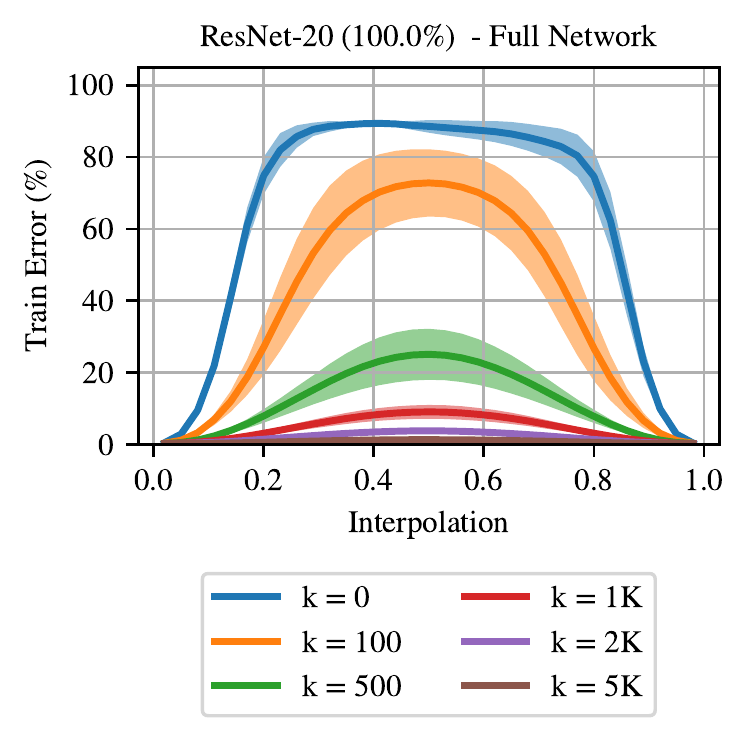}
\includegraphics[width=0.19\textwidth]{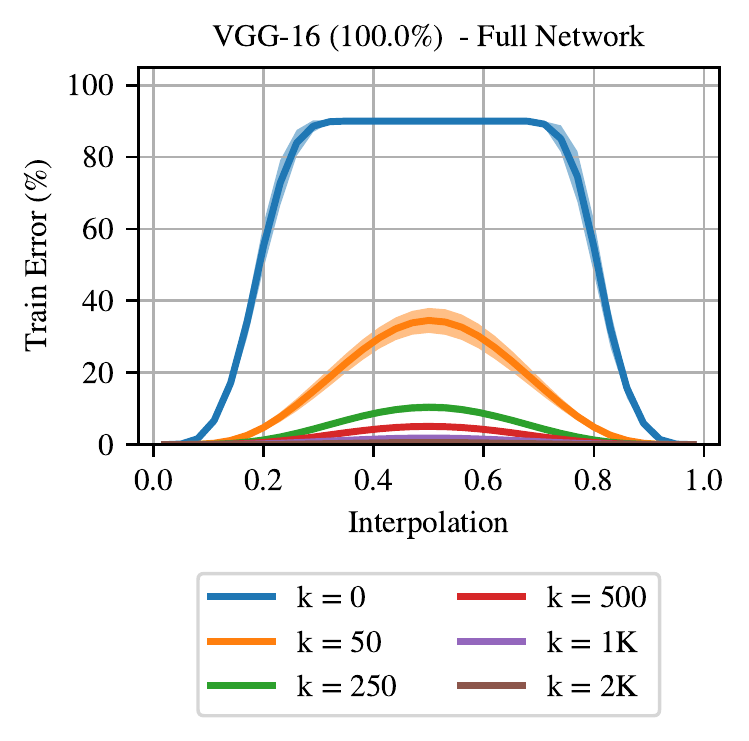}
\includegraphics[width=0.19\textwidth]{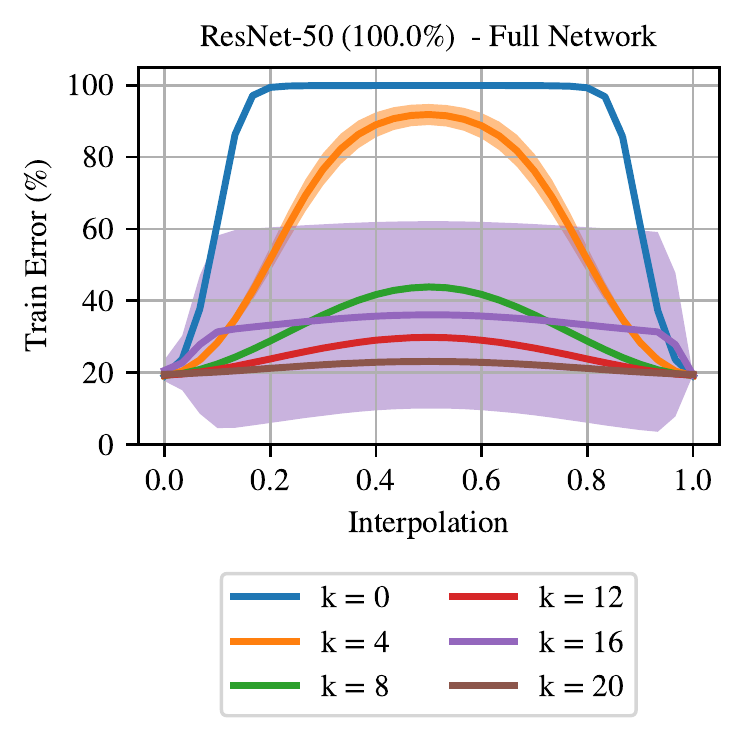}
\includegraphics[width=0.19\textwidth]{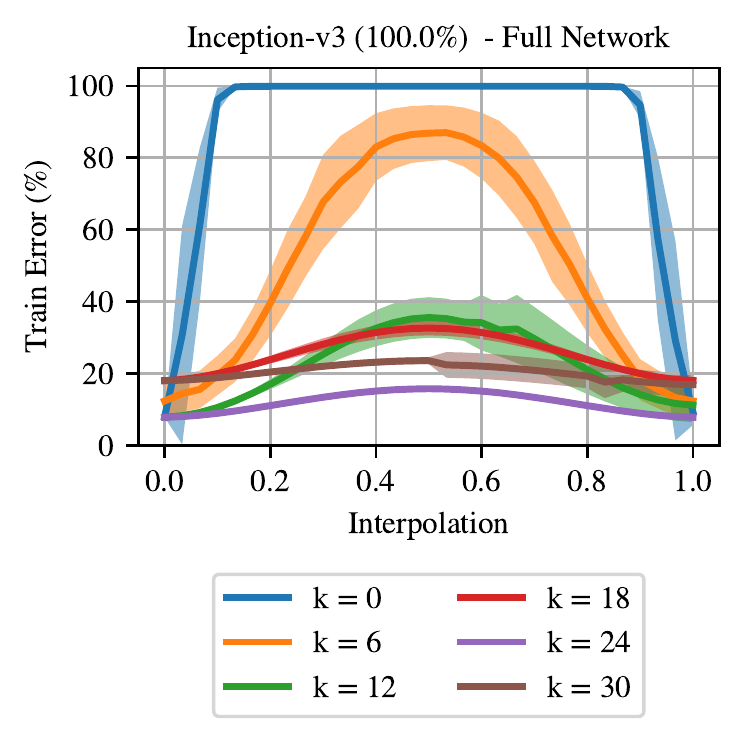}
\caption{The error when linearly interpolating between the minima found by randomly initializing a network, training to iteration $k$, and training two copies from there to completion using different data orders. Each line is the mean and standard deviation across three initializations and three data orders (nine samples in total). The errors of the trained networks are at interpolation = 0.0 and 1.0.}
\label{fig:hills-full}
\end{figure*}

\begin{figure*}
\centering
\includegraphics[width=0.19\textwidth]{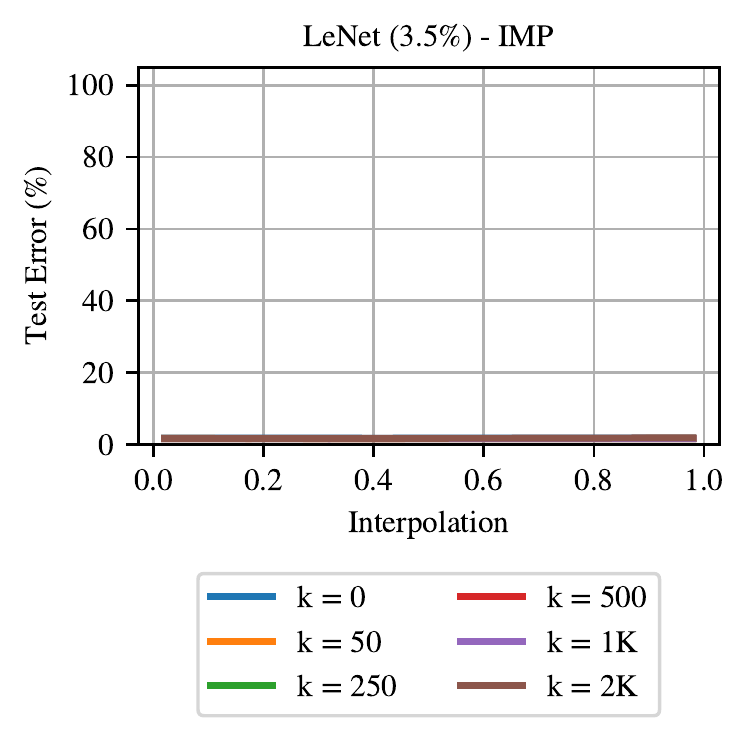}
\includegraphics[width=0.19\textwidth]{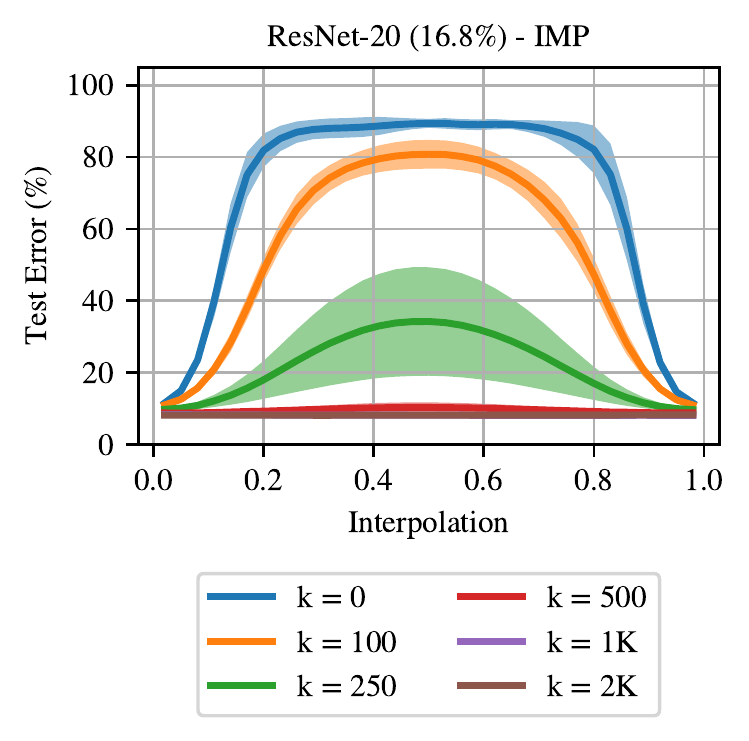}
\includegraphics[width=0.19\textwidth]{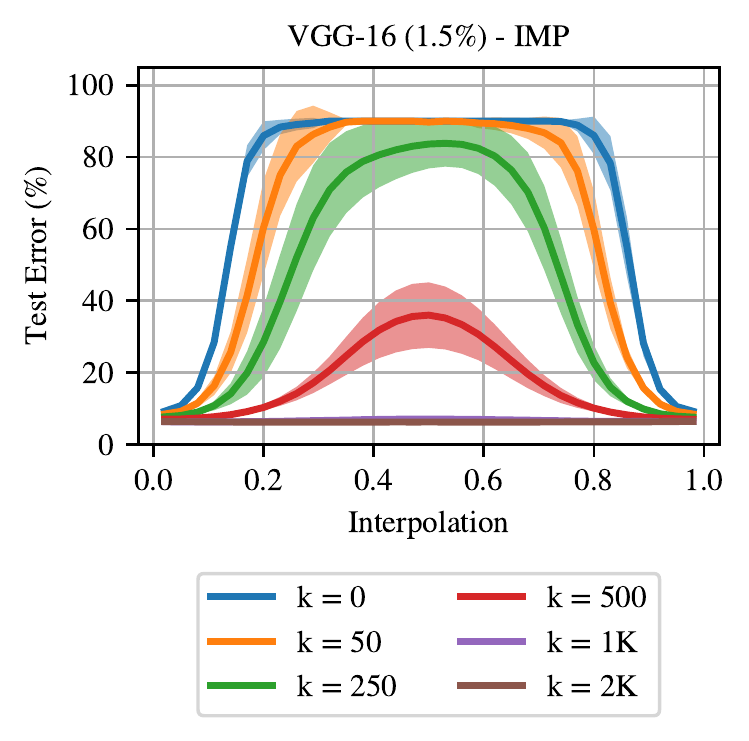}
\includegraphics[width=0.19\textwidth]{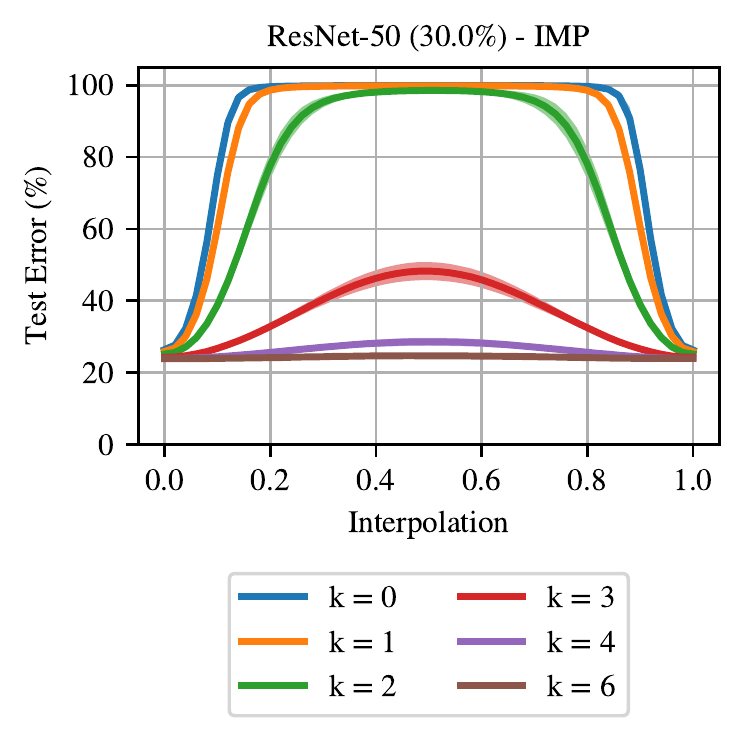}
\includegraphics[width=0.19\textwidth]{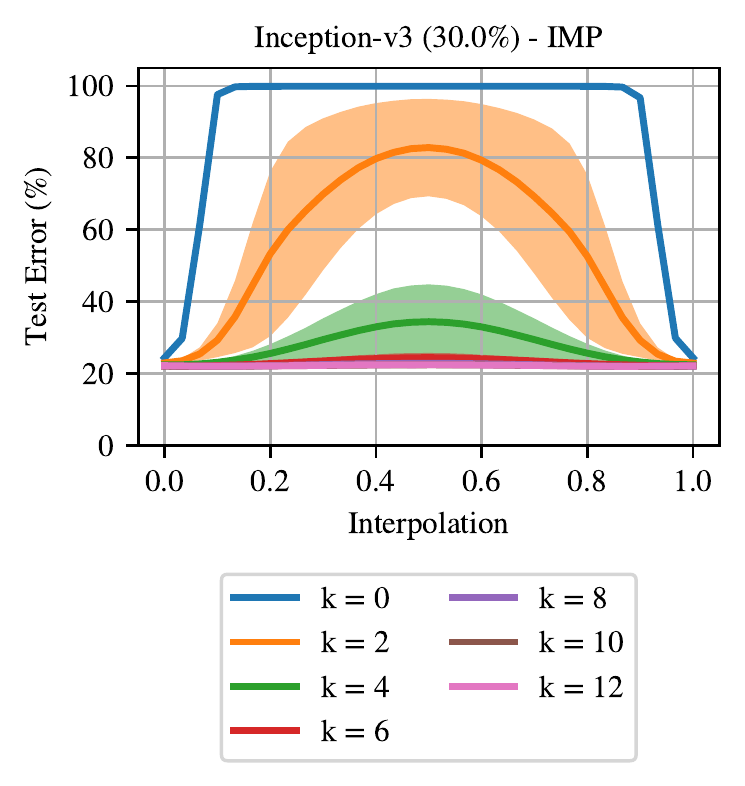}

\includegraphics[width=0.19\textwidth]{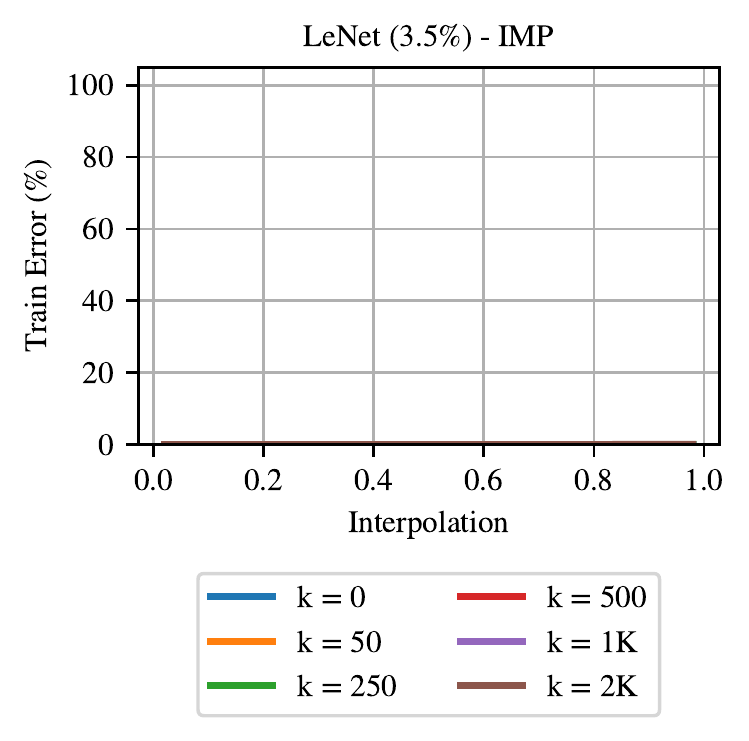}
\includegraphics[width=0.19\textwidth]{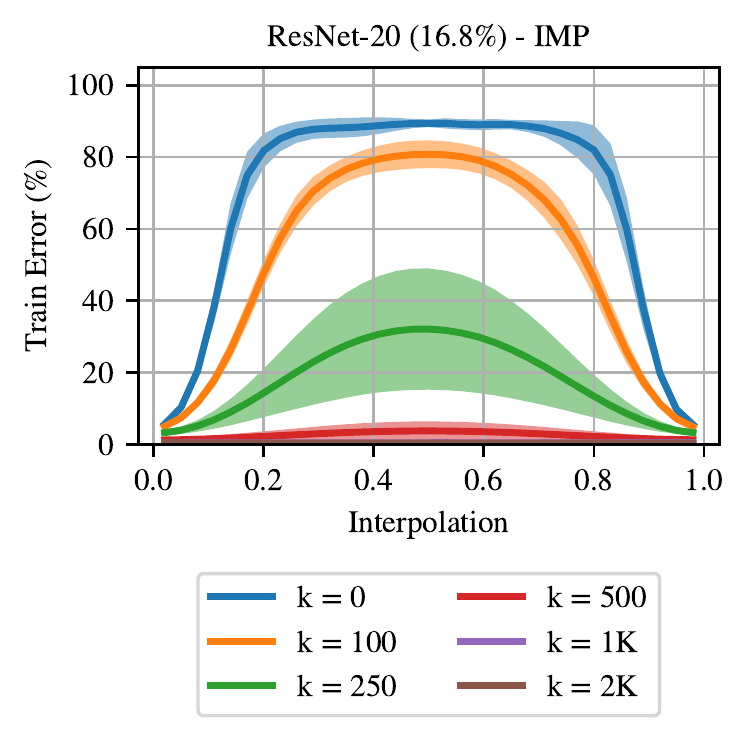}
\includegraphics[width=0.19\textwidth]{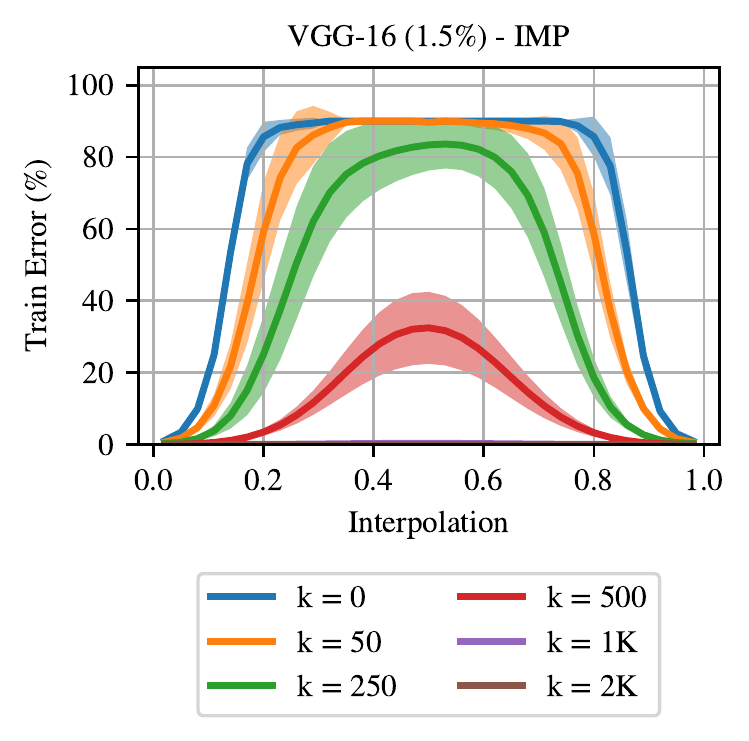}
\begin{minipage}{0.19\textwidth}~\end{minipage}
\begin{minipage}{0.19\textwidth}~\end{minipage}
\caption{The error when linearly interpolating between the minima found by randomly initializing a network, training to iteration $k$, pruning according to IMP, and training two copies from there to completion using different data orders. Each line is the mean and standard deviation across three initializations and three data orders (nine samples in total). The errors of the trained networks are at interpolation = 0.0 and 1.0. We did not interpolate using the training set for the ImageNet networks due to computational limitations.}
\label{fig:hills-imp}
\end{figure*}

\begin{figure*}
\centering
\includegraphics[width=0.19\textwidth]{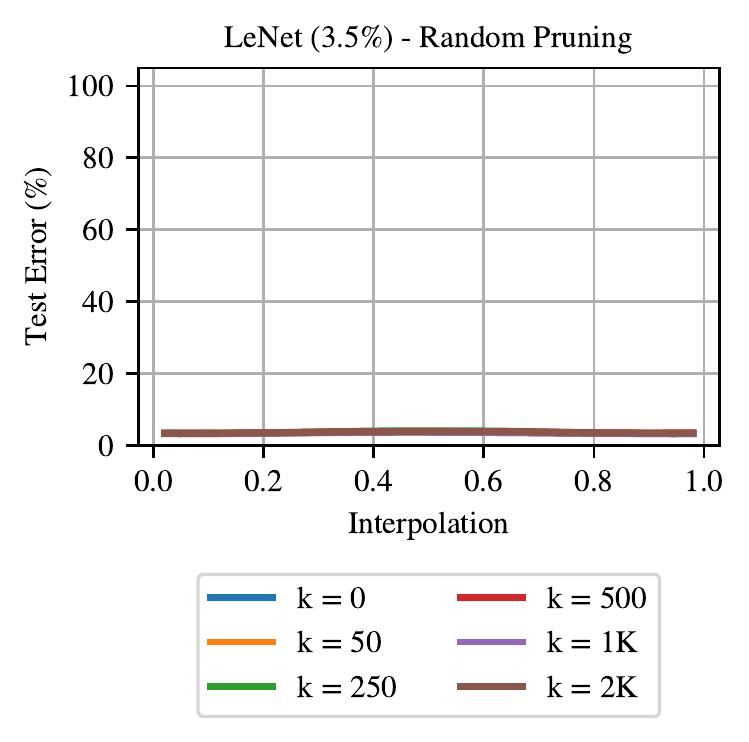}
\includegraphics[width=0.19\textwidth]{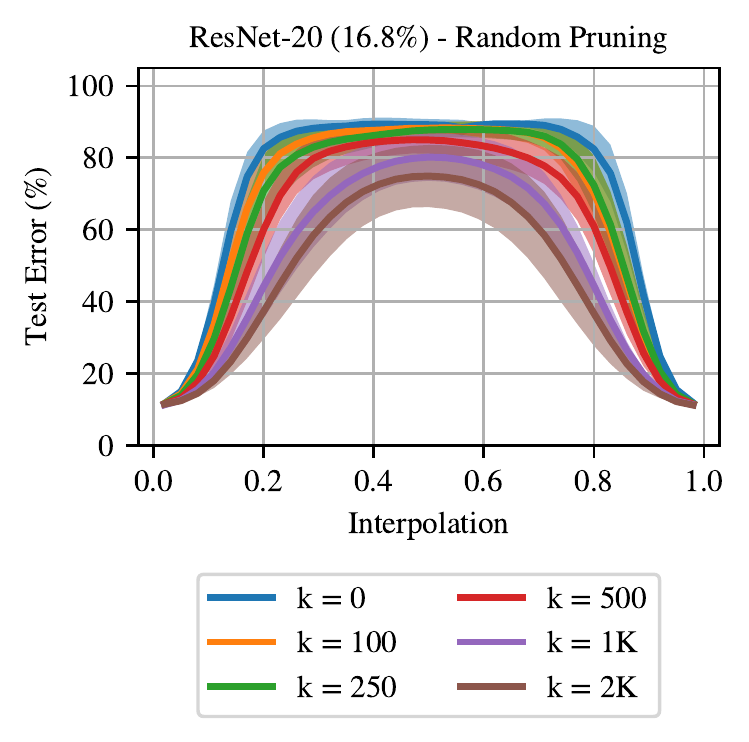}
\includegraphics[width=0.19\textwidth]{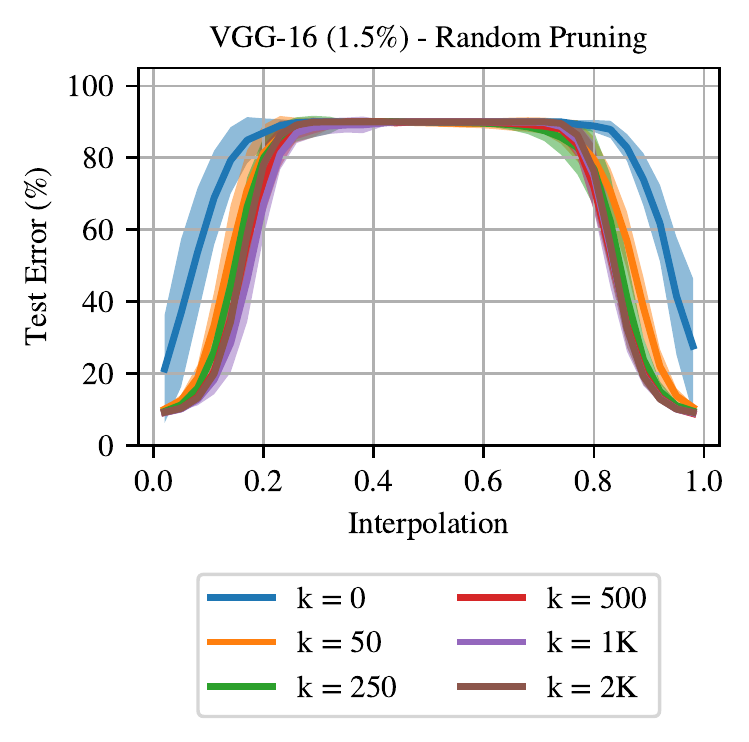}
\includegraphics[width=0.19\textwidth]{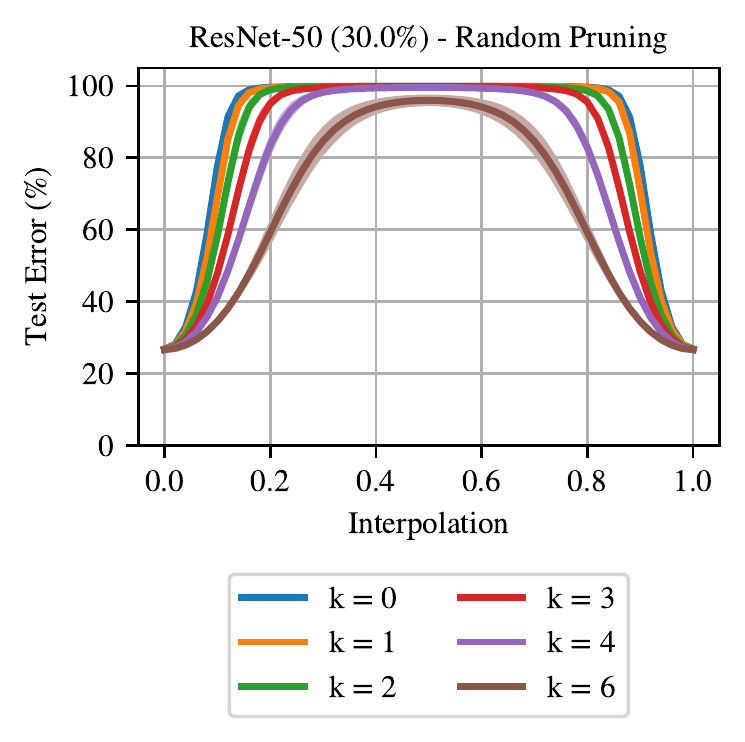}
\includegraphics[width=0.19\textwidth]{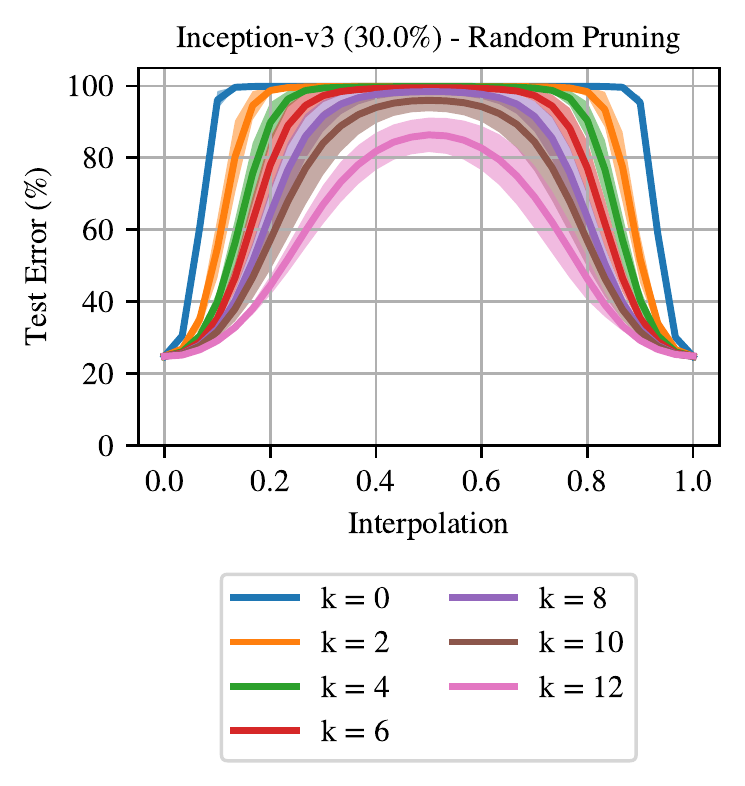}

\includegraphics[width=0.19\textwidth]{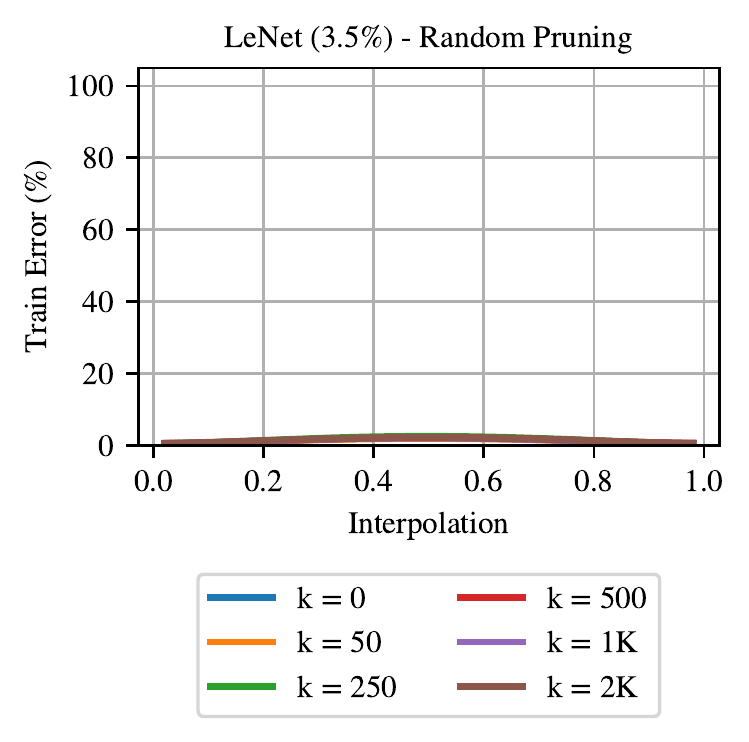}
\includegraphics[width=0.19\textwidth]{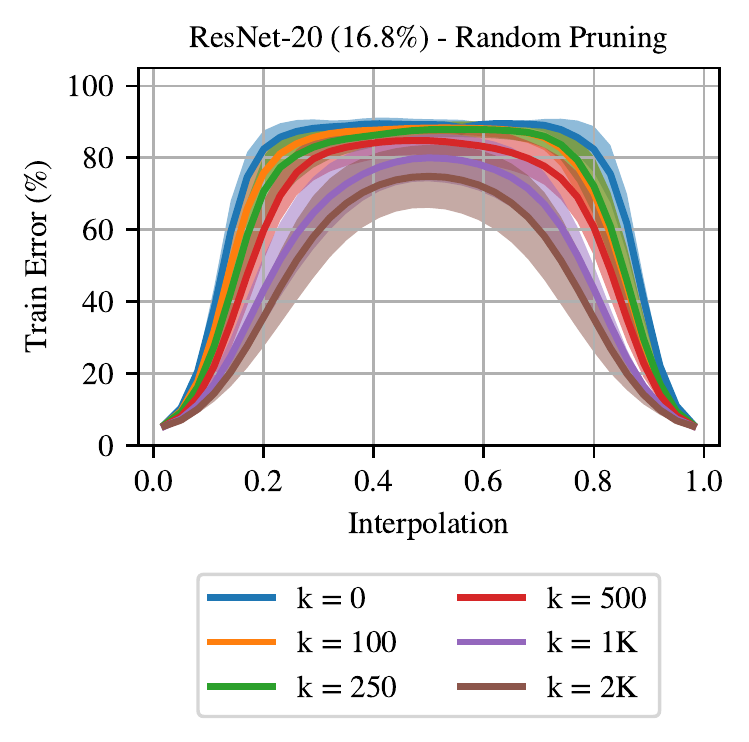}
\includegraphics[width=0.19\textwidth]{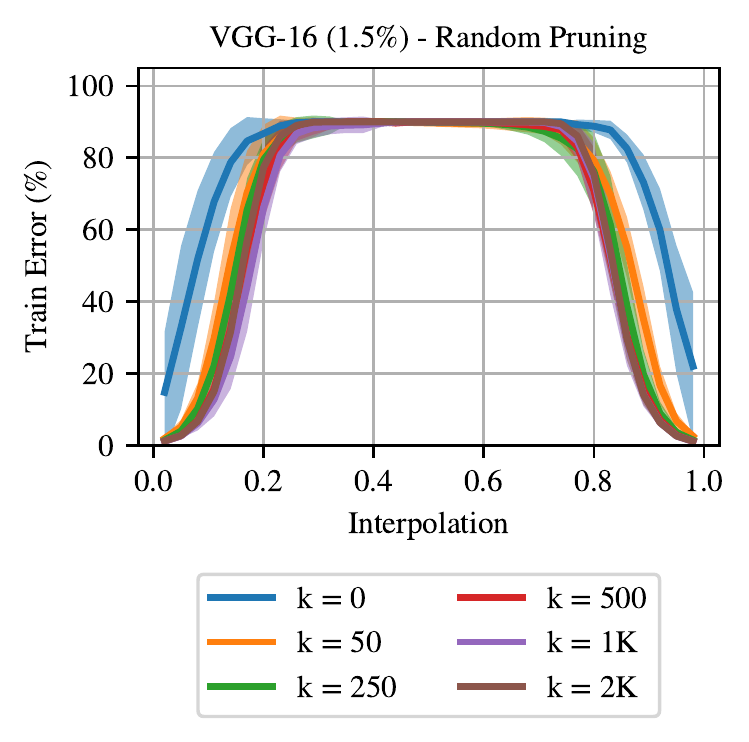}
\begin{minipage}{0.19\textwidth}~\end{minipage}
\begin{minipage}{0.19\textwidth}~\end{minipage}
\caption{The error when linearly interpolating between the minima found by randomly initializing a network, training to iteration $k$, pruning randomly in the same layerwise proportions as IMP, and training two copies from there to completion using different data orders. Each line is the mean and standard deviation across three initializations and three data orders (nine samples in total). The errors of the trained networks are at interpolation = 0.0 and 1.0. We did not interpolate using the training set for the ImageNet networks due to computational limitations.}
\label{fig:hills-rearr}
\end{figure*}

\begin{figure*}
\centering
\includegraphics[width=0.19\textwidth]{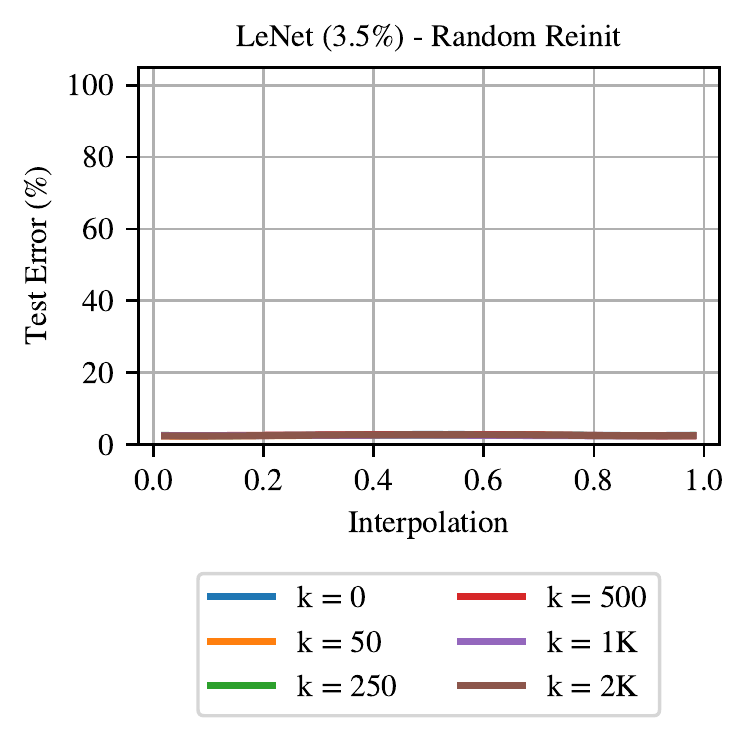}
\includegraphics[width=0.19\textwidth]{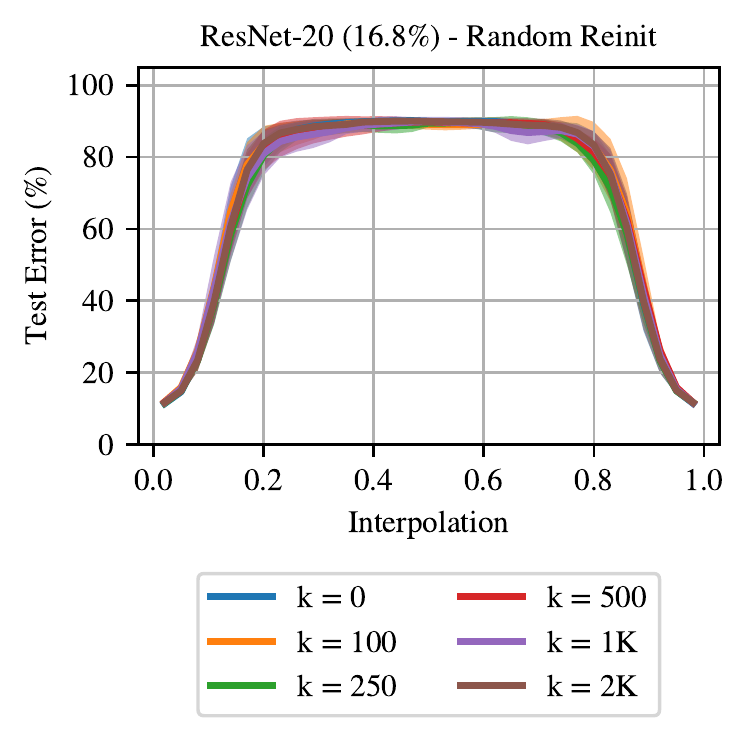}
\includegraphics[width=0.19\textwidth]{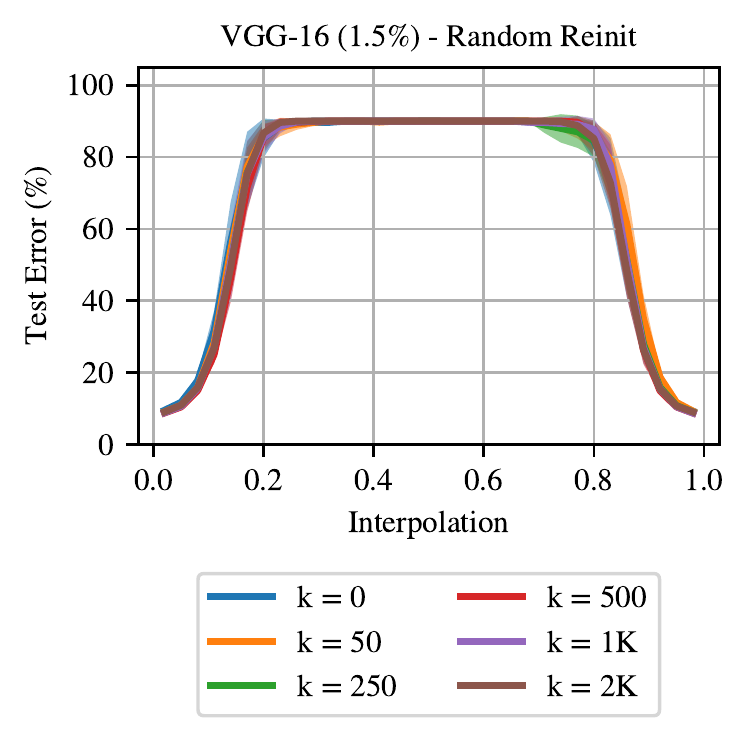}
\includegraphics[width=0.19\textwidth]{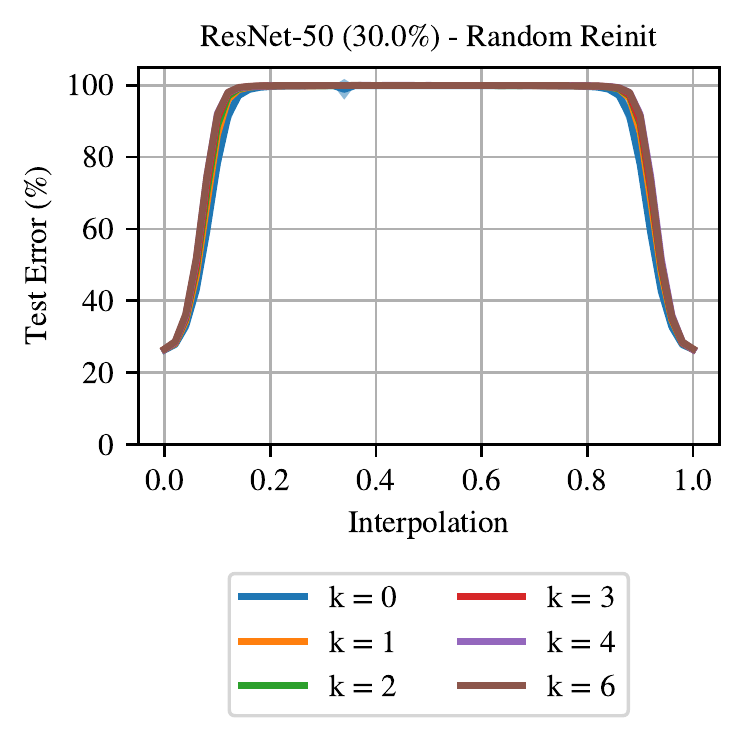}
\includegraphics[width=0.19\textwidth]{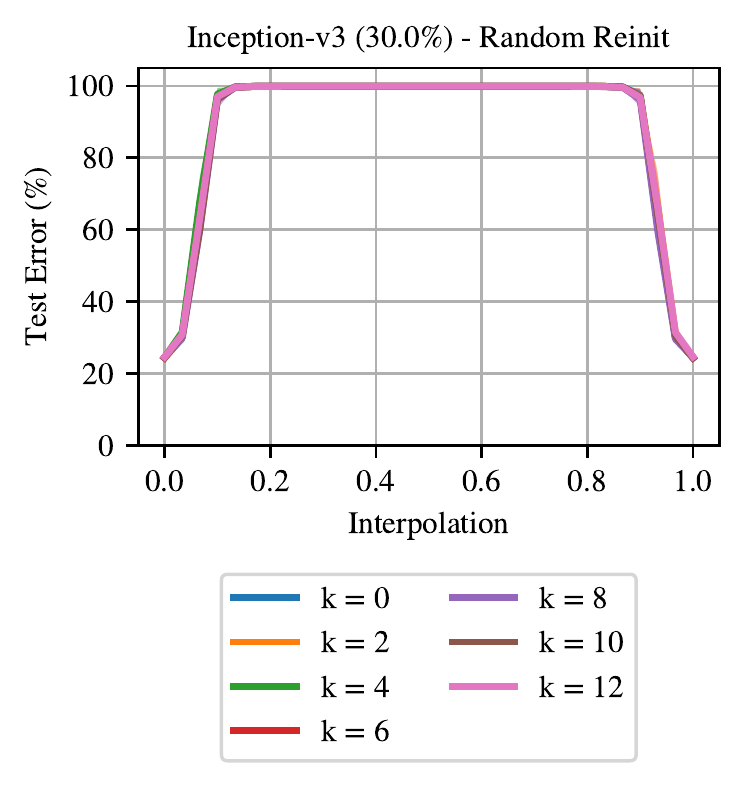}

\includegraphics[width=0.19\textwidth]{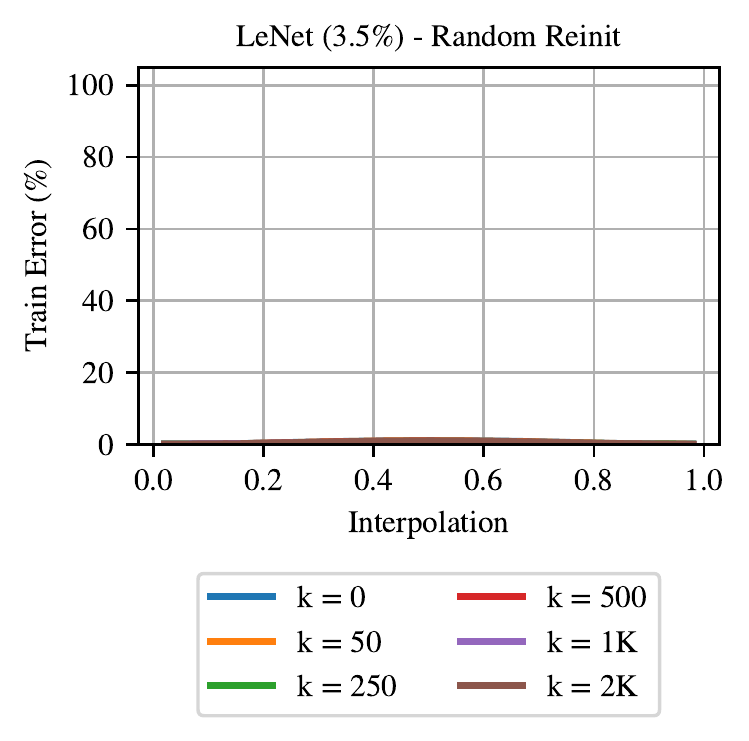}
\includegraphics[width=0.19\textwidth]{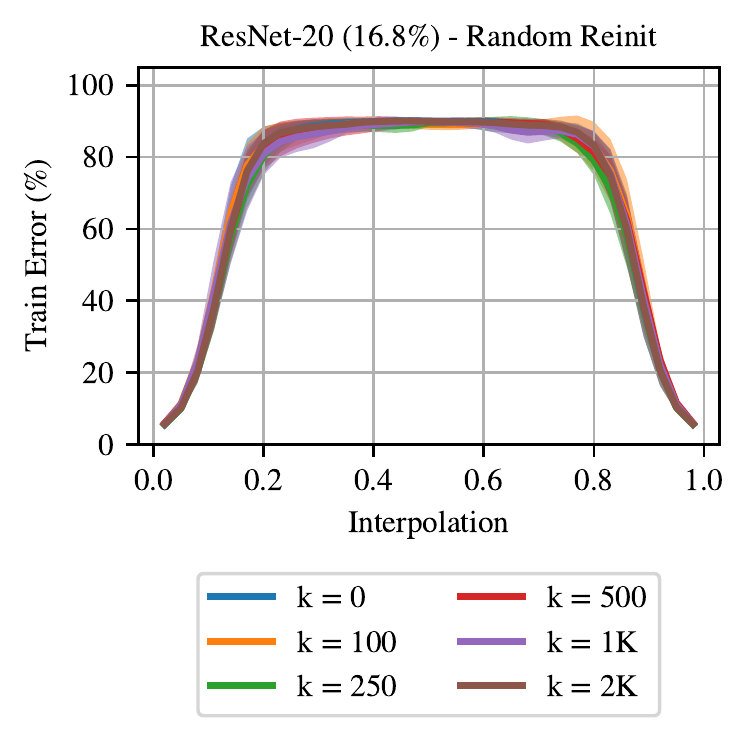}
\includegraphics[width=0.19\textwidth]{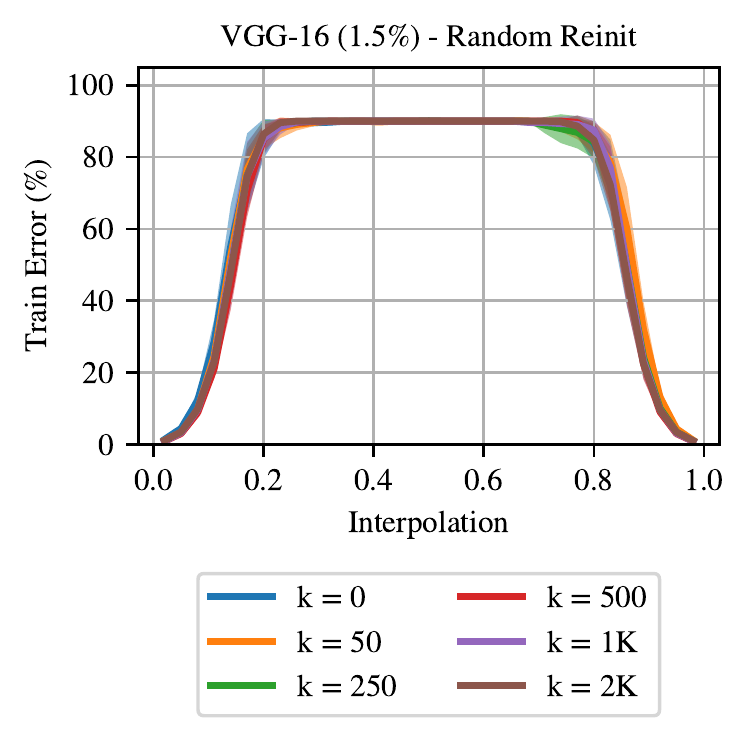}
\begin{minipage}{0.19\textwidth}~\end{minipage}
\begin{minipage}{0.19\textwidth}~\end{minipage}
\caption{The error when linearly interpolating between the minima found by randomly initializing a network, training to iteration $k$, pruning according to IMP, randomly reinitializing, and training two copies from there to completion using different data orders. Each line is the mean and standard deviation across three initializations and three data orders (nine samples in total). The errors of the trained networks are at interpolation = 0.0 and 1.0. We did not interpolate using the training set for the ImageNet networks due to computational limitations.}
\label{fig:hills-reinit}
\end{figure*}

In Figures \ref{fig:full-instability-later} and \ref{fig:sparse-instability-later}, we plot the instability value derived from linearly interpolating between copies of the same network or subnetwork trained on different data orders.
In this appendix, we plot the linear interpolation data from which we derived the instabilities in Figures \ref{fig:full-instability-later} and \ref{fig:sparse-instability-later}.
We plot this data for the unpruned networks (Figure \ref{fig:hills-full}), IMP subnetworks (Figure \ref{fig:hills-imp}), randomly pruned subnetworks (Figure \ref{fig:hills-rearr}), and the randomly reinitialized IMP subnetworks (Figure \ref{fig:hills-reinit}).

\section{Train Instability for Sparse Subnetworks}
\label{app:sparse-train-data}

\begin{figure*}
\begin{tikzpicture}[x=\textwidth,y=\textwidth, every node/.style = {anchor=north west}]
\node at (0.0, 0) {\includegraphics[width=0.19\textwidth]{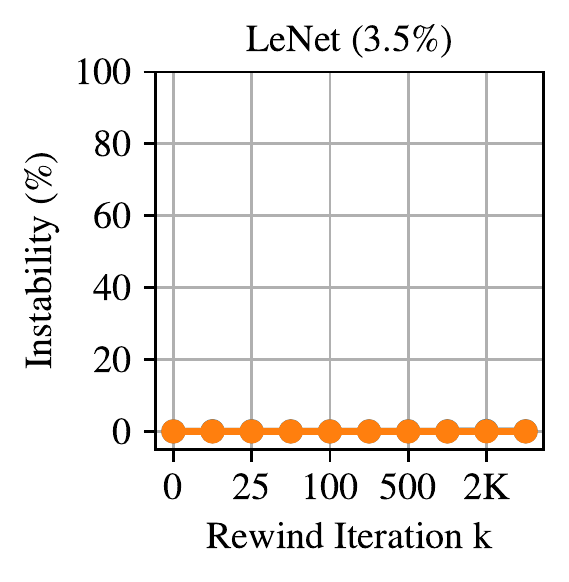}};
\node at (0.2, 0) {\includegraphics[width=0.19\textwidth]{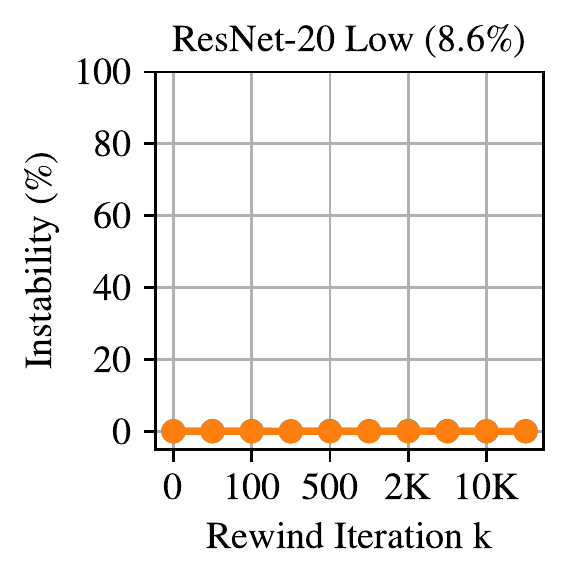}};
\node at (0.4, 0) {\includegraphics[width=0.19\textwidth]{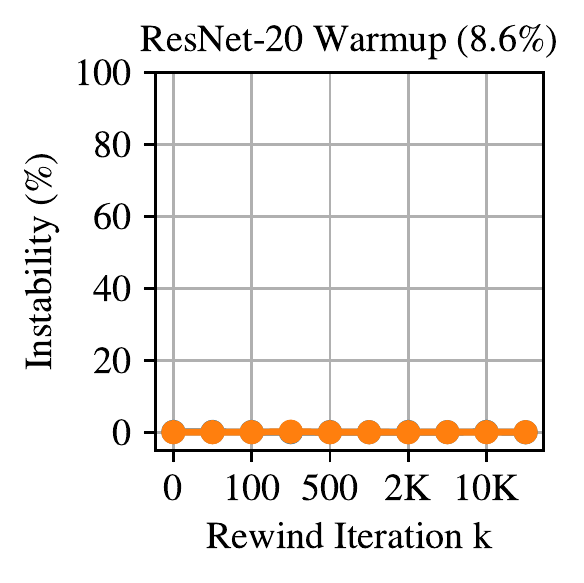}};
\node at (0.6, 0) {\includegraphics[width=0.19\textwidth]{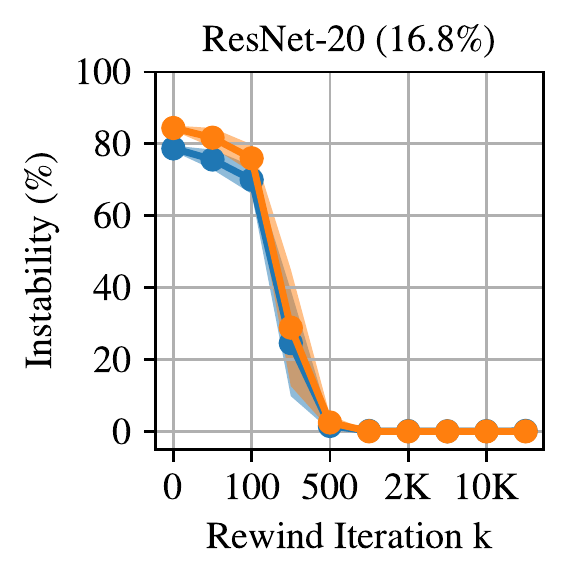}};
\node at (0.8, 0) {\includegraphics[width=0.19\textwidth]{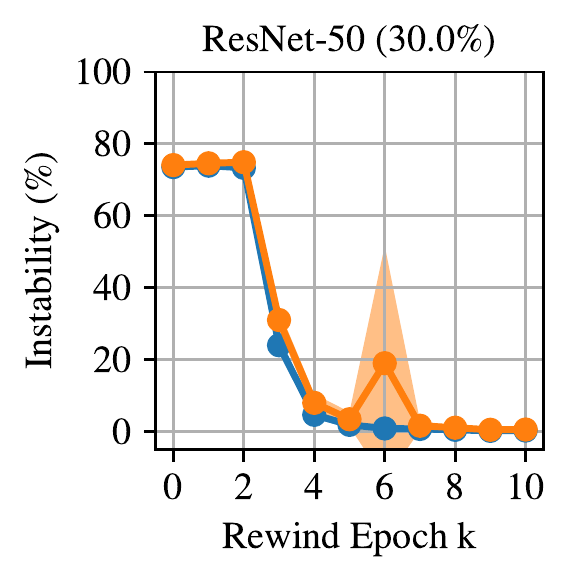}};

\node at (0.0,  -0.24) {\includegraphics[width=0.19\textwidth]{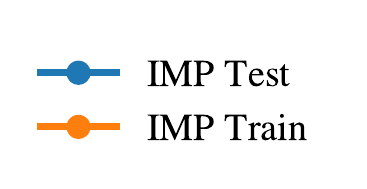}};
\node at (0.2,  -0.19) {\includegraphics[width=0.19\textwidth]{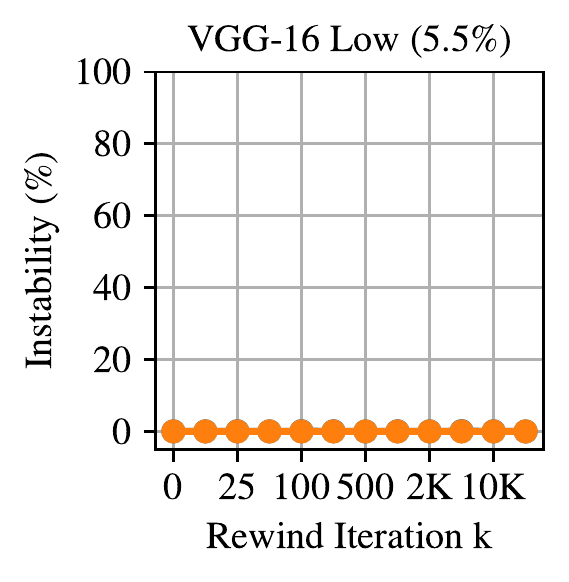}};
\node at (0.4,  -0.19) {\includegraphics[width=0.19\textwidth]{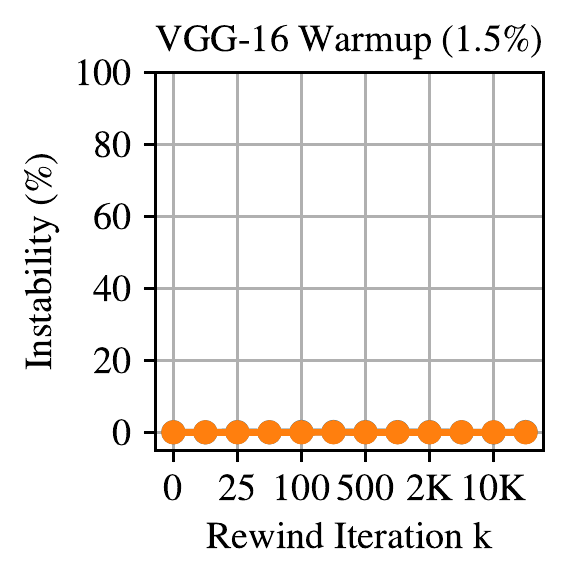}};
\node at (0.6,  -0.19) {\includegraphics[width=0.19\textwidth]{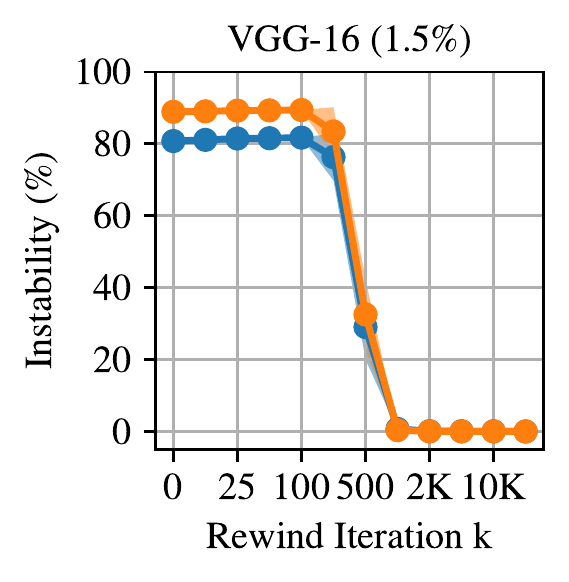}};
\node at (0.8,  -0.19) {\includegraphics[width=0.19\textwidth]{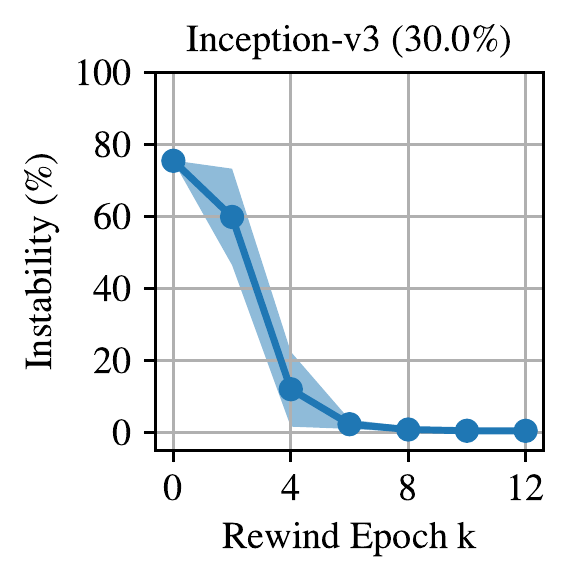}};

\end{tikzpicture}
\caption{The train and test set instability of subnetworks that are created by using the state of the full network at iteration $k$, applying the pruning mask found by performing IMP with rewinding to iteration $k$, and training on different data orders from there.
 Each line is the mean and standard deviation across three initializations and three data orders (nine samples in total). Percents are percents of weights remaining. We did not compute the train set quantities for Inception-v3 due to computational limitations.}
\label{fig:train-sparse-instability-later-stability}
\end{figure*}

\begin{figure*}
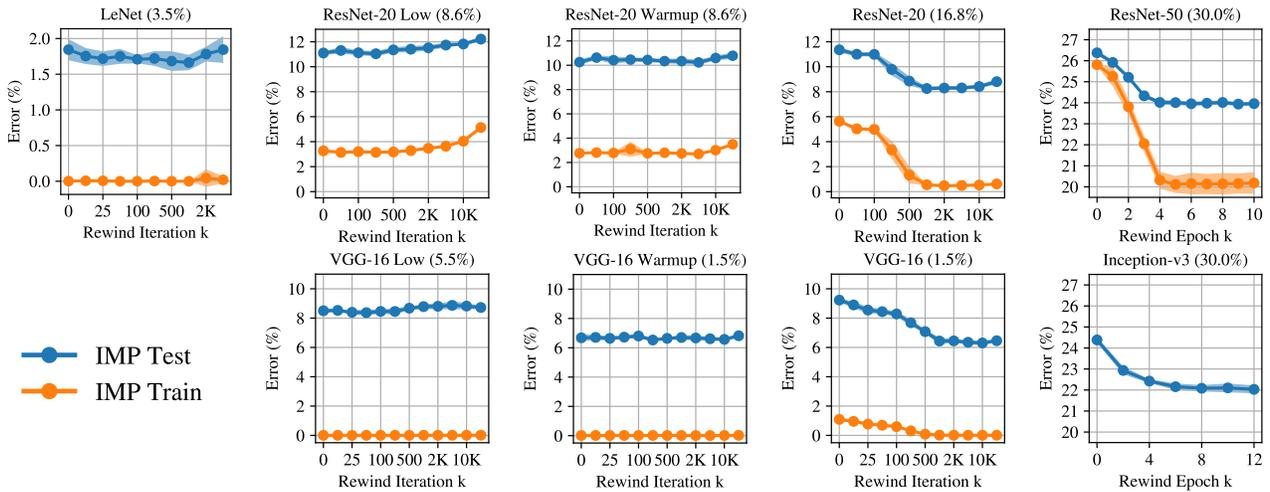

\begin{tikzpicture}[x=\textwidth,y=\textwidth, every node/.style = {anchor=north west}]
\node at (0.0, 0) {\includegraphics[width=0.19\textwidth]{figures/sparse-error-train/lenet-level15-dataorder}};
\node at (0.2, 0) {\includegraphics[width=0.19\textwidth]{figures/sparse-error-train/resnet20low-level11-dataorder}};
\node at (0.4, 0) {\includegraphics[width=0.19\textwidth]{figures/sparse-error-train/resnet20warmup-level11-dataorder}};
\node at (0.6, 0) {\includegraphics[width=0.19\textwidth]{figures/sparse-error-train/resnet20-level8-dataorder}};
\node at (0.8, 0) {\includegraphics[width=0.19\textwidth]{figures/sparse-error-train/resnet50-level70-dataorder}};

\node at (0.0,  -0.24) {\includegraphics[width=0.19\textwidth]{figures/sparse-error-train/lenet-level15-dataorder-legend}};
\node at (0.2,  -0.19) {\includegraphics[width=0.19\textwidth]{figures/sparse-error-train/vgg16low-level13-dataorder}};
\node at (0.4,  -0.19) {\includegraphics[width=0.19\textwidth]{figures/sparse-error-train/vgg16warmup-level19-dataorder}};
\node at (0.6,  -0.19) {\includegraphics[width=0.19\textwidth]{figures/sparse-error-train/vgg16-level19-dataorder}};
\node at (0.8,  -0.19) {\includegraphics[width=0.19\textwidth]{figures/sparse-error-train/inceptionv3-level70-dataorder}};

\end{tikzpicture}
\caption{The train and test set error of subnetworks that are created by using the state of the full network at iteration $k$, applying the pruning mask found by performing IMP with rewinding to iteration $k$, and training on different data orders from there.
 Each line is the mean and standard deviation across three initializations. Percents are percents of weights remaining. We did not compute the train set quantities for Inception-v3 due to computational limitations.}
\label{fig:train-sparse-instability-later-error2}
\end{figure*}

In Section \ref{sec:pruned-networks}, we only measure instability and error on the test set.
We make this choice for simplicity after observing in Section \ref{sec:full-networks} that train and test instability closely align.
In this appendix, we present the data from Section \ref{sec:pruned-networks} on the test set.
Figures \ref{fig:train-sparse-instability-later-stability} and \ref{fig:train-sparse-instability-later-error2} examine the instability and error of the same IMP subnetworks as Figure \ref{fig:sparse-instability-later}, but it shows both the train and test sets.
We did not compute the train set quantities for Inception-v3 due to computational limitations.

Train set and test set instability are nearly identical, just as we found in Section \ref{sec:full-networks}.
Interestingly, the two coincide more closely for IMP subnetworks of ResNet-50 than they do for the unpruned networks in Section \ref{sec:full-networks}.

For networks that are unstable at rewinding iteration 0, train error and test error follow similar trends, starting higher when the subnetworks are unstable and dropping when the subnetworks become stable.
In other words, the unstable IMP subnetworks are not able to fully optimize to 0\% train error, while the stable IMP subnetworks are.

\section{Alternate Distance Functions}
\label{app:alternate-distance-metrics}

Instability analysis involves training two copies of the same network on different data orders and comparing the networks that result.
In the main body of the paper, our method of comparison is linear interpolation, which we find to offer valuable new insights into neural network optimization and the lottery ticket hypothesis.
However, one could parameterize instability analysis with a wide range of other functions for comparing pairs of neural networks.
In this appendix, we discuss four alternate methods for which we collected data using the MNIST and CIFAR-10 networks.

\textbf{$L_2$ Distance.}
One simple way to compare neural networks is to measure the $L_2$ distance between the trained weights.
The limitation of this function is that there is not necessarily any relationship between $L_2$ distance and the functional similarity of networks or the structure of the loss landscape.
In other words, there is no clear interpretation of $L_2$ distance.

\begin{figure*}
\centering
\begin{tikzpicture}[x=\textwidth,y=\textwidth, every node/.style = {anchor=north west}]
\node at (0.0, 0) {\includegraphics[width=0.19\textwidth]{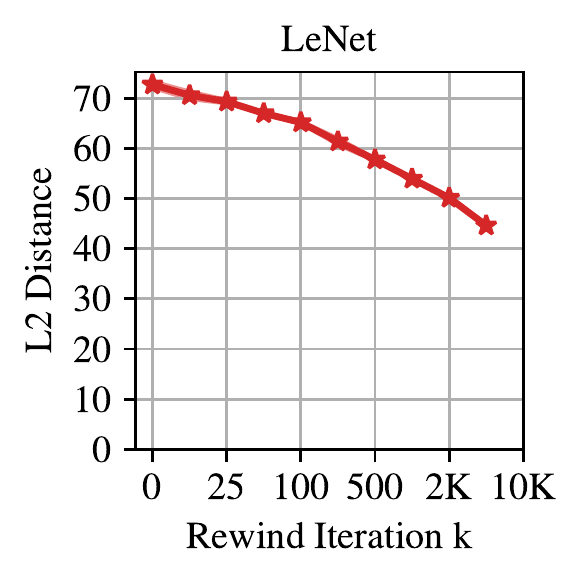}};
\node at (0.2, 0) {\includegraphics[width=0.19\textwidth]{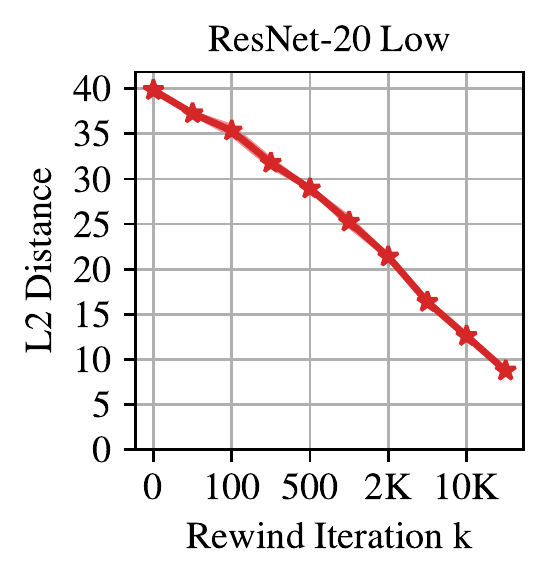}};
\node at (0.4, 0) {\includegraphics[width=0.19\textwidth]{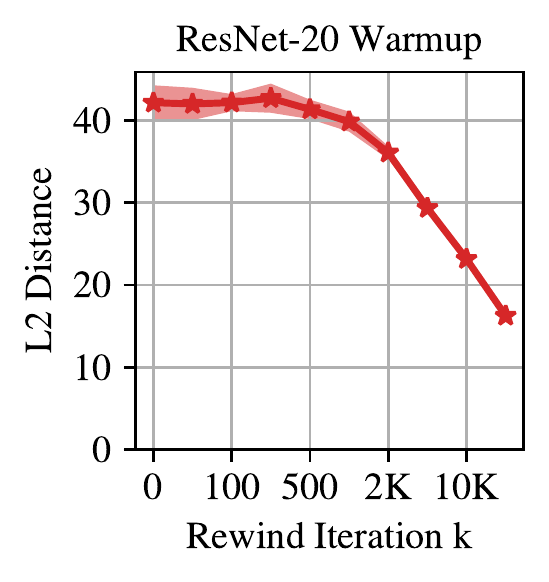}};
\node at (0.6, 0) {\includegraphics[width=0.19\textwidth]{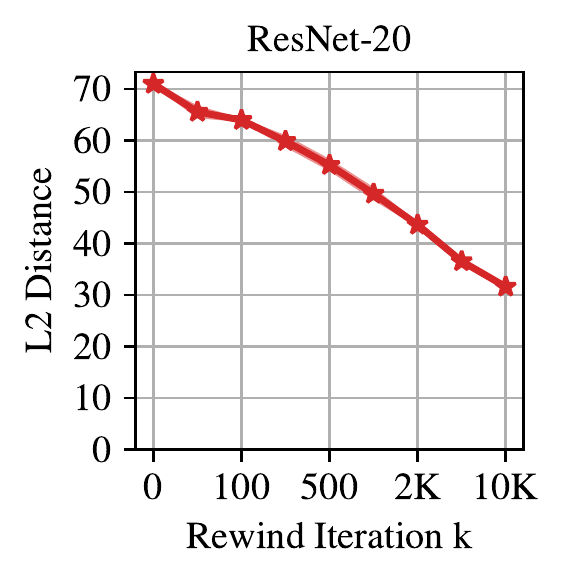}};

\node at (0.0,  -0.24) {\includegraphics[width=0.19\textwidth]{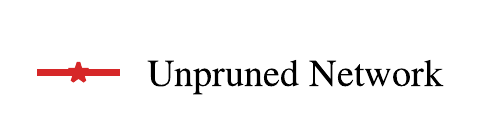}};
\node at (0.2,  -0.19) {\includegraphics[width=0.19\textwidth]{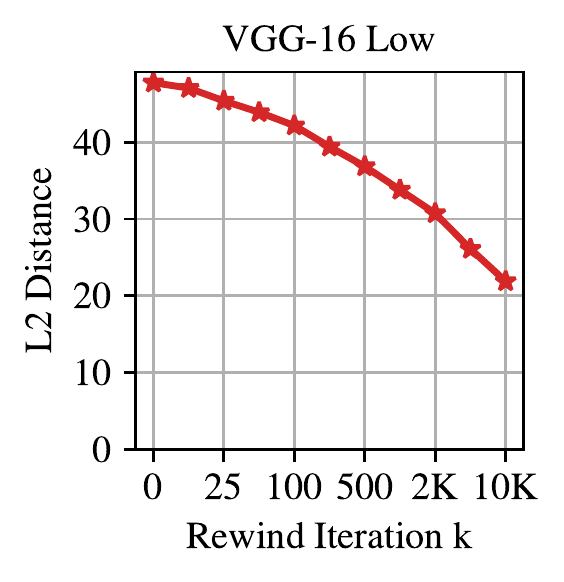}};
\node at (0.4,  -0.19) {\includegraphics[width=0.19\textwidth]{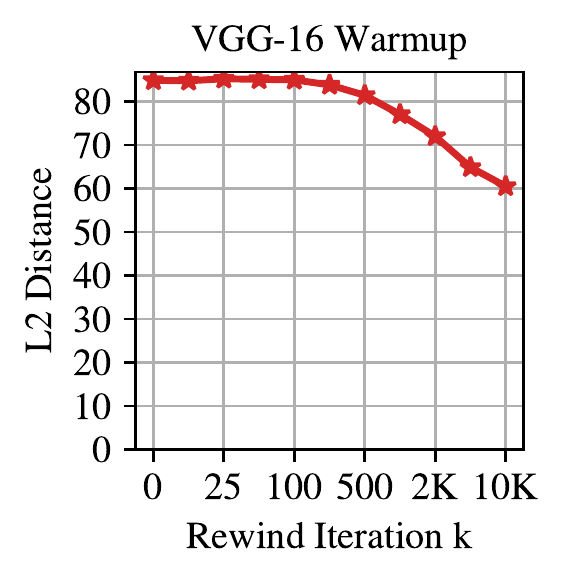}};
\node at (0.6,  -0.19) {\includegraphics[width=0.19\textwidth]{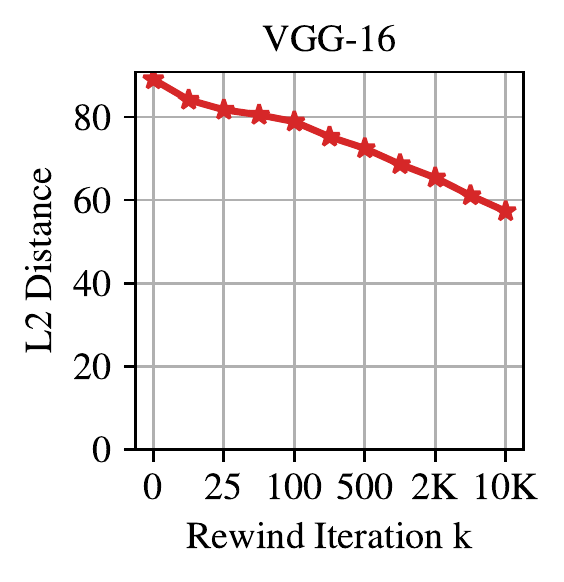}};

\end{tikzpicture}

\vspace{-0.5em}
\caption{The $L_2$ distance between networks that are created by trained to iteration $k$, making two copies, and training on different data orders from there.
 Each line is the mean and standard deviation across three initializations and three data orders (nine samples in total). Percents are percents of weights remaining. We did not compute the train set quantities for the ImageNet networks due to computational limitations.}
\label{fig:app-l2-dist-full}
\end{figure*}

\begin{figure*}
\centering
\begin{tikzpicture}[x=\textwidth,y=\textwidth, every node/.style = {anchor=north west}]
\node at (0.0, 0) {\includegraphics[width=0.19\textwidth]{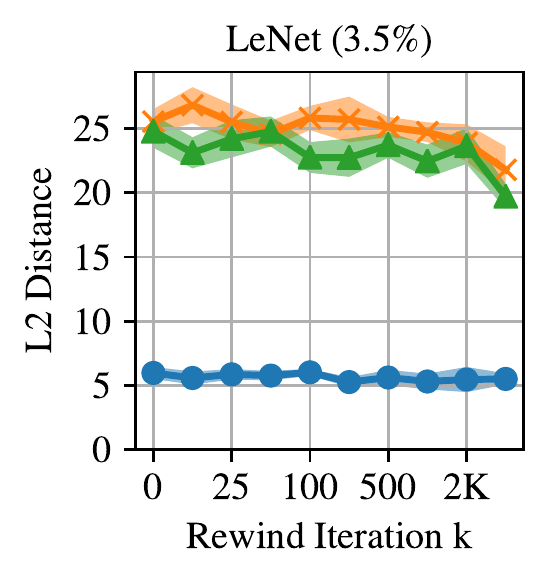}};
\node at (0.2, 0) {\includegraphics[width=0.19\textwidth]{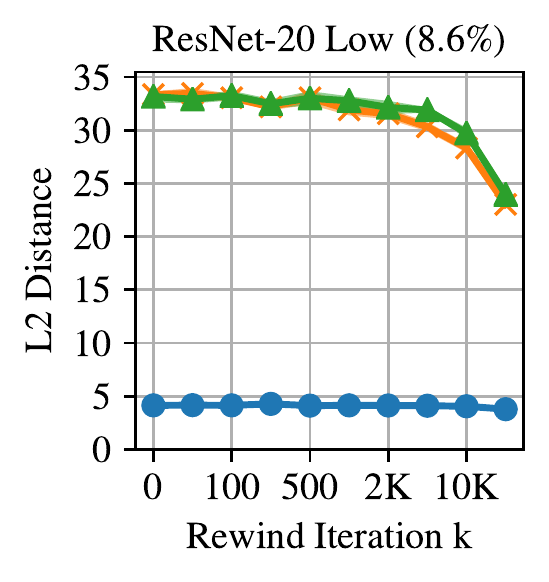}};
\node at (0.4, 0) {\includegraphics[width=0.19\textwidth]{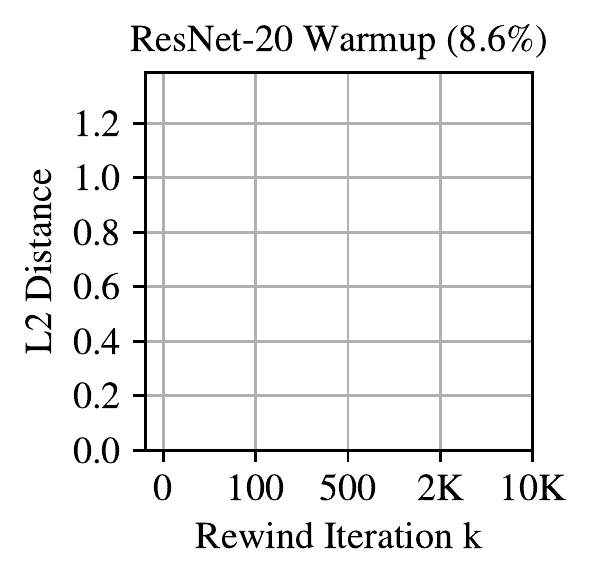}};
\node at (0.6, 0) {\includegraphics[width=0.19\textwidth]{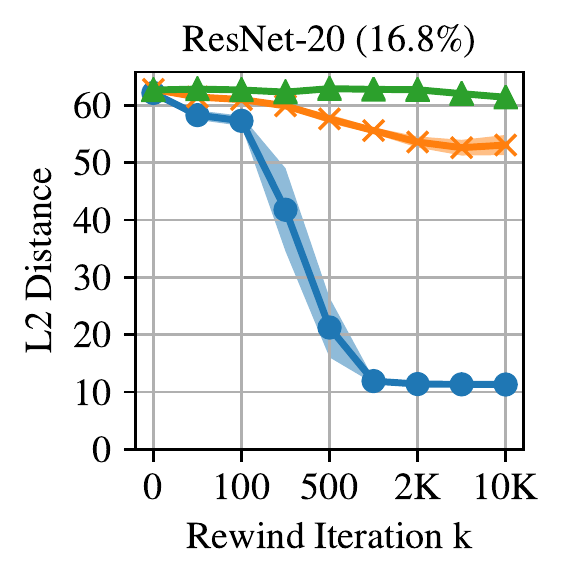}};

\node at (0.0,  -0.24) {\includegraphics[width=0.19\textwidth]{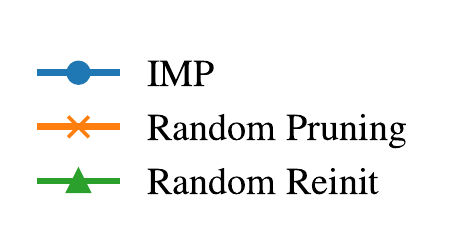}};
\node at (0.2,  -0.19) {\includegraphics[width=0.19\textwidth]{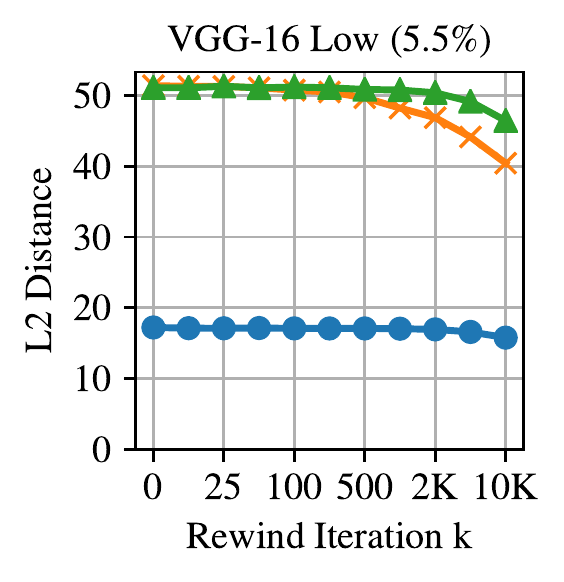}};
\node at (0.4,  -0.19) {\includegraphics[width=0.19\textwidth]{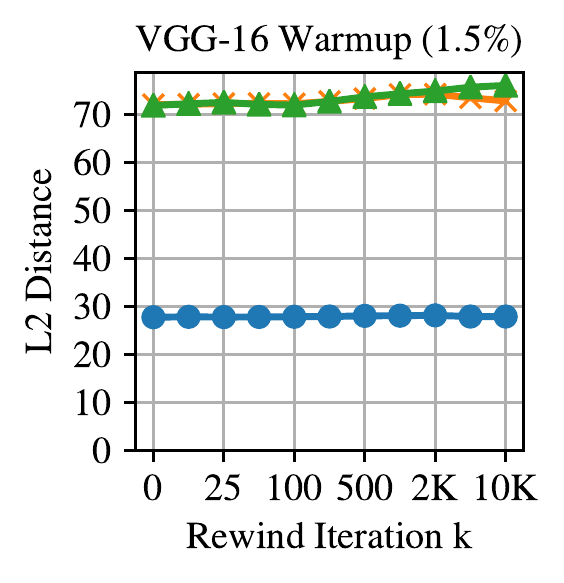}};
\node at (0.6,  -0.19) {\includegraphics[width=0.19\textwidth]{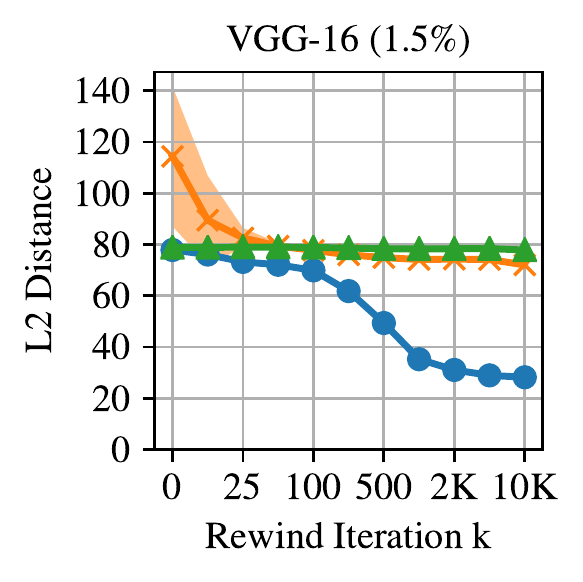}};

\end{tikzpicture}
\vspace{-0.5em}
\caption{The $L_2$ distance between subnetworks that are created by using the state of the full network at iteration $k$, applying a pruning mask, and training on different data orders from there.
 Each line is the mean and standard deviation across three initializations and three data orders (nine samples in total). Percents are percents of weights remaining. We did not compute the train set quantities for the ImageNet networks due to computational limitations.}
\label{fig:app-l2-dist}
\end{figure*}

In Figure \ref{fig:app-l2-dist-full}, we plot the $L_2$ distance function at all rewinding points for the unpruned networks. In Figure \ref{fig:app-l2-dist}, we plot the $L_2$ distance function at all rewinding points for all three classes of sparse networks.
We plot this data separately because $L_2$ distance is not necessarily comparable between sparse networks (which have fewer parameters) and dense networks (which have more parameters).

For the unpruned networks, distance decreases linearly as we logarithmically increase the rewinding iteration.
We see no distinct changes in behavior when the networks become stable, and the $L_2$ distance remains far from 0 at this point.

For the IMP subnetworks, $L_2$ distance mirrors the behavior of instability.
In cases where the IMP subnetworks are stable at all rewinding points (ResNet-20 low/warmup, VGG-16 low/warmup, and LeNet), the $L_2$ distance is at a lower level than the $L_2$ distance between the other baselines (random pruning and random reinitialization) and is consistent across rewinding points.
In cases where the IMP subnetworks are unstable at initialization but become stable later (ResNet-20 and VGG-16), the $L_2$ distance begins high (at the same level as the $L_2$ distance for the randomly pruned and randomly reinitialized baselines) and drops when the subnetworks become stable, settling at a lower level.

Although stable IMP subnetworks are closer in $L_2$ distance than unstable IMP subnetworks and the baselines, the $L_2$ distance remains far from zero.
In general, it is difficult to translate the results of this function into higher-level statements about the relationships between the networks.

\begin{figure*}
\centering
\begin{tikzpicture}[x=\textwidth,y=\textwidth, every node/.style = {anchor=north west}]
\node at (0.0, 0) {\includegraphics[width=0.19\textwidth]{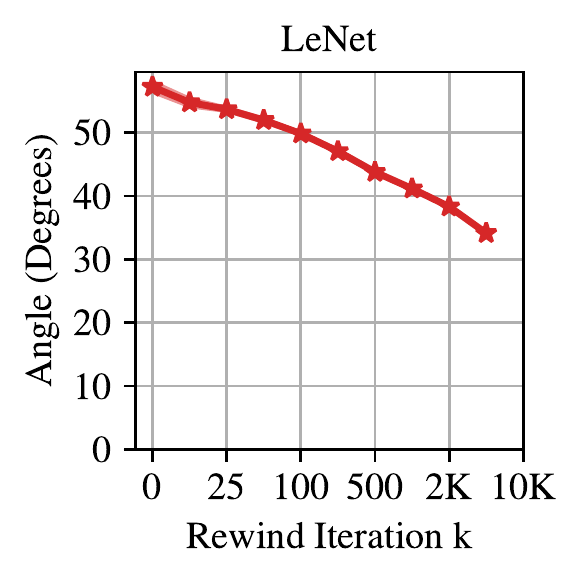}};
\node at (0.2, 0) {\includegraphics[width=0.19\textwidth]{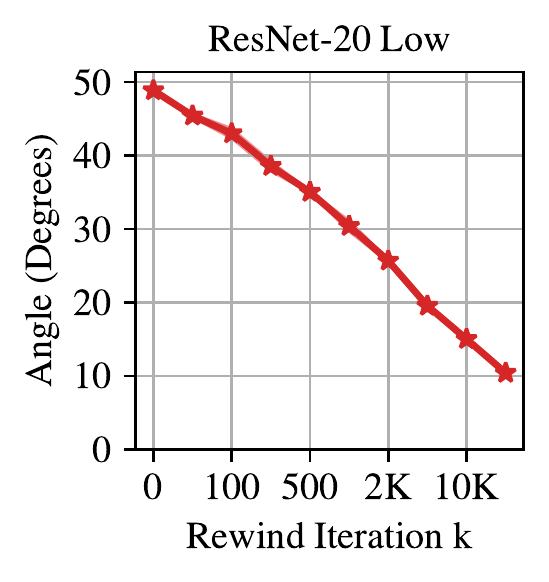}};
\node at (0.4, 0) {\includegraphics[width=0.19\textwidth]{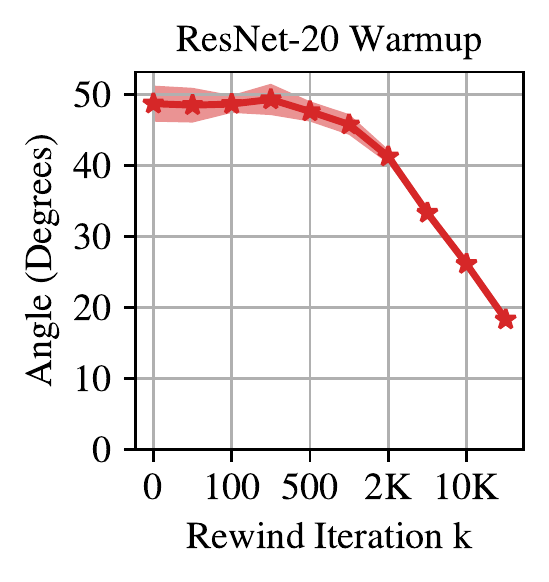}};
\node at (0.6, 0) {\includegraphics[width=0.19\textwidth]{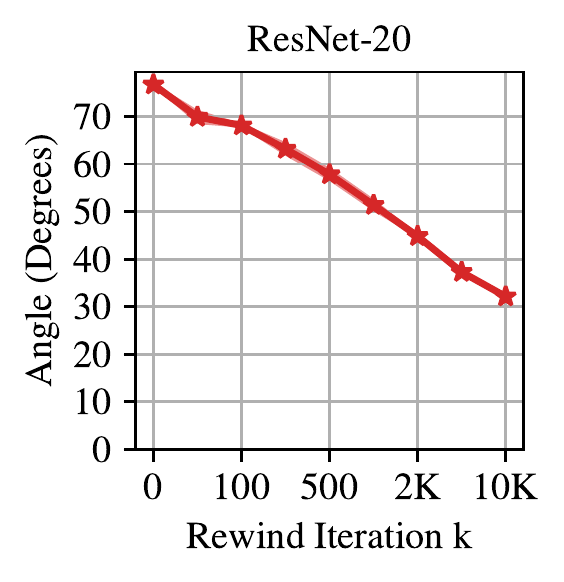}};

\node at (0.0,  -0.24) {\includegraphics[width=0.19\textwidth]{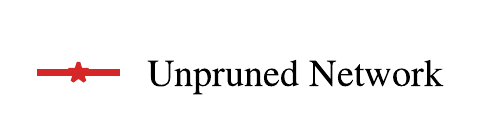}};
\node at (0.2,  -0.19) {\includegraphics[width=0.19\textwidth]{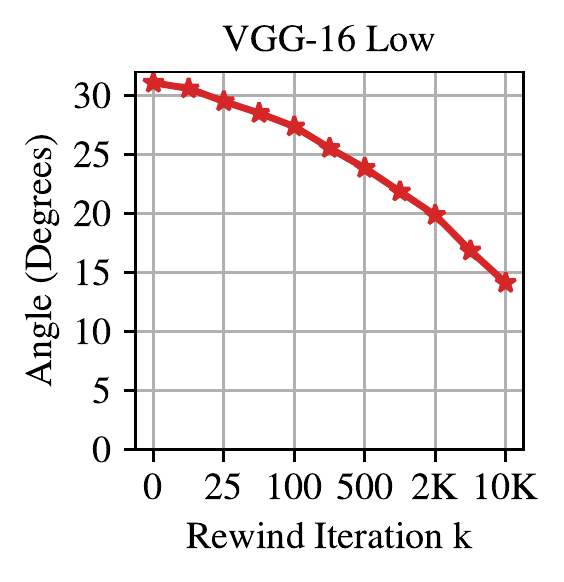}};
\node at (0.4,  -0.19) {\includegraphics[width=0.19\textwidth]{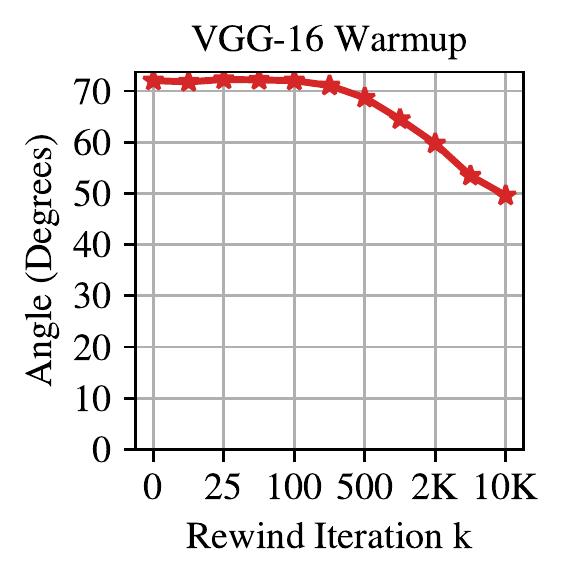}};
\node at (0.6,  -0.19) {\includegraphics[width=0.19\textwidth]{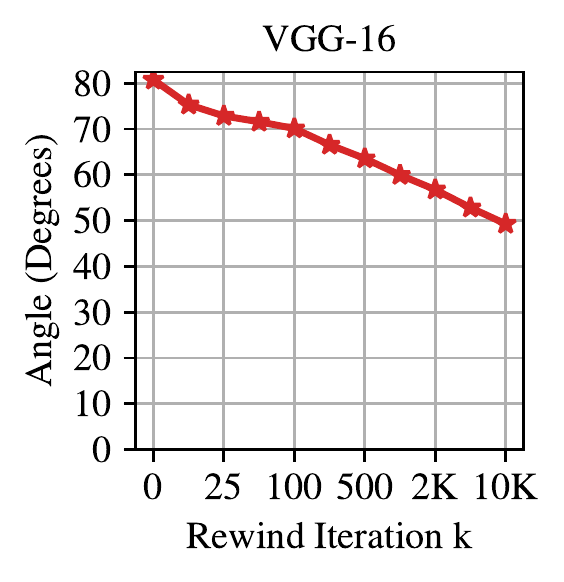}};

\end{tikzpicture}

\vspace{-0.5em}
\caption{The cosine distance between networks that are created by trained to iteration $k$, making two copies, and training on different data orders from there.
 Each line is the mean and standard deviation across three initializations and three data orders (nine samples in total). Percents are percents of weights remaining. We did not compute the train set quantities for the ImageNet networks due to computational limitations.}
\label{fig:app-cosine-dist-full}
\end{figure*}

\begin{figure*}
\centering
\begin{tikzpicture}[x=\textwidth,y=\textwidth, every node/.style = {anchor=north west}]
\node at (0.0, 0) {\includegraphics[width=0.19\textwidth]{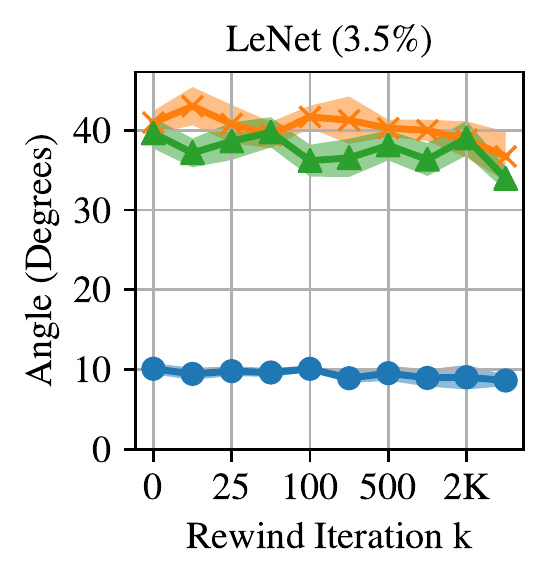}};
\node at (0.2, 0) {\includegraphics[width=0.19\textwidth]{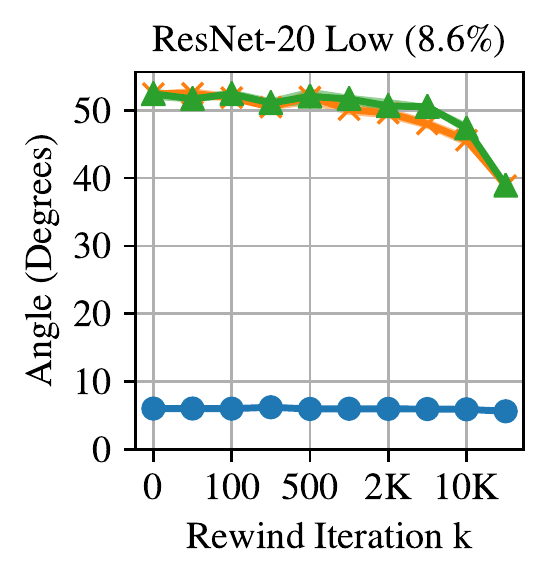}};
\node at (0.4, 0) {\includegraphics[width=0.19\textwidth]{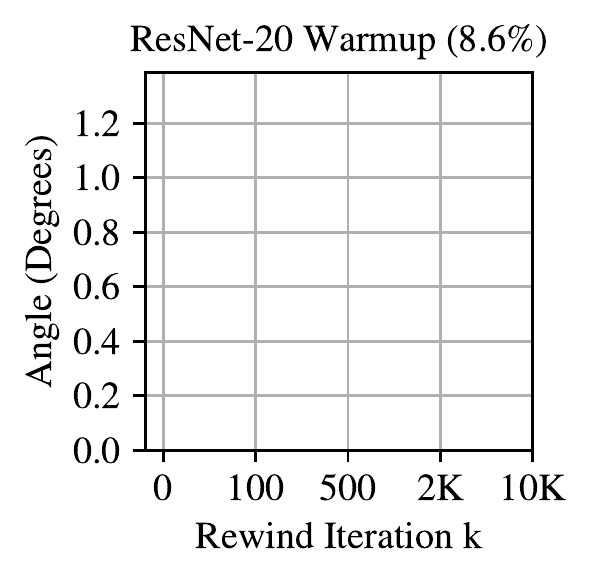}};
\node at (0.6, 0) {\includegraphics[width=0.19\textwidth]{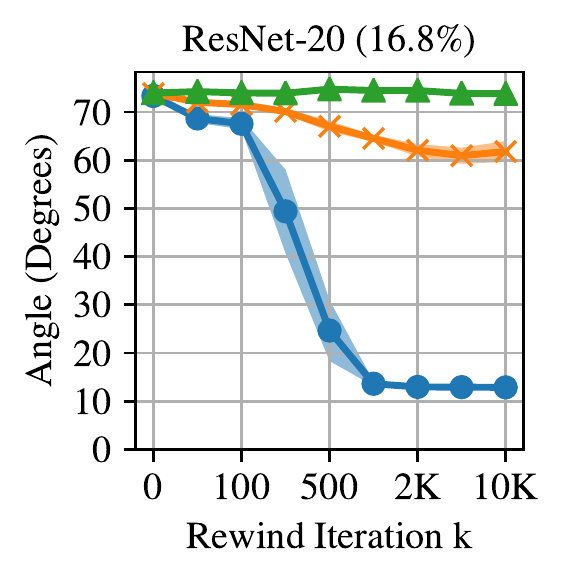}};

\node at (0.0,  -0.24) {\includegraphics[width=0.19\textwidth]{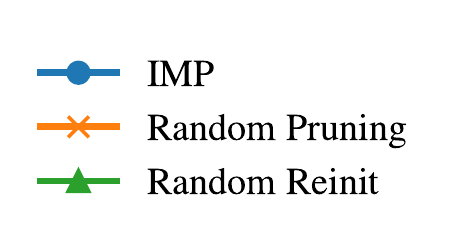}};
\node at (0.2,  -0.19) {\includegraphics[width=0.19\textwidth]{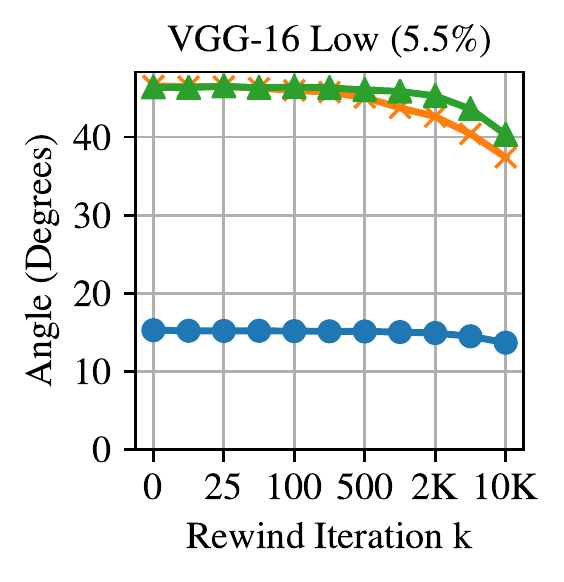}};
\node at (0.4,  -0.19) {\includegraphics[width=0.19\textwidth]{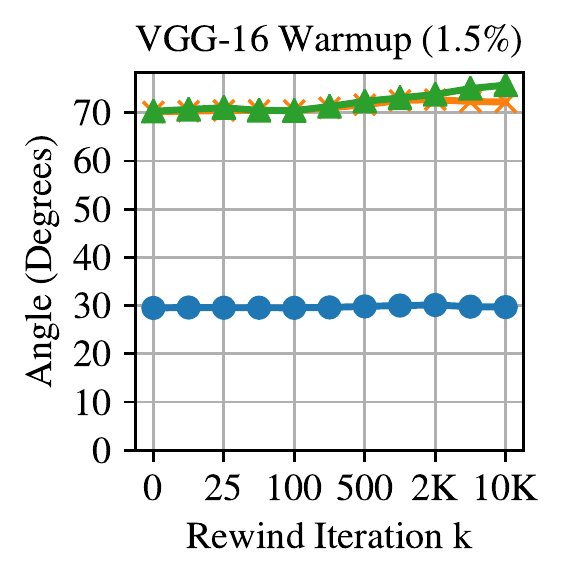}};
\node at (0.6,  -0.19) {\includegraphics[width=0.19\textwidth]{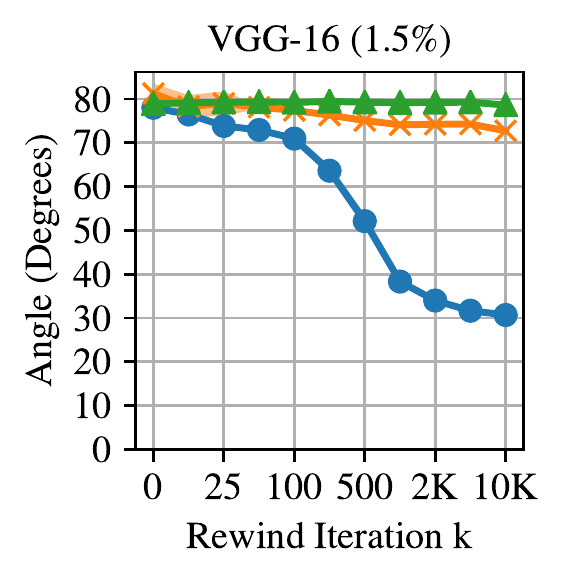}};

\end{tikzpicture}

\vspace{-0.5em}
\caption{The cosine distance between subnetworks that are created by using the state of the full network at iteration $k$, applying a pruning mask, and training on different data orders from there.
 Each line is the mean and standard deviation across three initializations and three data orders (nine samples in total). Percents are percents of weights remaining. We did not compute the train set quantities for the ImageNet networks due to computational limitations.}
\label{fig:app-cosine-dist}
\end{figure*}

\textbf{Cosine distance.}
In Figures \ref{fig:app-cosine-dist-full} (unpruned networks) and \ref{fig:app-cosine-dist} (sparse networks), we plot the cosine distance in a manner similar to $L_2$ distance.
The results are similar to those for $L_2$ distance, and the same interpretation applies.

\textbf{Classification differences.}
This distance function computes the number of examples that are classified differently by two networks.
Unlike linear interpolation and $L_2$/cosine distance, this function looks at the functional behavior of the networks rather than the parameterizations.
This function is particularly valuable because it allows us to compare the dense and sparse networks directly.

\begin{figure*}
\centering
\begin{tikzpicture}[x=\textwidth,y=\textwidth, every node/.style = {anchor=north west}]
\node at (0.0, 0) {\includegraphics[width=0.19\textwidth]{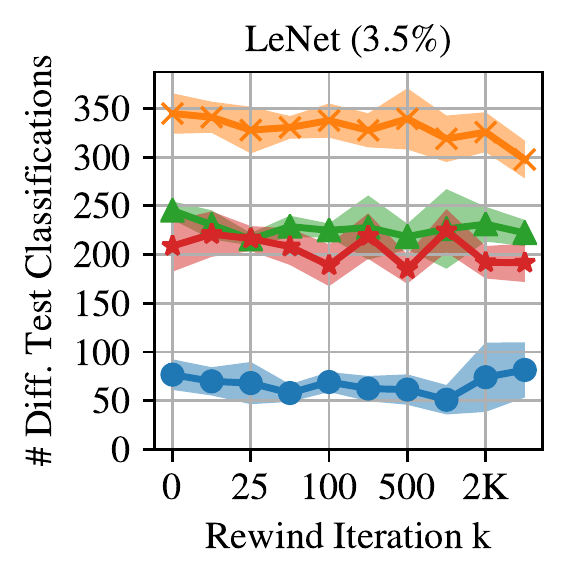}};
\node at (0.2, 0) {\includegraphics[width=0.19\textwidth]{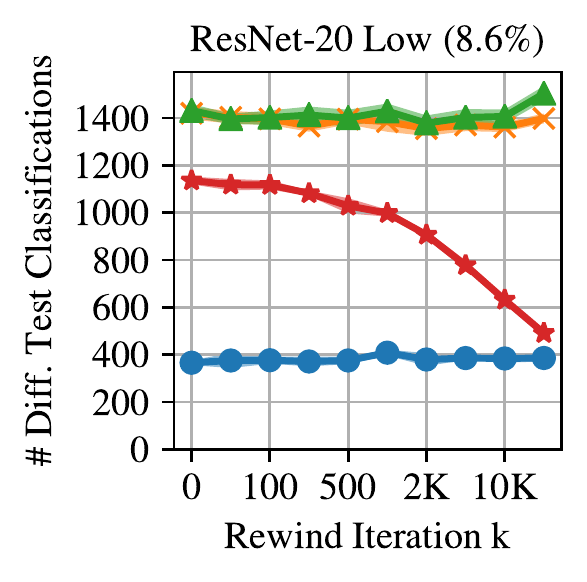}};
\node at (0.4, 0) {\includegraphics[width=0.19\textwidth]{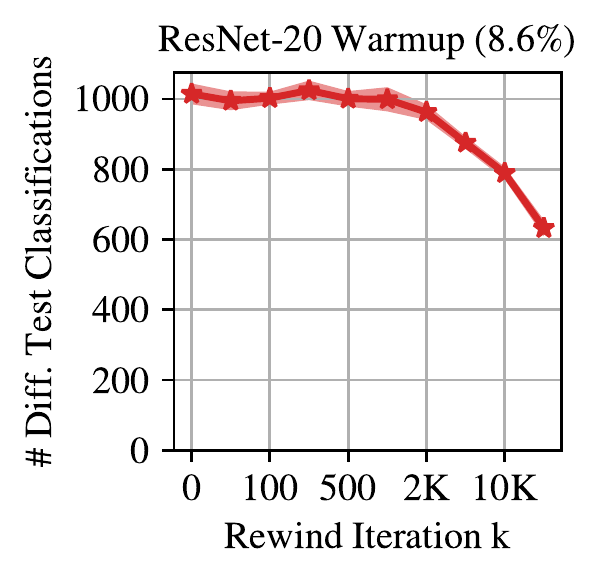}};
\node at (0.6, 0) {\includegraphics[width=0.19\textwidth]{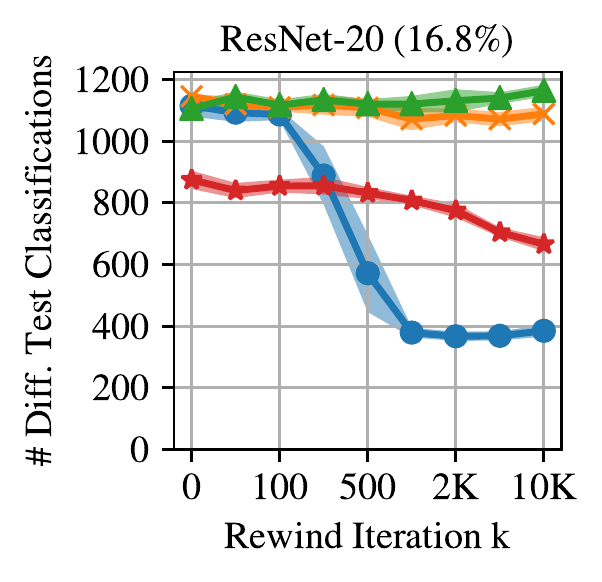}};

\node at (0.0,  -0.24) {\includegraphics[width=0.19\textwidth]{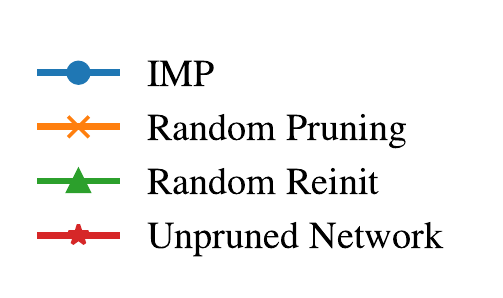}};
\node at (0.2,  -0.19) {\includegraphics[width=0.19\textwidth]{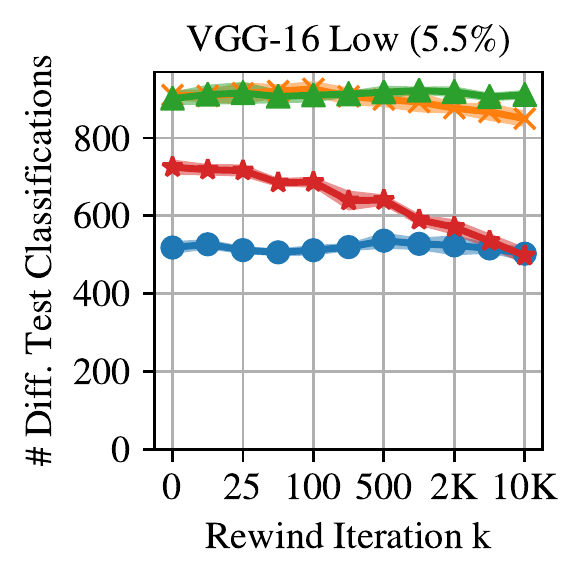}};
\node at (0.4,  -0.19) {\includegraphics[width=0.19\textwidth]{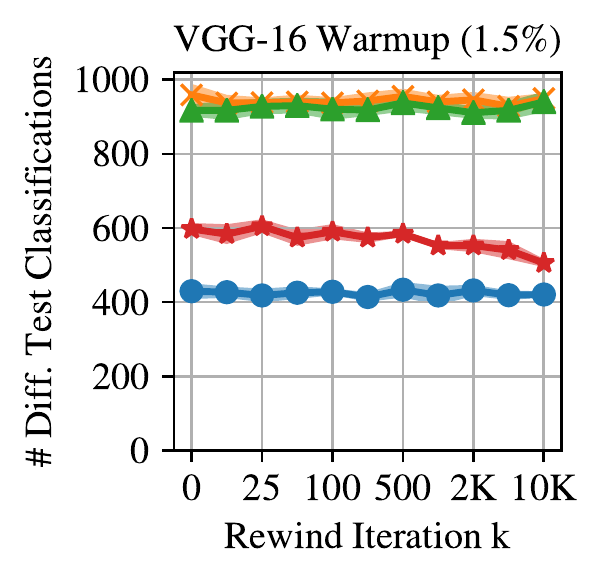}};
\node at (0.6,  -0.19) {\includegraphics[width=0.19\textwidth]{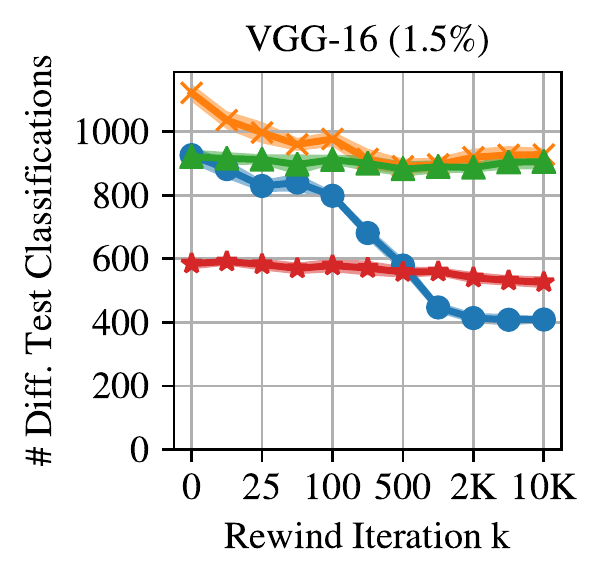}};

\end{tikzpicture}

\caption{The number of different test set classifications between networks that are created by training the full network to iteration $k$, optionally applying a pruning mask, and training on different data orders from there.
 Each line is the mean and standard deviation across three initializations and three data orders (nine samples in total). Percents are percents of weights remaining.}
\label{fig:app-classdiff-test}
\end{figure*}

\begin{figure*}
\centering
\begin{tikzpicture}[x=\textwidth,y=\textwidth, every node/.style = {anchor=north west}]
\node at (0.0, 0) {\includegraphics[width=0.19\textwidth]{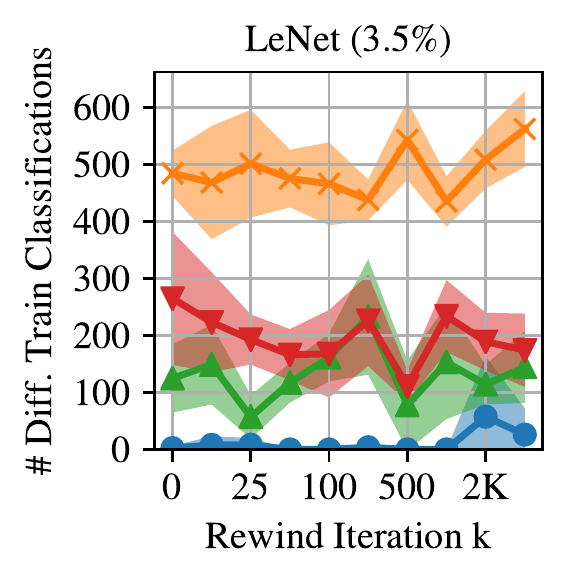}};
\node at (0.2, 0) {\includegraphics[width=0.19\textwidth]{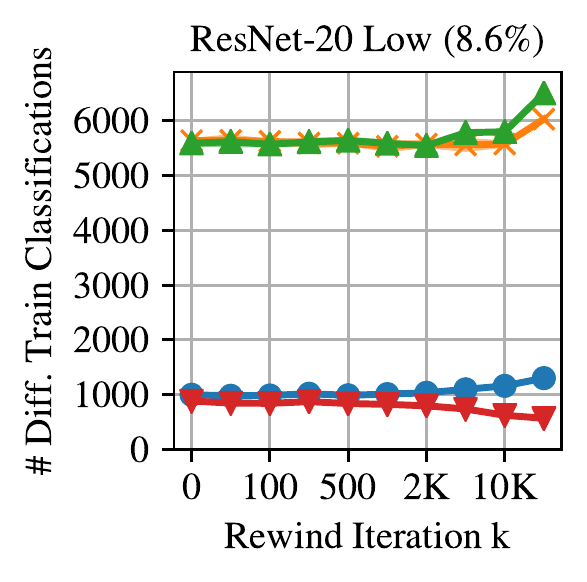}};
\node at (0.4, 0) {\includegraphics[width=0.19\textwidth]{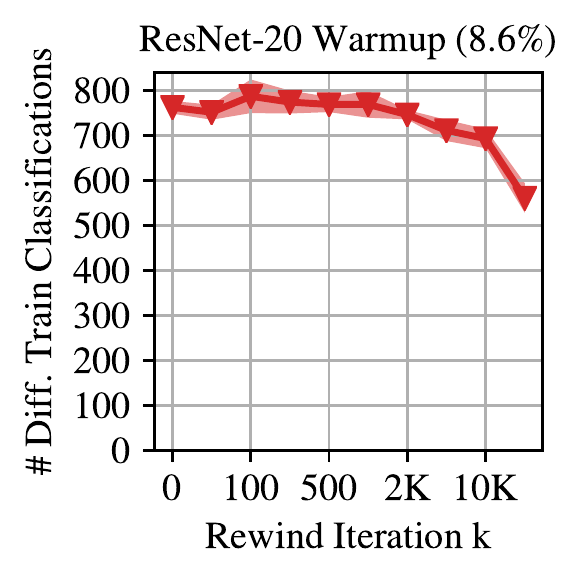}};
\node at (0.6, 0) {\includegraphics[width=0.19\textwidth]{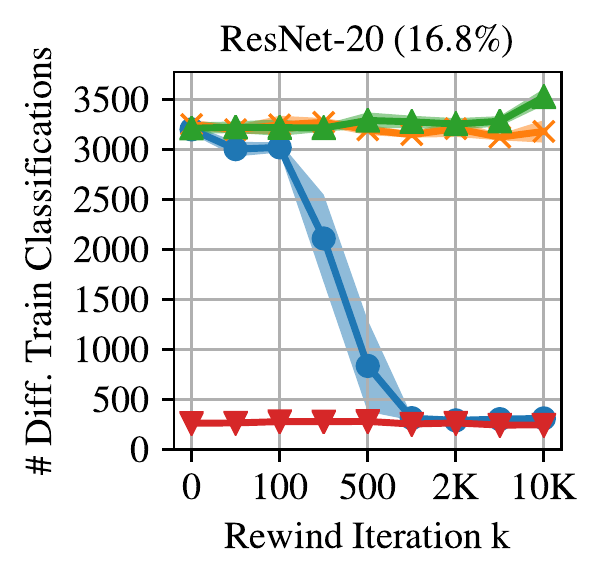}};

\node at (0.0,  -0.24) {\includegraphics[width=0.19\textwidth]{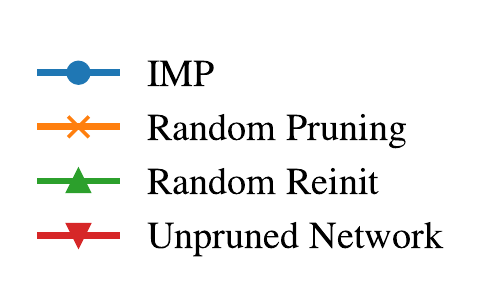}};
\node at (0.2,  -0.19) {\includegraphics[width=0.19\textwidth]{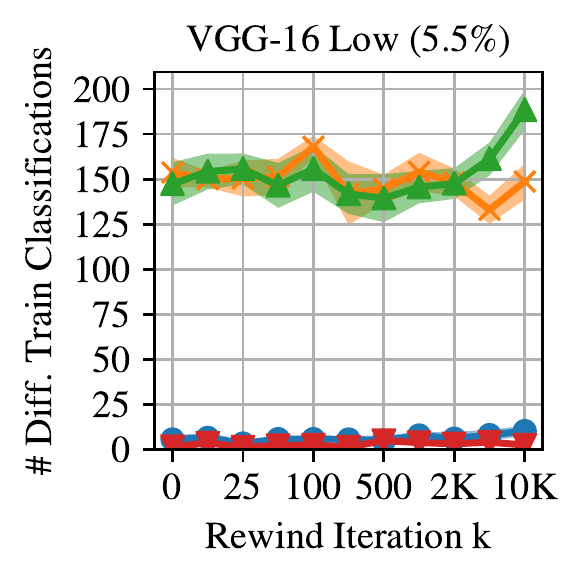}};
\node at (0.4,  -0.19) {\includegraphics[width=0.19\textwidth]{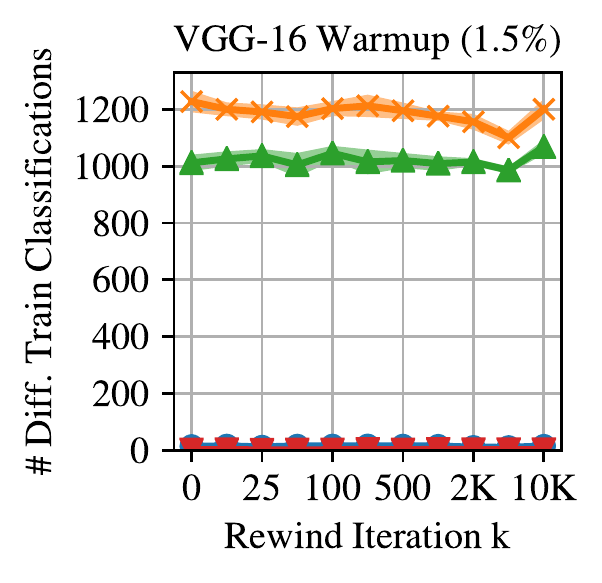}};
\node at (0.6,  -0.19) {\includegraphics[width=0.19\textwidth]{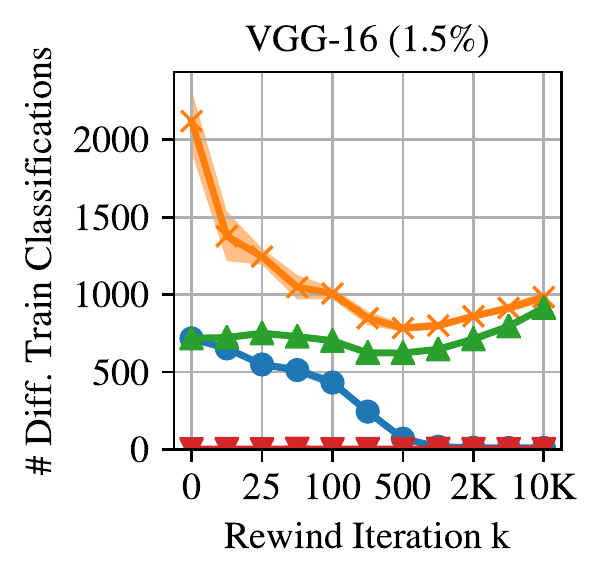}};

\end{tikzpicture}

\caption{The number of different train set classifications between networks that are created by training the full network to iteration $k$, optionally applying a pruning mask, and training on different data orders from there.
 Each line is the mean and standard deviation across three initializations and three data orders (nine samples in total). Percents are percents of weights remaining.}
\label{fig:app-classdiff-train}
\end{figure*}

In Figures \ref{fig:app-classdiff-test} (test set) and \ref{fig:app-classdiff-train} (train set), we plot this function for the unpruned and sparse networks across rewinding iterations.
The unpruned networks generally classify the same number of examples differently no matter the rewinding iteration, although the number of different classifications decreases gradually for the latest rewinding iterations for ResNet-20 low and warmup.
We see no relationship between this function and instability.

The behavior of the IMP sparse networks better matches instability.
IMP subnetworks that are stable from initialization (ResNet-20 low and warmup, VGG-16 low and warmup, LeNet) consistently have the same distance no matter the rewinding iteration.
This distance is lower than that for the randomly pruned and randomly reinitialized baselines.

IMP subnetworks that are unstable at iteration 0 (ResNet-20 and VGG-16) have the same number of different classifications as the baselines when rewinding to iteration 0.
When the networks become stable, the number of different classifications drops substantially to a lower level.

One challenge with using this distance function is that it is inherently entangled with accuracy.
As the accuracy of the networks improves, the number of different classifications might decrease simply because the networks will classify more examples correctly (and thereby, the same way).
Consider the IMP subnetworks of ResNet-20 on the CIFAR-10 test set (the graph in the upper right of Figure \ref{fig:app-classdiff-test}, blue line).
At rewinding iteration 0, the networks have about 11\% error on the test set, meaning there are at most 2200 examples they could classify differently.%
\footnote{In the worst case, all examples that one network misclassifies will be classified correctly by the other. Since each network misclassifies 1100 examples, 2200 examples will be classified differently in total.}
In Figure \ref{fig:app-classdiff-test}, we see that the networks are classifying about 1100 examples differently.

When ResNet-20 IMP subnetworks are stable, error decreases to 8.5\%, meaning at most 1700 examples can be classified differently.
However, in Figure \ref{fig:app-cosine-dist-full}, we see that only about 350 examples are being classified differently.
Although this number is lower than the 1100 differences at rewinding iteration 0 in absolute terms, accuracy has improved as well, so we must consider these differences in context.
At rewinding iteration 0, classification differences are 50\% of their maximum possible value, while at rewinding iteration 1000, classification differences are at 21\% of their maximum possible value.
In summary, as the IMP subnetworks become stable, they behave in a more functionally similar fashion, even considering accuracy improvements.

\begin{figure*}
\centering
\begin{tikzpicture}[x=\textwidth,y=\textwidth, every node/.style = {anchor=north west}]
\node at (0.0, 0) {\includegraphics[width=0.19\textwidth]{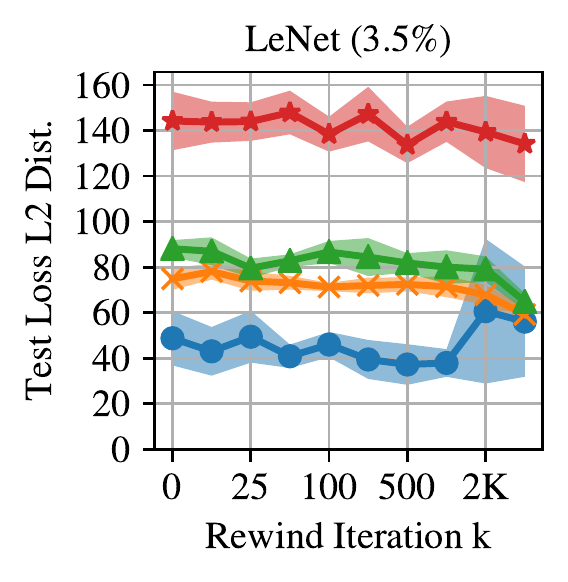}};
\node at (0.2, 0) {\includegraphics[width=0.19\textwidth]{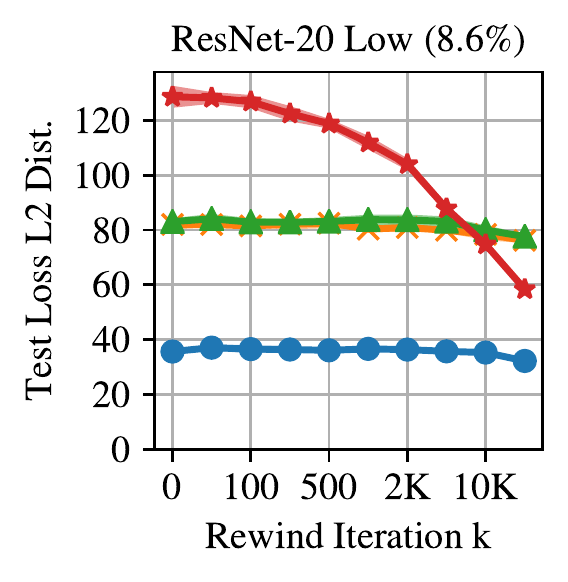}};
\node at (0.4, 0) {\includegraphics[width=0.19\textwidth]{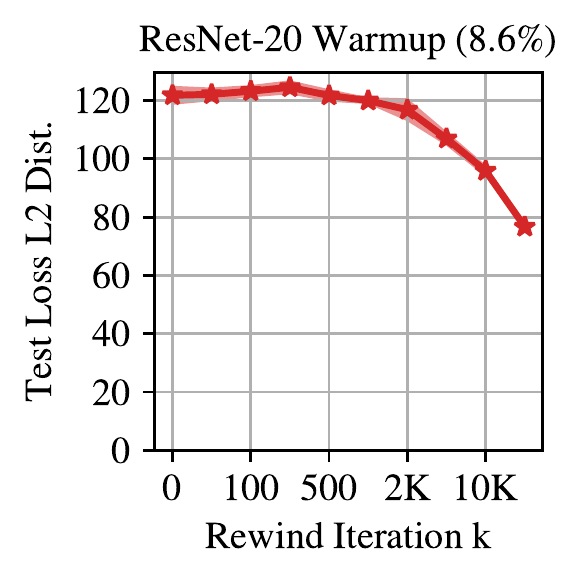}};
\node at (0.6, 0) {\includegraphics[width=0.19\textwidth]{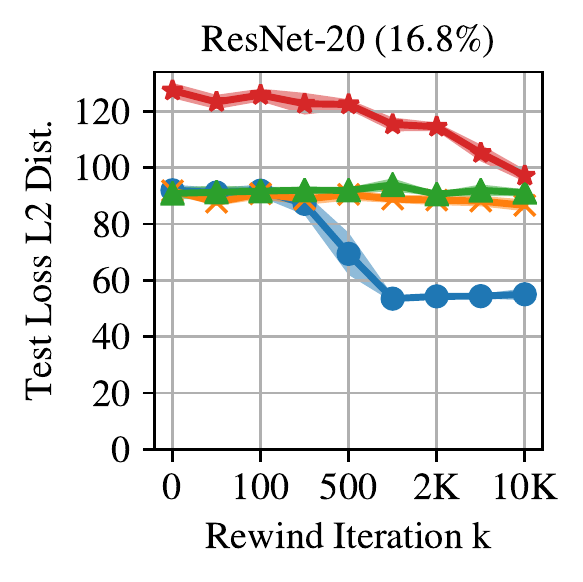}};

\node at (0.0,  -0.24) {\includegraphics[width=0.19\textwidth]{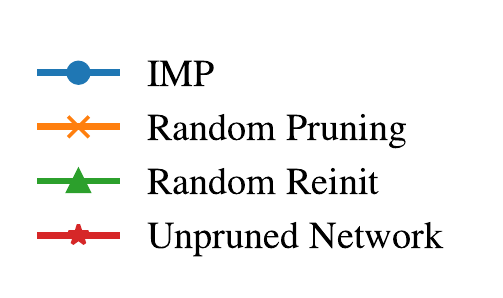}};
\node at (0.2,  -0.19) {\includegraphics[width=0.19\textwidth]{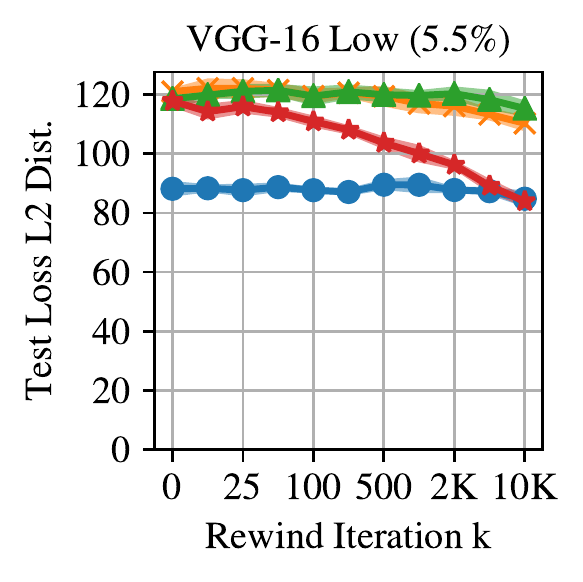}};
\node at (0.4,  -0.19) {\includegraphics[width=0.19\textwidth]{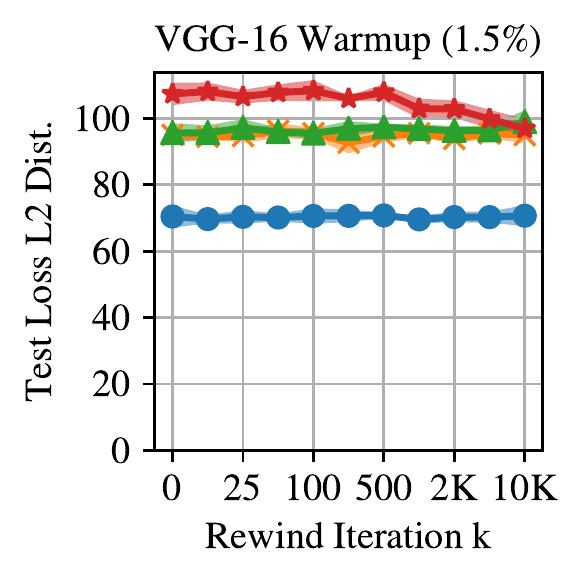}};
\node at (0.6,  -0.19) {\includegraphics[width=0.19\textwidth]{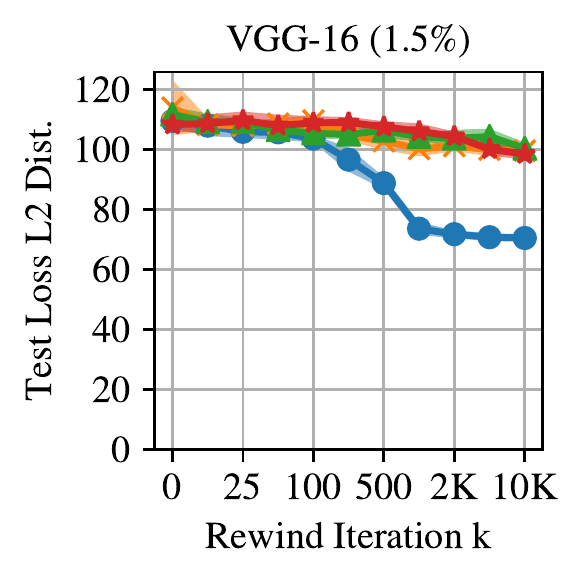}};
\end{tikzpicture}

\caption{The $L_2$ distance between the per-example losses on the test set for networks that are created by training the full network to iteration $k$, optionally applying a pruning mask, and training on different data orders from there.
 Each line is the mean and standard deviation across three initializations and three data orders (nine samples in total). Percents are percents of weights remaining. We did not compute the train set quantities for the ImageNet networks due to computational limitations.}
\label{fig:app-loss-dist-test}
\end{figure*}

\begin{figure*}
\centering
\begin{tikzpicture}[x=\textwidth,y=\textwidth, every node/.style = {anchor=north west}]
\node at (0.0, 0) {\includegraphics[width=0.19\textwidth]{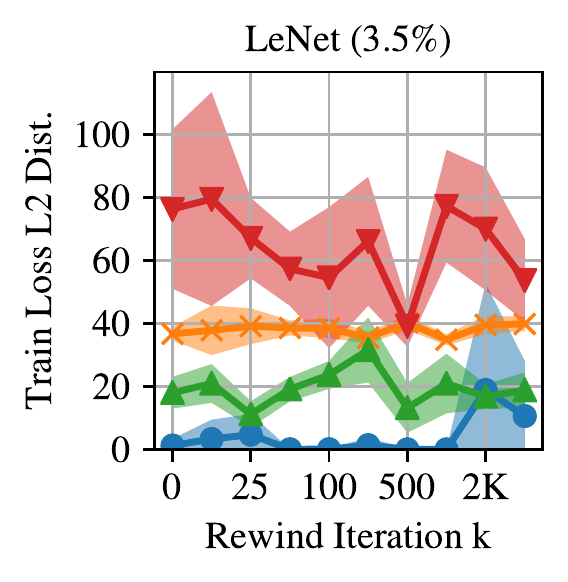}};
\node at (0.2, 0) {\includegraphics[width=0.19\textwidth]{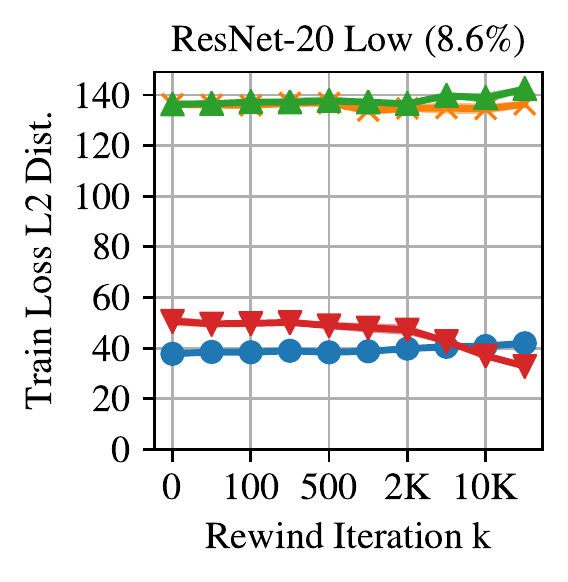}};
\node at (0.4, 0) {\includegraphics[width=0.19\textwidth]{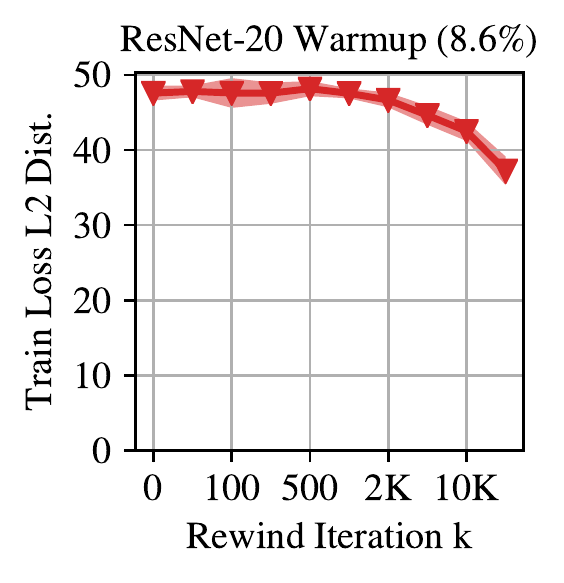}};
\node at (0.6, 0) {\includegraphics[width=0.19\textwidth]{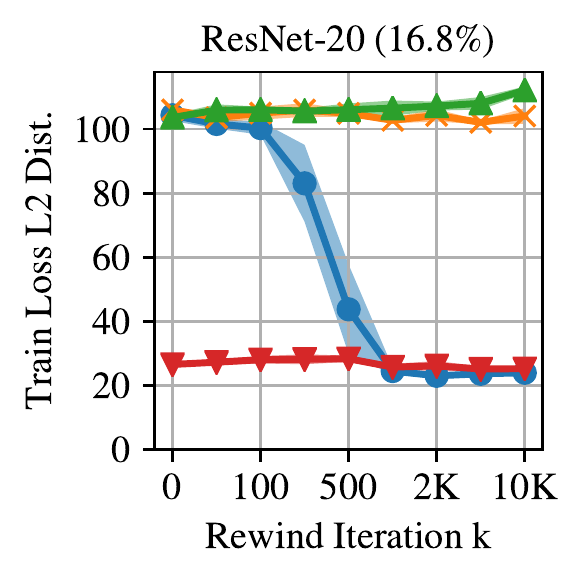}};

\node at (0.0,  -0.24) {\includegraphics[width=0.19\textwidth]{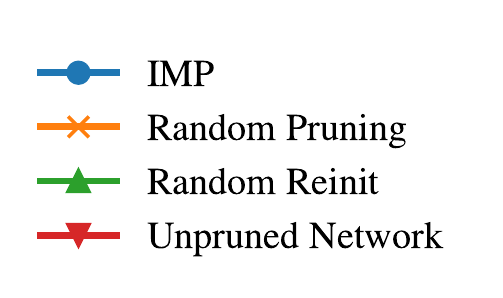}};
\node at (0.2,  -0.19) {\includegraphics[width=0.19\textwidth]{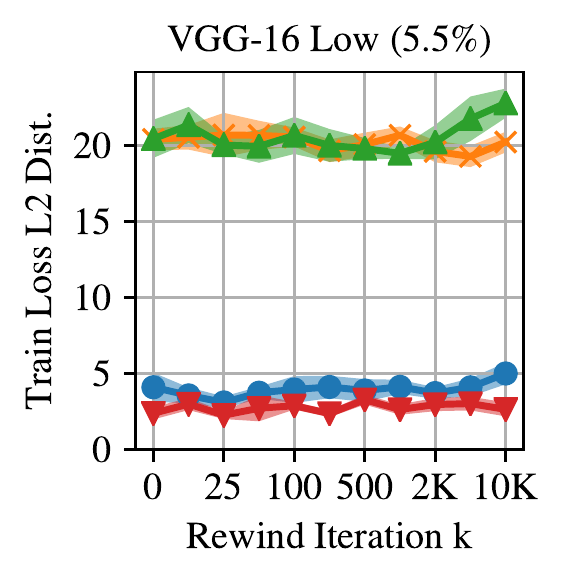}};
\node at (0.4,  -0.19) {\includegraphics[width=0.19\textwidth]{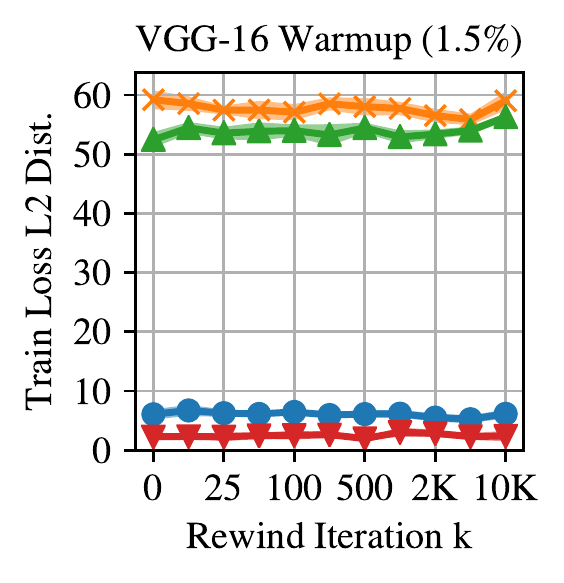}};
\node at (0.6,  -0.19) {\includegraphics[width=0.19\textwidth]{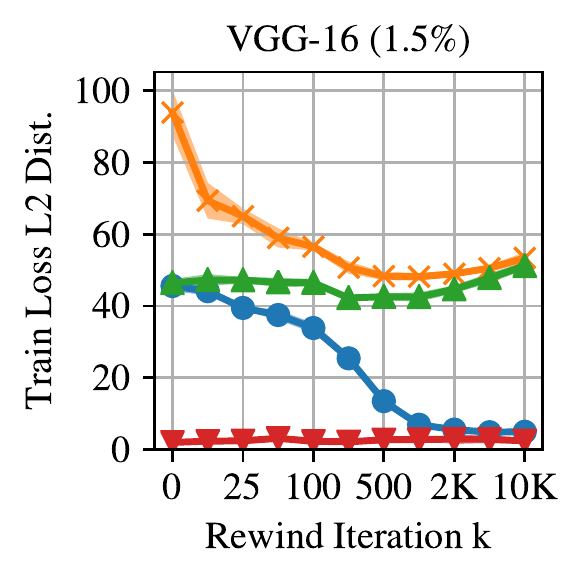}};
\end{tikzpicture}

\caption{The $L_2$ distance between the per-example losses on the train set for networks that are created by training the full network to iteration $k$, optionally applying a pruning mask, and training on different data orders from there.
 Each line is the mean and standard deviation across three initializations and three data orders (nine samples in total). Percents are percents of weights remaining. We did not compute the train set quantities for the ImageNet networks due to computational limitations.}
\label{fig:app-loss-dist-train}
\end{figure*}

\textbf{Loss $L_2$ distance.}
This distance function computes the $L_2$ distance between the vector of cross-entropy losses aggregated by computing the loss on each example.
This function again considers only the functional behavior of the networks, but it uses the per-example loss rather than the classification decisions, which may provide more information about the functional behavior of the networks.
We plot this data in Figures \ref{fig:app-loss-dist-test} (test set) and \ref{fig:app-loss-dist-train} (train set). It largely mirrors the behavior from the classification difference function, and the same interpretations apply.

\end{appendix}

\end{document}